\pdfoutput=1

\documentclass[final,1p,times]{elsarticle}

\newcommand{\paperTitle}[0]{Commonsense Visual Sensemaking for Autonomous Driving}
\newcommand{\paperSubTitle}[0]{}

\usepackage[colorinlistoftodos,bordercolor=white,backgroundcolor=red!60!white,linecolor=red!60!white,textsize=scriptsize]{todonotes}

\newcommand{\makeBlue}[1]{{\color{blue!70!black}#1}}

\newcommand{\appendixNote}[1]{{\color{black}\small\sffamily#1}}
\newcommand{\aspMeta}[1]{{\color{black}\bf #1}}
\newcommand{\appSkip}[0]{\smallskip}

\usepackage{amssymb}

\usepackage{url}
\usepackage{xcolor}

\usepackage{graphicx}
\usepackage{amsmath}
\usepackage{booktabs}
\usepackage[english]{babel}

\usepackage[export]{adjustbox}

\usepackage{colortbl}
\makeatletter
\def\hlinewd#1{%
  \noalign{\ifnum0=`}\fi\hrule \@height #1 \futurelet
   \reserved@a\@xhline}
\makeatother

\usepackage{latexsym}
\usepackage{amsmath}
\usepackage{amssymb}

\usepackage{subcaption}
\usepackage{caption}

\usepackage{enumitem}

\usepackage[ruled,vlined,linesnumbered]{algorithm2e}

\usepackage[frozencache=true,cachedir=.]{minted}

\usepackage{natbib}

\title{\large\sffamily\textbf{\paperTitle}\\[8pt]{\normalsize On Generalised Neurosymbolic Online Abduction Integrating Vision and Semantics}}

\usepackage[pdftitle={\paperTitle: \paperSubTitle},
pdfauthor={Jakob Suchan, Mehul Bhatt, and Srikrishna Varadarajan},
linkbordercolor=white,
bookmarksopen=false,
bookmarksnumbered=false]{hyperref}

\hypersetup{
    pdftitle={\paperTitle: \paperSubTitle}, 						
    pdfauthor={Jakob Suchan, Mehul Bhatt, and Srikrishna Varadarajan},     					
    pdfsubject={Artificial Intelligence; Computer Vision; Cognitive Vision; Autonomous Driving; Explainable AI; Neurosymbolism}, 									
    pdfkeywords={Artificial Intelligence; Computer Vision; Cognitive Vision; Autonomous Driving; Explainable AI}, 						
    unicode=false,          									
    pdftoolbar=false,        									
    pdfmenubar=true,        								
    pdffitwindow=false,     									
    pdfstartview={FitH},    								
    pdfnewwindow=true,      								
    bookmarks=true,         								
    colorlinks=true,       									
    linkcolor=red!80!black,          							
    citecolor=blue!70black,        							
    filecolor=blue!80black,      									
    urlcolor=red!80!black,          						
}

\let\oldnl\nl
\newcommand{\nonl}{\renewcommand{\nl}{\let\nl\oldnl}}

\newcommand{\predTh}[1]{{\operatorname{\mathsf{#1}}}}
\newcommand{\predThF}[1]{{\operatorname{\mathsf{#1}}}}

\newcommand{\timePoints}[0]{$\mathcal{T}$}

\newcommand{\objectSortNormal}[0]{$\mathcal{O}$}
\newcommand{\objectSort}[0]{$\mathcal{O}$}

\newcommand{\motionTrack}[0]{$\mathcal{MT}$}

\newcommand{\spatialPrimitivesNormal}[0]{$\mathcal{E}$}
\newcommand{\spatialPrimitives}[0]{$\mathcal{E}$}
\newcommand{\spatialRelations}[0]{$\mathcal{R}$}

\definecolor{YellowGreen}{RGB}{160,200,40}
\definecolor{mathcolor}{RGB}{7,72,110}

\journal{Artificial Intelligence Journal}

\begin{document}

\begin{frontmatter}



\author{{\sffamily \textbf{Jakob Suchan}$~~~$--$~~~$\textbf{Mehul Bhatt}$~~~$--$~~~$\textbf{Srikrishna Varadarajan}}}

\address{\upshape{University of Bremen  (Germany)$~~~$---$~~~$\"{O}rebro University (Sweden)}\\[8pt]\textbf{CoDesign Lab (EU)}$~~~$$\mathbf{>}$$~~$Cognitive Vision\\{\upshape{\href{https://codesign-lab.org}{www.codesign-lab.org}}} $~~$\textbf{/}$~~$ {\upshape{\href{info@codesign-lab.org}{info@codesign-lab.org}}}}

\begin{abstract}{{\small\sffamily We  demonstrate the need and potential of systematically integrated \emph{vision} and \emph{semantics} solutions for visual sensemaking in the backdrop of autonomous driving. 
A general neurosymbolic method for \emph{online} visual sensemaking using answer set programming (ASP) is systematically formalised and fully implemented. The method integrates state of the art in visual computing, and is developed as a modular framework that is generally usable within hybrid architectures for realtime perception and control. We evaluate and demonstrate with community established benchmarks {KITTIMOD}, {MOT-2017}, and  {MOT-2020}. As use-case, we focus on the significance of human-centred visual sensemaking ---e.g., involving semantic representation and explainability, question-answering, commonsense interpolation--- in safety-critical autonomous driving situations. The developed neurosymbolic framework is domain-independent, with the case of autonomous driving designed to serve as an exemplar for online visual sensemaking in diverse cognitive interaction settings in the backdrop of select human-centred AI technology design considerations.}

$~$

}
\end{abstract}


\end{frontmatter}


\medskip
\medskip
\medskip

{\normalsize\sffamily\textbf{PUBLICATION NOTE}}.

{\sffamily \footnotesize

This is a preprint / review version of an accepted contribution to be published as part of the Artificial Intelligence Journal (AIJ).$^{*}$ The article is an extended version of an IJCAI 2019 publication \citep{out-of-sight-2019}. The overall scientific agenda (pertaining to Cognitive Vision and Deep Semantics \citep{BhattECAI2020}) driving this research is available at:

\medskip

$~$\quad\quad\textbf{CoDesign Lab (EU)}$~~~$$>$$~~~$Cognitive Vision$~~~$/$~~~$\url{https://codesign-lab.org/cognitive-vision/}

$~$\quad\quad Related select publications:$~~$\url{https://codesign-lab.org/select-papers/#cognitive_vision}

\vfill

{\scriptsize $^{*}$The AIJ published version is final; it also fully incorporates all reviewer feedback.}
}

\newpage

\section{\uppercase{Motivation}}\label{sec:motivation}
Autonomous driving research has received enormous academic \& industrial interest in recent years (Sec \ref{sec:related-work}). This surge has coincided with (and been driven by) advances in \emph{deep learning} based computer vision research. Although end-to-end {deep learning} based vision \& control has (arguably) been successful for self-driving vehicles, we posit that there is a clear need and tremendous potential for hybrid visual sensemaking solutions that integrate \emph{vision and semantics} towards fulfilling essential legal and ethical responsibilities involving explainability, human-centred AI (Artificial Intelligence), and industrial standardisation (e.g, pertaining to representation, realisation of rules and norms, fulfilling statutory obligations).

\medskip

\textbf{Autonomous Vehicles:$~~$``Standardisation and Regulation''}\quad 

As the self-driving vehicle industry develops further, it will be necessary to have an articulation and community consensus on aspects such as representation, interoperability, human-centred performance benchmarks, and data archival \& retrieval mechanisms. Within autonomous driving, the need for standardisation and ethical regulation has most recently garnered interest internationally, e.g.,  with the Federal Ministry of Transport and Digital Infrastructure in Germany (BMVI) taking a lead in eliciting 20 key propositions\footnote{The  $20$ key propositions elicited by the German federal ministry BMVI highlight a range of factors pertaining to safety, utilitarian considerations, human rights, statutory liability, technological transparency, data management and privacy etc   \citep{ethicalGermany2018}.} (with legal implications) for the fulfilment of ethical commitments for automated and connected driving systems \citep{ethicalGermany2018}. In spite of major investments in self-driving vehicle research, issues related to human-centred'ness, human collaboration, and standardisation have been barely addressed, with  the current focus in driving research primarily being on two basic considerations: \emph{how fast to drive, and which way and how much to steer}. This is necessary, but inadequate if autonomous vehicles are to become commonplace and function with humans \citep{moral-machine2018,Bonnefon1573}. Ethically driven standardisation and regulation will require addressing challenges in foundational human-centred AI technology design, e.g., pertaining to semantic visual interpretation, natural / multimodal human-machine interaction, high-level data analytics (e.g., for post hoc diagnostics, dispute settlement). This will necessitate ---amongst other things--- human-centred qualitative benchmarks and design \& evaluation of multifaceted hybrid solutions integrating diverse methodologies in Artificial Intelligence, Machine Learning, Cognitive Science, Design Science etc.

\begin{figure}[t]

\centering
\includegraphics[width=0.13\textwidth]{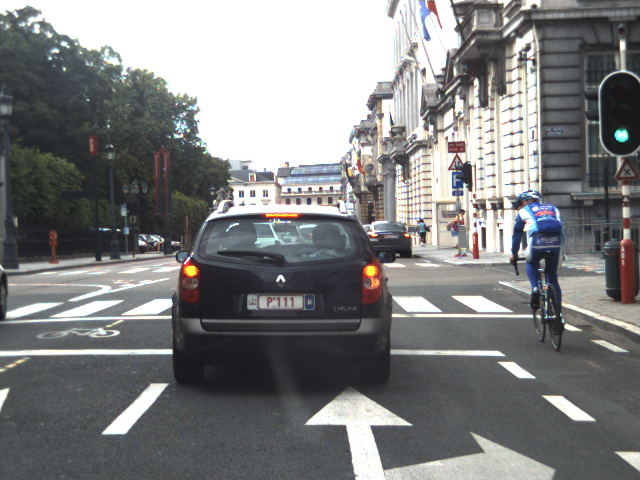}
\includegraphics[width=0.13\textwidth]{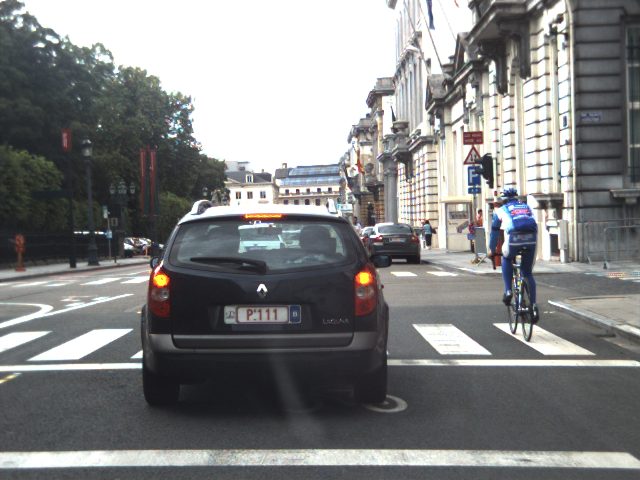}
\includegraphics[width=0.13\textwidth]{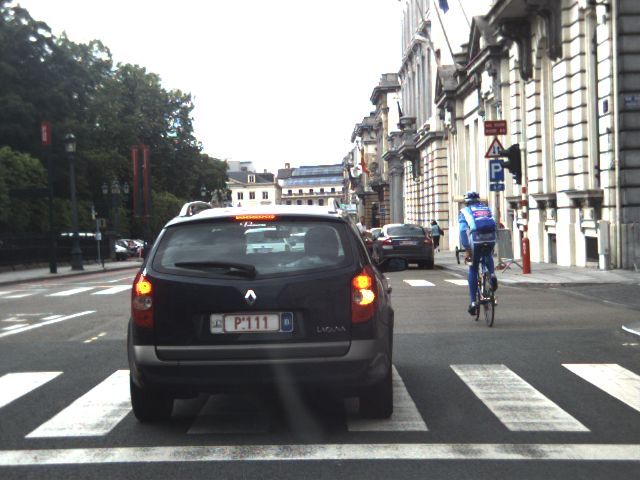}
\includegraphics[width=0.13\textwidth]{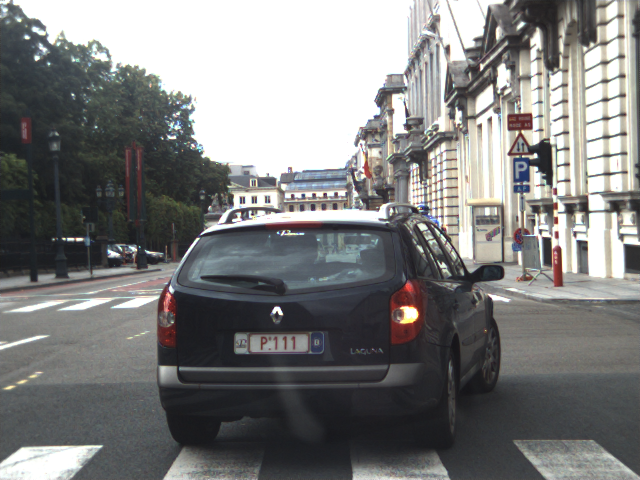}
\includegraphics[width=0.13\textwidth]{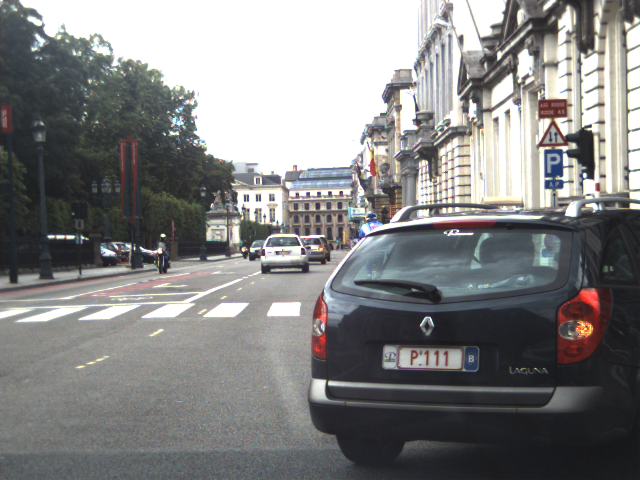}
\includegraphics[width=0.13\textwidth]{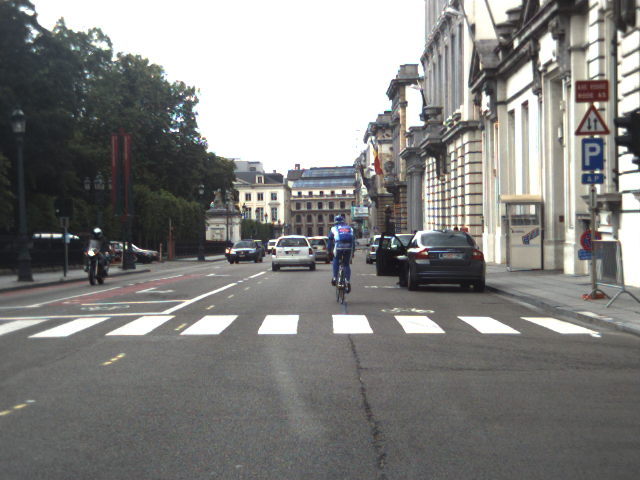}
\includegraphics[width=0.13\textwidth]{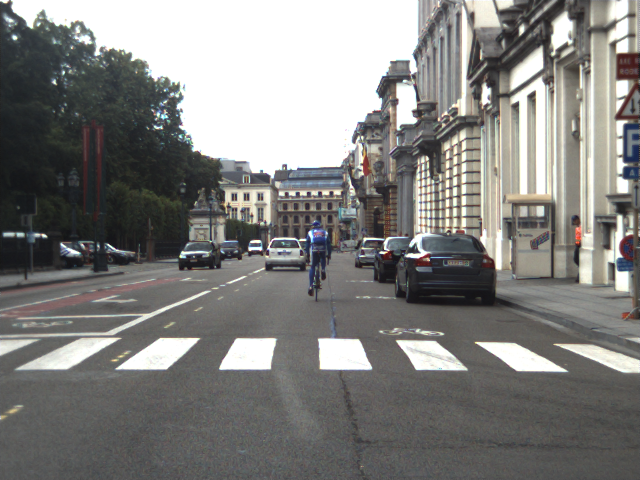}


\center
\includegraphics[width=0.9\columnwidth]{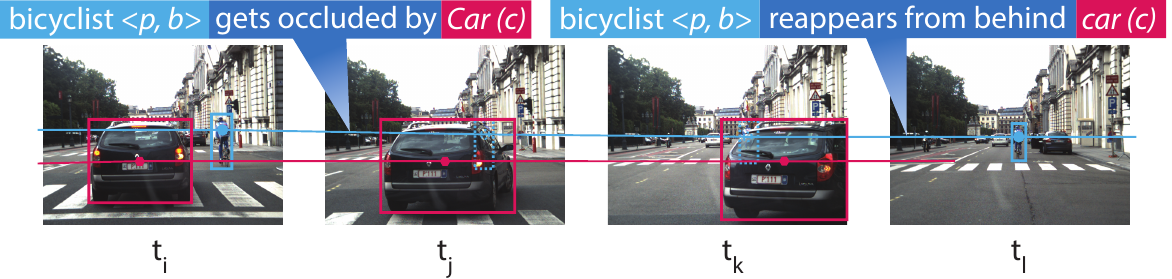}

\medskip
\includegraphics[width=0.9\columnwidth]{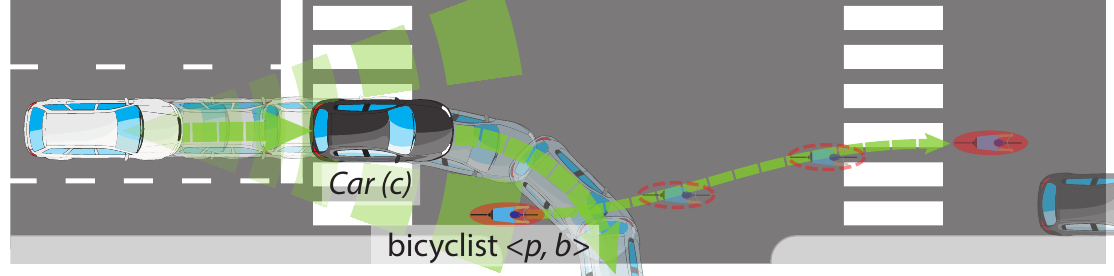}
\caption{\textbf{Out of sight but not out of mind}; the case of hidden entities: e.g., an occluded cyclist.}
\label{fig:first-page-scanario}
\end{figure}

\medskip

\textbf{Neurosymbolism: Visual Sensemaking Needs Both ``Vision and Semantics''}\quad  

Visual sensemaking requires a systematically developed general and modular integration of high-level techniques concerned with ``commonsense and semantics'' with low-level neural methods capable of computing primitive features of interest in visual data. Towards this, this research demonstrates the significance of semantically-driven methods rooted in knowledge representation and reasoning (KR) in addressing research questions pertaining to explainability and human-centred AI particularly from the viewpoint of (perceptual) sensemaking of dynamic visual imagery. This is done in the backdrop of the autonomous driving domain; as an example, consider the \emph{occlusion scenario} in Fig. \ref{fig:first-page-scanario}:

\begin{quote}
{\small\sffamily
\noindent Car ($c$) is {\color{blue!80!black}in-front}, and indicating to {\color{blue!80!black}turn-right}; {\color{blue!80!black}during} this time, person ($p$) is {\color{blue!80!black}on} a bicycle ($b$) and positioned {\color{blue!80!black}front-right} of $c$ and {\color{blue!80!black}moving-forward}. Car $c$ turns-right, during which the bicyclist $<p, b>$ is not {\color{blue!80!black}visible}. {\color{blue!80!black}Subsequently}, bicyclist $<p, b>$ {\color{blue!80!black}reappears}.
}
\end{quote}

\noindent The occlusion scenario of Fig. \ref{fig:first-page-scanario} is one of range of (seemingly) mundane safety-critical moments that one may regularly experience while driving a vehicle (Fig. \ref{fig:MT-examples-scene-safety}, Fig. \ref{fig:safety-critical-sits}-\ref{fig:safety-critical-sits_examples} and Table \ref{tbl:safety-critical-situations} include additional examples). This scenario is sufficiently indicative of  several challenges concerning \emph{epistemological} and \emph{phenomenological} aspects relevant to a wide range of \emph{dynamic spatial systems} \citep{Bhatt08-SCC-Dynamics,QSTR-Emerging-2011,Bhatt:RSAC:2012}: 

\begin{itemize}
	\item \textbf{projection and interpolation} of missing information (e.g., what could be hypothesised about {\small\sffamily bicyclist $<p, b>$} when it is \emph{occluded}; how can this hypothesis enable in planning 
	\item object \textbf{identity maintenance} at a semantic level, e.g., in the presence of occlusions, missing and noisy quantitative data, error in detection and tracking
	\item ability to make \textbf{default assumptions}, e.g., pertaining to persistence objects and/or object attributes  
	\item maintaining \textbf{consistent beliefs} respecting (domain-neutral) commonsense criteria, e.g., related to compositionality \& indirect effects, space-time continuity, positional changes resulting from motion
	\item inferring / computing \textbf{counterfactuals}, in a manner akin to human cognitive ability to perform mental simulation for purposes of introspection, performing ``what-if'' reasoning tasks to determine 
an immediate next step). 
	
\end{itemize}
		
Addressing such challenges ---be it realtime or post-hoc--- in view of human-centred AI concerns pertaining to representations rooted to natural language, explainability, ethics and regulation requires a systematic (neurosymbolic) integration of {\textbf{Semantics and Vision}}, i.e., robust commonsense representation \& inference about spacetime dynamics on the one hand, and powerful low-level visual computing capabilities, e.g., pertaining to object detection and tracking on the other.

\medskip

\textbf{Deep Semantics: (Systematically) ``Integrating AI and Vision''}\quad  

The development of domain-independent computational models of {perceptual sensemaking} ---e.g., encompassing capabilities such as visuospatial Q/A, spatio-temporal relational learning, visuospatial abduction--- with multimodal human behavioural stimuli such as RGB(D), video, audio, eye-tracking requires the representational and inferential mediation of commonsense and spatio-linguistically rooted abstractions of space, motion, actions, events and interaction. We characterise \textbf{Deep Semantics} \citep{BhattECAI2020} within a declarative AI setting as: 

\smallskip

{\tiny$\blacktriangleright$}\quad general methods for the processing and semantic interpretation of dynamic visuospatial imagery with an emphasis on the ability to {\textbf{abstract, learn, and reason}}  with cognitively rooted structured characterisations of commonsense knowledge about {\textbf{space and motion}}.

{\tiny$\blacktriangleright$}\quad the existence of declarative models --e.g., pertaining to space, space-time, motion, actions \& events, spatio-linguistic conceptual knowledge (e.g., Table \ref{tbl:relations})-- and corresponding formalisation supporting (domain-neutral) \textbf{reasoning capabilities} (e.g.,  visual Q/A and learning, non-monotonic visuospatial abduction)

\smallskip
Formal semantics and computational models of deep semantics manifest themselves in declarative AI settings such as {constraint logic programming}, {inductive logic programming}, and {answer set programming}. Naturally, a practical illustration of the intergated ``AI and Vision'' method requires a tight but modular integration of the (declarative) commonsense  spatio-temporal abstraction and reasoning  with robust low-level visual computing foundations (primarily) driven by state of the art visual computing techniques (e.g., for visual feature detection, tracking).

\medskip
\medskip

\textbf{\uppercase{Key Contributions}} 

This research is situated within the broader auspices of the scientific agenda of cognitive vision and perception, which addresses visual, visuospatial and visuo-locomotive perception and interaction from the viewpoints of language, logic, spatial cognition and artificial intelligence \citep{BhattECAI2020} (Sec \ref{sec:related-work}). The key contribution of this paper is to develop a general and systematic declarative visual sensemaking method capable of \emph{online abduction}: {\small\textbf{realtime, incremental, commonsense}} question-answering and belief maintenance over dynamic visuospatial imagery. Supported are {\small\bf(1--3)}:

\medskip

{\small\bf(1)}.\quad {\small\bf Human-Centred Representation for Space and Motion}

Declaratively modelled ontological characterisation of human-centric relational representations that are semantically rooted to commonsense spatio-linguistic primitives pertaining to space and motion as they occur in natural language \citep{Bhatt-Schultz-Freksa:2013,mani-james-motion}.

\smallskip

{\small\bf(2)}.\quad {\small\bf Systematic High-level Abductive Reasoning}

Driven by Answer Set Programming {(ASP)} \citep{Brewka:2011:ASP}, the ability to abductively compute commonsense interpretations and explanations in a range of (a)typical everyday driving situations, e.g., concerning safety-critical decision-making;\quad the declarative model of space and motion, in addition to supporting abductive reasoning about space and change, is also naturally amenable to high-level semantic interpretation (e.g., by question answering) for post-hoc analytical purposes (e.g., as might be relevant in situations requiring diagnosis et al for litigation, insurance claims).

\smallskip

{\small\bf(3)}.\quad {\small\bf Online Performance of Modularly Integrated Vision and Semantics}

Online performance --in an ``active vision'' context--  of the overall framework modularly integrating high-level commonsense reasoning component with state of the art low-level (deep learning based) visual computing for practical application in real world settings (with autonomous driving serving as a solid demonstration platform). 

\medskip
\medskip

\textbf{\uppercase{Organisation of the Paper}}.\quad 

\emph{The rest of the article is organised as follows}:

\begin{itemize}

	\item \textbf{Section \ref{sec:ontology}} presents the ontological and formal representational foundations of the developed visual sensemaking framework; main focus is on  the commonsense representation aspects pertaining to the modelling of space, space-time, motion, events, and other aspects relevant to modelling and reasoning about spatio-temporal dynamics.
	
	\item \textbf{Section \ref{sec:vissenmaking}} presents the overall visual sensemaking framework and its technical implementation with a central focus on the general answer set programming based method for online abduction; we elaborate on the declarative model directly vis-a-vis the ASP implementation.

	\item \textbf{Section \ref{sec:application-sec}} demonstrates \& empirically evaluates the core online abduction component with community established real-world datasets and benchmarks, namely: {KITTIMOD} \citep{Geiger2012CVPR}, {MOT-17} \citep{MOT16-Benchmark}, and MOT-20 \cite{MOTChallenge20}.
	
	
	\item 	\textbf{Section \ref{sec:related-work}} discusses related works primarily from the viewpoints of knowledge representation, and visual computing as pursued in computer vision research.

	\item 	\textbf{Section \ref{sec:summar-and-outlook}} concludes with a brief summary of our work, together with pointers to immediate research questions for follow-up, as well as more broad-based directions that this work aims to open up.
	
\end{itemize}


\textbf{Appendices A--C}.\quad \ref{appA} provides annotations of select Answer Set Programming source code relevant to the declarative model presented in Section \ref{sec:vissenmaking}. \ref{appB} presents additional examples chosen from community benchmark datasets together with sample data; it also includes an elaborated version of a running example used in the paper. \ref{app:data} provide a succinct view of (select) data corresponding to (select) scenes.

\section{\uppercase{Commonsense $~$--$~$ Space $~$--$~$ Motion}:\\\uppercase{Ontological and Representational Aspects}}\label{sec:ontology}

We present the ontological and formal representational foundations of the developed visual sensemaking framework while focussing on  the commonsense representational aspects pertaining to the modelling of space, space-time, motion, events, and other aspects relevant to modelling and reasoning about spatio-temporal dynamics. Towards this, Table \ref{tbl:ontology} summarises the individual constituents of $\Sigma_{st}$ (spatiotemporal primitives) and  $\Sigma_{dyn}$ (spatiotemporal dynamics), and Table \ref{tbl:relations} elaborates the supported commonsense relations for the abstraction of space, motion, and (inter)action. Figure \ref{fig:s-t-entities} is a (non-exhaustive) collection of generic / domain-neutral spacetime motion patterns supported; Figures \ref{fig:MT-DrivingDomain} and \ref{fig:MT-examples-scene-safety} include concrete instance of such generic motion patterns: Fig. \ref{fig:MT-DrivingDomain} illustrates motion patterns for \emph{approach}, \emph{occlusion}, and \emph{connected motion}; and Fig. \ref{fig:MT-examples-scene-safety} illustrates the motion patterns underlying a security-critical scenario involved an elaborate lane changing episode.

\subsection{{Commonsense Abstractions for Space and Motion}}
Commonsense spatio-temporal relations and patterns (e.g., \emph{left}, \emph{touching}, \emph{part of}, \emph{during}, \emph{collision}) offer a human-centered and cognitively
adequate formalism for semantic grounding and automated reasoning for everyday (embodied) multimodal interactions \citep{Bhatt-Schultz-Freksa:2013,mani-james-motion}.
Qualitative, multi-domain\footnote{Multi-domain refers to more than one aspect of space, e.g., topology, orientation, direction, distance, shape; this requires a mixed domain ontology involving points, line-segments, polygons, and regions of space, time, and space-time \citep{ASPMTQS2015-Bhatt,ruleml-space-time2018,Hazarika:2005:thesis}.} representations
of spatial, temporal, and spatio-temporal relations and motion patterns (e.g., Fig \ref{fig:MT-DrivingDomain}-\ref{fig:s-t-entities}), and their mutual transitions can provide a mapping between high-level semantic models of actions and events on one hand, and low-level /  quantitative trajectory data emanating from visual computing algorithms on the other. For instance, by spatio-linguistically grounding complex trajectory data --e.g., pertaining to on-road moving objects-- to a formal framework of space and motion,  generalized (activity-based) commonsense reasoning about dynamic scenes, spatial relations, and motion trajectories denoting single and multi-object path \& motion predicates can be supported. For instance, such predicates can be abstracted within a region-based 4D space-time framework \citep{Hazarika:2005:thesis,DBLP:conf/ecai/BennettCTH00,ruleml-space-time2018}, object interactions \citep{DBLP:journals/ai/Davis08,DBLP:journals/ai/Davis11}, or even spatio-temporal narrative knowledge. An adequate commonsense spatio-temporal representation can, therefore, connect with low-level quantitative data, and also help to ground symbolic descriptions of actions and objects to be queried, reasoned about, or even manipulated in the real world.

\begin{table}[t]
\renewcommand{\arraystretch}{1.5}
\footnotesize
\begin{center}
\begin{tabular}{>{\columncolor[gray]{0.92}} p{4.6cm}  l  r}

\hlinewd{1pt}

\textbf{\color{blue!70!black}{\sffamily\uppercase{Ontology$~~$/$~~$Space \& Motion}}} & {\sffamily\textbf{REPRESENTATION}} &    \\\hline
 \hline

\multicolumn{2} {l} {\textbf{Spatio-Temporal Ontology} ({$\Sigma_{st}$})}\\
 
 \hline


Domain Objects   & \objectSort$~~$= $~~$$\{o_1, ... , o_n\}$ &  e.g., \emph{cars, people, cyclists} \\ [10pt]

Spatial Entities   & \spatialPrimitives$~~$= $~~$$\{\varepsilon_{1}, ..., \varepsilon_{n}\}$ &  \emph{points, line-segments, rectangles} \\ [10pt]

Time & \timePoints$~~$= $~~$$\{t_{1}, ..., t_{n}\}$ &  \emph{time-points, time-intervals} \\ [10pt]

Motion & {$\mathcal{MT}_{o_i}$}$~~$=$~~$($\varepsilon_{t_{s}}, ..., \varepsilon_{t_{e}}$) &  \emph{motion tracks / space-time histories} \\ [10pt]

Spatio-Temporal Relationships & \spatialRelations$~~$ &  \emph{e.g., topology, orientation, distance} \\ [10pt]

\hline

\multicolumn{2} {l} {\textbf{Spatio-Temporal Dynamics} ({$\Sigma_{dyn}$})}\\
 
 \hline

Fluents & ${\Phi}$~~$=$~~$\{\phi_1, ... , \phi_n\}$ &  \emph{e.g., visibility, hidden\_by, clipped} \\ [10pt]

Events & ${\Theta}$~~$=$~~$\{\theta_1, ... , \theta_n\}$  &  \emph{e.g., hides\_behind, missing\_detections} \\ \hline 





\multicolumn{2} {l} {\textbf{Problem Specification}}\\
 
 \hline

Visual Observations   & ${\mathcal{VO}_t}$~~$=$~~$\{obs_1, ... , obs_n\}$ &  \emph{e.g., \spatialPrimitives{} corresponding to object detections} \\ [10pt]

Predictions & ${\mathcal{P}_t}$~~$=$~~$\{p_{trk_1}, ... , p_{trk_n}\}$  &  \emph{e.g.,  \spatialPrimitives{} for predicted track} \\ 

Matching Likelihood & ${\mathcal{ML}_t}$~~$=$~~$\{ml_{trk_1, obs_1}, ... , ml_{trk_n, obs_m}\}$  &  \emph{e.g., IoU between tracks and detections} \\ 


\hline 


\multicolumn{2} {l} {\textbf{Hypothesis}}\\
 
 \hline

Assignments   & $\mathcal{H}^{assign}$ &  \emph{abduced assignments} \\ [10pt]

Events & $\mathcal{H}^{events}$~~$=$~~$\{\theta_1, ... , \theta_n\}$  &  \emph{abduced event sequence} \\ 

Explanations &$\mathcal{EXP} ~~ \leftarrow~~ <\mathcal{H}^{events}, \mathcal{MT}>$& \emph{scene dynamics; abduced events}\\[-2pt]
&& \emph{and corresponding motion tracks}\\

\hline 

\hlinewd{1pt}



\end{tabular}
\end{center}
\caption{Commonsense -- Space -- Motion: Ontological and Representational Aspects}
\label{tbl:ontology}
\end{table}%

\begin{table}[t]
\centering
{
\scriptsize\sffamily
\renewcommand{\arraystretch}{1.5}

\begin{tabular}{>{\columncolor[gray]{0.92}}l p{5.8 cm} p{3 cm}}

\hlinewd{1pt}
\textbf{\sffamily\color{blue!70!black}\uppercase{Spatio-Temporal Domain} ($\mathcal{QS}$)}   & {\sffamily\textbf{Spatial, Time, Motion Relations}$~$ ({\color{blue!70!black}$\mathcal{R}$})} & {\sffamily\textbf{Entities} $~$({\color{blue!90!black}$\mathcal{E}$})} \\
\hline

{\bf\sffamily Mereotopology} & 
{\scriptsize\sffamily disconnected (dc), external contact (ec), partial overlap (po), tangential proper part (tpp), non-tangential proper part (ntpp), proper part (pp), part of (p), discrete (dr), overlap (o), contact (c)} & 
arbitrary rectangles, circles, polygons, cuboids, spheres \\[20pt]

{\bf\sffamily Incidence} & 
{\scriptsize\sffamily interior, on boundary, exterior, discrete, intersects} & 
2D point with rectangles, circles, polygons; 3D point with cuboids, spheres \\[20pt]

{\bf\sffamily Orientation} & 
{\scriptsize\sffamily left, right, collinear, front, back, on, facing towards, facing away, same direction, opposite direction} & 
2D point, circle, polygon with 2D line\\[20pt]

{\bf\sffamily Distance, Size} &
{\scriptsize\sffamily adjacent, near, far, smaller, equi-sized, larger} & 
rectangles, circles, polygons, cuboids, spheres\\[20pt]

{\bf\sffamily Motion} & 
{\sffamily {moving}: towards, away, parallel; growing / shrinking:  vertically, horizontally; splitting / merging; rotation: left, right, up, down, clockwise, couter-clockwise} &
rectangles, circles, polygons, cuboids, spheres\\[20pt]

{\bf\sffamily Time} &
{\scriptsize\sffamily before, after, meets, overlaps, starts, during, finishes, equals} & 
time-points, time intervals\\
\hlinewd{1pt}


\end{tabular}
}
  \caption{Commonsense Relations for Abstract Representation of Space, Motion, Interaction}
\label{tbl:relations}
\end{table}

\subsection{{Space, Motion, Objects, Events, Change: Ontology and Formal Model}}\label{subsec:onto-formal-model}
\label{sub_sec:space_motion}
Reasoning about spatio-temporal dynamics is based on high-level representations of objects, and their respective motion \& mutual interactions in spacetime. Foundational ontological primitives for commonsense representation and reasoning about spatio-temporal dynamics are: 

\begin{itemize}

	\item {$\Sigma_{st}$} corresponds to primitives for representing space, time, motion and scene-level relational spatiotemporal structure

	\item {$\Sigma_{dyn}$} corresponds to the domain-independent commonsense theory for representing and reasoning about change.
	
\end{itemize}	

\medskip


$\mathbf{\Sigma_{st}}$ {$<$\objectSort, \spatialPrimitives, \timePoints, \motionTrack, \spatialRelations$>$}  and  $\mathbf{\Sigma_{dyn}}$ {$<${$\Phi$}}, {{$\Theta$}$>$} are as follows (Tables \ref{tbl:ontology} and \ref{tbl:relations}):


\begin{figure}
\centering
\subcaptionbox*{a) Moving Towards and Occluding (Two cars crossing eachother)}{\includegraphics[width=0.45\textwidth]{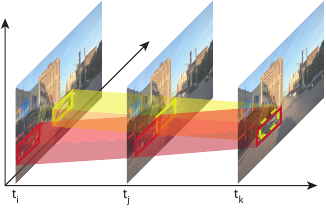}}
\quad
\subcaptionbox*{b) Connected Motion (Person on a Bicycle)}{\includegraphics[width=0.45\textwidth]{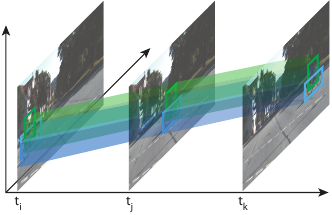}}
\caption{Space-Time Histories in Context: Motion Tracks Under Conditions of Occlusion and Partial Overlapp}
\label{fig:MT-DrivingDomain}
\end{figure}


\begin{figure*}[t] 
\centering
\subcaptionbox*{$\predTh{discrete}(o_1,o_2)$}{\includegraphics[height = 0.5in]{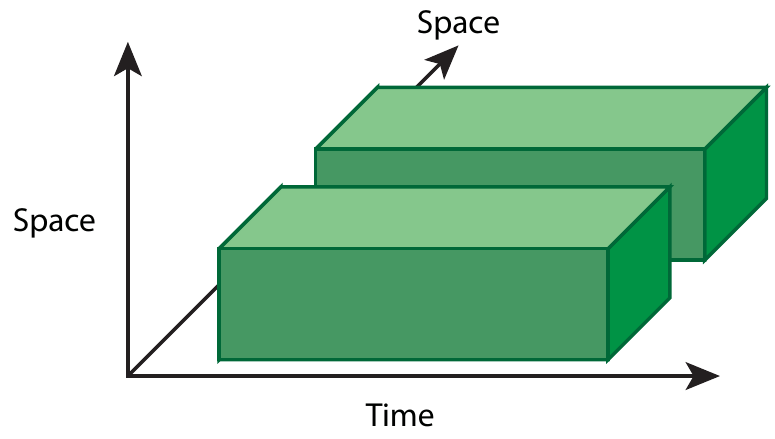}}\hfill
\subcaptionbox*{$\predTh{overlapping}(o_1,o_2)$}{\includegraphics[height = 0.5in]{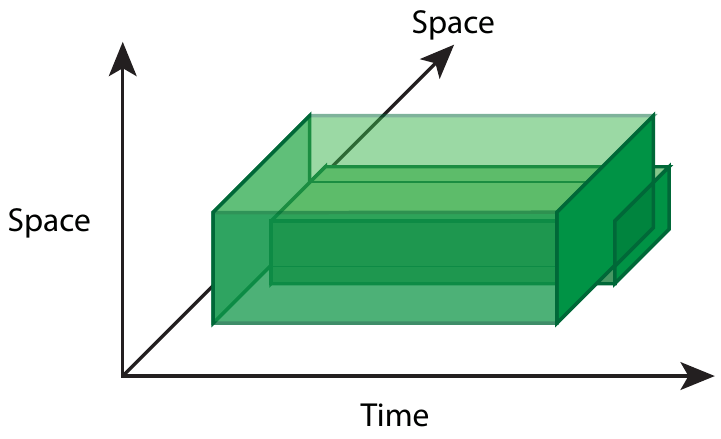}}\hfill
\subcaptionbox*{$\predTh{inside}(o_1,o_2)$}{\includegraphics[height = 0.5in]{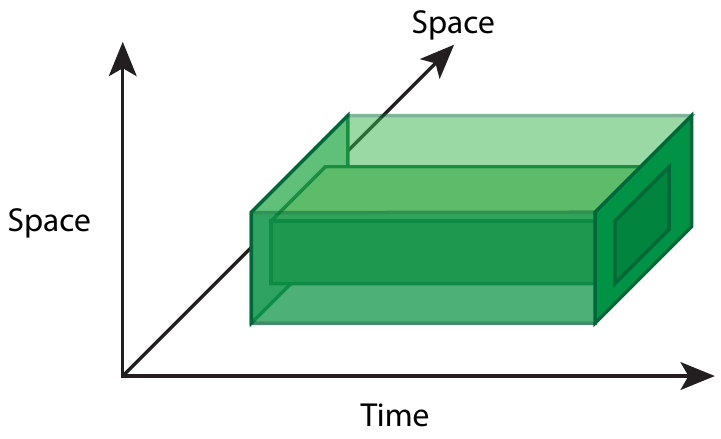}}\hfill
\subcaptionbox*{$\predTh{moving}(o)$}{\includegraphics[height = 0.5in]{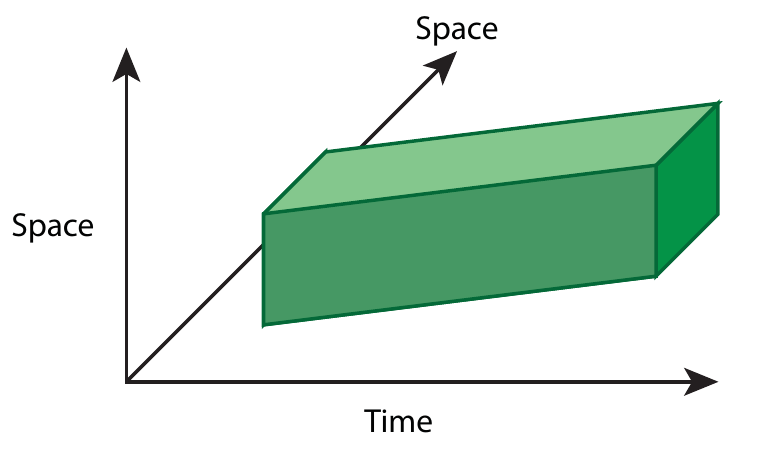}}\hfill
\subcaptionbox*{$\predTh{stationary}(o)$}{\includegraphics[height = 0.5in]{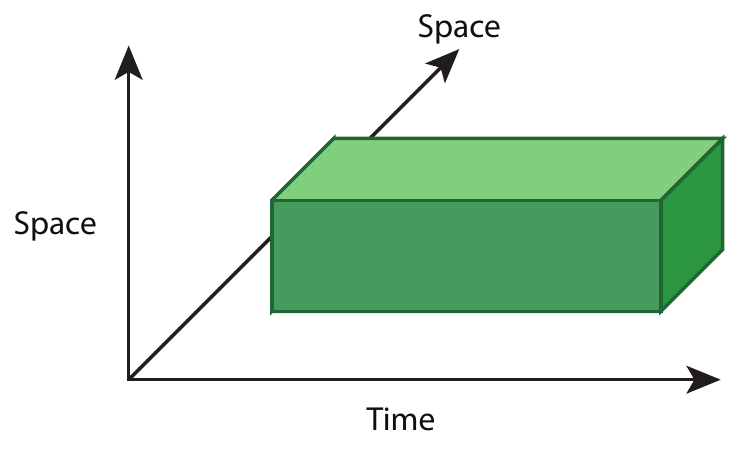}}

\vspace{8pt}
\subcaptionbox*{$\predTh{growing}(o)$}{\includegraphics[height = 0.5in]{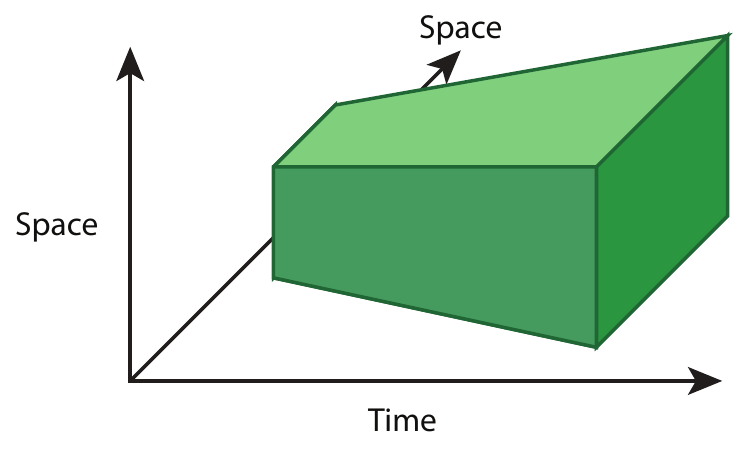}}\hfill
\subcaptionbox*{$\predTh{shrinking}(o)$}{\includegraphics[height = 0.5in]{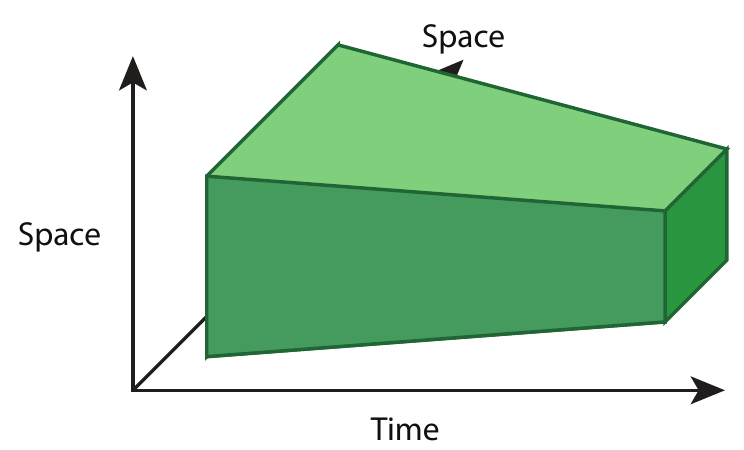}}\hfill
\subcaptionbox*{$\predTh{parallel}(o_1,o_2)$}{\includegraphics[height = 0.5in]{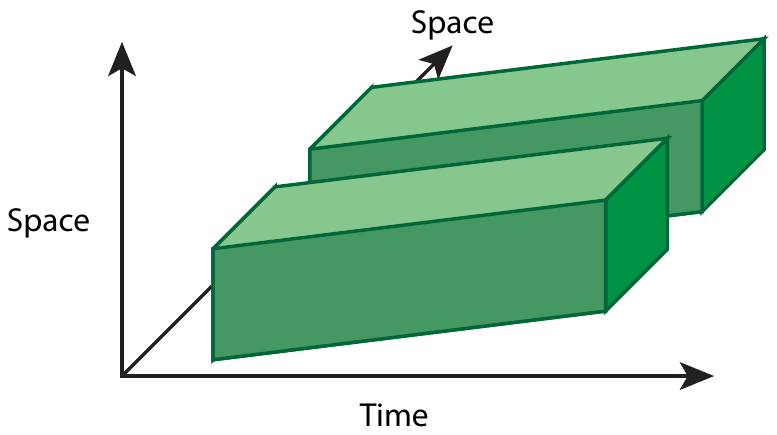}}\hfill
\subcaptionbox*{$\predTh{merging}(o_1,o_2)$}{\includegraphics[height = 0.5in]{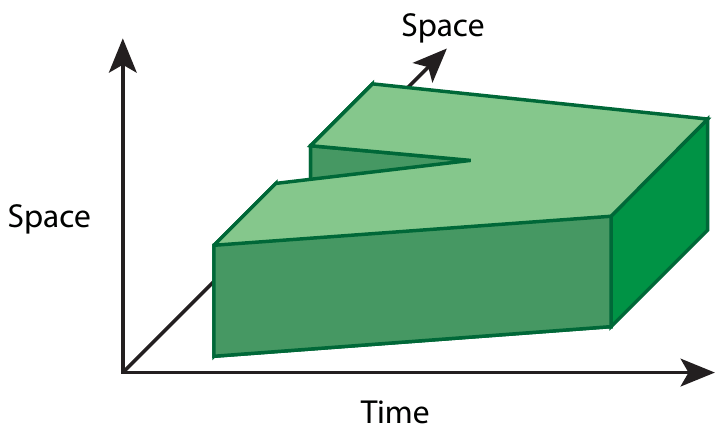}}\hfill
\subcaptionbox*{$\predTh{splitting}(o_1,o_2)$}{\includegraphics[height = 0.5in]{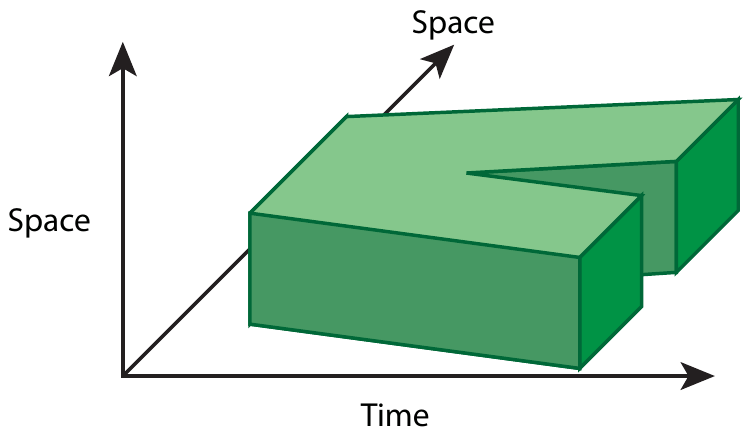}}

\vspace{8pt}
\subcaptionbox*{$\predTh{curved}(o)$}{\includegraphics[height = 0.5in]{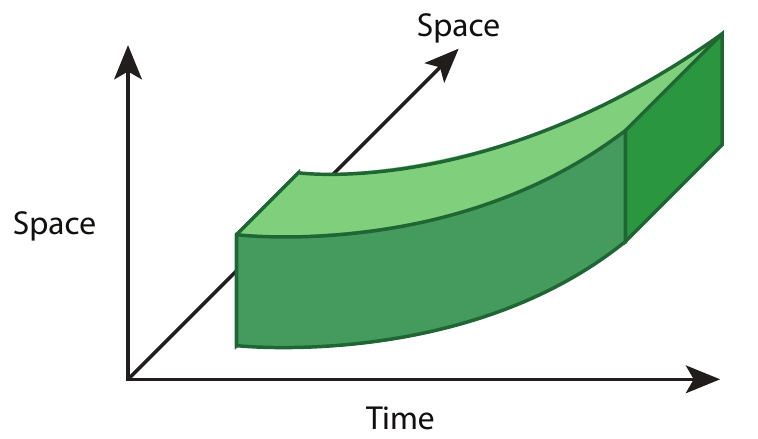}}\hfill
\subcaptionbox*{$\predTh{cyclic}(o)$}{\includegraphics[height = 0.5in]{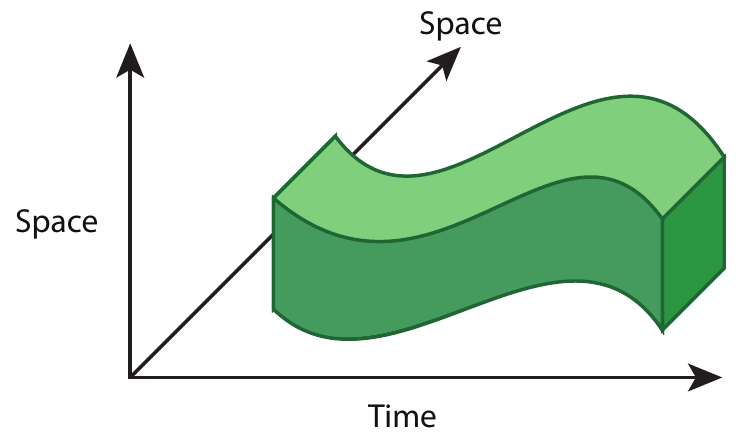}}\hfill
\subcaptionbox*{$\predTh{moving\_into}(o_1,o_2)$}{\includegraphics[height = 0.5in]{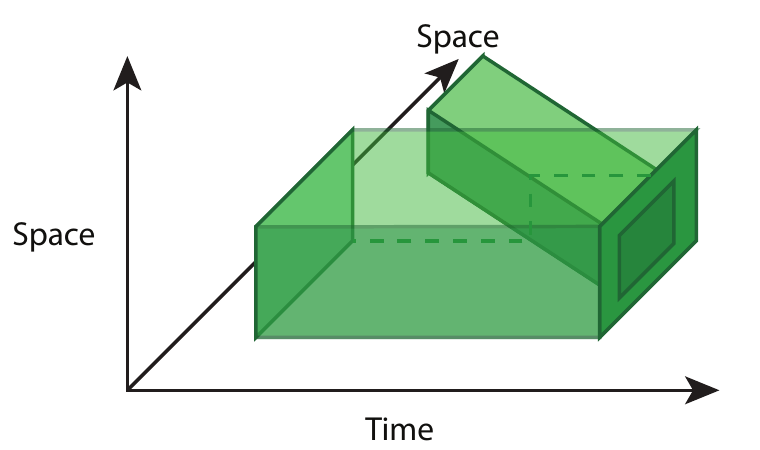}}\hfill
\subcaptionbox*{$\predTh{moving\_out}(o_1,o_2)$}{\includegraphics[height = 0.5in]{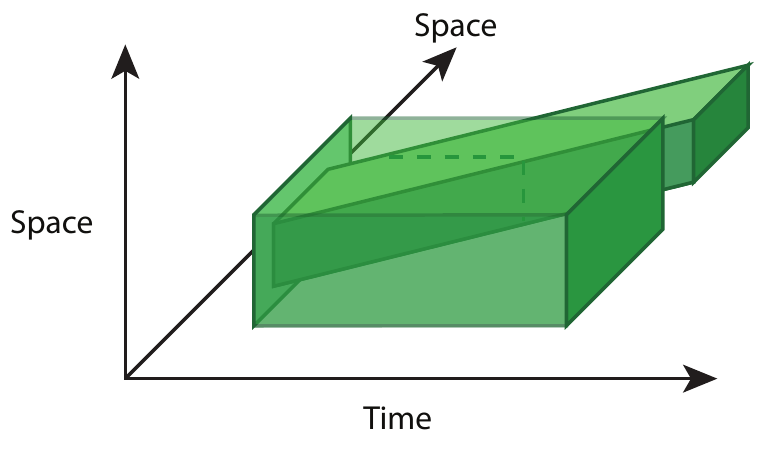}}\hfill
\subcaptionbox*{$\predTh{attached}(o_1,o_2)$}{\includegraphics[height = 0.5in]{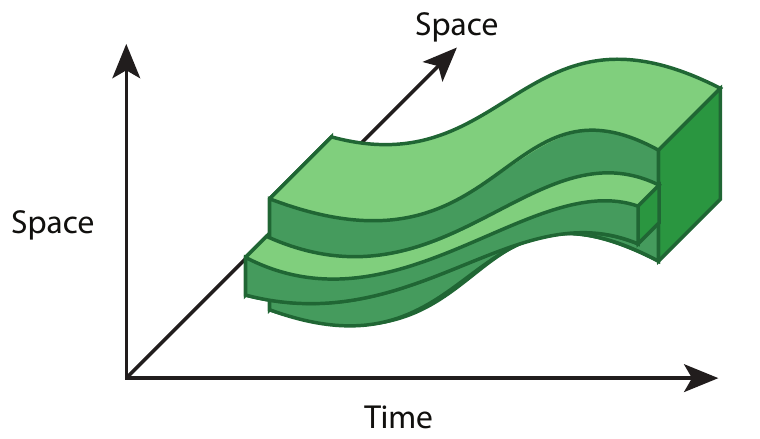}}

\caption{{Commonsense Spatial Reasoning with Spatio-Temporal Entities. Illustrated are:  Space-Time Histories for Spatio-temporal Patterns and Events}}
\label{fig:s-t-entities}
\end{figure*}

\begin{figure}
\centering
\includegraphics[width=\textwidth]{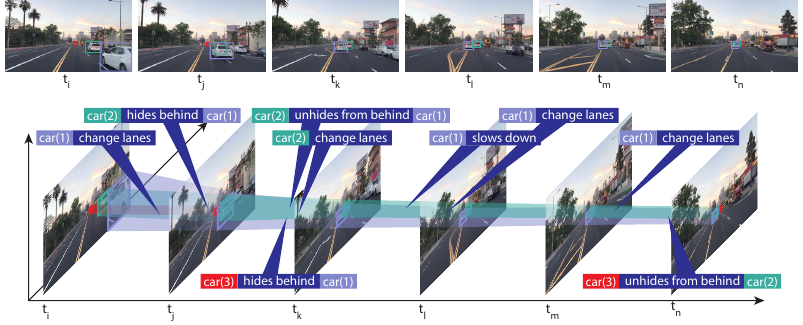}

\caption{Space-Time Histories of Moving Objects: Safety-Criticality Case of a Close Encounter$~~$/$~~${\footnotesize\sffamily Car(1) is {\color{blue!80!black}moving towards} car(2) on the right lane, and {\color{blue!80!black}changes} to the left lane to perform an {\color{blue!80!black}overtaking} action; {\color{blue!80!black}subsequently}, car(2) also {\color{blue!80!black}changes} to left lane to {\color{blue!80!black}overtake} car(3) that {\color{blue!80!black}stopped} and is blocking the right lane. To avoid a {\color{blue!80!black}collision} car(1) performs an {\color{blue!80!black}emergency break} and {\color{blue!80!black}leaves} the left lane to the left, {\color{blue!80!black}entering} the lane for the oncoming traffic. 
}
}
\label{fig:MT-examples-scene-safety}
\end{figure}

\begin{itemize}

\item {\textbf{Domain Objects}} (\objectSort).\quad  The high-level, domain-dependent visual elements in the scene, e.g., road-side stakeholders such as \emph{people}, \emph{cars}, \emph{cyclists}, constitute domain objects. Domain objects are denoted by \objectSort\ = $\{o_1, ... , o_n\}$; elements in \objectSortNormal\ are geometrically interpreted as \emph{spatial entities}.

\item \textbf{Spatial Entities} (\spatialPrimitives).\quad Spatial entities correspond to abstractions of domain objects by way of \emph{points}, \emph{line-segments} or (axis-aligned) \emph{rectangles} based on their spatial properties (and a particular reasoning task at hand). Spatial entities are denoted by \spatialPrimitives\ = $\{\varepsilon_{1}, ..., \varepsilon_{n}\}$.

\item \textbf{Time} (\timePoints).\quad The temporal dimension is represented by \textbf{time points}, denoted as \timePoints\ $= \{t_{1}, ..., t_{n}\}$. 

\item \textbf{Motion Tracks} ({$\mathcal{MT}$}).\quad Motion-tracks represent the spacetime motion trajectories (e.g., Fig. \ref{fig:MT-DrivingDomain}) of abstract spatial entities (\spatialPrimitives) corresponding to domain object (\objectSort) of interest.  {$\mathcal{MT}_{o_i}$} = ($\varepsilon_{t_{s}}, ..., \varepsilon_{t_{e}}$) represents the \textbf{motion track} of a single object $o_i$, where $t_s$ and $t_e $ denote the start and end time of the track and $\varepsilon_{t_s}$ to $\varepsilon_{t_e}$ denotes the spatial entity (\spatialPrimitivesNormal) ---e.g., the \emph{axis-aligned bounding box}---corresponding to the object $o_i$ at time points $t_s$ to $t_e $. Whereas Figures \ref{fig:MT-DrivingDomain} and \ref{fig:MT-examples-scene-safety} presents one example of a space-time trajectory, Fig. \ref{fig:s-t-entities} is a general (but non-exhaustive) set of patterns supported by our reasoning framework. 

\item \textbf{Spatio-Temporal Relationships} (\spatialRelations).\quad The spatial configuration of the scene and changes thereof are characterised based on the {spatio-temporal relationships} (\spatialRelations; Table \ref{tbl:relations}) between abstract representations (\spatialPrimitives) of the domain objects (\objectSort). For the running and demo examples of this paper, positional relations on axis-aligned rectangles based on the {Rectangle Algebra} (RA) \cite{Balbiani1999} suffice; RA uses the relations of Interval Algebra (IA) \cite{Allen1983} {\small $\mathcal{R}_{\mathsf{IA}} \equiv {\{}\mathsf{before},$ $\mathsf{after},$ $\mathsf{during},$ $\mathsf{contains},$ $\mathsf{starts},$ $\mathsf{started\_by},$ $\mathsf{finishes},$ $\mathsf{finished\_by},$ $\mathsf{overlaps},$ $\mathsf{overlapped\_by},$ $\mathsf{meets},$ $\mathsf{met\_by},$  $\mathsf{equal}{\}}$} to relate two objects by the \emph{interval relations} projected along each modelled dimension separately (e.g., horizontal and vertical dimensions). 

\item \textbf{Dynamics$~$/$~$Fluents and Events}.\quad The set of \textbf{fluents} ${\Phi} = \{\phi_1, ... , \phi_n\}$ and \textbf{events} ${\Theta} = \{\theta_1, ... , \theta_n\}$ respectively characterise the dynamic properties of the objects in the scene and high-level abducibles (e.g., Tables \ref{tbl:events} and \ref{tbl:fluents}). For reasoning about dynamics (with {$<${$\Phi$}}, {{$\Theta$}$>$}), we use the epistemic generalisation of the event calculus \cite{eventCalculus1989} as per the formalisation in \cite{Ma2014-ASP-based_Epistemic_Event_Calculus,Miller2013-Epistemic_Event_Calculus}; in particular, for examples of this paper, the Functional Event Calculus (FEC) fragment of \citet{Ma2014-ASP-based_Epistemic_Event_Calculus} suffices.\footnote{Main axioms relevant for this paper pertain to $\predTh{occurs-at}(\theta, t)$ denoting that an event occurred at time $t$ and $\predTh{holds-at}(\phi, v, t)$ denoting that $v$ holds for a fluent $\phi$ at time $t$. It it worth noting that in so far as the approach to reason about changes is concerned, our modular framework is by no means limited to the specific approach being utilised. In principle, any method capable of modelling \emph{dynamic spatial systems} \citep{Bhatt08-SCC-Dynamics} encompassing \emph{space, actions, and change} \cite{Bhatt:RSAC:2012,QSTR-Emerging-2011} is usable; basic considerations guiding choice of an action theory pertain to expressivity, modular elaboration tolerance, and support for basic epistemological aspects such as \emph{frame} and \emph{ramification} \citep{Solving-the-Frame-Shanahan-1997}. For instance, other epistemic settings for abductive inference with ASP too may be utilised \citep{postdictive-asp1,postdictive-asp2}.}


\end{itemize}

\newpage

\textbf{Problem Specification and Hypothesis}.\quad 

\begin{itemize}

\item {\bf Problem Specification $<\mathcal{VO}_t, \mathcal{P}_t, \mathcal{ML}_t >$ }. \quad 
The abduction for each time point  is given by the visual observations ($\mathcal{VO}_t$) consisting of spatial entities \spatialPrimitives{}, i.e., bounding boxes for the detected objects, 
spatial entities \spatialPrimitives{} of object detections; the predicted locations ($\mathcal{P}_t$) for each track at time point $t$ given as spatial entities \spatialPrimitives{}; and the matching likelihood ($\mathcal{ML}_t $), i.e., based on the IoU between detected objects and tracks, providing an estimate of how likely a detection belongs to a track,.

\item {\bf Hypothesis} \quad Abduced hypothesis consist of assignments ($\mathcal{H}^{assign}$) of detections to tracks and high-level events ($\mathcal{H}^{events}$) explaining object motion, e.g., occlusion of an object, caused by the object passing behind an other object.
The online abduction results in abduced visuo-spatial dynamics ($\mathcal{EXP}$) consisting of motion tracks ($\mathcal{MT}$) (generated using the abduced assignments in $\mathcal{H}^{assign}$) and the events ($\mathcal{H}^{events}$) explaining the motion tracks.

\end{itemize}

\section{\uppercase{Visual Sensemaking}:$~$ \uppercase{A General Method Driven by\\Answer Set Programming}}\label{sec:vissenmaking}

Rooted in answer set programming, the developed framework is general, modular, and designed for integration as a reasoning engine within (hybrid) architectures designed for real-time decision-making and control where visual perception is needed as one of the several components. In such large scale AI systems the declarative model of the scene dynamics resulting from the presented framework can be used for semantic question-answering (Q/A), inference etc to support decision-making.

\begin{figure*}[t]
\center
\includegraphics[width = 1.0\textwidth]{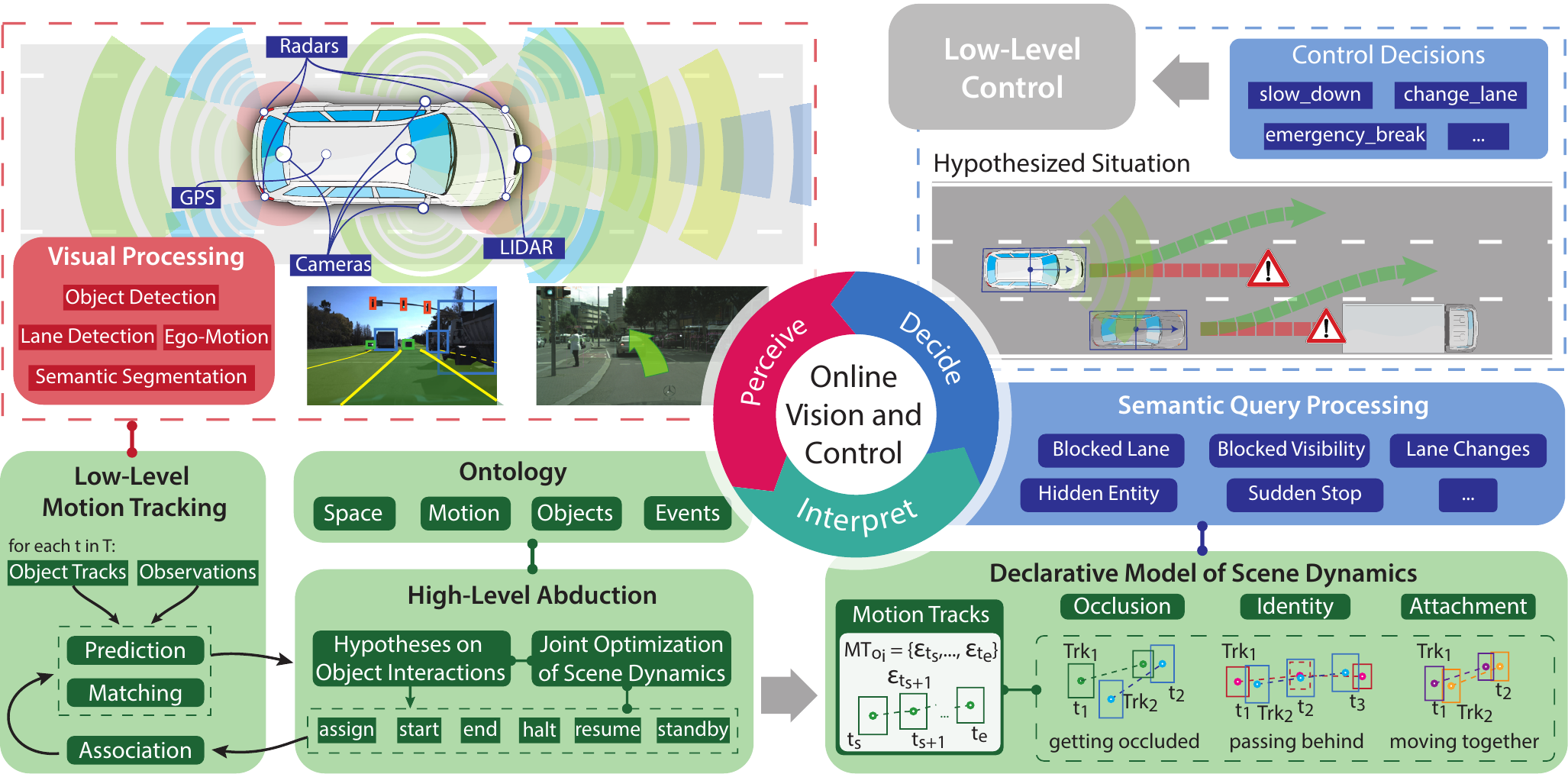}
\caption{{ A General Online Abduction Framework$~$ / $~$Conceptual Overview}}
\label{fig:system_overview}
\end{figure*}

\subsection{{Tracking as Abduction}}\label{sec:tracking-as-abbduction}

Our proposed framework, in essence, jointly solves the problem of assignment of \emph{detections} to \emph{tracks} and explaining overall scene dynamics (e.g. {\small\sffamily appearance}, {\small\sffamily disappearance}) in terms of high-level \emph{events} within an online integrated low-level visual computing and high-level abductive reasoning framework (Fig. \ref{fig:system_overview}).

Scene dynamics are tracked using a \emph{detect and track} approach: we tightly integrate low-level visual computing (for detecting scene elements) with high-level {ASP}-based abduction to solve the assignment of observations to object tracks in an \emph{incremental} manner. For each time point $t$ we generate a \emph{problem specification} consisting of the object tracks and visual observations and use {ASP} to abductively solve the corresponding assignment problem incorporating the ontological structure of the domain / data (abstracted with {$\Sigma$}).

{\bf Steps 1--3} (Alg. \ref{alg:VA} \& Table \ref{tbl:computational_steps}) consist of:\footnote{In the context of Alg. \ref{alg:VA} / Table \ref{tbl:computational_steps}, note that we utilise Clingo v5.3.0 \cite{Gebser2014-Clingo} for the grounding and solving of the answer set program.} \\[2pt]

{\bf 1)} Formulating the ASP problem specification consisting of the visual observations, prediction of motion of each object, and a measure for the likelyhood that a detection is associated with a track. Further the problem specification contains the state of the world, given by the sequence of events ($\mathcal{H}^{events}$) before time point $t$.

\medskip

{\bf 2)} Associating detections to tracks, by jointly abducing matchings between object detections and tracks, together with the high-level events explaining these matches.  

\medskip

{\bf 3)} Finding the hypothesis and corresponding associations best explaining the visual observations using optimization, i.e., maximizing matching likelihood and minimizing event costs. 

\medskip

In the following we describe each step in detail:

\medskip
\medskip

\noindent {{\color{blue!80!black}\textbf{Step 1.}}\quad{\textbf{Formulating the Problem Specification}}}\quad 

The { ASP} problem specification for each time point $t$ is given by the tuple {$<\mathcal{VO}_t, \mathcal{P}_t, \mathcal{ML}_t >$} and the sequence of events ($\mathcal{H}^{events}$) before time point $t$.

\medskip

\noindent $\bullet~~${\textbf{Visual Observations}}\quad Scene elements derived directly from the visual input data are represented as spatial entities $\mathcal{E}$, i.e., $\mathcal{{VO}}_t$ = $\{\varepsilon_{obs_1}, ... , \varepsilon_{obs_n}\}$ is the set of observations at time $t$  (Table \ref{tbl:computational_steps}). For the examples and empirical evaluation in this paper (Sec. \ref{sec:application-sec}) we focus on \emph{Obstacle / Object Detections} -- detecting {\small\sffamily cars, pedestrians, cyclists, traffic lights} etc using {YOLOv3} \cite{Redmon2018_YOLOv3}. Further we generate scene context using \emph{Semantic Segmentation} -- segmenting the {\small\sffamily road, sidewalk, buildings, cars, people, trees}, etc. using {DeepLabv3+} \cite{deeplabv3plus2018}, and \emph{Lane Detection} -- estimating lane markings, to detect {\small\sffamily lanes} on the road, using {SCNN} \cite{Pan2018_scnn}. Type and confidence score for each observation is given by $type_{obs_i}$ and $conf_{obs_i}$.

\medskip

\noindent $\bullet$\quad{\textbf{Movement Prediction}}\quad For each track $trk_i$ changes in \emph{position} and \emph{size} are predicted using kalman filters; this results in an estimate of the spatial entity $\varepsilon$ for the next time-point $t$ of each motion track  $\mathcal{P}_t$ = $\{\varepsilon_{trk_1}, ... , \varepsilon_{trk_n}\}$.

\medskip

\noindent $\bullet$\quad{\textbf{Matching Likelihood}}\quad For each pair of tracks and observations $\varepsilon_{trk_i}$ and $\varepsilon_{obs_j}$, where $\varepsilon_{trk_i}\in\mathcal{P}_t$ and $\varepsilon_{obs_j}\in\mathcal{VO}_t$, we compute the likelihood $\mathcal{ML}_t = \{ml_{trk_1,obs_1}, ..., ml_{trk_i,obs_j}\}$ that $\varepsilon_{obs_j}$ belongs to $\varepsilon_{trk_i}$. The intersection over union (IoU) provides a measure for the amount of overlap between the \emph{spatial entities} $\varepsilon_{obs_j}$ and $\varepsilon_{trk_i}$.

\begin{figure}[t]
\centering
\noindent\begin{minipage}{0.75\columnwidth}

{

\SetNlSty{texttt}{\color{blue}}{\quad}
\IncMargin{1.0em}

\begin{algorithm}[H]
\small

\KwData{
Visual imagery ({$\mathcal{V}$}), and \\background knowledge $\Sigma$ $~\equiv_{def}~$  $\Sigma_{dyn}$ $~\cup~$  $\Sigma_{st}$ 
	}
	  \medskip

\KwResult{
Visual Explanations ({$\mathcal{EXP}$})\hfill {(Refer Table \ref{tbl:computational_steps})}
}

\medskip

$\mathcal{MT} \leftarrow \varnothing$,  $\mathcal{H}^{events} \leftarrow \varnothing$

\medskip

\For{$t \in T$}
{

\medskip

	$\mathcal{VO}_{t} \leftarrow observe(\mathcal{V}_t)$
	
	\medskip
	
	$\mathcal{P}_{t} \leftarrow \varnothing$, $\mathcal{ML}_{t}  \leftarrow \varnothing$
	
	\medskip
	
	\For{$trk \in \mathcal{MT}_{t-1}$}
	{
		$p_{trk} \leftarrow kalman\_predict(trk)$
		
		$\mathcal{P}_{t} \leftarrow \mathcal{P}_{t} \cup p_{trk}$
		
		\medskip
		
		\For{$obs \in \mathcal{VO}_{t}$}
		{
			$ml_{trk, obs} \leftarrow calc\_IoU(p_{trk}, obs)$
			
			$\mathcal{ML}_{t} \leftarrow \mathcal{ML}_{t} \cup ml_{trk, obs}$
		}
		
	}
	
	\medskip

	$abduce(<\mathcal{H}^{assign}_{t},~\mathcal{H}^{events}_{t}>)$ \hfill(Refer Step 2)

	\medskip

	$\mathcal{H}^{events} \leftarrow \mathcal{H}^{events} \cup \mathcal{H}^{events}_{t}$

	\medskip
	
	$\mathcal{MT}_{t} \leftarrow update(\mathcal{MT}_{t-1}, \mathcal{VO}_{t}, \mathcal{H}_{assign})$

}

\Return{$~\mathcal{EXP} ~~ \leftarrow~~ <\mathcal{H}^{events}, \mathcal{MT}>$}

\caption{$~~${\color{blue!80!black}{Online\_Abduction}($\mathcal{V}$, $\Sigma$)}
\label{alg:VA}
}
\end{algorithm}
}

\end{minipage}
\end{figure}


\begin{table}
\centering

\noindent\begin{minipage}{1.0\columnwidth}

{
\centering
\small
\begin{tabular}{|p{\columnwidth}|}
\hline

\textbf{For each} $t \in T$ 

\medskip



%
%


{\color{blue!80!black}\textbf{Step 1.}}\quad\textbf{Formulating the Problem Specification} \quad  $<\mathcal{VO}_t, \mathcal{P}_t, \mathcal{ML}_t >$ \quad  

\begin{tabular}{p{0.45\columnwidth}c}

\begin{minipage}[c]{\linewidth}

\vspace{2pt}
{\bf(1)} detect Visual Observations ($\mathcal{VO}_t$) e.g., \emph{People, Cars, Objects, Roads, Lanes}, 

\vspace{2pt}
{\bf(2)} Predictions ($\mathcal{P}_t$) of next position and size of object tracks using kalman filters, and

\vspace{2pt}
{\bf(3)} calculate Matching Likelihood ($\mathcal{ML}_t $) based on Intersection over Union (IoU) between predictions and detections.

\end{minipage}
&
\quad\includegraphics[valign=m,width = 0.45\textwidth]{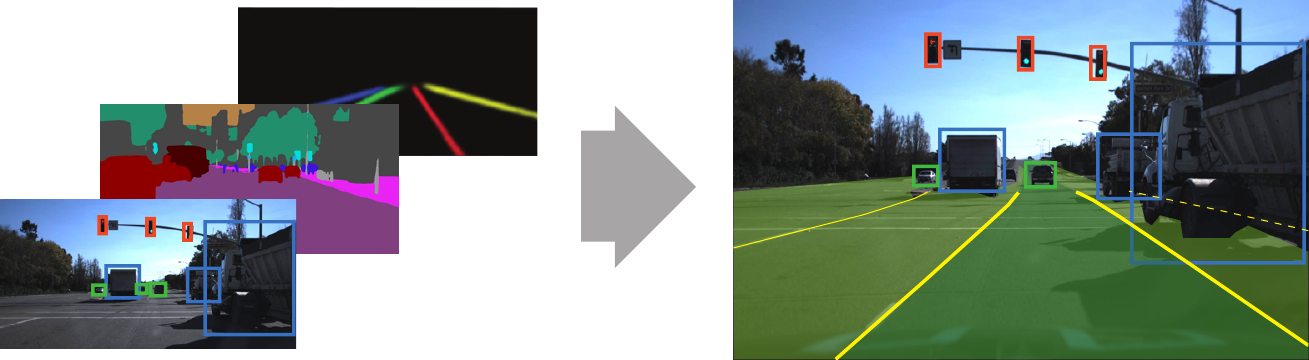}

\end{tabular}

\begin{center}
\vspace{-6pt}
\footnotesize
\begin{minted}[bgcolor=blue!5!white]{prolog}
obs(obs_0,car,99). obs(...). ... box2d(obs_16,1078,86,30,44). ...
trk(trk_0,car). trk(...). ... box2d(trk_0,798,146,113,203). ...
iou(trk_0,obs_0,83921). iou(...). ... iou(trk_23,obs_16,0). ...
\end{minted}
\vspace{-15pt}


%
\end{center}

\medskip

{\color{blue!80!black}\textbf{Step 2.}}\quad\textbf{Abduction based Association} \quad 
generate hypothesis for {\bf(1)}  matching of tracks and observations ($\mathcal{H}^{assign}_{t}$), and {\bf(2)}  and high-level events  ($\mathcal{H}^{events}_{t}$) explaining {\bf(1)}.

\begin{center}
\includegraphics[width = 0.9\textwidth]{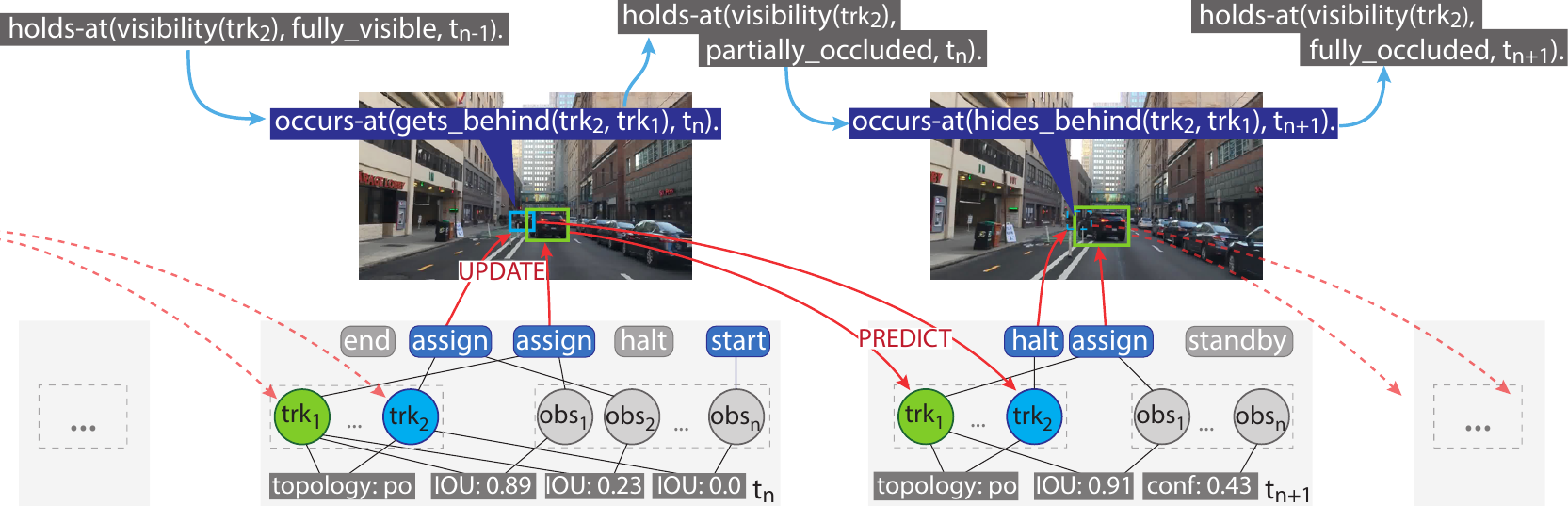}
\end{center}

{\color{blue!80!black}\textbf{Step 3.}}\quad\textbf{Finding the Optimal Hypothesis} \quad Jointly optimize $\mathcal{H}^{assign}_{t}$ and $\mathcal{H}^{events}_{t}$ by maximizing matching likelihood $\mathcal{ML}_t$ and minimizing event costs.
	\\[2pt] \hline\hline
{\color{blue!80!black}\textbf{RESULT.}}\quad\textbf{Visuo-Spatial Scene Semantics} \quad 
Resulting motion tracks ($\mathcal{MT}$) and the corresponding event sequence ($\mathcal{H}^{events}$) explaining the low-level motion:

\medskip

\includegraphics[width = 1.0\textwidth]{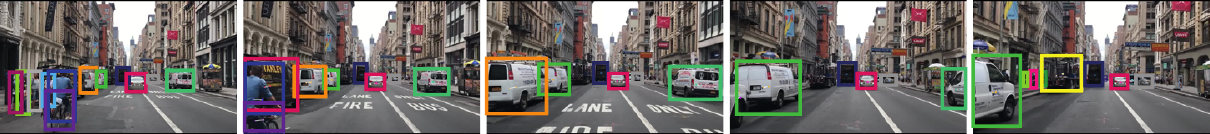}
\begin{center}
\footnotesize
\vspace{-7pt}
\begin{minted}[bgcolor=green!25!white]{prolog}
... occurs_at(missing_detections(trk_10),35) ...  occurs_at(recover(trk_10),36)
... occurs_at(lost(trk_18),41) ... occurs_at(hides_behind(trk_9,trk_10),41)
... occurs_at(...) ... occurs_at(unhides_from_behind(trk_9,trk_10),42) ...
\end{minted}
\vspace{-20pt}
\end{center}

\\\hline
\end{tabular}
\caption{{\footnotesize\sffamily Computational Steps for Online Visual Abduction}}
\label{tbl:computational_steps}
}

\end{minipage}
\end{table}

\medskip

\noindent {{\color{blue!80!black}\textbf{Step 2.}}\quad{\textbf{Abduction based Association }}}\quad 
Following perception as logical abduction most directly in the sense of \citet{Shanahan05}, we define the
task of abducing visual explanations as finding an association ($\mathcal{H}^{assign}_{t}$) of observed scene elements ($\mathcal{VO}_t$) to the motion tracks of objects ($\mathcal{MT}$) given by the predictions $\mathcal{P}_t$, together with a high-level explanation ($\mathcal{H}^{events}_{t}$), such that  $[\mathcal{H}^{assign}_{t} \wedge \mathcal{H}^{events}_{t}]$  is consistent with the background knowledge and the previously abduced event sequence $\mathcal{H}^{events}$, and entails the perceived scene given by $<\mathcal{VO}_t, \mathcal{P}_t, \mathcal{ML}_t >$:

\medskip

\centerline{$\Sigma ~ \wedge ~ \mathcal{H}^{events} ~ \wedge ~ [\mathcal{H}^{assign}_{t} ~ \wedge ~ \mathcal{H}^{events}_{t}] ~~\models~~ \mathcal{VO}_{t} ~ \wedge ~ \mathcal{P}_{t} ~ \wedge ~ \mathcal{ML}_t $}

\medskip

where $\mathcal{H}^{assign}_{t}$ consists of the assignment of detections to object tracks, and $\mathcal{H}^{events}_{t}$ consists of the high-level \emph{events} {{$\Theta$}} explaining the assignments.

\medskip

\noindent $\bullet~~${\textbf{Associating Objects and Observations}}\quad Finding the best match between observations ($\mathcal{VO}_t$) and object tracks ($\mathcal{P}_t$) is done by generating all possible assignments and then maximising a matching likelihood $ml_{trk_i,obs_j}$ between pairs of spatial entities for matched observations  $\varepsilon_{obs_j}$ and predicted track region $\varepsilon_{trk_i}$ (See Step 3). Towards this we use \emph{choice rules} \cite{Gebser2014-Clingo} (i.e., one of the heads of the rule has to be in the stable model) for $\varepsilon_{obs_j}$ and $\varepsilon_{trk_i}$, 
generating all possible assignments in terms of assignment actions:  \emph{assign, start, end, halt, resume, ignore\_det, ignore\_trk}.

%
%


\begin{table}
\renewcommand{\arraystretch}{1.5}
\begin{center}
\footnotesize
\begin{tabular}{>{\columncolor[gray]{0.92}}l p{2.4in}}
\hlinewd{1pt}
{\sffamily\color{blue!70!black}\textbf{EVENTS}} & \textbf{Description} \\\hline

$\predThF{enters\_fov(Trk)}$ & Track $\mathsf{Trk}$ enters the field of view.\\[2pt]

$\predThF{leaves\_fov(Trk)}$ & Track  $\mathsf{Trk}$ leaves the field of view.\\[2pt]

$\predThF{hides\_behind(Trk_1, Trk_2)}$ & Track $\mathsf{Trk_1}$ hides behind track $\mathsf{Trk_2}$.\\[2pt]

$\predThF{unhides\_from\_behind(Trk_1, Trk_2)}$ & Track $\mathsf{Trk_1}$ unhides from behind track $\mathsf{Trk_2}$.\\[2pt]

$\predThF{missing\_detections(Trk)}$ & Missing detections for track $\mathsf{Trk}$.\\[2pt]

\hlinewd{1pt}
\end{tabular}

\caption{{{\bf ABDUCIBLES}; Events Relevant to Explaining (Dis)Appearance}}
\label{tbl:events}
\end{center}
\end{table}%

\begin{table}
\renewcommand{\arraystretch}{1.5}
\begin{center}
\footnotesize
\begin{tabular}{>{\columncolor[gray]{0.92}}l p{1.05in} p{1.7in}}
\hlinewd{1pt}
{\sffamily\color{blue!70!black}\textbf{FLUENTS}} & \textbf{Values} & \textbf{Description} \\\hline

$\predThF{in\_fov(Trk)}$ & \{true;false\} & Track  $\mathsf{Trk}$ is in the field of view.\\

$\predThF{hidden\_by(Trk1, Trk2)}$ & \{true;false\} & Track  $\mathsf{Trk1}$ is hidden by $\mathsf{Trk2}$.\\

$\predThF{visibility(Trk)}$ & \{fully\_visible; \newline partially\_occluded; fully\_occluded\} & Visibility state of track  $\mathsf{Trk}$.\\

$\predThF{clipped(Trk)}$ & \{true;false\} & Track  $\mathsf{Trk}$ is interrupted, e.g., missing detection(s).\\

\hlinewd{1pt}

\end{tabular}
\caption{{{\bf ABDUCIBLES}; Fluents Relevant to Explaining (Dis)Appearance}}
\label{tbl:fluents}
\end{center}
\end{table}%

\footnotesize
\begin{minted}[bgcolor=blue!5!white]{prolog}
1{ 
    assign(Trk, Det): det(Det, _, _);
    end(Trk);  
    ignore_trk(Trk);  
    halt(Trk);
    resume(Trk, Det): det(Det, _, _)  
}1  
:-  trk(Trk, _).
\end{minted}
\vspace{-20pt}
\begin{minted}[bgcolor=blue!5!white]{prolog}
1{ 
    assign(Trk, Det): trk(Trk, _);
    start(Det);  
    ignore_det(Det);
    resume(Trk, Det): trk(Trk, _)
}1  
:-  det(Det, _, _).
\end{minted}

\normalsize

For each assignment action we define \emph{integrity constraints}\footnote{\sffamily Integrity constraints restrict the set of answers by eliminating stable models where the body is satisfied.} that restrict the set of answers generated by the choice rules, e.g., the following constraints are applied to assigning an observation $\varepsilon_{obs_j}$ to a track $trk_i$, applying thresholds on the $IoU_{trk_i,obs_j}$  and the confidence of the observation $conf_{obs_j}$, further we define that the type of the observation has to match the type of the track it is assigned to (e.g., also see Fig. \ref{fig:ijcai-overview-technical}):

%
%
%

\begin{figure}[t]
    \centering
    \includegraphics[width=\textwidth]{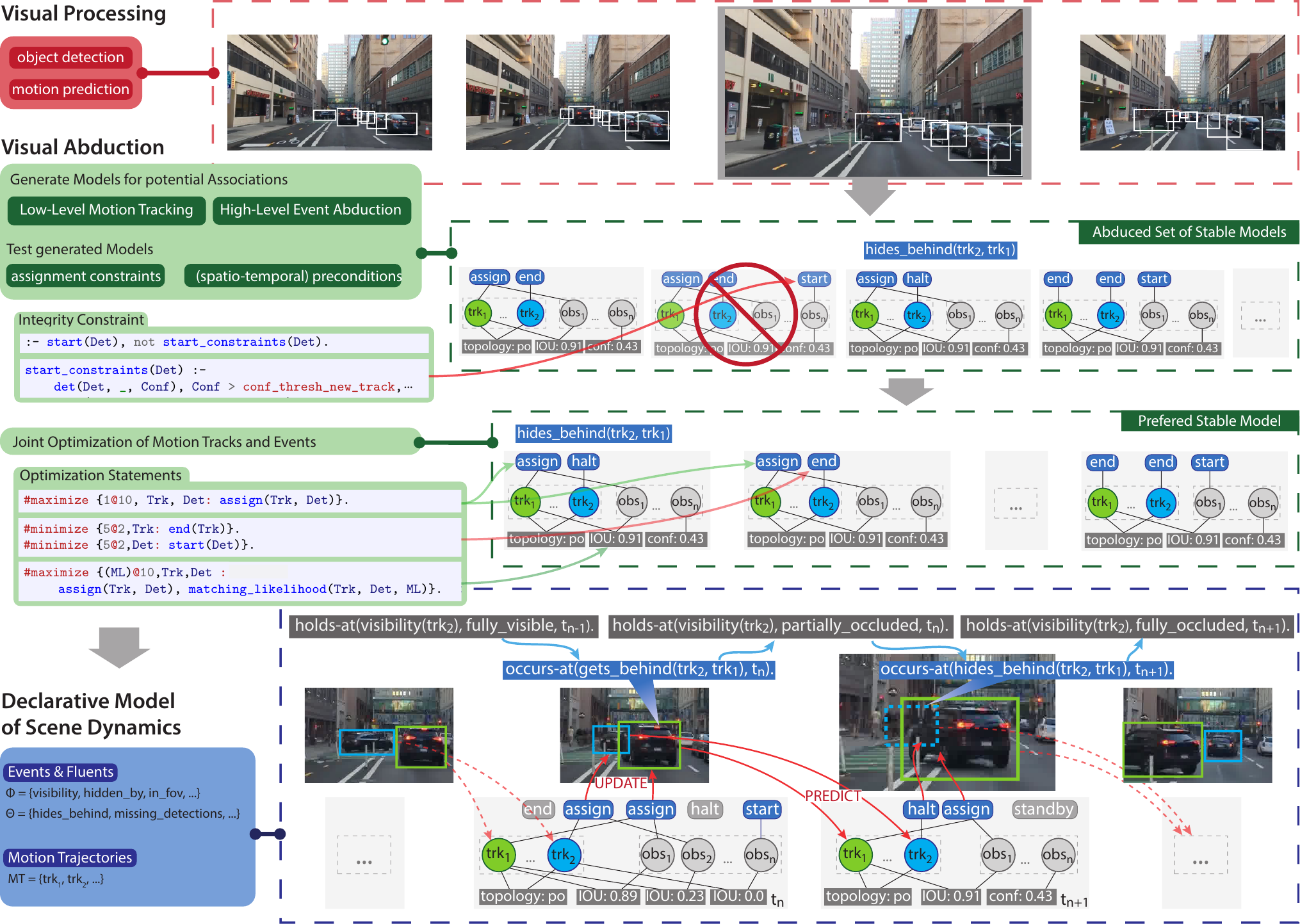}
    \caption{{\sffamily\footnotesize Commonsense Visual Explainability in Active Vision \& Control (for Autonomous Driving); The Case of Hidden Entities.}}
    \label{fig:ijcai-overview-technical}
\end{figure}

\footnotesize
\begin{minted}[bgcolor=blue!5!white]{prolog}
:- assign(Trk, Det), not assignment_constraints(Trk, Det).
\end{minted}
\vspace{-20pt}
\begin{minted}[bgcolor=blue!5!white]{prolog}
assignment_constraints(Trk, Det) :-
    trk(Trk, Trk_Type),  trk_state(Trk, active), 
    det(Det, Det_Type, Conf), Conf > conf_thresh_assign,
    match_type(Trk_Type, Det_Type),
    iou(Trk, Det, IOU), IOU > iou_thresh.
\end{minted}
\normalsize

\noindent $\bullet~~${\textbf{Abducible High-Level Events}}\quad  
For the length of this paper, we restrict to high-level visuo-spatial abducibles pertaining to \emph{object persistence} and \emph{visibility} (Table \ref{tbl:events}): {\bf(1)}. \emph{Occlusion}: Objects can disappear or reappear as result of occlusion with other objects; 
{\bf(2)}.~\emph{Noise and Missing Observation}: (Missing-)\-observations can be the result of faulty detections.

\medskip

Lets take the case of \emph{occlusion}:  functional fluent {\small\sffamily visibility} could be denoted $fully\_visible$,\\$partially\_occluded$ or $fully\_occluded$:

%
%
%
%
%

\footnotesize
\begin{minted}[bgcolor=blue!5!white]{prolog}
fluent(visibility(Trk)) :- trk(Trk, _).
\end{minted}
\vspace{-20pt}
\begin{minted}[bgcolor=blue!5!white]{prolog}
possVal(visibility(Trk), fully_visible) :- trk(Trk, _).
possVal(visibility(Trk), partially_visible) :- trk(Trk,_).
possVal(visibility(Trk), not_visible) :- trk(Trk, _).
\end{minted}
\normalsize

We define the event $hides\_behind/2$, stating that an object hides behind another object by defining the conditions that have to hold for the event to possibly occur, and the effects the occurrence of the event has on the properties of the objects, i.e., the value of the visibility fluent changes to $fully\_occluded$.

%
%
%
%
%
%
%
%

\footnotesize
\begin{minted}[bgcolor=blue!5!white]{prolog}
event(hides_behind(Trk1,Trk2)) :- trk(Trk1,_),trk(Trk2,_).
\end{minted}
\vspace{-20pt}
\begin{minted}[bgcolor=blue!5!white]{prolog}
causesValue(hides_behind(Trk1, Trk2), visibility(Trk1), not_visible, T) :-
    trk(Trk1,_), trk(Trk2,_), time(T).
\end{minted}
\vspace{-20pt}
\begin{minted}[bgcolor=blue!5!white]{prolog}
:- occurs_at(hides_behind(Trk1, Trk2), curr_time), not poss(hides_behind(Trk1, Trk2)).

poss(hides_behind(Trk1, Trk2)) :-
    trk(Trk1, _), trk(Trk2, _),
    position(overlapping_top, Trk1, Trk2),
    not holds_at(visibility(Trk1), not_visible, curr_time),
    not holds_at(visibility(Trk2), not_visible, curr_time).
\end{minted}
\normalsize

For abducing the occurrence of an event we use choice rules that connect the event with assignment actions, e.g., a track getting halted may be explained by the event that the track hides behind another track.

%
%
%
%
%

\footnotesize
\begin{minted}[bgcolor=blue!5!white]{prolog}
1{  
    occurs_at(hides_behind(Trk, Trk2), curr_time):  trk(Trk2,_); 
    ... 
}1  
:-   halt(Trk).
\end{minted}
\normalsize

\noindent {{\color{blue!80!black}\textbf{Step 3.}}\quad{\textbf{Finding the Optimal Hypothesis}}}\quad To ensure an \emph{optimal assignment}, we use { ASP} based optimization to maximize the matching likelihood between matched pairs of tracks and detections. Towards this, we first define the matching likelihood based on the Intersection over Union (IoU) between the observations and the predicted boxes for each track as described in \cite{Bewley2016_sort}:

%
%
%
%
%

\footnotesize
\begin{minted}[bgcolor=blue!5!white]{prolog}
matching_likelihood(Trk, Det, IOU) :-
    det(Det, _, _), trk(Trk, _), iou(Trk, Det, IOU).
\end{minted}
\normalsize

We then maximize the matching likelihood for all assignments, using the build in \emph{maximize} statement:

\footnotesize
\begin{minted}[bgcolor=blue!5!white]{prolog}

A#maximize {(ML)@10,Trk,Det : assign(Trk, Det), matching_likelihood(Trk, Det, ML)}.
\end{minted}
\normalsize

To find the best set of hypotheses with respect to the observations, we \emph{minimize} the occurrence of certain events and association actions, e.g., the following optimization statements minimize starting and ending tracks; the resulting assignment is then used to update the motion tracks accordingly.

%
%
%
%
%
%

\footnotesize
\begin{minted}[bgcolor=blue!5!white]{prolog}
A#minimize {5@2,Trk: end(Trk)}. 
A#minimize {5@2,Det: start(Det)}.
\end{minted}
\normalsize

It is important here to note that: {\textbf{(1)}}. by jointly abducing the object dynamics and high-level events we can impose constraints on the assignment of detections to tracks, i.e., an assignment is only possible if we can find an explanation supporting the assignment; and {\textbf{(2)}}. the likelihood that an event occurs guides the assignments of observations to tracks. Instead of independently tracking objects and interpreting the interactions, this yields to event sequences that are consistent with the abduced object tracks, and noise in the observations is reduced (See evaluation in Sec. \ref{sec:application-sec}).

%
%
%
%
%

\begin{figure}[t]
\centering
\begin{subfigure}[c]{0.48\textwidth}
\includegraphics[width=\textwidth]{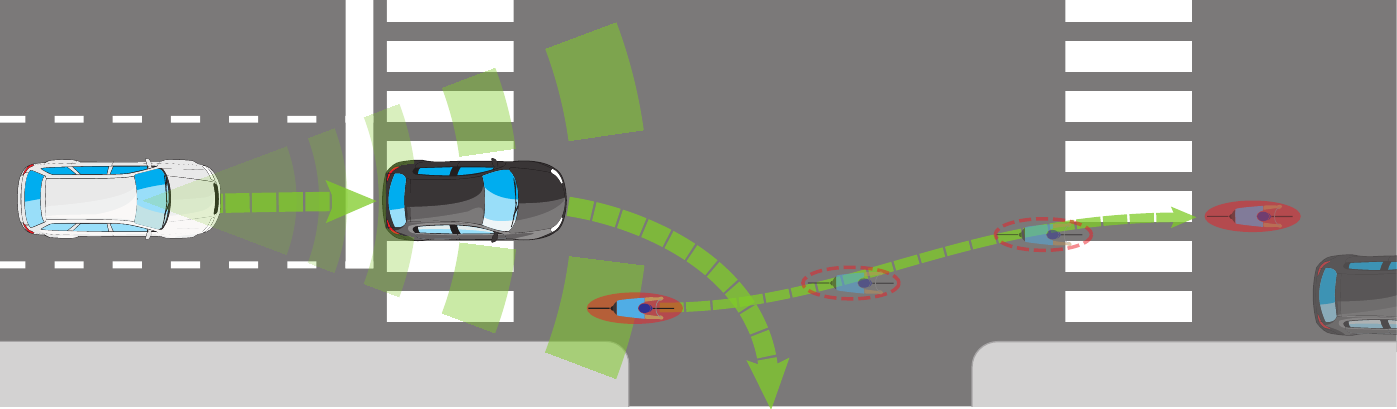}
\subcaption{}
\end{subfigure}
\quad
\begin{subfigure}[c]{0.48\textwidth}
\includegraphics[width=\textwidth]{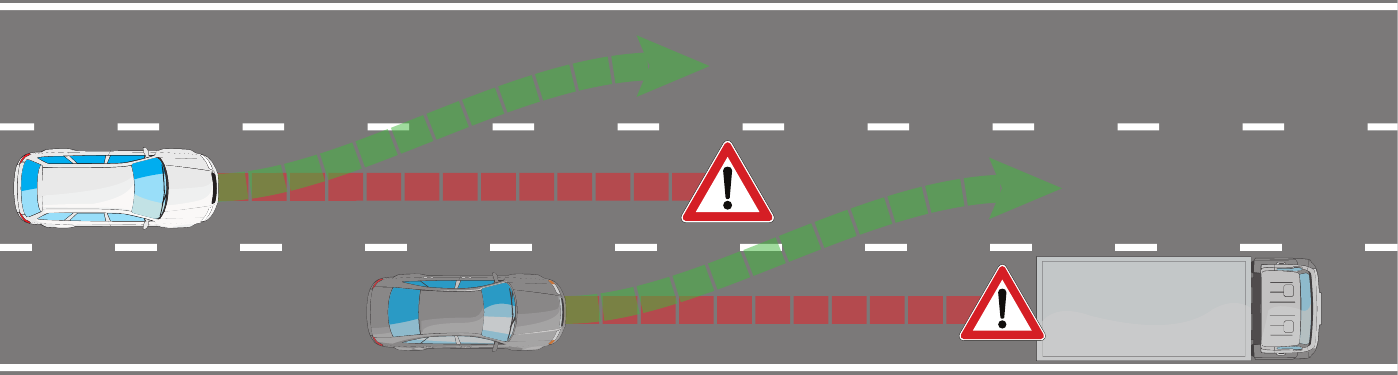}
\subcaption{}
\end{subfigure}

\begin{subfigure}[c]{0.48\textwidth}
\includegraphics[width=\textwidth]{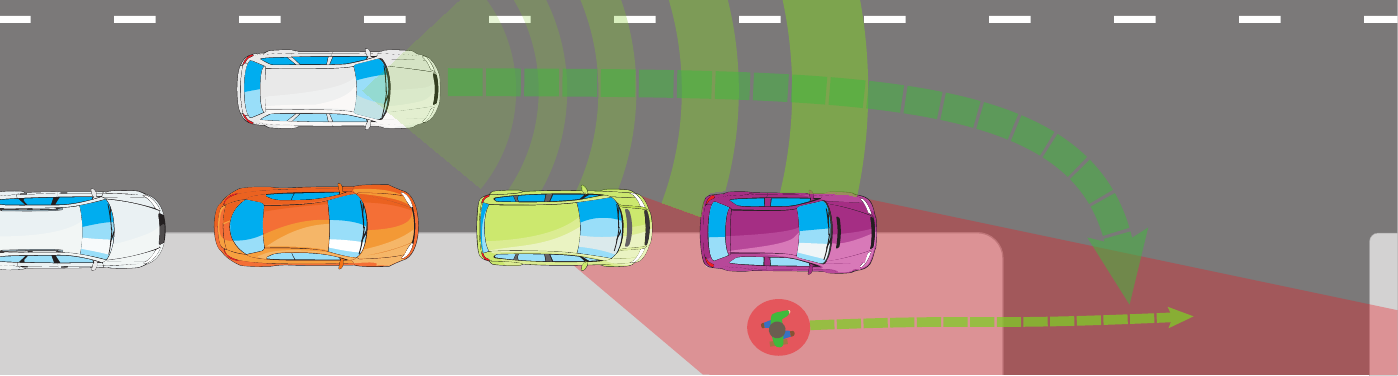}
\subcaption{}
\end{subfigure}
\quad
\begin{subfigure}[c]{0.48\textwidth}
\includegraphics[width=\textwidth]{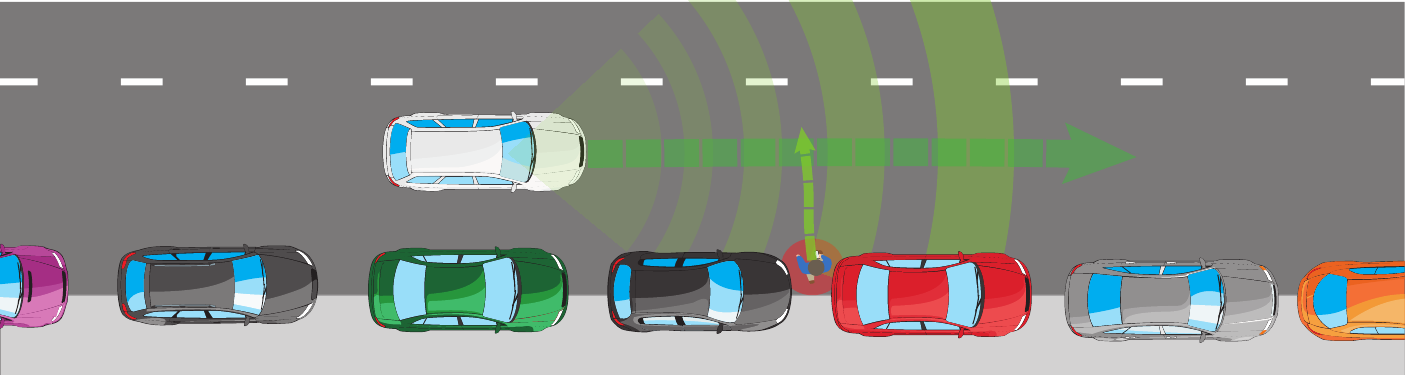}
\subcaption{}
\end{subfigure}


\caption{{\sffamily\footnotesize Safety-Critical Situation (select prototypes):\quad  \textbf{(a)}. momentarily occluded / hidden entities; \textbf{(b)}. overtaking / lane-crossing situation; \textbf{(c)}. blocked visibility}; and  \textbf{(d)} suddenly appearing objects.}
\label{fig:safety-critical-sits}
\end{figure}

{
\begin{table}[t]
\renewcommand{\arraystretch}{1.5}
\begin{center}
\footnotesize
\begin{tabular}{>{\columncolor[gray]{0.92}}l l p{5.2cm}}
\hlinewd{1pt}
{\sffamily\color{blue!70!black}\textbf{SITUATION}} & \textbf{Objects} & \textbf{Description} \\\hline

OVERTAKING & \emph{vehicle, vehicle} & vehicle is overtaking another vehicle in front of the car.\\[2pt]


HIDDEN\_ENTITY & \emph{entity, object} & traffic participant may be hidden by an obstacle, e.g. another car or van.\\[2pt]

REDUCED\_VISIBILITY & \emph{object} & visibility is reduced by some object in front of car.\\[2pt]

SUDDEN\_STOP & \emph{vehical} & vehicle in front of the car is suddenly stopping.\\[2pt]

BLOCKED\_LANE & \emph{lane, object} & lane of the road is blocked by some object.\\[2pt]

EXITING\_PARKED\_VEHICLE & \emph{person, vehicle} & person is exiting a parked vehicle.\\
\hlinewd{1pt}
\end{tabular}
\caption{{\sffamily\small Select Safety-Critical Situations}}
\label{tbl:safety-critical-situations}
\end{center}
\end{table}%
}

%

 \begin{figure}[t]
\centering


\begin{subfigure}[c]{\textwidth}
\centering
\includegraphics[width=0.19\textwidth]{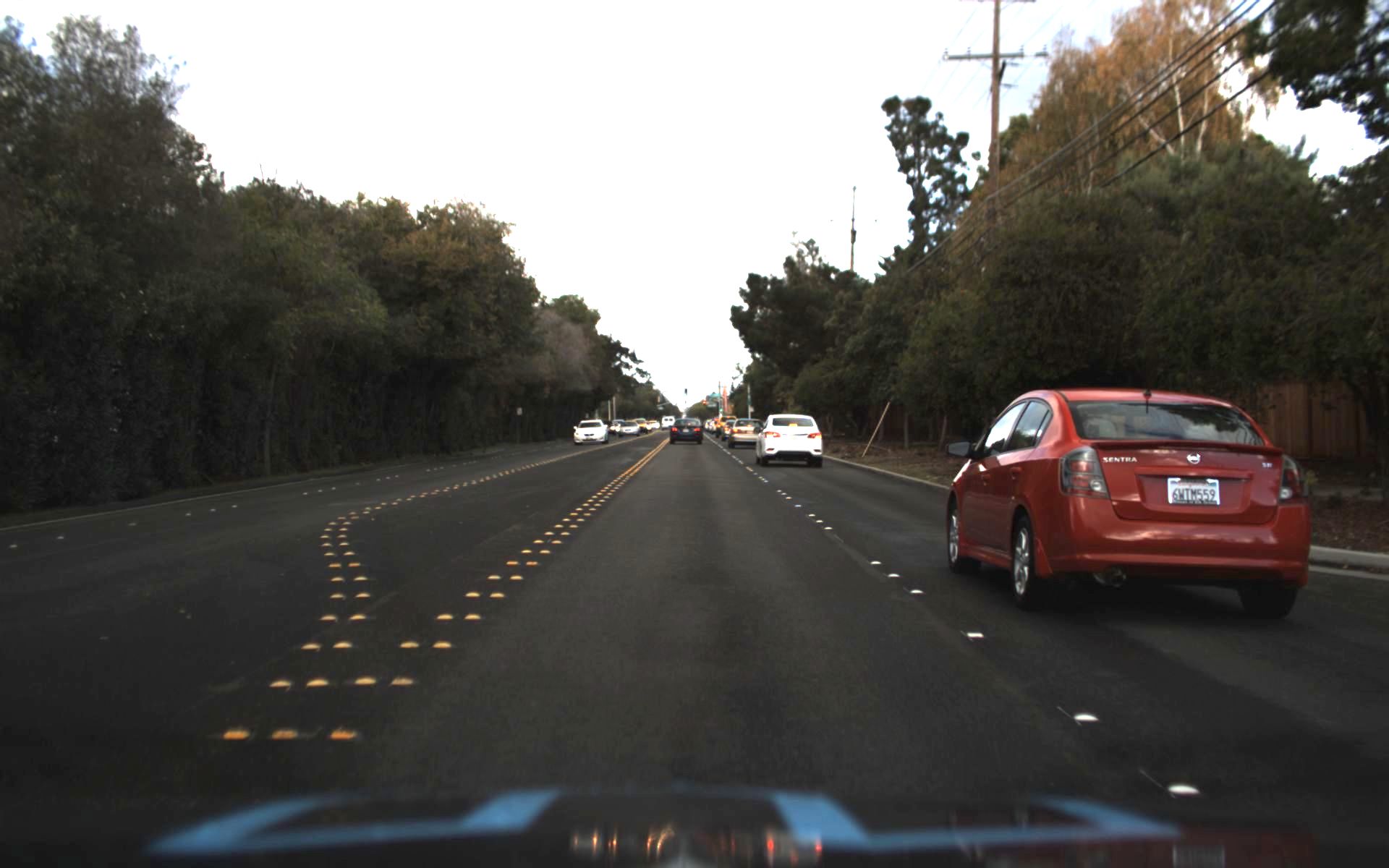}
\includegraphics[width=0.19\textwidth]{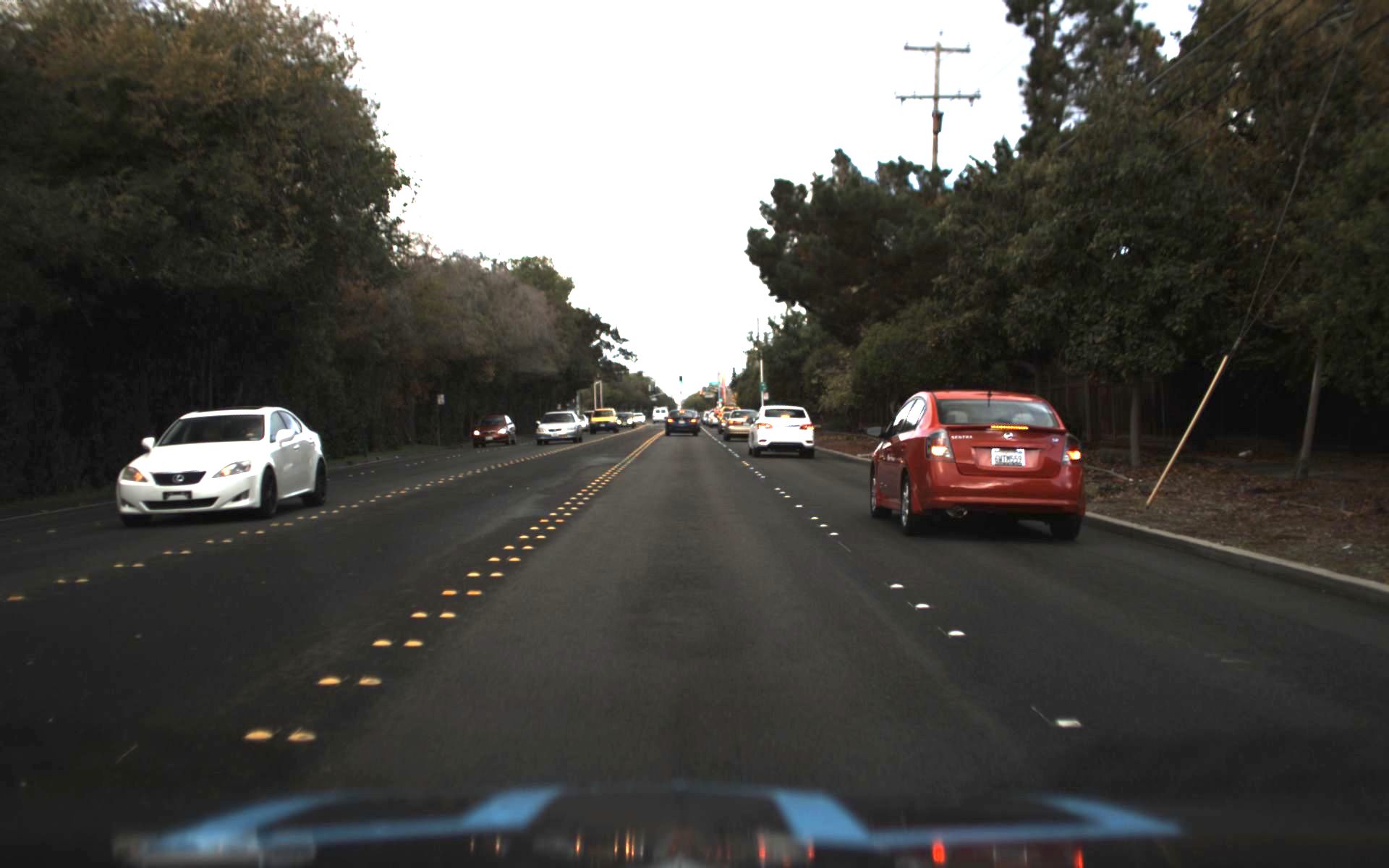}
\includegraphics[width=0.19\textwidth]{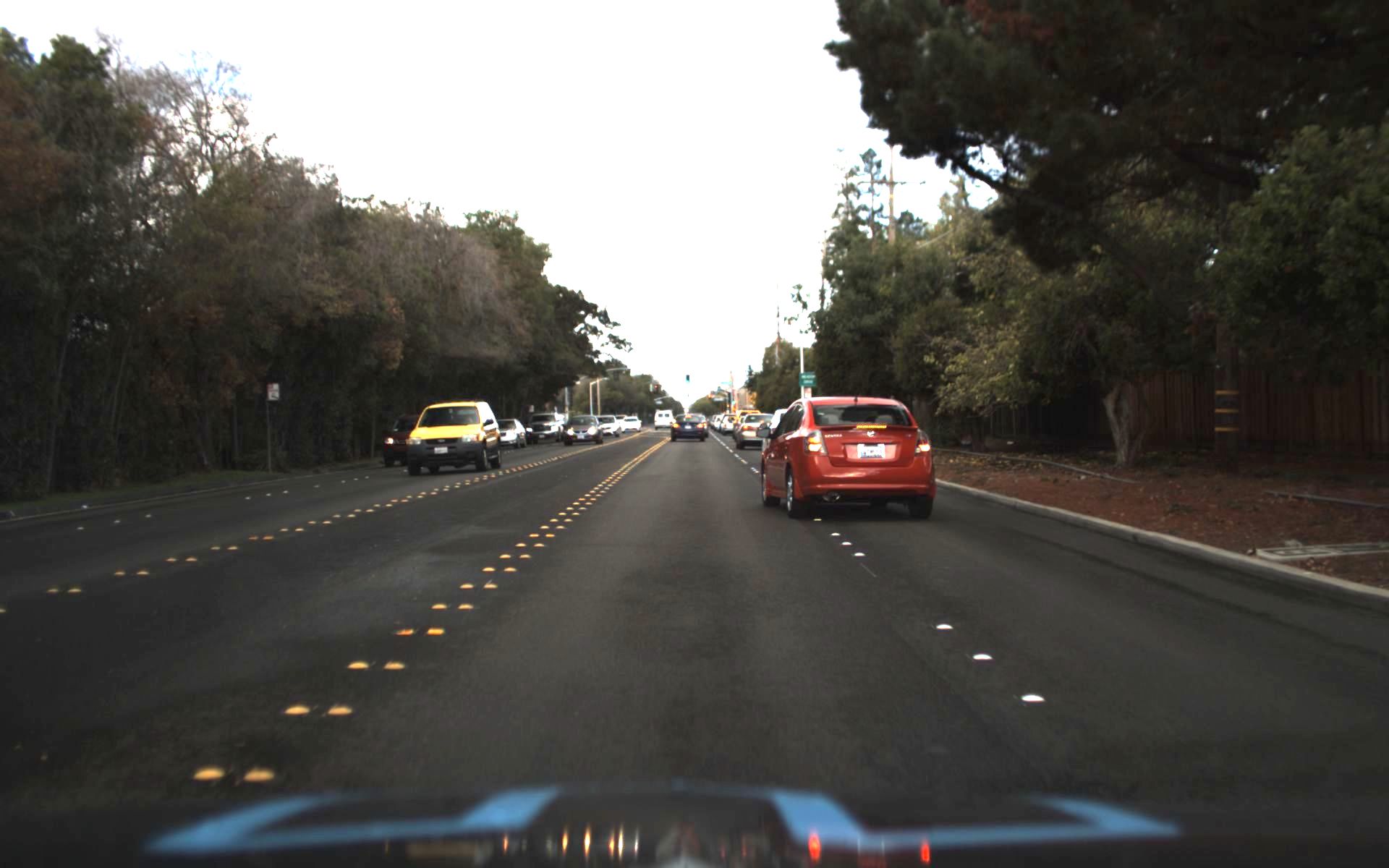}
\includegraphics[width=0.19\textwidth]{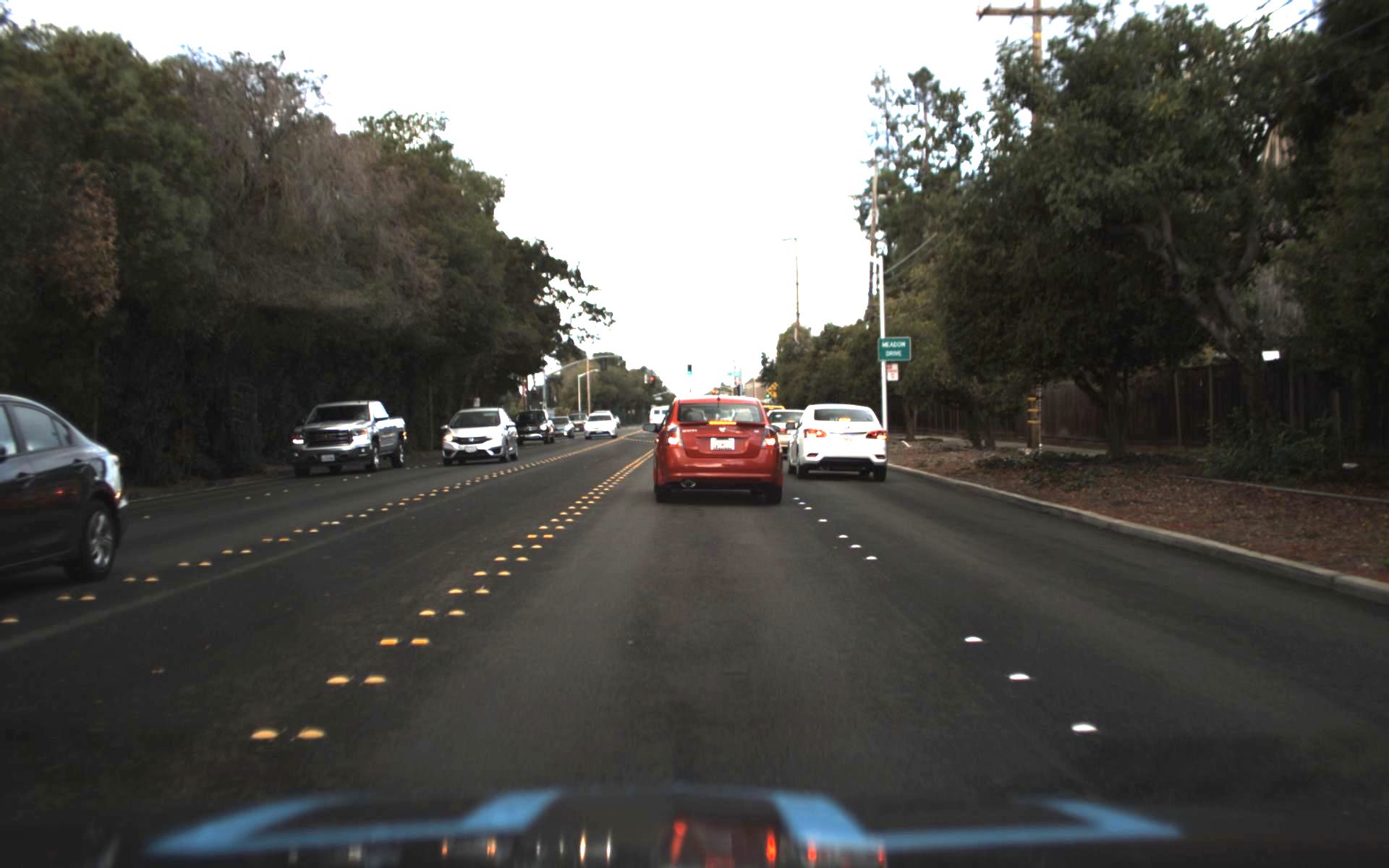}
\includegraphics[width=0.19\textwidth]{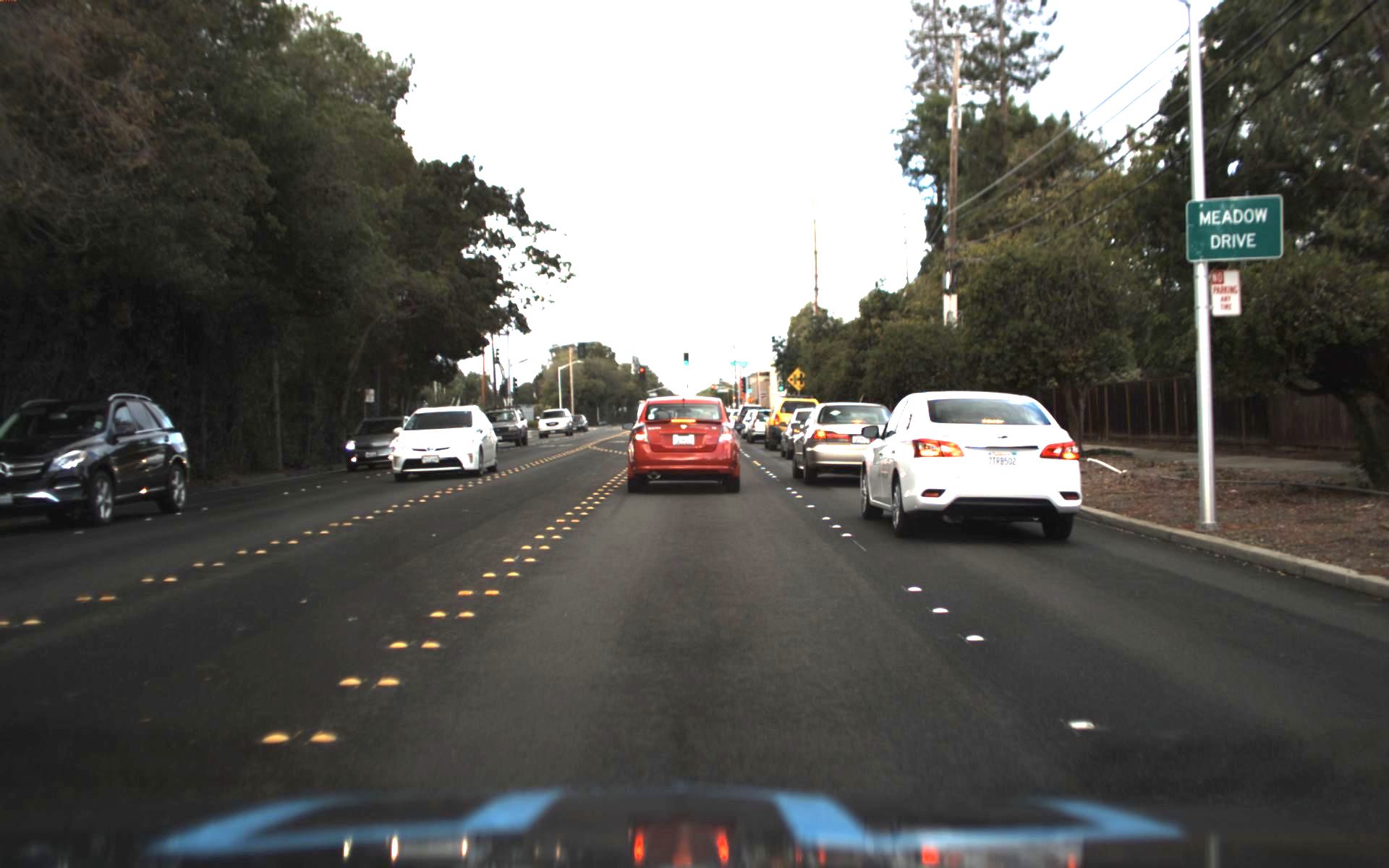}
\subcaption{}
\end{subfigure}

\begin{subfigure}[c]{\textwidth}
\centering
\includegraphics[width=0.19\textwidth]{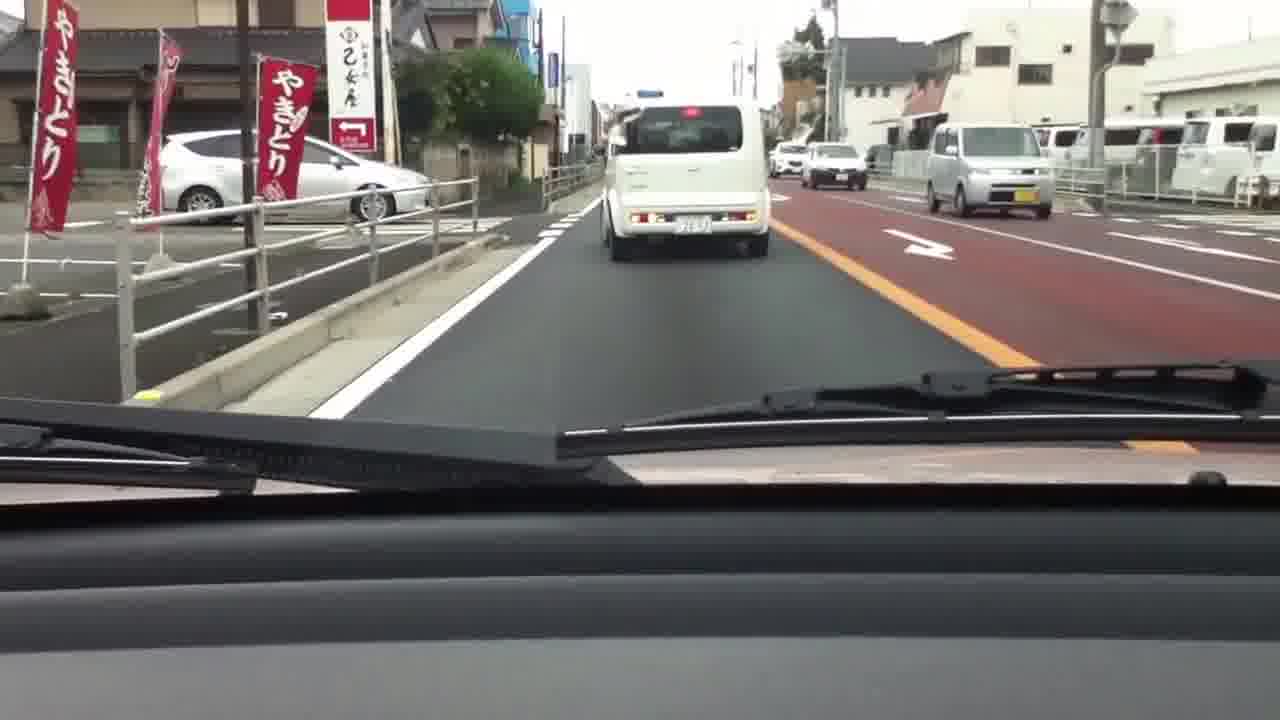}
\includegraphics[width=0.19\textwidth]{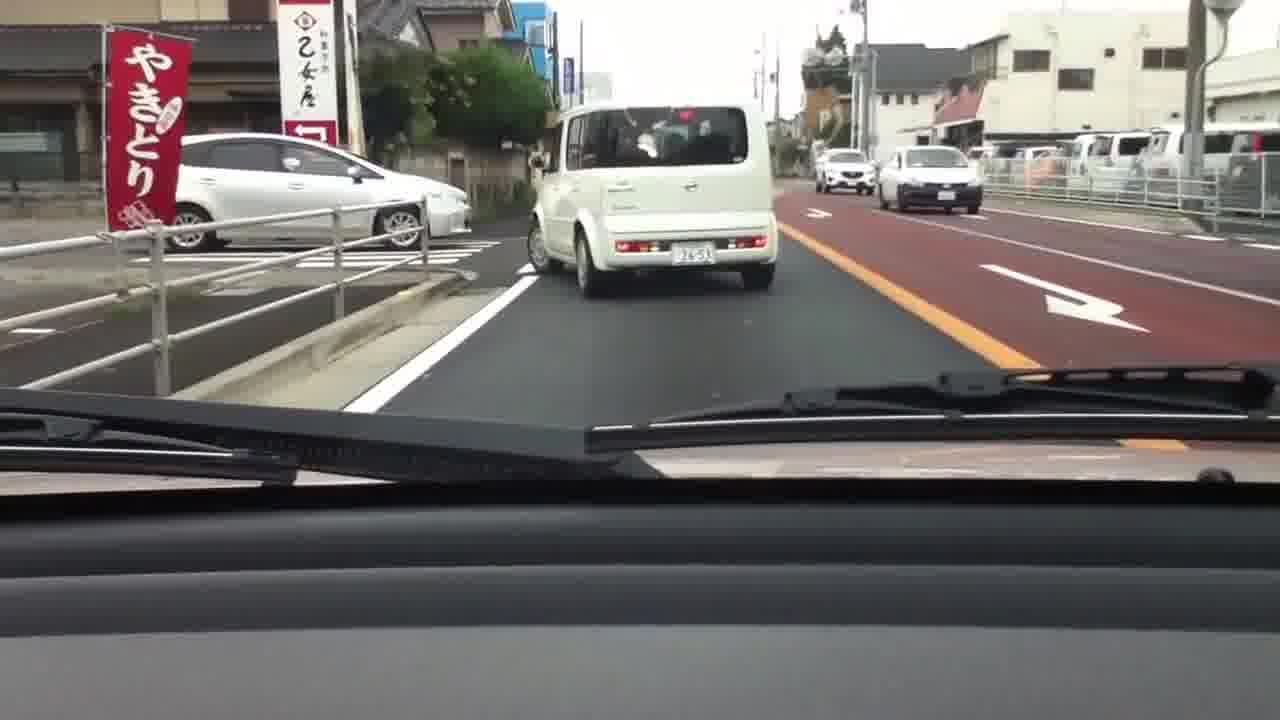}
\includegraphics[width=0.19\textwidth]{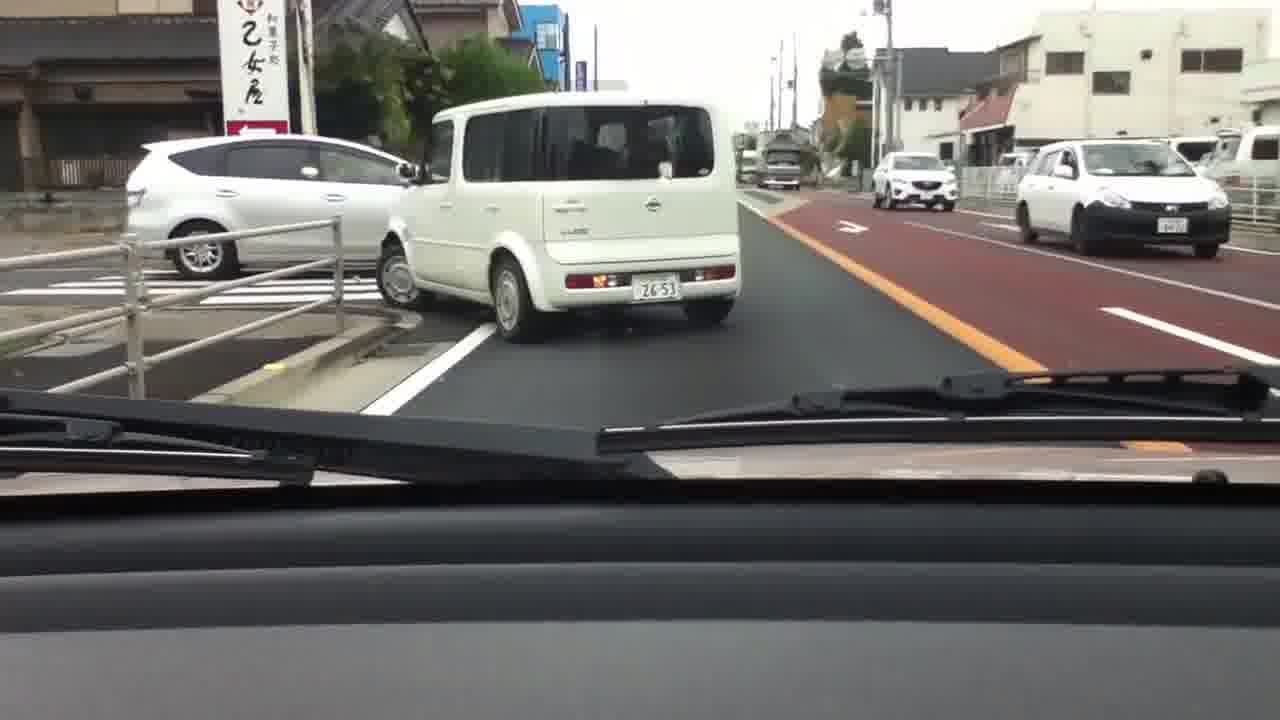}
\includegraphics[width=0.19\textwidth]{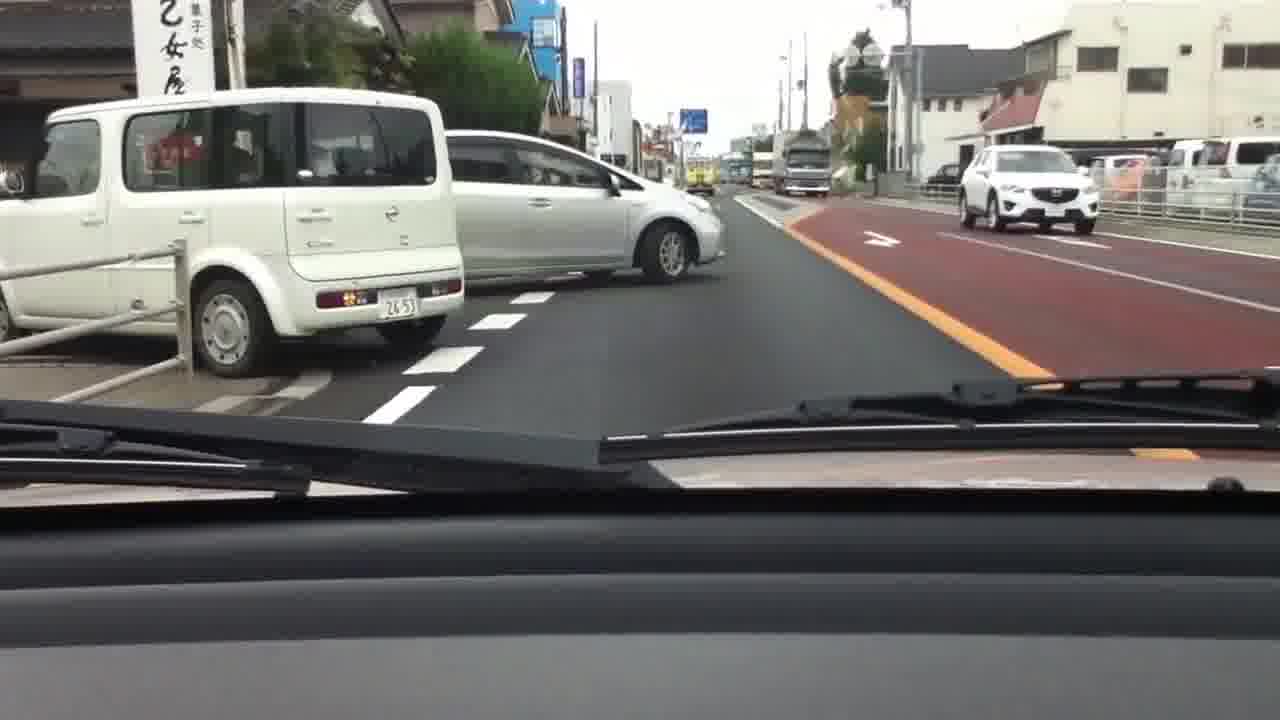}
\includegraphics[width=0.19\textwidth]{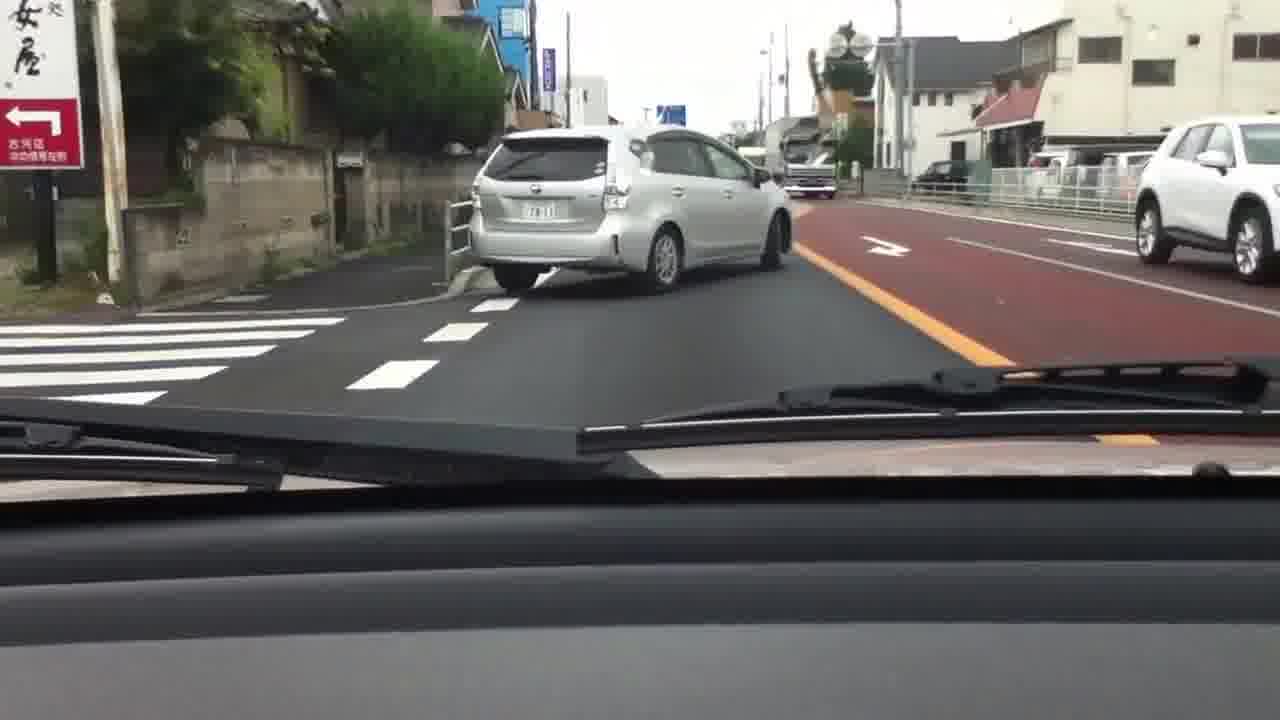}
\subcaption{}
\end{subfigure}

\begin{subfigure}[c]{\textwidth}
\centering
\includegraphics[width=0.19\textwidth]{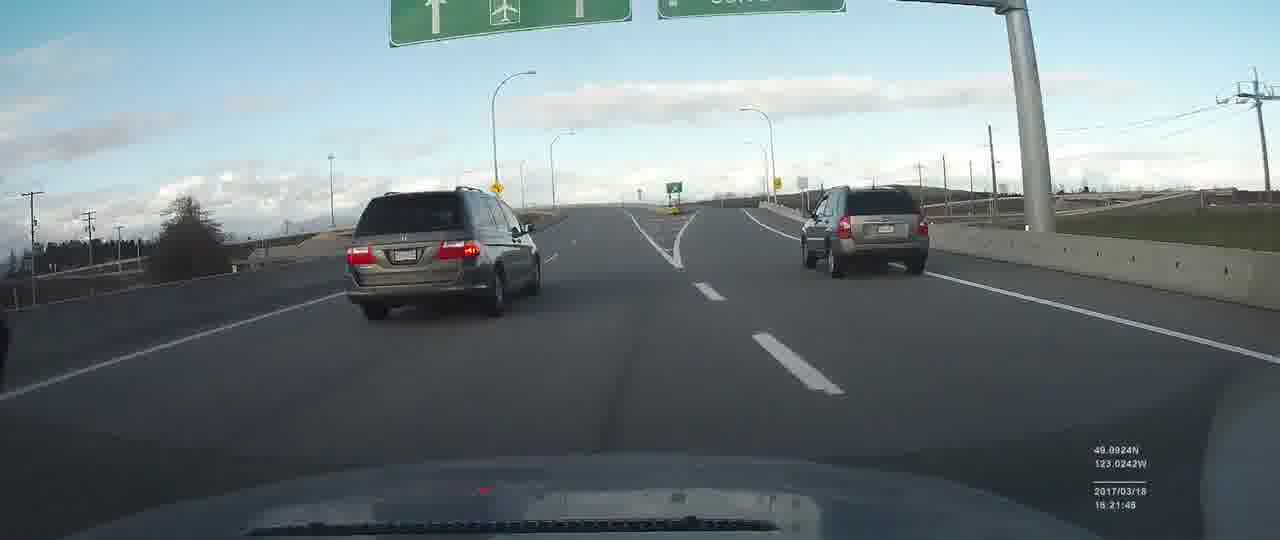}
\includegraphics[width=0.19\textwidth]{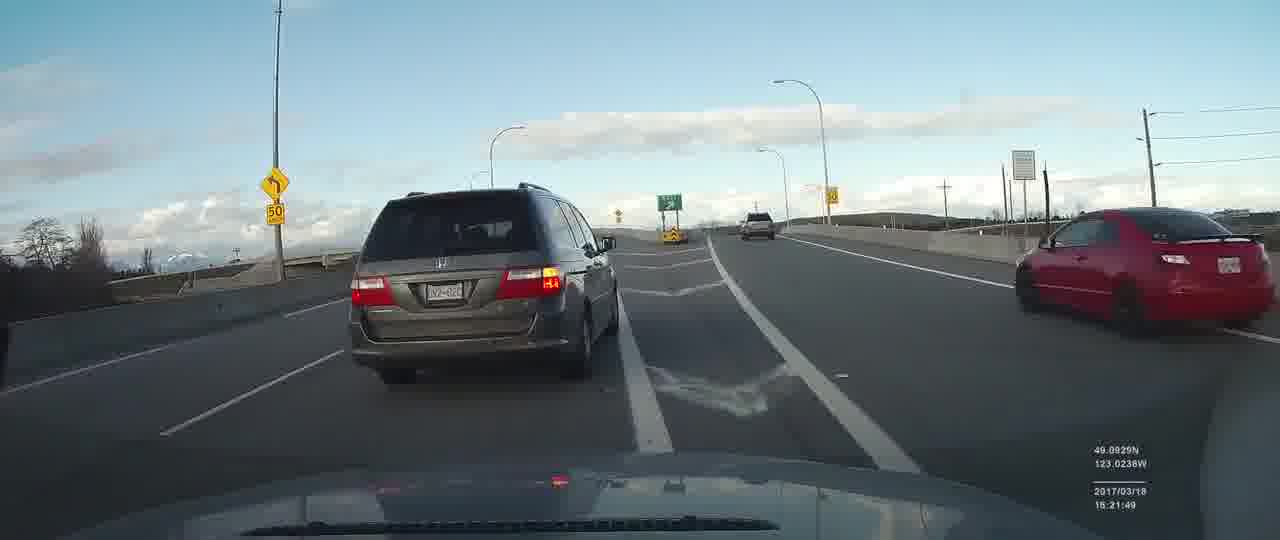}
\includegraphics[width=0.19\textwidth]{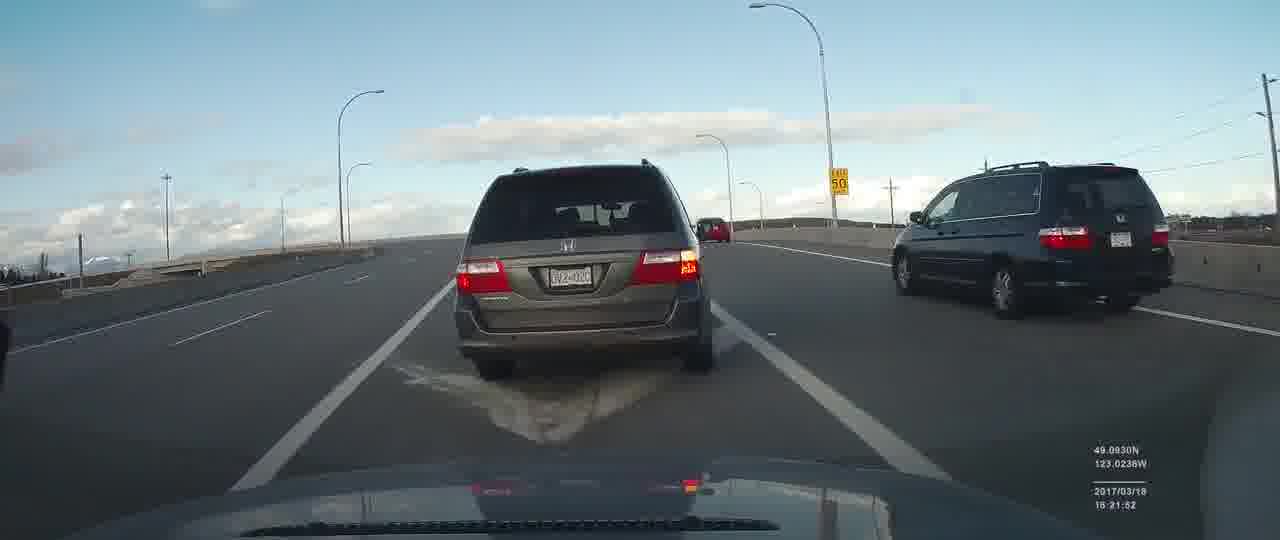}
\includegraphics[width=0.19\textwidth]{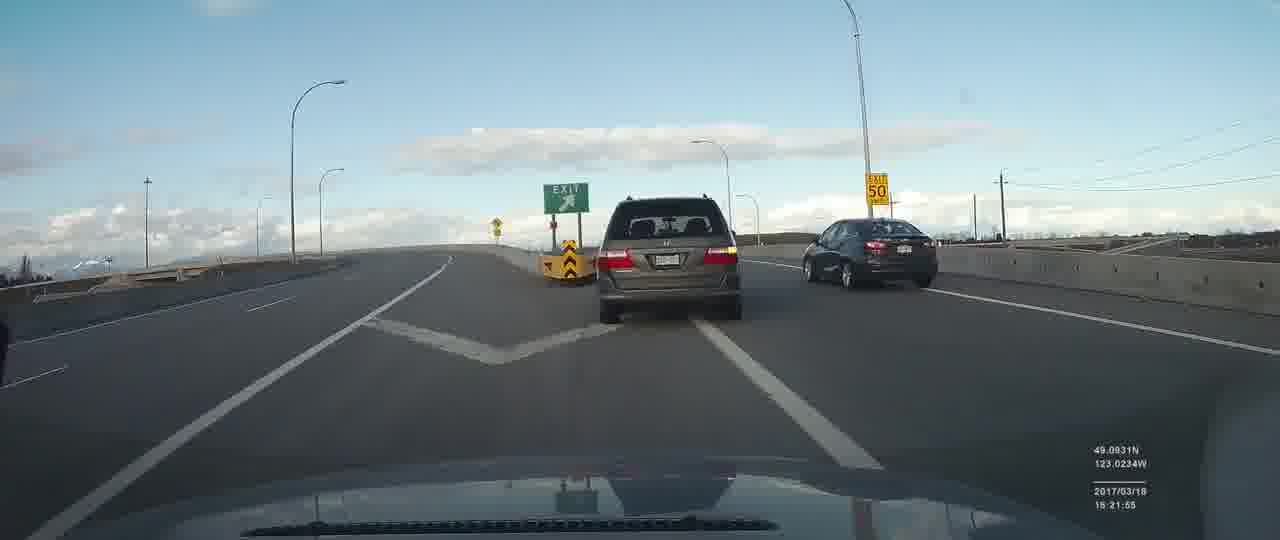}
\includegraphics[width=0.19\textwidth]{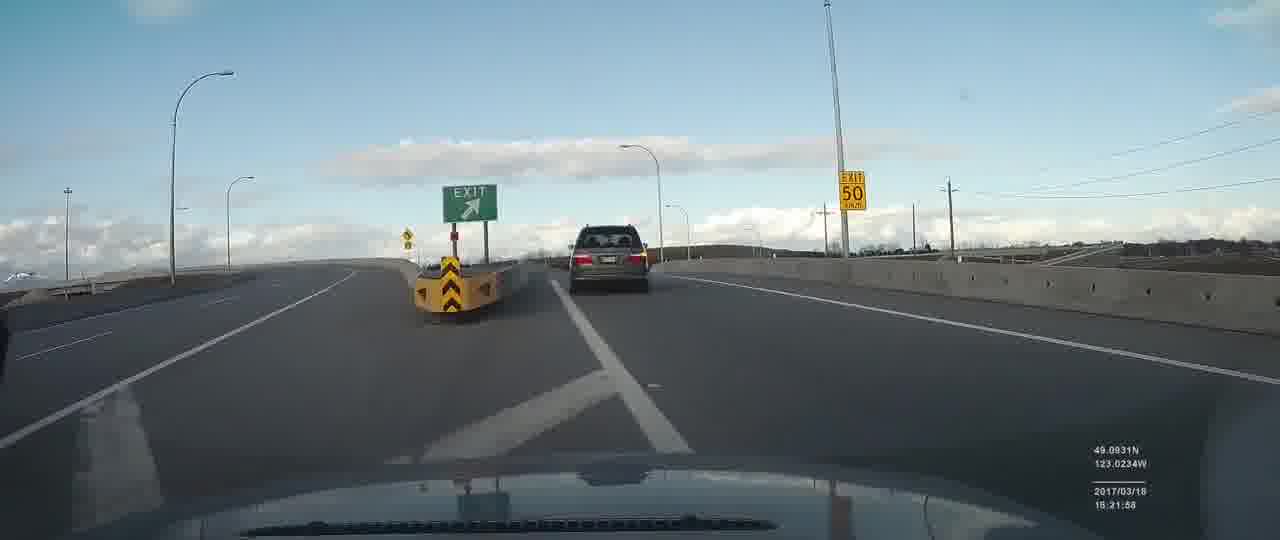}
\subcaption{}
\end{subfigure}

\begin{subfigure}[c]{\textwidth}
\centering
\includegraphics[width=0.19\textwidth]{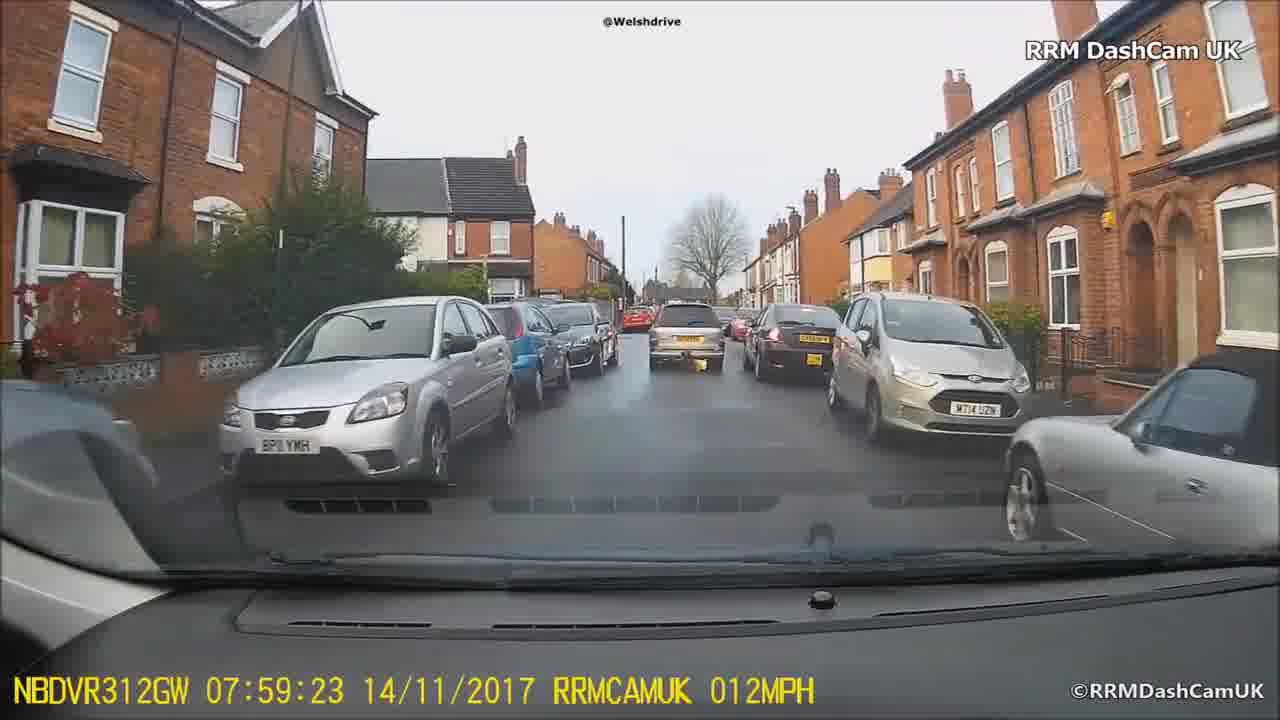}
\includegraphics[width=0.19\textwidth]{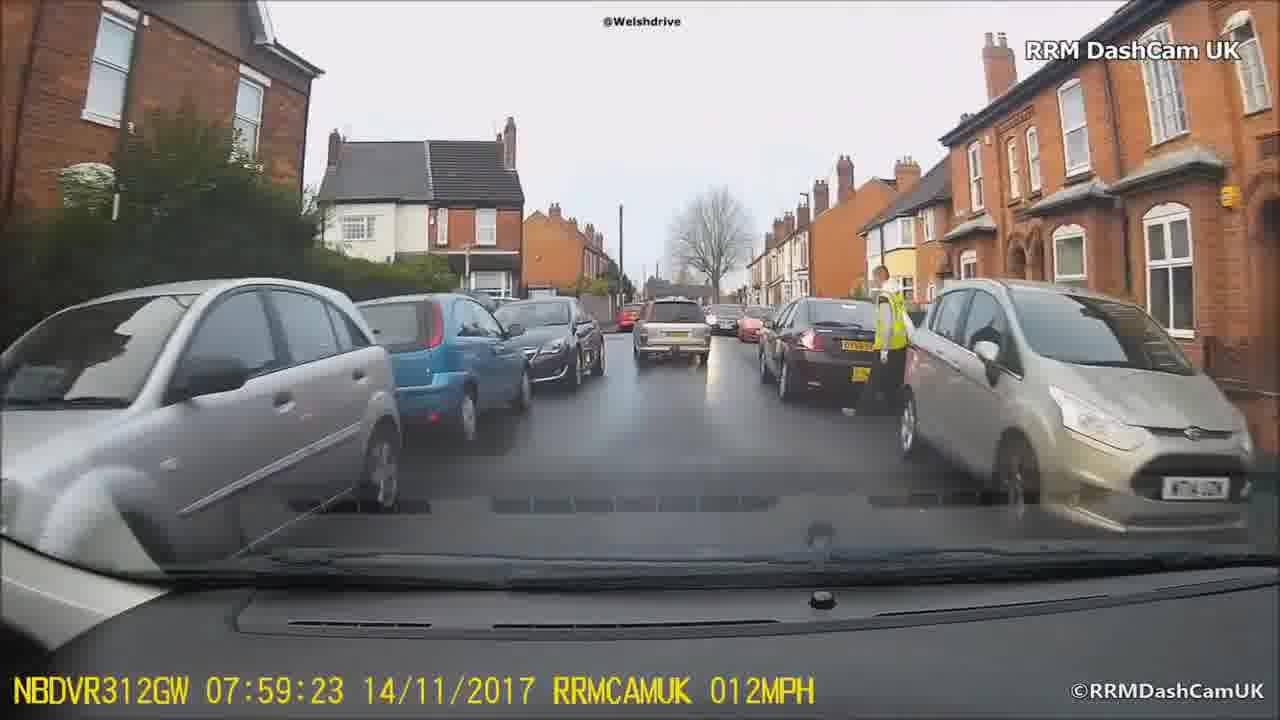}
\includegraphics[width=0.19\textwidth]{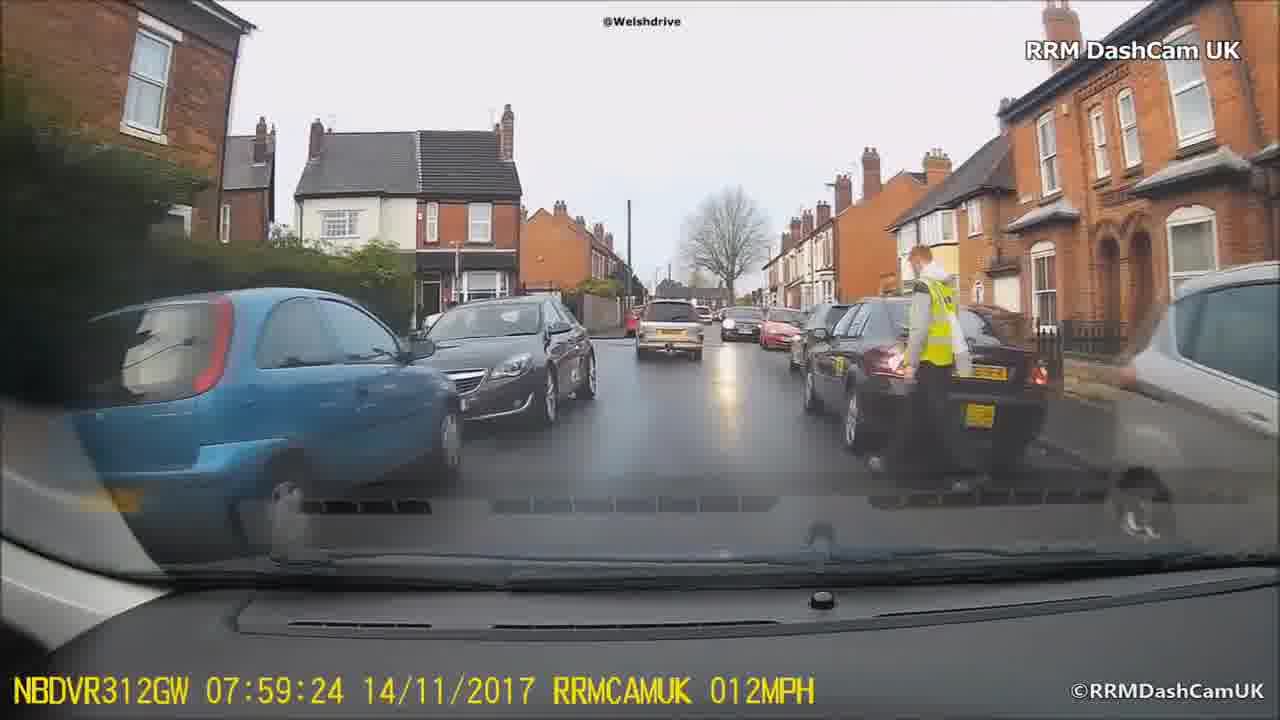}
\includegraphics[width=0.19\textwidth]{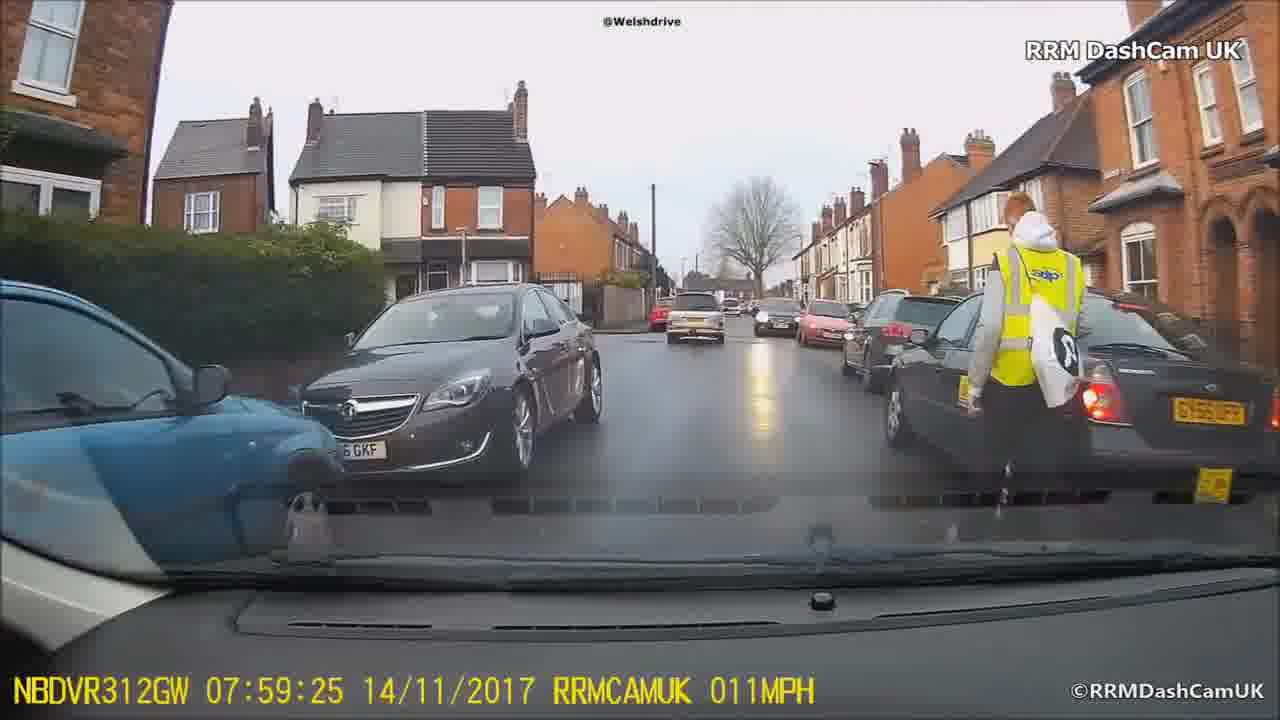}
\includegraphics[width=0.19\textwidth]{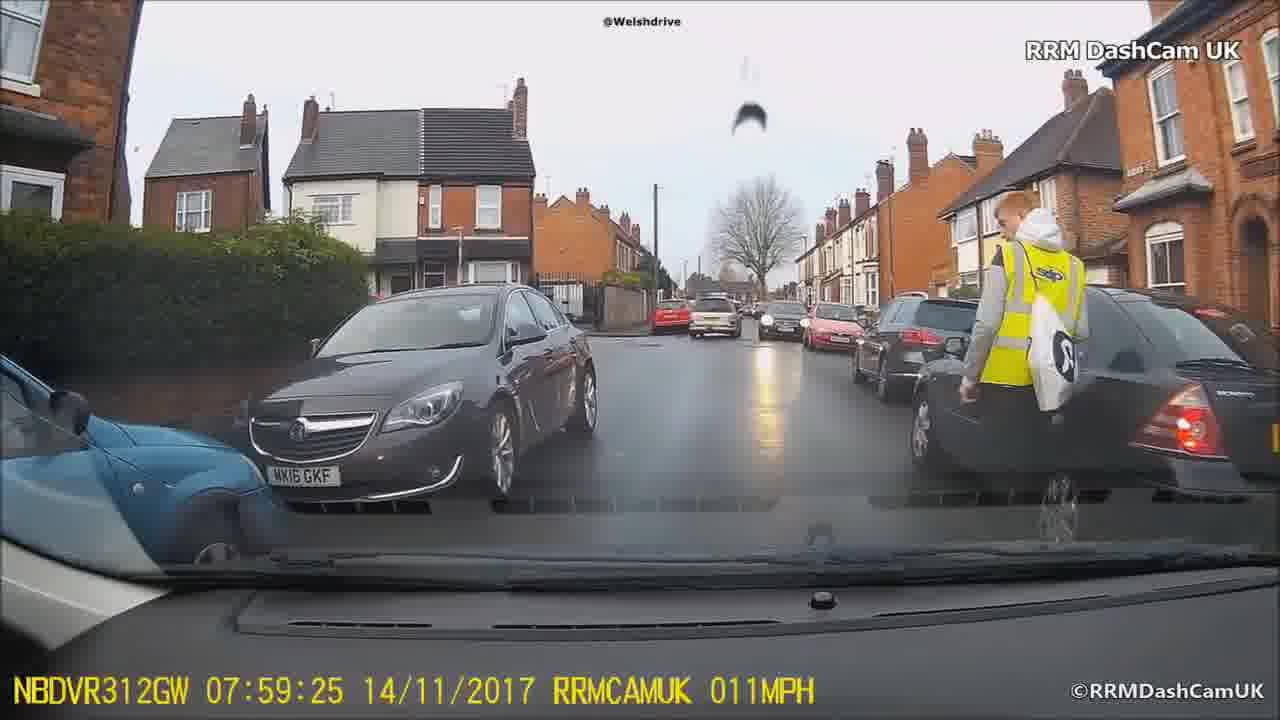}
\subcaption{}
\end{subfigure}

\begin{subfigure}[c]{\textwidth}
\centering
\includegraphics[width=0.19\textwidth]{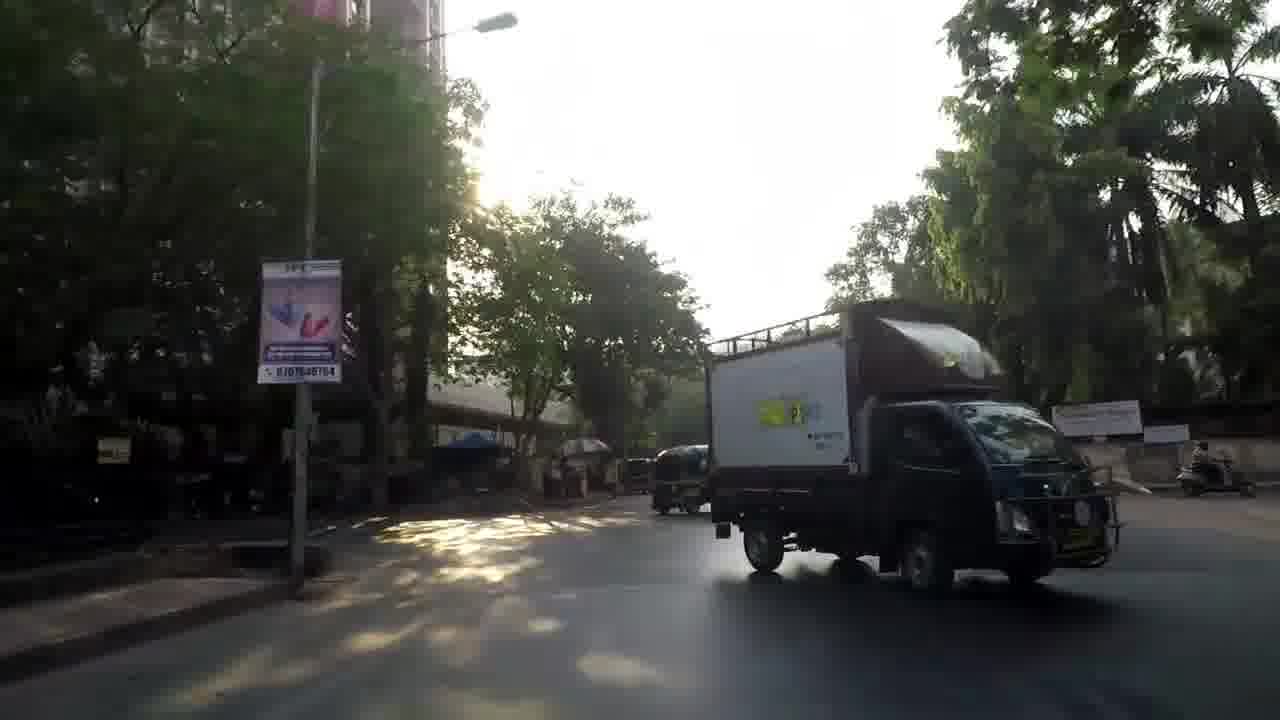}
\includegraphics[width=0.19\textwidth]{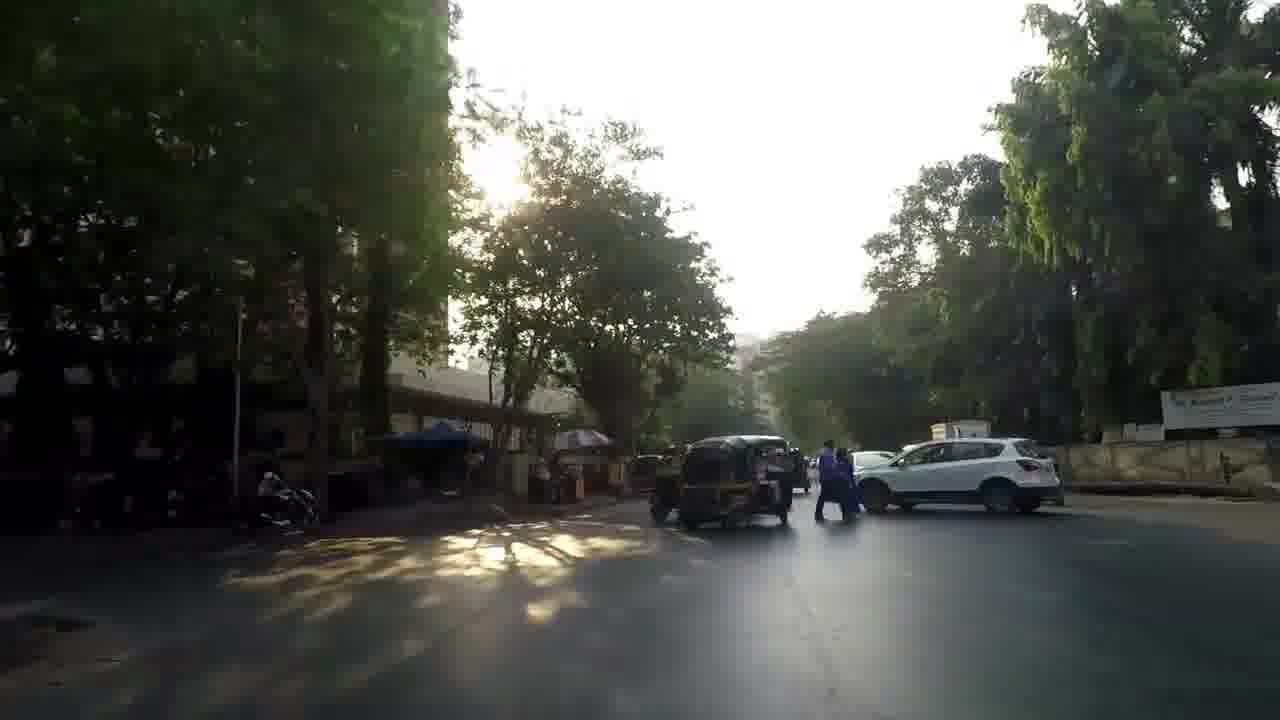}
\includegraphics[width=0.19\textwidth]{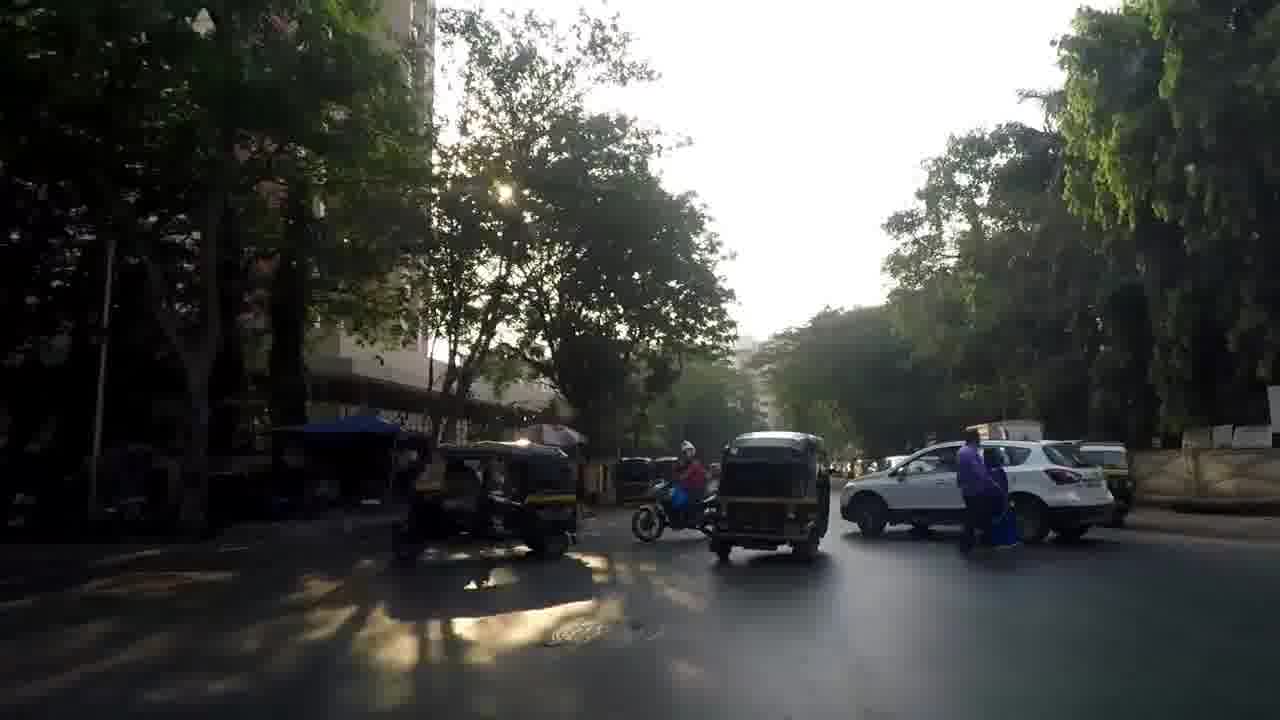}
\includegraphics[width=0.19\textwidth]{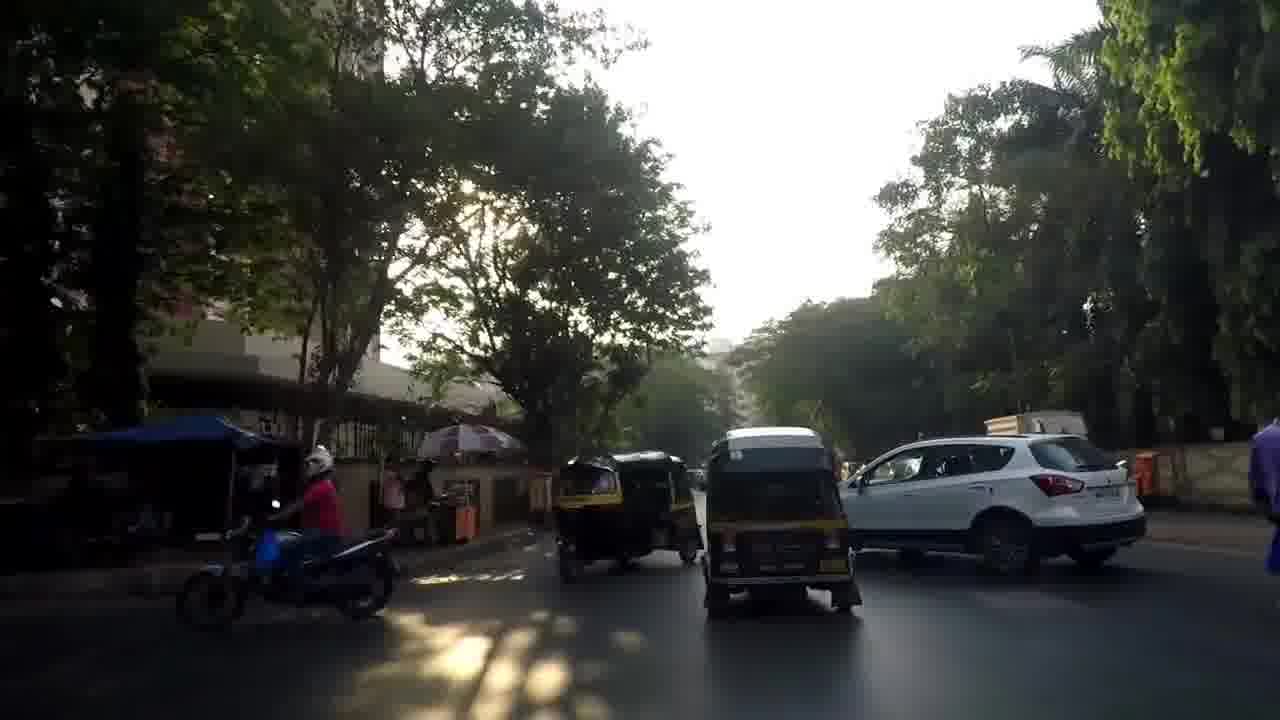}
\includegraphics[width=0.19\textwidth]{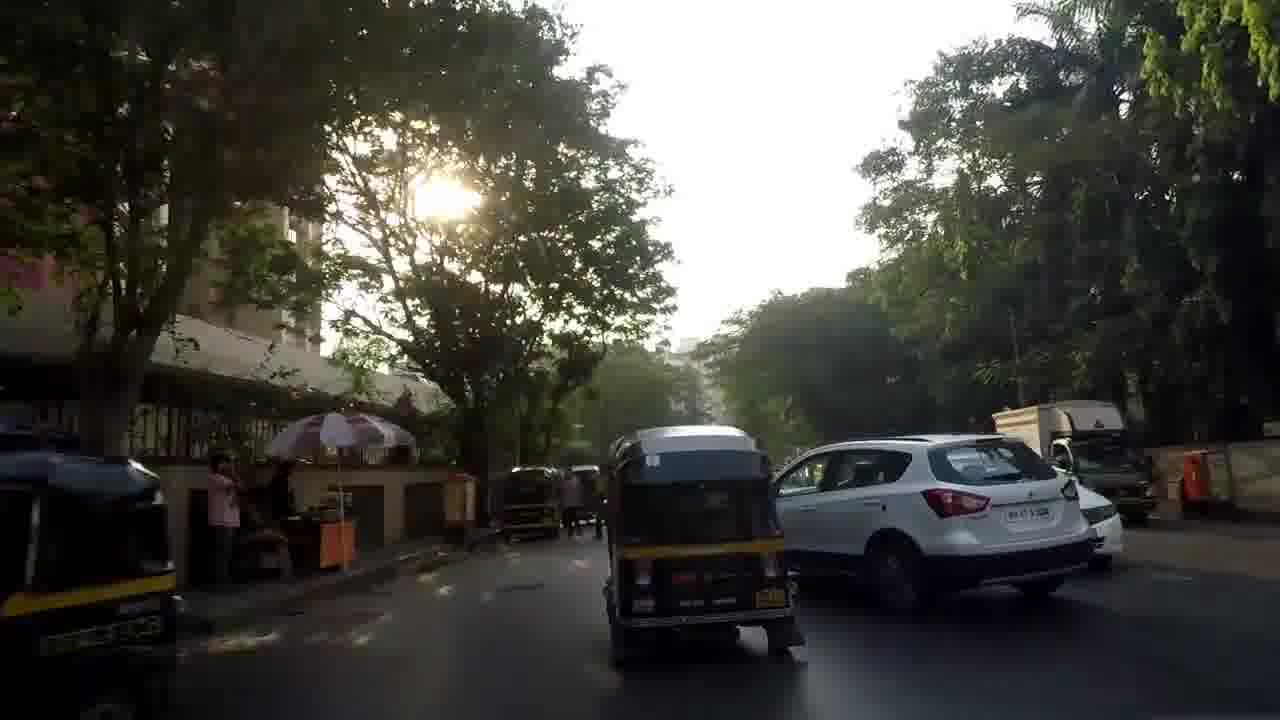}
\subcaption{}
\end{subfigure}

\caption{{\sffamily\footnotesize \textbf{Sample Safety-Critical Episodes}:\quad  \textbf{(a)}. overtaking event in front of the car; \textbf{(b)}. occlusion while turning left; \textbf{(c)}. abrupt lane change on the highway;  \textbf{(d)}. pedestrian suddenly appearing from between two parked cars; and \textbf{(e)}. (relatively) crowded and chaotic inner city traffic}.}
\label{fig:safety-critical-sits_examples}
\end{figure}

\section{\uppercase{Evaluation:$~$ Application and Empirical Performance Analysis}}\label{sec:application-sec}
We demonstrate applicability towards identifying and interpreting \emph{safety-critical situations} (e.g., Table \ref{tbl:safety-critical-situations}; Figures \ref{fig:safety-critical-sits}, \ref{fig:safety-critical-sits_examples}; Fig. \ref{fig:MT-examples-scene-safety}); these encompass those scenarios where interpretation of spacetime dynamics, driving behaviour, environmental characteristics is necessary to anticipate and avoid potential dangers. We also provide an empirical evaluation of the active sensemaking framework in the context of community benchmark datasets.

\subsection{Application: Visual Perception by Abduction}\label{sec:application-driving}

\begin{figure}[t]
\centering

\centering
\includegraphics[width=0.16\textwidth]{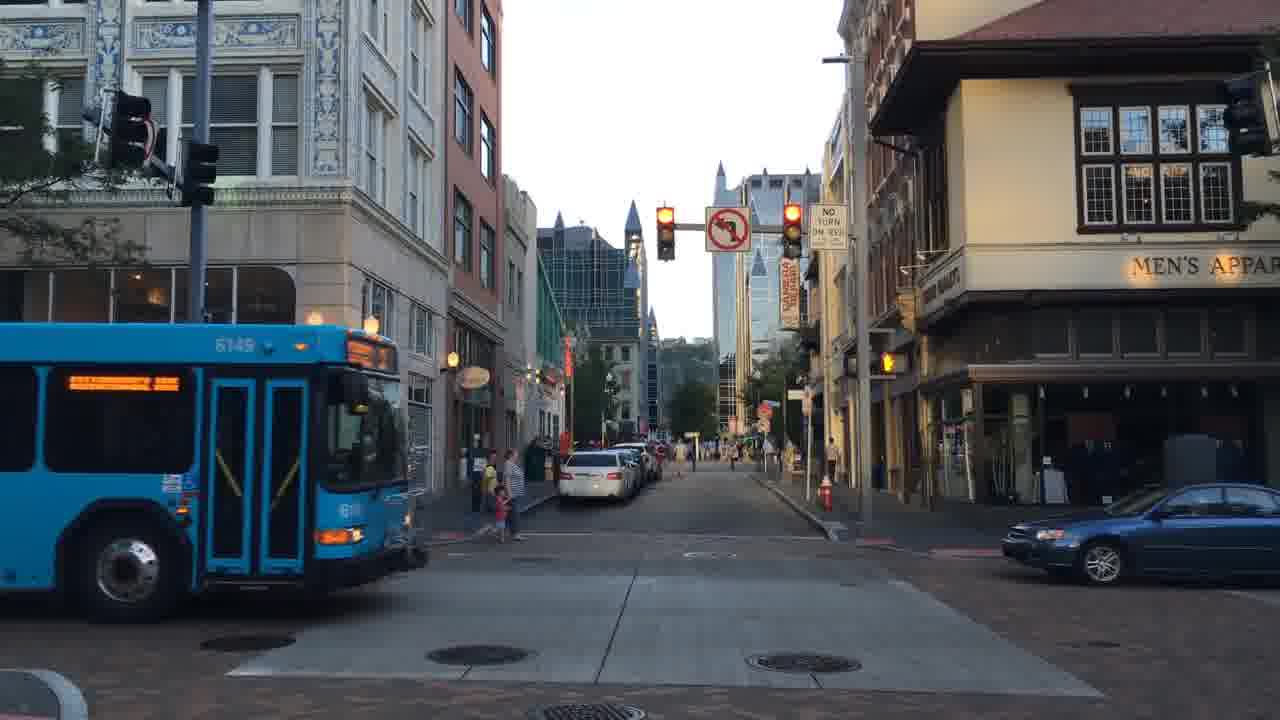}
\includegraphics[width=0.16\textwidth]{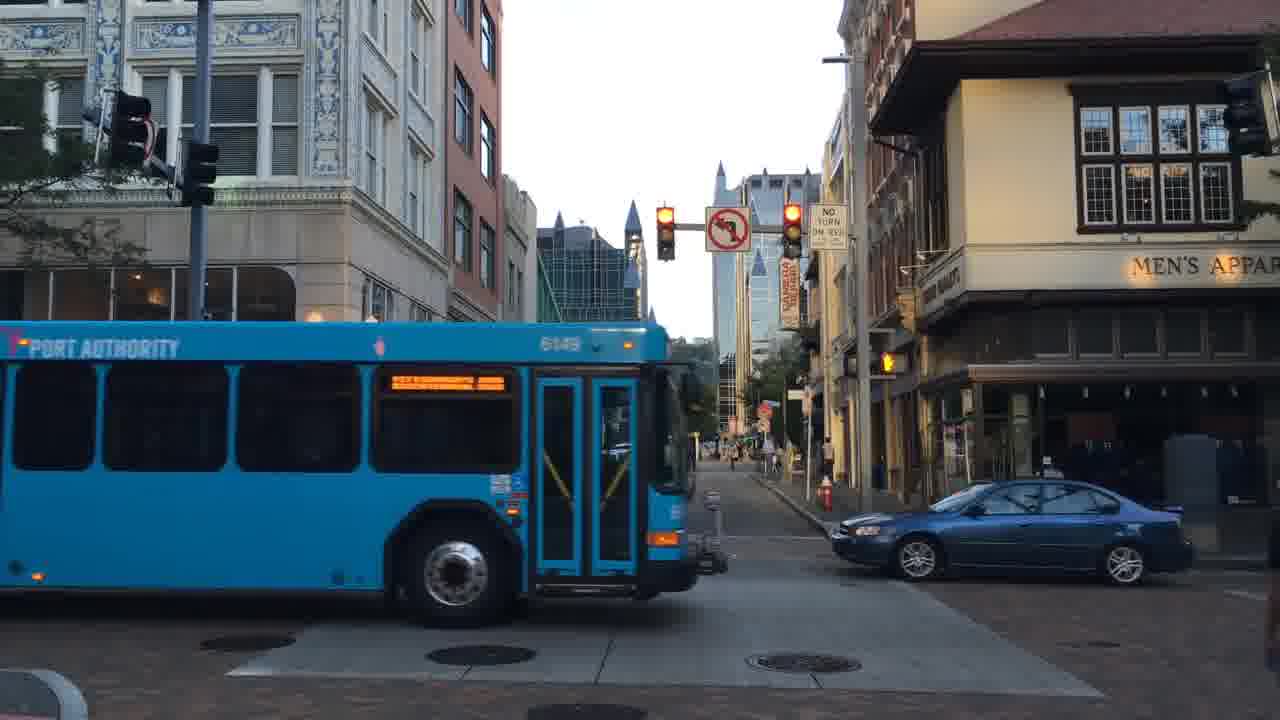}
\includegraphics[width=0.16\textwidth]{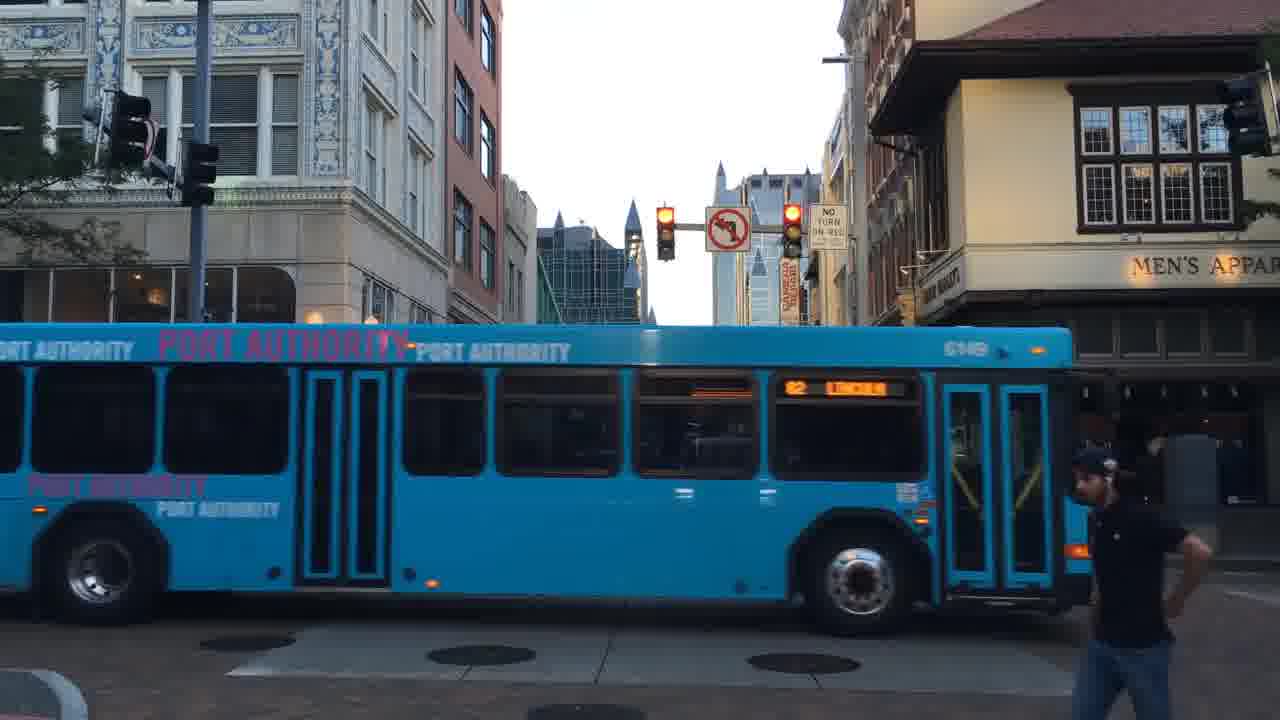}
\includegraphics[width=0.16\textwidth]{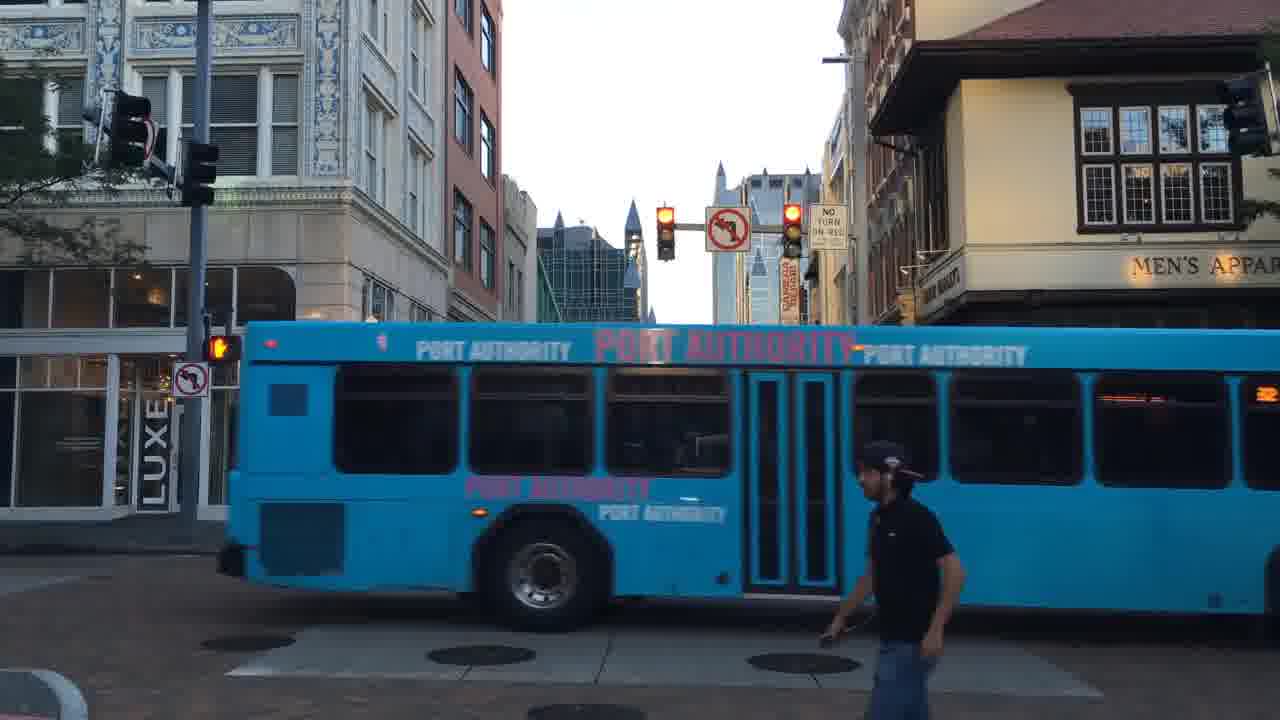}
\includegraphics[width=0.16\textwidth]{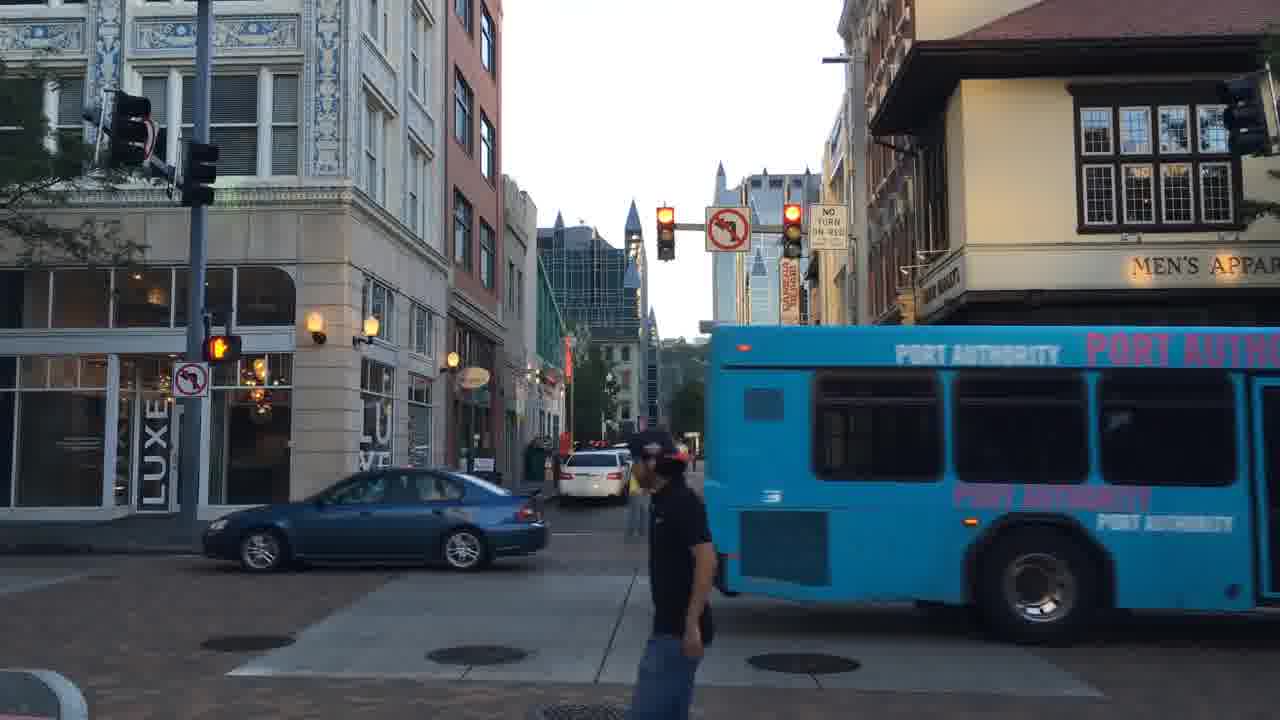}
\includegraphics[width=0.16\textwidth]{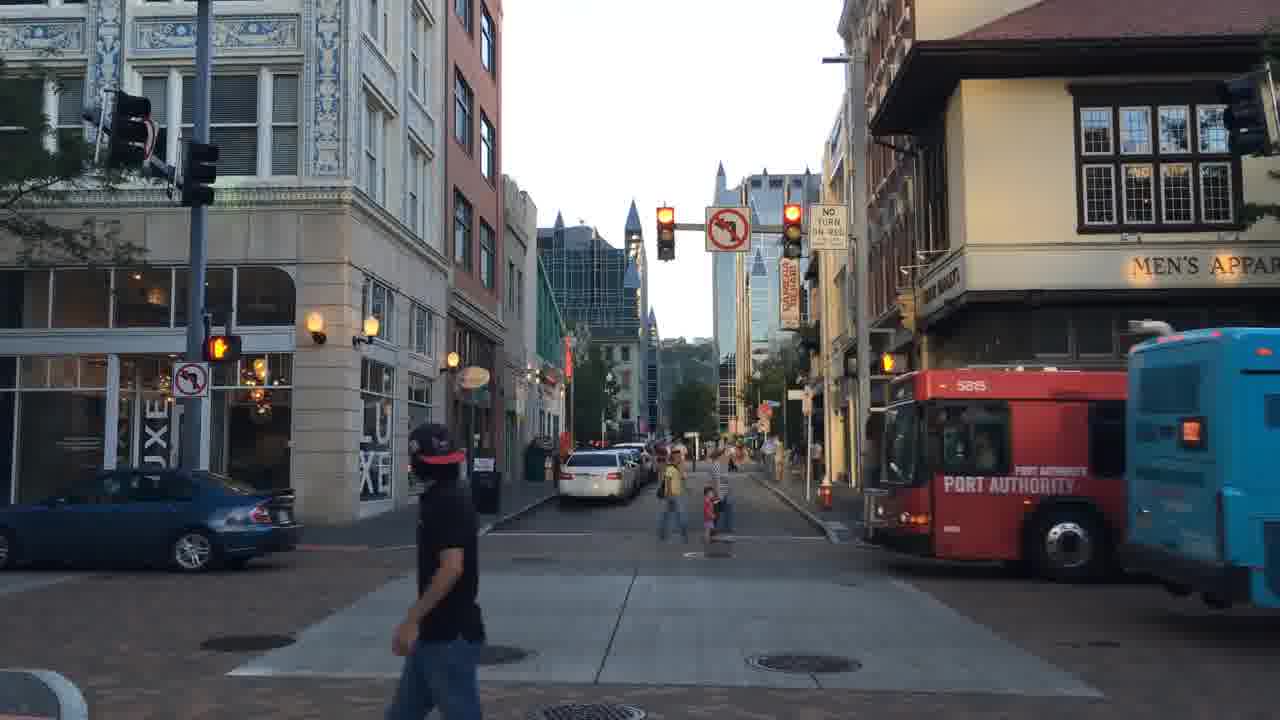}

\vspace{6pt}

\includegraphics[width = 1.0\columnwidth]{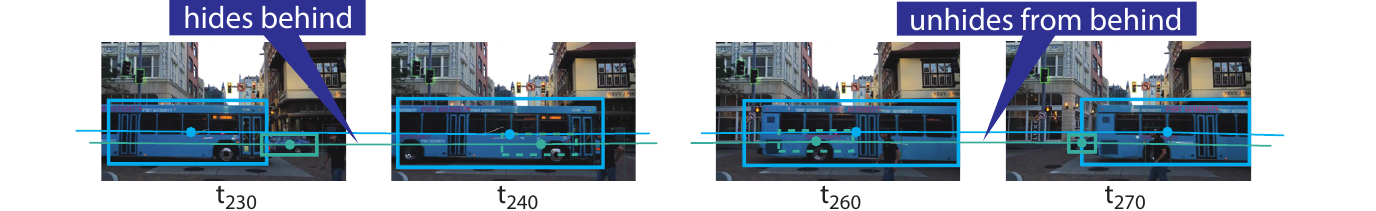}

\caption{Abducing 'Hiding Behind' Event}
\label{fig:app:example_pgh_02}
\end{figure}

\subsubsection{Abducing Explanations: Appearance and Disappearance}

%

Consider the scene in Fig. \ref{fig:app:example_pgh_02}, where a car is passing behind a bus and is getting hidden during this. When the car hides behind the bus (time point 235), the track $trk\_13$ gets $halted$ and the event $hides\_behind$ is abduced to explain why the car is not detected anymore and the corresponding track is halted.


The problem specification for time point $235$ ($<\mathcal{VO}_{235}, \mathcal{P}_{235}, \mathcal{ML}_{235} >$) is given as follows: 

$\bullet$ \quad $\mathcal{VO}_{235}$ the visual observation, consisting of the object detections, given by the bounding box, the type and the confidence:

\footnotesize
\begin{minted}[bgcolor=blue!5!white]{prolog}
det(det_0, person, 99). det(det_1, bus, 99). det(det_2, traffic_light, 86). 
det(det_3, traffic_light, 81). det(det_4, traffic_light, 78). 
det(det_5, traffic_light, 59). 

box2d(det_0, 1114, 450, 148, 270). box2d(det_1, 8, 305, 992, 333). 
box2d(det_2, 656, 205, 21, 56). box2d(det_3, 179, 137, 42, 75). 
box2d(det_4, 108, 89, 46, 86). box2d(det_5, 784, 202, 21, 44).
\end{minted} 
\normalsize

$\bullet$ \quad $\mathcal{P}_{235}$ the predictions for each track, given by the predicted bounding box, the state in which the track currently is, and the type of the tracked object:

\footnotesize
\begin{minted}[bgcolor=blue!5!white]{prolog}
trk(trk_3, traffic_light). trk_state(trk_3, active). trk(trk_7, traffic_light). 
trk_state(trk_7, active). trk(trk_8, traffic_light). trk_state(trk_8, active). 
trk(trk_12, bus). trk_state(trk_12, active). trk(trk_13, car). 
trk_state(trk_13, active). trk(trk_15, person). trk_state(trk_15, active).

box2d(trk_3, 178, 136, 43, 73). box2d(trk_7, 105, 90, 49, 82). 
box2d(trk_8, 655, 205, 21, 55). box2d(trk_12, 48, 294, 915, 350).
box2d(trk_13, 904, 473, 181, 108). box2d(trk_15, 1111, 427, 156, 310).
\end{minted} 
\normalsize 

%

$\bullet$ \quad And $\mathcal{ML}_{235}$ the matching likelihood for each track with each detection, here given by the IoU between the detection bounding box and the predicted bounding box for the track\footnote{Note, only those IoUs are stated which are bigger than 0.}:

\footnotesize
\begin{minted}[bgcolor=blue!5!white]{prolog}
iou(trk_15,det_0,82426). iou(trk_12,det_1,88079). iou(trk_13,det_1,3022). 
iou(trk_8,det_2,98532). iou(trk_3,det_3,94981). iou(trk_7,det_4,90457).
\end{minted} 
\normalsize

Solving the assignment of detections to tracks can now be done based on the \emph{choice rules} for associating objects and observations, detailed in Section \ref{sec:tracking-as-abbduction} Step 2. 

To restrict the assignment we can impose constraints on the matching, by stating integrity constraints, e.g., for ensuring that only tracks and detections with the same type are matched, we could state the following integrity constraint. Stating that any stable model where the body is satisfied can not be in the set of answers, i.e., any model assigning a track and a detection which are not of the same type can not be an answer. Further, the track has to be active, the confidence of the detection has to be above a threshold, and the IoU between the track and the detection has to be above a threshold:

\footnotesize
\begin{minted}[bgcolor=blue!5!white]{prolog}
:- assign(Trk, Det), not assignment_constraints(Trk, Det).

assignment_constraints(Trk, Det) :-
    trk(Trk, Trk_Type), det(Det, Det_Type, Conf),
    trk_state(Trk, active), 
    match_type(Trk_Type, Det_Type),
    Conf > conf_thresh_assign,
    iou(Trk, Det, IOU), IOU > iou_thresh.

\end{minted}
\normalsize

By maximizing the matching likelihood we get the optimal assignment of detections to tracks, in our example the bus is detected by detection $det\_1$ which gets assigned to the corresponding track $trk\_12$, but as the car is hiding behind the bus, there is no corresponding detection, thus the track of the car $trk\_13$ gets halted:

\footnotesize
\begin{minted}[bgcolor=green!25!white]{prolog}
halt(trk_13) assign(trk_15,det_0) assign(trk_12,det_1) assign(trk_8,det_2) 
assign(trk_3,det_3) assign(trk_7,det_4)
\end{minted}
\normalsize 

The assignment actions are linked with high-level events for explaining the assignments, i.e., the halted track $trk\_13$ can be explained either by missing detections or by the track hiding behind another track. In this case track $trk\_13$ is hiding behind track $trk\_12$, this can be abduced based on possible events, which in this case is the $hides\_behind$ event.

For the event $hides\_behind$\textbackslash2 the predicted tracks have to be overlapping. This is ensured by (spatial) preconditions of the event, given by the predicate $poss$\textbackslash1:    

\footnotesize
\begin{minted}[bgcolor=blue!5!white]{prolog}
poss(hides_behind(Trk1, Trk2)) :-
    trk(Trk1, _), trk(Trk2, _),
    position(overlapping_top, Trk1, Trk2),
    not holds_at(visibility(Trk1), not_visible, curr_time),
    not holds_at(visibility(Trk2), not_visible, curr_time).
\end{minted}
\normalsize

In our example we can now abduce that the track $trk\_13$ representing the car is ended, because the car got hidden by the bus represented by track $trk\_12$. In the formal representation of event calculus this is represented by the predicate $occurs\_at$\textbackslash2 as follows: 

\footnotesize
\begin{minted}[bgcolor=green!25!white]{prolog}
occurs_at(hides_behind(trk_13,trk_12),235)
\end{minted}
\normalsize

At time point $268$ the car reappears, after passing behind the bus.
Due to the previously abduced event $hides\_behind$\textbackslash2, the visibility fluent for the track of the car $trk\_13$ has now the value $not\_visible$. 

For the detection $det\_1$ we can then abduce that track $trk\_13$ unhides from behind track $trk\_12$ based on the following event definition, stating that the event $unhides\_from\_behind$\textbackslash2 is possible when Trk1 is $not\_visible$ and Trk2 is not $not\_visible$:

\footnotesize
\begin{minted}[bgcolor=blue!5!white]{prolog}
poss(unhides_from_behind(Trk1, Trk2)) :-
    trk(Trk1, _), trk(Trk2, _),
    holds_at(visibility(Trk1), not_visible, curr_time),
    not holds_at(visibility(Trk2), not_visible, curr_time).
\end{minted}
\normalsize 


\footnotesize
\begin{minted}[bgcolor=green!25!white]{prolog}
resume(trk_13,det_1) assign(trk_15,det_0) assign(trk_12,det_2) assign(trk_7,det_3) 
assign(trk_8,det_4) assign(trk_3,det_5)
\end{minted}
\normalsize

\footnotesize
\begin{minted}[bgcolor=green!25!white]{prolog}
occurs_at(unhides_from_behind(trk_13,trk_12),268))
\end{minted}
\normalsize

\begin{figure}[t]
\centering

\centering
\includegraphics[width=0.19\textwidth]{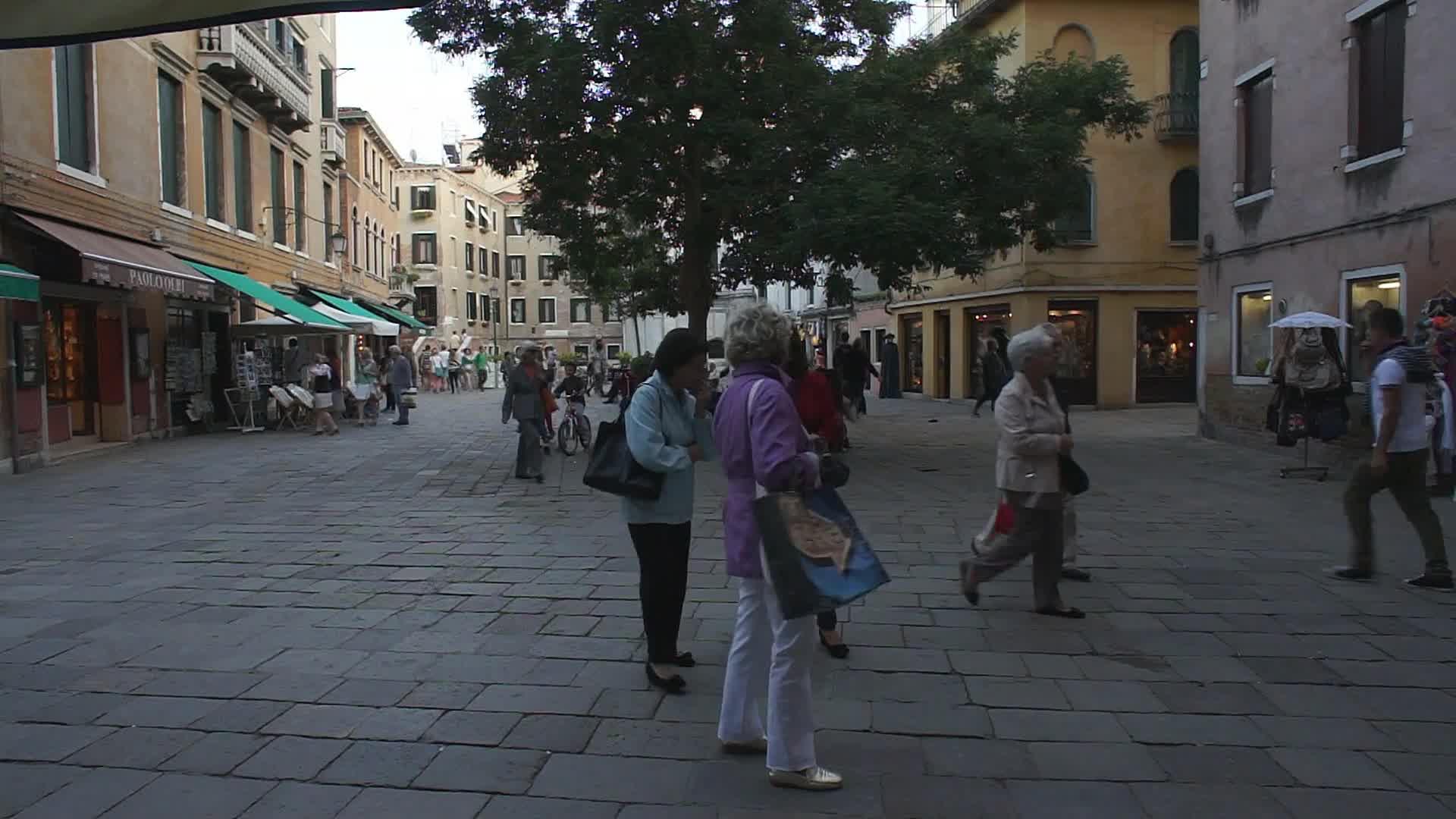}
\includegraphics[width=0.19\textwidth]{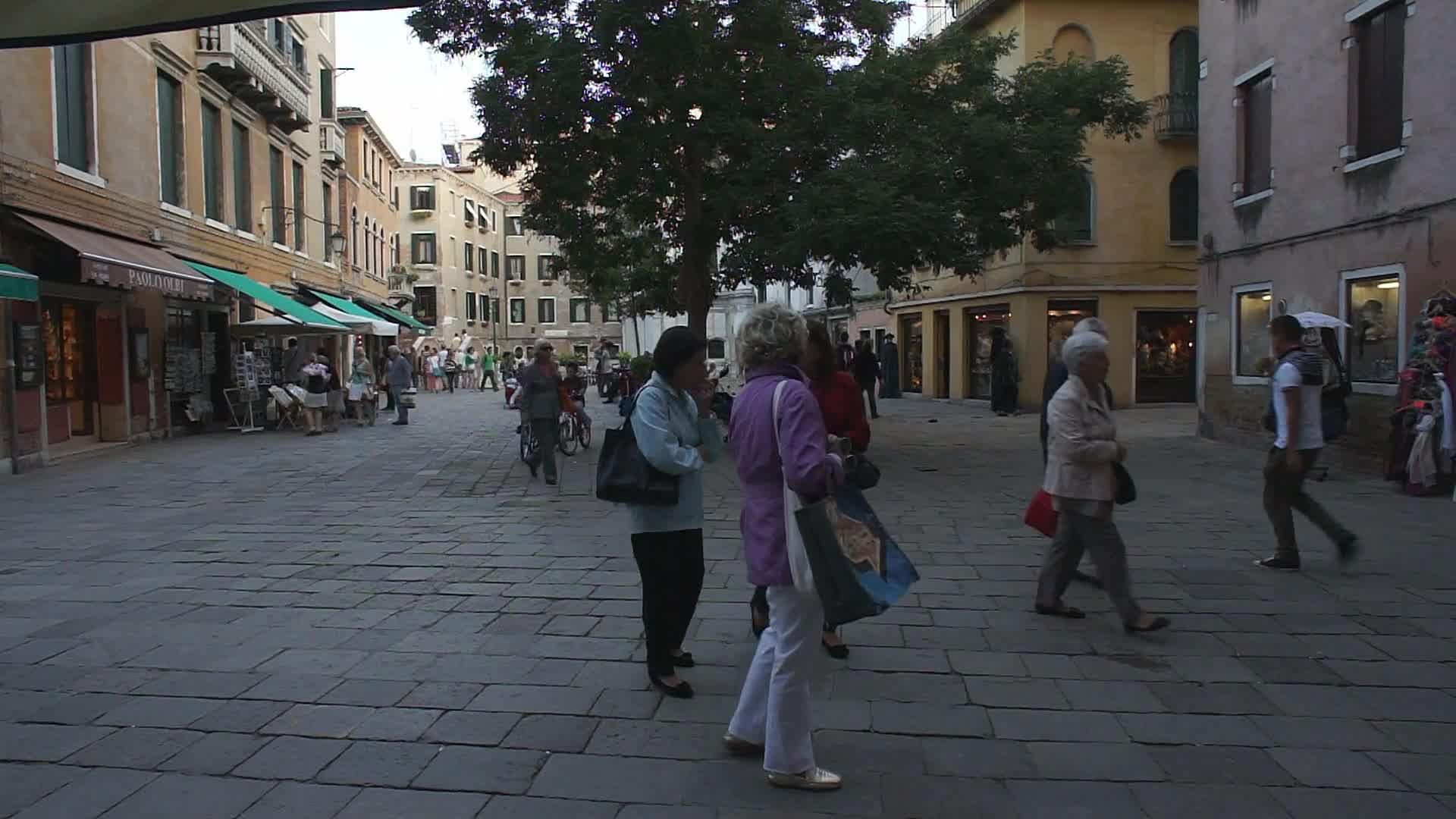}
\includegraphics[width=0.19\textwidth]{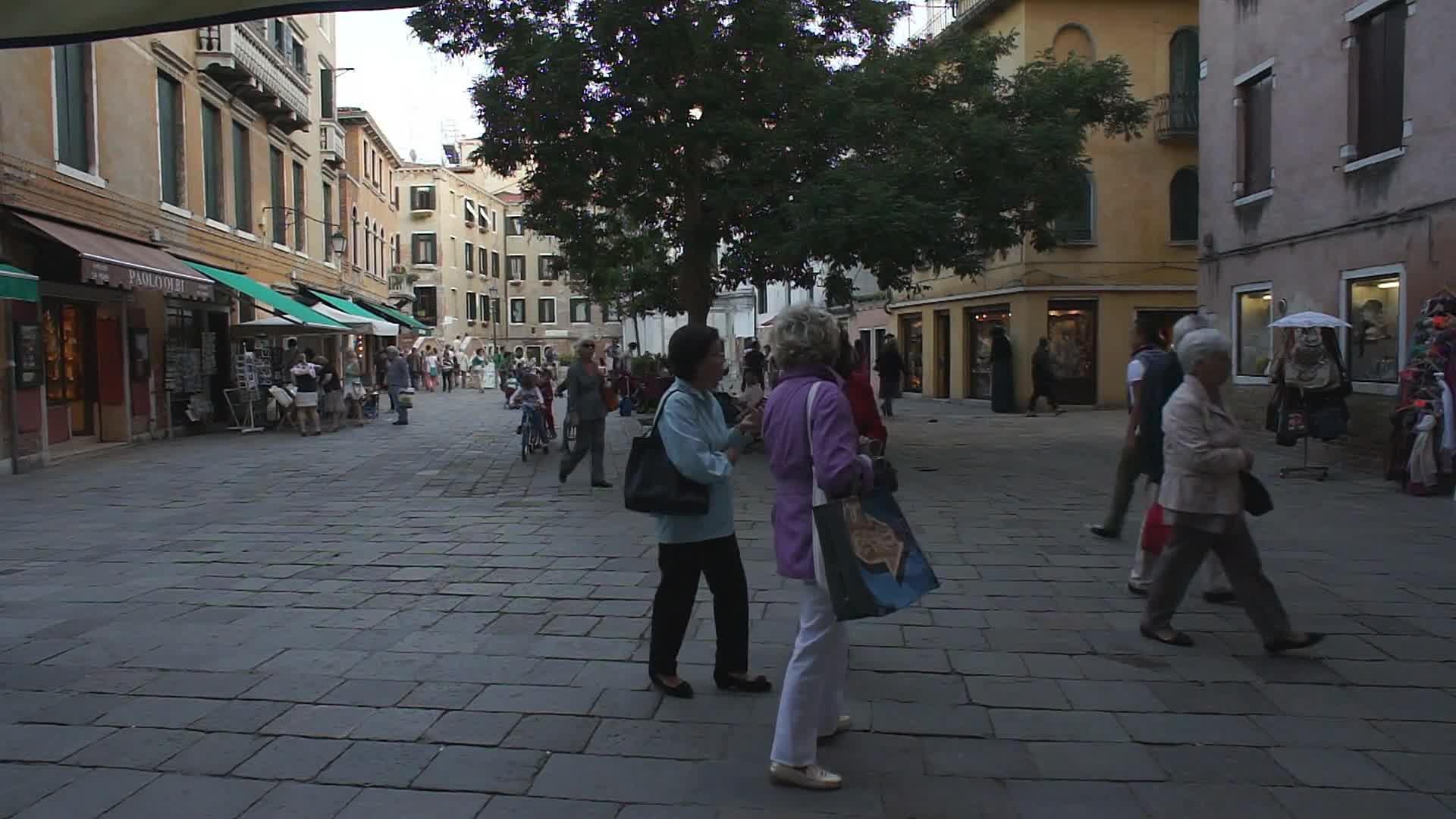}
\includegraphics[width=0.19\textwidth]{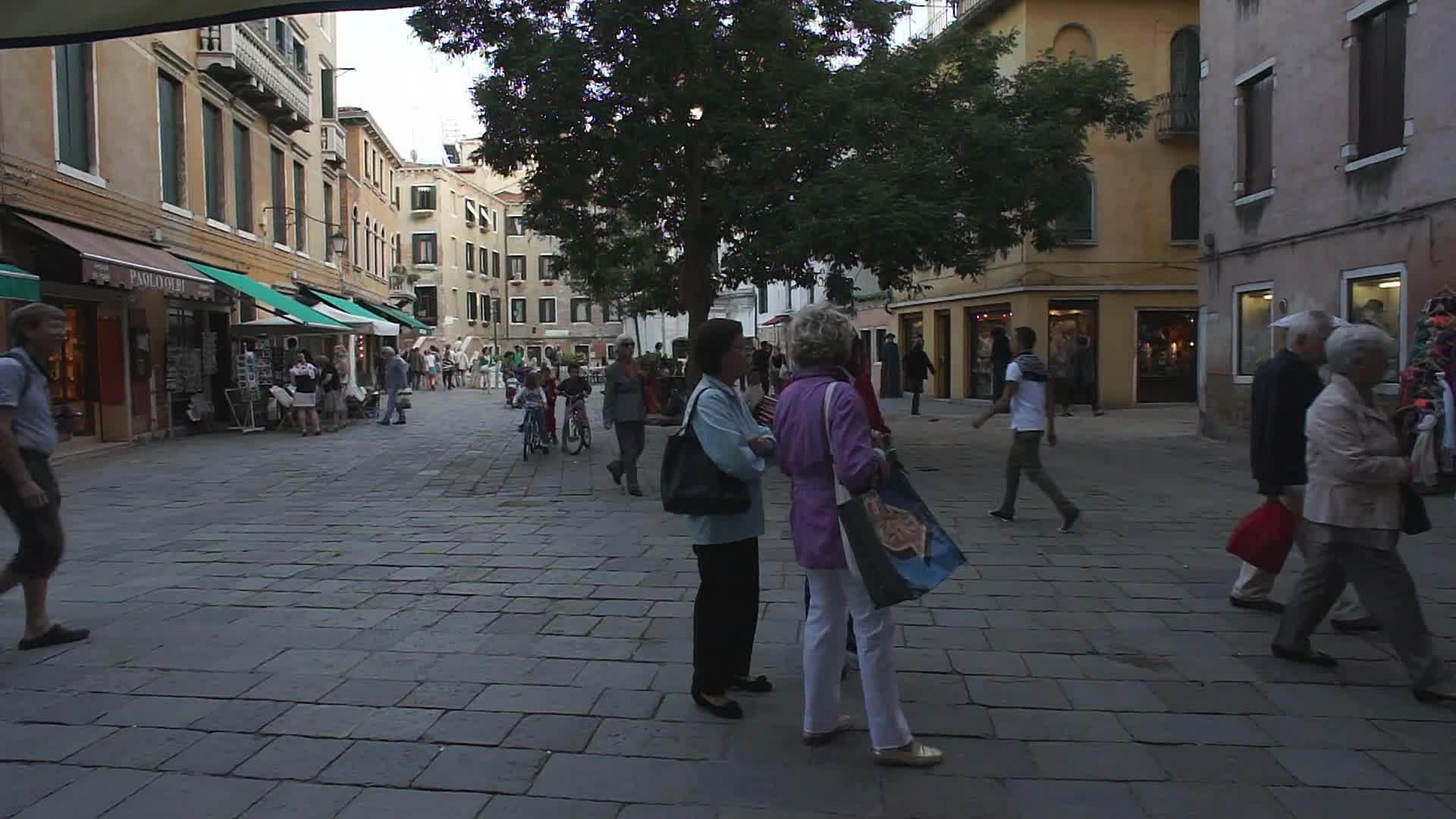}
\includegraphics[width=0.19\textwidth]{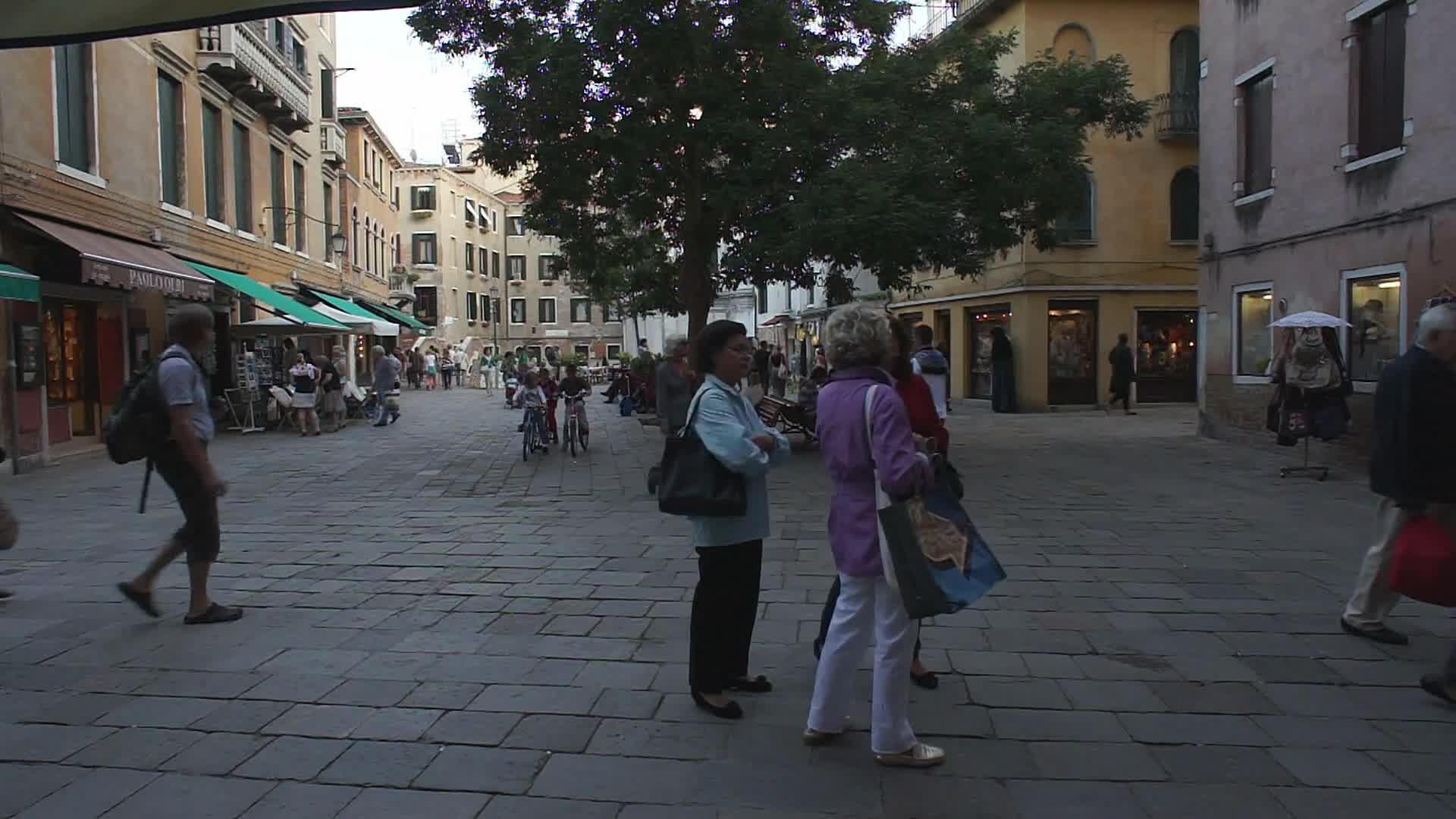}

\vspace{6pt}

\includegraphics[width = 1.0\columnwidth]{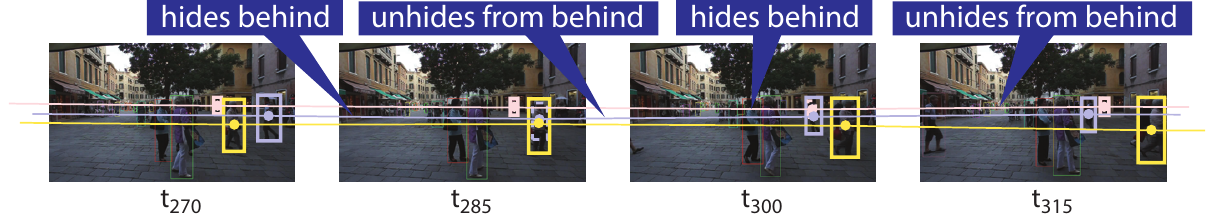}

\caption{{\sffamily\footnotesize Abducing Event Sequences (Example from MOT Dataset)}}
\label{fig:example_mot}
\end{figure}

Similarly, when looking at a slightly more complex scene, like the one depicted in Fig. \ref{fig:example_mot}, we get an event sequence describing the interactions happening in the scene:

 \footnotesize
\begin{minted}[bgcolor=green!25!white]{prolog}
...
 occurs_at(hides_behind(trk_34,trk_16),283) 
 occurs_at(unhides_from_behind(trk_34,trk_16),293)
 occurs_at(hides_behind(trk_37,trk_34),296) 
 occurs_at(unhides_from_behind(trk_37,trk_34),311)
 ...
\end{minted}
\normalsize 

This event sequence explains the visuospatial dynamics of the scene and can be used for reasoning about the scene.

\subsubsection{Reasoning about Hidden Entities}\label{sec:reasoning-hidden-entities}

Consider the situation of Fig. \ref{fig:example_occlusion}: a {\sffamily\small car} gets occluded by another car turning left and reappears \emph{in front of} the autonomous vehicle. Using online abduction for abducing high-level interactions of scene objects we can hypothesize that the {\sffamily\small car} got \emph{occluded} and anticipate its reappearance based on the perceived scene dynamics. 

The predictions for each track is given by the predicted bounding box, the state in which the track currently is, and the type of the tracked object:


\footnotesize
\begin{minted}[bgcolor=blue!5!white]{prolog}
trk(trk_3, car). trk_state(trk_3, active). 
...
trk(trk_41, car). trk_state(trk_41, active). 
...

box2d(trk_3, 660, 460, 134, 102). 
...
box2d(trk_41, 631, 471, 40, 47). 
...
\end{minted} 
\normalsize

{
\begin{figure}[t]
\center

\centering
\includegraphics[width=0.19\textwidth]{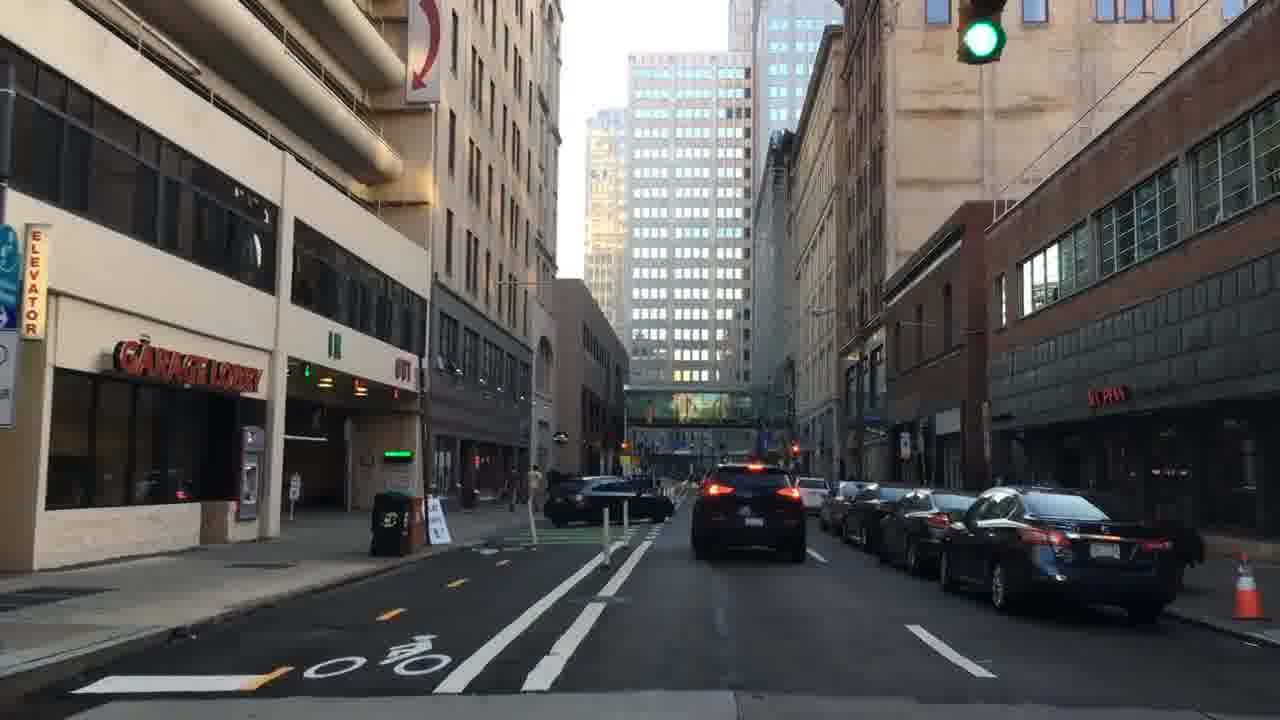}
\includegraphics[width=0.19\textwidth]{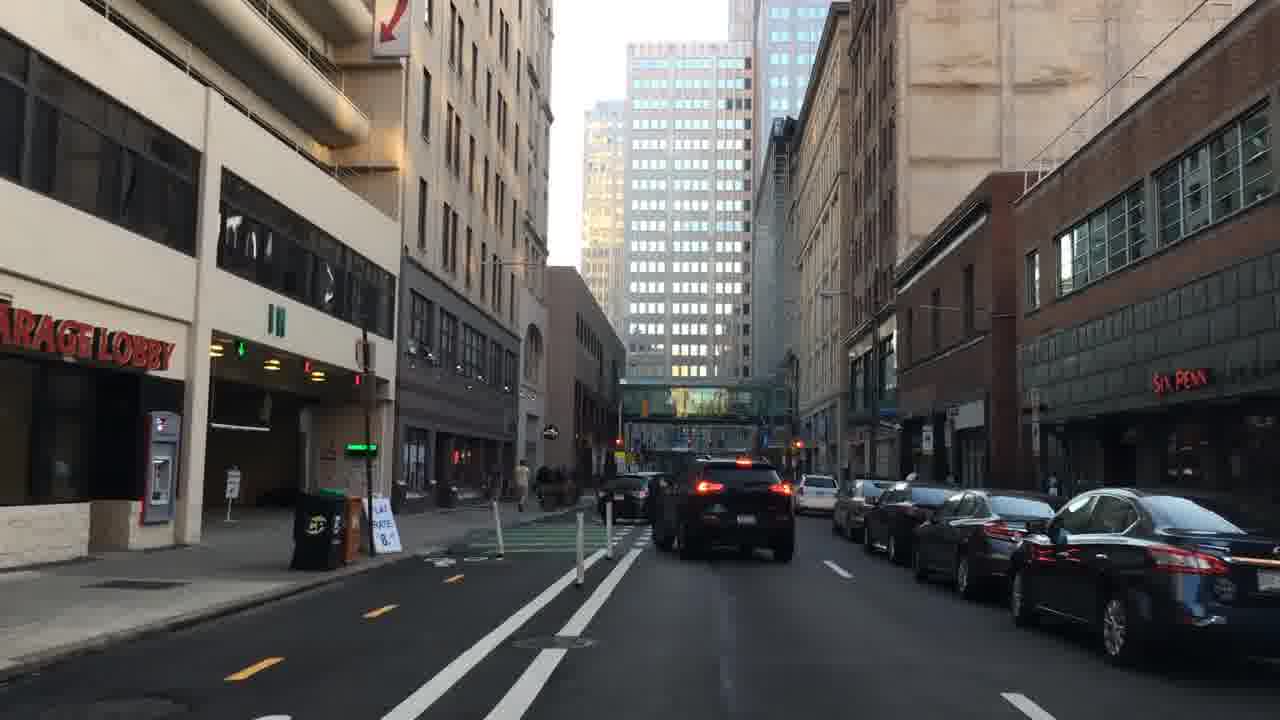}
\includegraphics[width=0.19\textwidth]{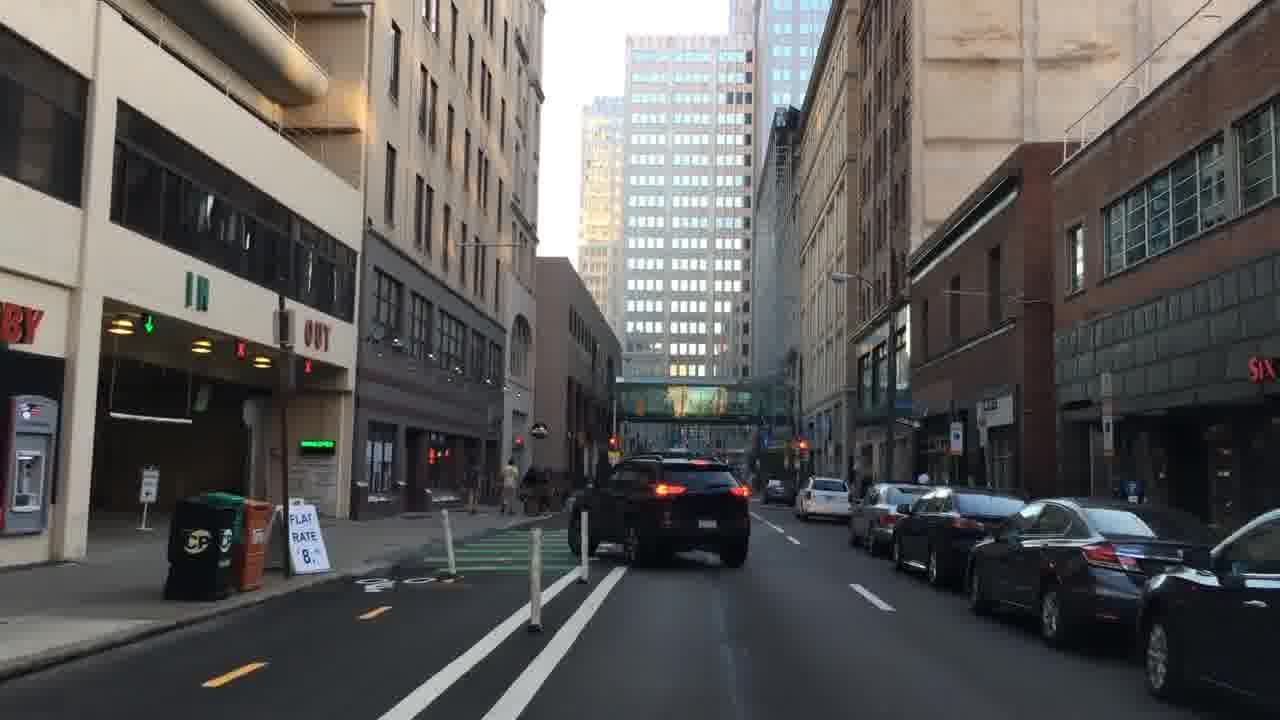}
\includegraphics[width=0.19\textwidth]{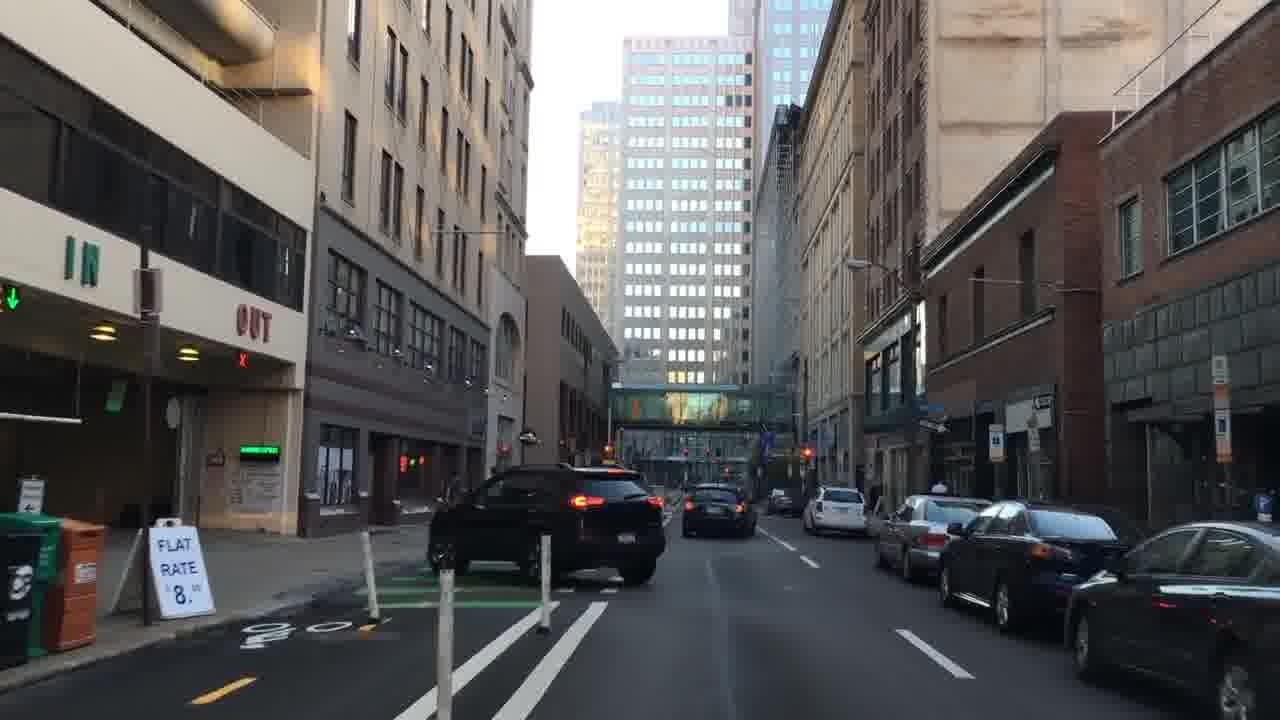}
\includegraphics[width=0.19\textwidth]{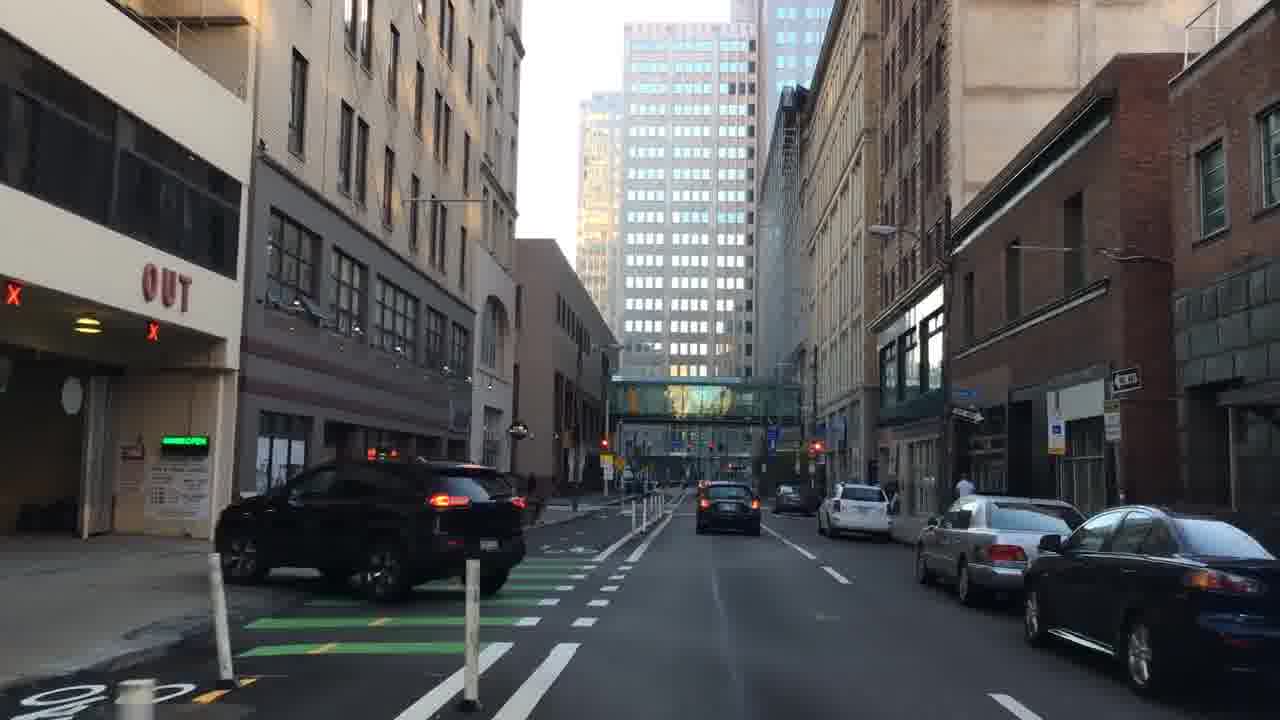}

\vspace{6pt}

\includegraphics[width = 1.0\columnwidth]{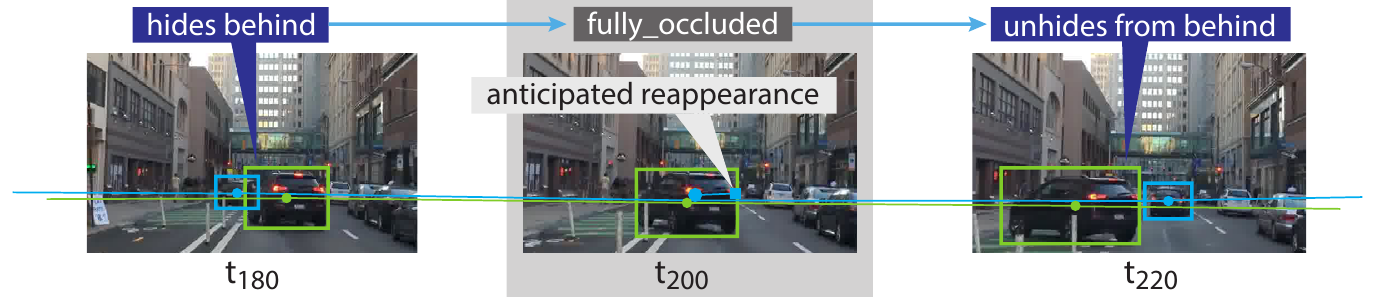}
\caption{{Abducing Occlusion to Anticipate Reappearance}}
\label{fig:example_occlusion}
\end{figure}
}

%
%

Based on this problem specification for time point  $179$, the event $hides\_behind(trk\_41,trk\_3)$ is abduced, as there is no detection that could be associated with $trk\_41$ and  $trk\_3$ is partially overlapping with $trk\_41$:

\footnotesize
\begin{minted}[bgcolor=green!25!white]{prolog}
... occurs_at(hides_behind(trk_41,trk_3),179) ...
\end{minted}
\normalsize 

The abduced explanation together with the object dynamics may then be used for visual reasoning and anticipation of events, which can serve for decision support. 
Towards this we define a rule stating that a \emph{hidden} {\sffamily\small object} may \emph{unhide} from behind the object it is hidden by and anticipate the time point $t$ based on the object \emph{movement} as follows: 

%
%
%
%

\footnotesize
\begin{minted}[bgcolor=blue!5!white]{prolog}
anticipate(unhides_from_behind(Trk1, Trk2), T) :-
    time(T), curr_time < T,
    holds_at(hidden_by(Trk1, Trk2), curr_time),
    topology(proper_part, Trk1, Trk2),
    movement(moves_out_of, Trk1, Trk2, T).
\end{minted}
\normalsize

We then interpolate the objects position at time point $t$ to predict where the object may \emph{reappear}:

%
%
%

\footnotesize
\begin{minted}[bgcolor=blue!5!white]{prolog}
point2d(interpolated_position(Trk, T), PosX, PosY) :-
    time(T), curr_time < T, T1 = T-curr_time,
    box2d(Trk1, X, Y,_,_), trk_mov(Trk1, MovX, MovY),
    PosX = X+MovX*T1, PosY = Y+MovX*T1.
\end{minted}
\normalsize

For the occluded {\sffamily\small car} in our example we get the following prediction for time $t$ and position $x,y$:

%
%
%

\footnotesize
\begin{minted}[bgcolor=green!25!white]{prolog}
anticipate(unhides_from_behind(trk_41, trk_2), 202) 
point2d(interpolated_position(trk_41, 202), 738, 495)
\end{minted}
\normalsize

Based on this prediction we can then define a rule that gives a warning if a hidden entity may reappear in front of the vehicle, which could be used by the control mechanism, e.g., to adapt driving and slow down in order to keep safe distance:

%

\footnotesize
\begin{minted}[bgcolor=blue!5!white]{prolog}
warning(hidden_entity_in_front(Trk1, T)) :-
    time(T), T-curr_time < anticipation_threshold,
    anticipate(unhides_from_behind(Trk1, _), T),
    position(in_front, interpolated_pos(Trk1, T)).
\end{minted}
\normalsize

{
\begin{table*}[t]
\renewcommand{\arraystretch}{1.2}
\begin{center}
\scriptsize

\footnotesize

\begin{tabular}{>{\columncolor[gray]{0.92}}l l c c c c c c c c}

\hlinewd{1pt}

{\sffamily\color{blue!70!black}\textbf{BENCHMARK}} & \textbf{MOTA}$\uparrow$ & \textbf{MOTP}$\uparrow$ & ML$\downarrow$ & MT$\uparrow$ & FP$\downarrow$ & FN$\downarrow$ & ID sw.$\downarrow$  & Frag.$\downarrow$ \\\hline 
\hline


\multicolumn{9} {l} {{\sffamily KITTI tracking } -- \emph{Cars} \quad { \scriptsize (8008 frames, 636 targets)} }\\

\hline
-- baseline& 
45.72 \% &
76.89 \% &
19.14 \% &
23.04 \% &
785 &
11182 &
1097 &
1440 \\

-- with Abd.& 
{\bf 50.5 \%} & 
74.76 \% & 
20.21 \% &
23.23 \% &
1311 &
10439 &
165 &
490 \\




\hline

\multicolumn{9} {l} {{\sffamily KITTI tracking } -- \emph{Pedestrians} \quad { \scriptsize (8008 frames, 167 targets)}}\\


\hline

-- baseline & 
28.71 \% &
71.43 \% &
26.94 \% &
9.58 \% &
1261 &
6119 &
539 &
833 \\

-- with Abd.& 
{\bf 32.57 \%} & 
70.68 \% & 
22.15 \% &
14.37 \% &
1899 &
5477 &
115 &
444 \\




\hline

\multicolumn{9} {l} {{\sffamily MOT 2017} \quad { \scriptsize (5316 frames, 546 targets)}}\\


\hline
-- baseline& 
41.4 \% &
88.0 \% &
35.53 \% &
16.48 \% &
4877 &
60164 &
779 &
741 \\

-- with Abd.& 
{\bf 46.2 \%} & 
87.9 \% & 
31.32 \% &
20.7 \% &
5195 &
54421 &
800 &
904 \\



\hline

\multicolumn{9} {l} {{\sffamily MOT 2020} \quad { \scriptsize (8931 frames, 2332 targets)}}\\

%
%
%


\hline
-- baseline& 
49.5 \% &
87.1 \% &
17.79 \% &
18.19 \% &
5271 &
531529 &
36560 &
39874 \\

-- with Abd.& 
{\bf 50.7 \%} & 
87.2 \% & 
18.65 \% &
17.16 \% &
4120 &
537427 &
17658 &
38346 \\

\hlinewd{1pt}


\end{tabular}

\end{center}
\caption{{ { Evaluation of Tracking Performance}; accuracy (MOTA), precision (MOTP), mostly tracked (MT) and mostly lost (ML) tracks, false positives (FP), false negatives (FN), identity switches (ID Sw.), and fragmentation (Frag.). 
}}
\label{tbl:MOT_result}
\end{table*}
}

\subsection{Empirical Performance Analysis} 

For online sensemaking, evaluation focusses on accuracy of abduced motion tracks, real-time performance, and the tradeoff between performance and accuracy. Our evaluation uses the \textbf{{ KITTI}} \emph{object tracking dataset}  \cite{Geiger2012CVPR}, which is a community established benchmark dataset for autonomous cars: it consists of $21$ training and $29$ test scenes, and provides accurate track annotations for $8$ object classes (e.g., {\small\sffamily  car, pedestrian, van, cyclist}). We also evaluate tracking results using the more general cross-domain \textbf{ Multi-Object Tracking} (MOT) dataset \cite{MOT16-Benchmark} established as part of the \emph{MOT Challenge}; We evaluate on MOT 2017 consisting of $7$ training and $7$ test scenes which are highly unconstrained videos filmed with both static and moving cameras, and MOT 2020 consisting of $4$ training and $4$ test scenes filmed in crowded environments. We evaluate on the available groundtruth for training scenes of both KITTI using YOLOv3 detections, and MOT17 / MOT20 using the provided faster RCNN (Region Based Convolutional Neural Network \citep{DBLP:conf/nips/RenHGS15}) detections.

\medskip

\subsubsection{Evaluating Object Tracking}
For evaluating \emph{accuracy} {(MOTA)} and \emph{precision} {(MOTP)} of abduced object tracks we follow the Clear {MOT} \cite{Bernardin2008} evaluation schema. 

\begin{itemize}

\item\textit{MOTA}\quad describes the accuracy of the tracking, taking into account the number of missed objects / false negatives (FN), the number of false positives (FP), and the number of miss-matches (MM).

\item\textit{MOTP}\quad describes the precision of the tracking based on the distance of the hypothesised track to the ground truth of the object it is associated to.

\end{itemize}

\noindent These metrics are used to assess how well the generated visual explanations describe the low-level motion in the scene.

{Results} (Table \ref{tbl:MOT_result}) show that jointly abducing high-level object interactions together with low-level scene dynamics increases the accuracy of the object tracks, i.e, we consistently observe an improvement of about $5\%$ on KITTI and MOT 2017. On KITTI MOTA improves from $45.72\%$ to $50.5\%$ for \emph{cars} and $28.71 \%$ to $32.57 \%$ for \emph{pedestrians}, and on MOT 2017 it improves from $41.4\%$ to $46.2\%$. On MOT 2020 we still observe an improvement of  1.2\% from 49.5\% to 50.7\%. This relatively small improvement is mainly because of the different nature of the dataset, i.e., the focus on crowded scenes filmed from a slightly above perspective, which leads to only few targets that get fully occluded by others, and thus there are fewer corrected tracks when the using abductive sensemaking compared to the scenes in KITTI and MOT 2017.

%

%

\medskip

\subsubsection{Online Performance and Scalability}
\begin{figure}[t]
\begin{center}
\footnotesize
\center
\includegraphics[width = 0.35\columnwidth,valign=c]{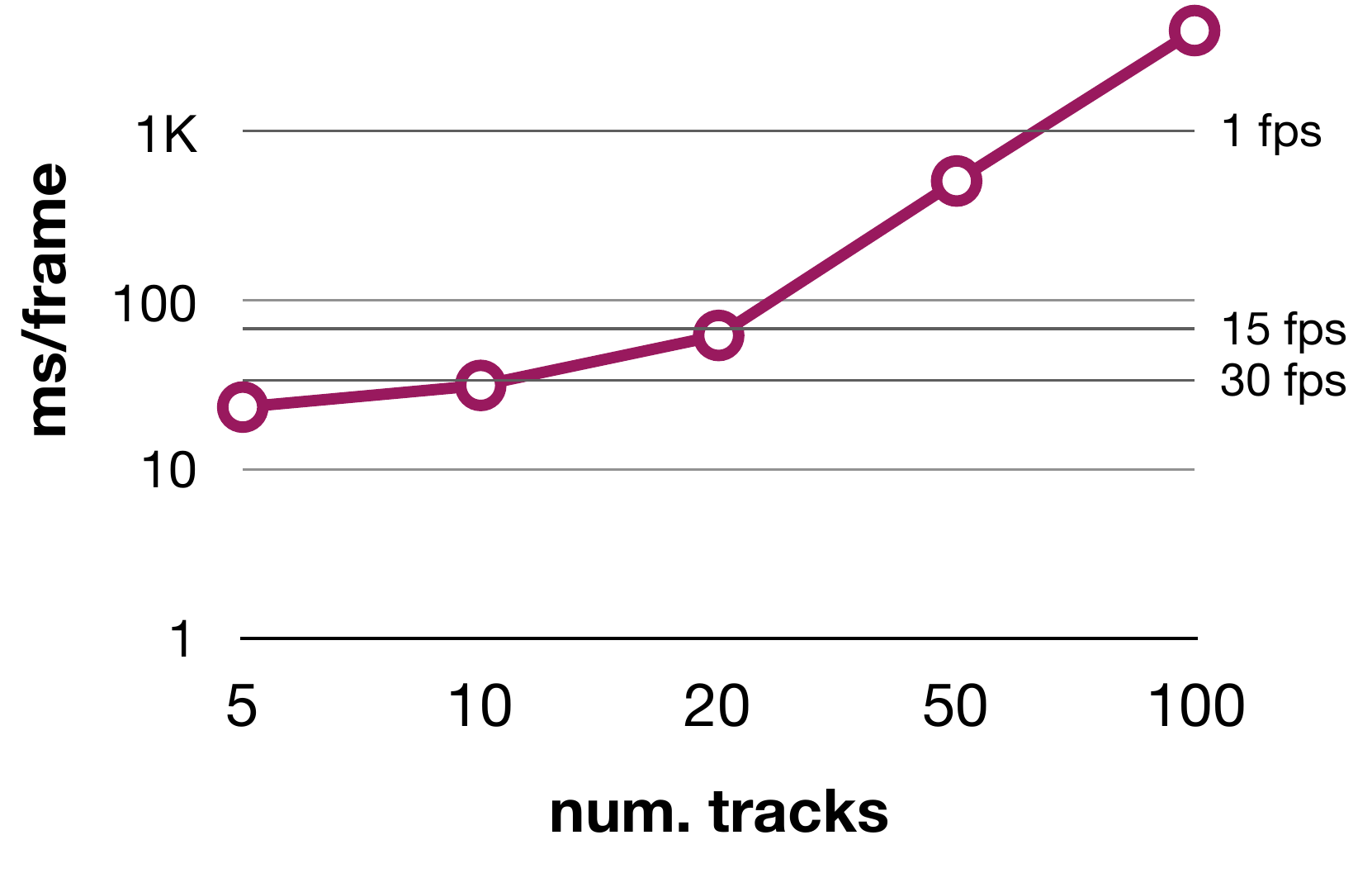} \quad\quad\quad\quad\quad\quad
{
\footnotesize
\begin{tabular}{>{\columncolor[gray]{0.92}}c r r c c c c}
\hlinewd{1pt}
{\sffamily\color{blue!70!black}\textbf{No. of TRACKS}} & $ms/frame$ & $fps$ \\\hline
{5}& 23.33 & 42.86 \\
{10}& 31.36 & 31.89  \\
{20}& 62.08 & 16.11 \\
{50}& 511.83 & 1.95 \\
{100}& 3996.38 & 0.25 \\
\hlinewd{1pt}
\end{tabular}
}

\end{center}
\caption{{Scalability}; processing time relative to the no. of tracks on synthetic dataset.}
\label{fig:real-time_result2}
\end{figure}

Performance of online abduction is evaluated with respect to its real-time capabilities.\footnote{Evaluation using a dedicated Intel Core i7-6850K 3.6GHz 6-Core Processor, 64GB RAM, and a NVIDIA Titan V GPU 12GB.}
{\bf (1).} We compare the time \& accuracy of online abduction for state of the art (real-time) detection methods: {YOLOv3}, 
{ SSD} \cite{Liu16-ssd}, and { Faster RCNN} \cite{DBLP:conf/nips/RenHGS15} (Fig. \ref{fig:real-time_result}). 
{\bf (2).} We evaluate scalability of the ASP based abduction on a synthetic dataset with controlled number of tracks and percentage of overlapping tracks per frame. Results  (Fig. \ref{fig:real-time_result2}) show that online abduction can perform with above $30$ frames per second for scenes with up to $10$ highly overlapping object tracks, and more than 50 tracks with $1$fps (for the sake of testing, it is worth noting that even for $100$ objects per frame it only takes about an average of $4$ secs per frame). Importantly, for realistic scenes such as in the {KITTI} dataset, abduction runs realtime at $33.9$fps using {YOLOv3}, and $46.7$ using {SSD} with lower accuracy but providing good precision.

\begin{figure}[t]
\begin{center}
\footnotesize
\begin{tabular}{>{\columncolor[gray]{0.92}}c c c c c c c }
\hlinewd{1pt}
{\sffamily\color{blue!70!black}\textbf{DETECTOR}} & Recall & MOTA & MOTP & $fps_{det}$ & $fps_{abd}$ \\\hline
{\sffamily   YOLOv3}&
0.690 &
50.5 \%&
74.76 \% &
45 &
33.9 \\
{\sffamily  SSD}&
0.599 &
30.63 \%&
77.4 \% &
8 &
46.7 \\
{\sffamily  FRCNN}&
0.624 &
37.96 \%&
72.9 \% &
5 &
32.0 \\
\hlinewd{1pt}

\end{tabular}

\end{center}
\caption{{Online Performance}; performance for pretrained detectors (DET.) on the \emph{'cars'} class of KITTI dataset}
\label{fig:real-time_result}
\end{figure}

\subsubsection{Discussion of Empirical Results}  Results show that integrating high-level abduction and object tracking improves the resulting object tracks and reduce the noise in the visual observations. For the case of online visual sense-making, {ASP} based abduction provides the required performance: even though the complexity of {ASP} based abduction increases quickly, with large numbers of tracked objects the framework can track up to $20$ objects simultaneously with $30 fps$ and achieve real-time performance on the {KITTI} benchmark dataset. It is also important to note that the tracking approach in this paper is based on \emph{tracking by detection} using a naive measure, i.e, the IoU (Sec. \ref{sec:tracking-as-abbduction}; Step 1), to associate observations and tracks, and it is not using any visual information in the prediction or association step. Naturally, this results in a lower accuracy, in particular when used with noisy detections and when tracking fast moving objects in a benchmark dataset such as {KITTI}. That said, due to the modularity of the implemented framework, extensions with different methods for predicting motion (e.g., using particle filters or optical flow based prediction) are straightforward: i.e., improving tracking is not the aim of our research. 


\section{\uppercase{Discussion and Related Work}}\label{sec:related-work}

Answer Set Programming is now widely used as a foundational declarative language and robust methodology for a range of (non-monotonic) knowledge representation and reasoning tasks \citep{Brewka:2011:ASP,ASP-SI-2018,Gebser2012-ASP,Clingo1-2011,Gebser2014-Clingo}. With {ASP} as a foundation, and driven by semantics, commonsense and explainability \cite{DBLP:journals/cacm/DavisM15,DBLP:journals/jair/Davis17}, this research aims to bridge the gap between high-level formalisms for logical visual reasoning (e.g., by abduction) and low-level visual processing by tightly integrating semantic abstractions of \emph{space and change} with their underlying numerical representations. More broadly, this goal is pursued within the larger agenda of \emph{cognitive vision and perception} \citep{BhattECAI2020}, which is an emerging line of research bringing together a novel \& unique combination of methodologies from Artificial Intelligence, Vision and Machine Learning, Cognitive Science and Psychology, Visual Perception, and Spatial Cognition and Computation. Research in cognitive vision and perception addresses visual, visuospatial and visuo-locomotive perception and interaction from the viewpoints of language, logic, spatial cognition and artificial intelligence \citep{out-of-sight-2019,DBLP:conf/ijcai/SuchanB16,DBLP:conf/aaai/SuchanBWS18,DBLP:conf/ilp/SuchanBS16,CogSys-Symmetry-2018}. In this broader context, the principal motivation and developmental goal of this research follows a one-point agenda, namely:

	\begin{quote}
	to develop a systematic, general, and modular integration of (methods in) Computer Vision and AI, particularly emphasising the integration of high-level knowledge representation and reasoning techniques  with low-level (i.e., quantitatively) based visual computing techniques (which in the present scientific status quo are primarily driven by end-to-end, black-box deep learning pipelines).
	\end{quote}

The integration of Vision and AI addressed in our research is motivated by the need to realise human-centred criteria pertinent to the design and implementation of high-level visual sensemaking technology, e.g.,  within autonomous driving systems where such criteria emanating from standardisation and regulation considerations are of utmost priority. Although this paper selectively focusses on the needs and challenges of active / online sensemaking in autonomous driving, the generality and modularly of the developed framework ensures foundational applicability in diverse applied contexts requiring perception, interaction and control; e.g., a case in point here being the fact that the demonstrated application and evaluation also directly function with general datasets such as MOT concerned with moving objects (Sec \ref{sec:application-sec}). Of at least equal importance are the modularity and elaboration tolerance of the framework, enabling seamless integration and experimention with advances in fast evolving computer vision methods, as well as experimenting with different forms of formal methods for \emph{reasoning about space, actions, and change} \citep{Bhatt:RSAC:2012,Bhatt08-SCC-Dynamics} that could either be embedded directly within answer set programming, or possibly be utilised independently as part of other declarative frameworks for knowledge representation and reasoning.

\textbf{Perception and Abduction: A KR Perspective}.\quad Within KR, the significance of  high-level (abductive) explanations in a range of contexts is long established:  planning \& process recognition \cite{Kautz1986,Kautz1991}, vision \& abduction \cite{Shanahan05}, probabilistic abduction \cite{Blythe11}, reasoning about spatio-temporal dynamics \cite{Bhatt08-SCC-Dynamics}, reasoning about continuous \emph{spacetime} change \cite{muller1998,Hazarika2002} etc. \citet{Dubba15} formalises abductive reasoning in an inductive-abductive loop within inductive logic programming {(ILP)}. \citet{aditya2015visual} formalise general rules for image interpretation with {ASP}. Closely related to this research is \citep{Suchan2018_visual_explanation}, which uses a two-step approach (with one huge \emph{problem specification}), first tracking and then explaining (and fixing) tracking errors; such an approach is not runtime / realtime capable. Within computer vision research there has recently been an interest to synergise with cognitively motivated methods; in particular, e.g., for perceptual grounding and inference \citep{compositional-grounding-jair2015}, and combining video analysis with textual information for understanding events and answering queries about video data \citep{DBLP:journals/ieeemm/TuMLCZ14}.


\medskip

\textbf{Perception in Autonomous Driving}.\quad The present industrial relevance and market potential\footnote{\textbf{Industrial initiatives in autonomous driving}.\quad Autonomous driving research within industry is now well established: there exist cab-sharing companies like Uber and Lyft attempting to replace human drivers with ``fully autonomous self-driving'' vehicles. Companies such as Baidu, Comma AI and organisations like Udacity are creating an open source platform for various technologies of the self driving stack. Manufacturing giants such as GM, Toyota, Ford, Daimler, Bosch are also taking steps to offer varying levels of autonomy to consumer and industrial vehicles either directly or indirectly; GM acquired Cruise Automation while Toyota has invested in and collaborates with pony.ai. Ford and Volkswagen has partnered with Argo AI to bring self driving capabilities to their respective vehicles. Waymo, a subsidiary of Alphabet has already deployed autonomous ride-sharing operations in two cities. Last, but not the least, is Tesla with its competitive advantage of already having close to 500000 cars on the road collecting data with (Level 2 assistance) AutoPilot enabled.}
 of autonomous driving technology can be primarily attributed to recent advances in deep learning driven computer vision. A typical engineering stack for autonomous driving consists of perception, prediction, planning and control modules \citep{Zeng_2019_CVPR}: perception gives the location, pose of the objects in the world while prediction forecasts the motion of the objects; planning involves creating a trajectory for the motion of the vehicle which is then executed by the controller. In object detection, \citet{pmlr-v97-tan19a} introduced EfficientDet which achieves order-of-magnitude better efficiency than previous works \citep{Redmon2018_YOLOv3,DBLP:conf/nips/RenHGS15,Liu16-ssd,pmlr-v97-tan19a} without any drop in performance.  Large datasets and self-supervised methods \citep{Chen_2019_ICCV,Zhou2017UnsupervisedLO} enable end to end joint learning of flow, depth and camera pose estimation more accurately, exploiting the inherent relation between each other.  More specialised research on object detection has investigated specific cases relevant in driving, such as detection of smaller objects \cite{7780603}, partially occluded pedestrians \cite{pang2019maskguided} and 3D object detection \cite{wang2019frustum,lehner2019patch,zhu2019classbalanced}. Recent advancements in object tracking involves neural methods like \cite{tracktor_2019_ICCV} which employ a tracking by detection paradigm and predicts the next object position using a simple neural network. Multi-object tracking is also extended to multi-object tracking and segmentation by \cite{Voigtlaender2019CVPR}.   Semantic and Instance segementation of the object \cite{Cordts2016Cityscapes} \cite{ImprovingSemanticSeg2019-cvpr} \cite{yuan2019objectcontextual} \cite{takikawa2019gatedscnn} provides accurate boundaries. Advancements in robust lane detection \cite{hou2019learning} make it possible to extend automatic lane keeping and lane switching. Recent neural methods estimate visual odometry, ego motion, depth and flow through a set of multi-task learning methodologies. \cite{Zhou2017UnsupervisedLO} \cite{Mahjourian_2018} show that depth and ego-motion can be learnt in a joint manner. Flow and depth are also learnt using a multi-task approach \cite{Chen_2019_ICCV} \cite{Zou_2018} \cite{Wang_2019_CVPR}.
 
\smallskip

\textbf{Hybrid Methods to Meet Multi-Faceted Challenges}.\quad Critical challenges in driving, e.g., pertaining to perception, prediction, planning and control modules \citep{Zeng_2019_CVPR}, are researched and developed individually which leads to a sub-optimal overall performance. End-to-end driving methodologies \citep{DBLP:journals/corr/BojarskiTDFFGJM16,Zeng_2019_CVPR} are constructed in such a way that the sensor outputs such as images, LiDAR (Light Detection and Ranging) are directly used to predict control signals like steering and acceleration. Furthermore, these methods are generally black-box and are unable to model the complex multi-faceted nature of autonomous driving encompassing dimensions of human factors and usability, (natural) roadside multimodal interaction \citep{KondyliSTAIRS20,Kondyli-DHM-2020} etc, or support the range of human-centred AI considerations related to declarative explainability, queryability etc that have been the principal impulse underlying the aims of the methods developed in this paper.

Our research achieves a systematic integration of KR and Vision methods hitherto developed, evaluated, and applied in completion isolation of one another; we believe that our resulting framework can serve as a one possible interpretation and exemplar for the neurosymbolic integration of relational AI and neural (visual) feature detectors. Furthermore, it offers a novel potential for a multifaceted but integrated applied evaluation and benchmarking of visual sensemaking technologies: e.g., it is common practice within computer vision research to evaluate and benchmark visual computing capabilities, e.g., for object detection, tracking, using absolute performance benchmarks either solely or primarily centred on incremental improvements in accuracy. Naturally, this is necessary for fundamental progress in vision research, but such an evaluation metric misses out on other crucial requirements as they pertain to human-centred AI considerations in applications domains such as autonomous driving. For instance, in light of ethically driven standardisation and regulatory considerations (Section \ref{sec:motivation}), this research has been motivated and directly addresses interpretability and explainability challenges {\small(\textbf{C1 -- C4})}:

\begin{itemize}

	\item[{\footnotesize\textbf{\color{blue!80!black}C1}}.] \emph{Active visual sensemaking}, e.g., involving (real-time) commonsense visuospatial abduction and (simulated) prediction of grounded percepts
	
	\item[{\footnotesize\textbf{\color{blue!80!black}C2}}.] \emph{Posthoc analysis of quantitative archives},  e.g., requiring semantic search / retrieval / visualisation for diagnosis, dispute settlement, inspection

	\item[{\footnotesize\textbf{\color{blue!80!black}C3}}.] \emph{Natural human-machine interaction}, e.g., involving natural language interfaces for (explanatory) communication between vehicle and passengers (or other stakeholders)

	\item[{\footnotesize\textbf{\color{blue!80!black}C4}}.] \emph{Standardisation} for  vehicular licensing \& validation, e.g., involving creation of diverse, naturalistic datasets usable in testing of autonomous vehicle performance; how to access the quality and distribution of training datasets utilised? (Sec \ref{sec:summar-and-outlook})

\end{itemize}

Our research, by its integrative approach, makes it possible to explicitly address ``human-centred interpretability and explainability challenges'' such as in (C1--C4)  for autonomous driving systems at the practical level of methods and tools. This is especially beneficial and timely since not everything in autonomous vehicles is about realtime control / decision-making; several human-machine interaction requirements (e.g., for interpretable diagnostic communication, universal design) also exist. The Federal Ministry of Transport and Digital Infrastructure in Germany (BMVI) has taken a lead in eliciting $20$ key propositions (with possible legal implications) for the fulfilment of ethical commitments for automated and connected driving systems \cite{ethicalGermany2018}. The BMVI report highlights a range of factors pertaining to safety, utilitarian considerations, human rights, statutory liability, technological transparency, data management and privacy etc. We claim that what appears as spectrum of complex challenges (in autonomous driving) that may possibly delay technology adoption is actually rooted to one fundamental methodological consideration that needs to be prioritised, namely: the design and implementation of human-centred technology based on a \emph{confluence} of techniques and perspectives from AI+ML, Cognitive Science \& Psychology, Human-Machine Interaction, and Design Science. Like in many applications of AI, such an integrative approach has so far not been explored also within autonomous driving research.


\section{\uppercase{Summary and Outlook}}\label{sec:summar-and-outlook}

We have developed a novel neurosymbolic abduction-driven \emph{online} (i.e., realtime, incremental) visual sensemaking framework: general, systematically formalised, modular, and fully implemented. Integrating robust state-of-the-art methods in \emph{knowledge representation} and \emph{computer vision}, the framework has been evaluated and demonstrated with established community benchmarks. We highlight application prospects of semantic vision for autonomous driving, a domain of emerging and long-term significance for research in Artificial Intelligence and Machine Learning. From the applied viewpoint of autonomous driving, our work is motivated by interpretability and explainability benchmarks (e.g., in active visual sensemaking, posthoc analysis, natural human-machine interaction, standardisation for licensing \& validation; Sec \ref{sec:application-driving}) that go far beyond basic considerations in contemporary autonomous driving research, namely: \emph{how fast to drive and which way to steer}, and testing performance by \emph{clocking mileage} alone by the use of deep learning based methods in training and testing phases.

\medskip

\textbf{Technical Extensions}\quad 

Our development of a systematic, modular, and general visual sensemaking methodology opens up several possibilities for further technical developments / extensions:

\begin{itemize}

	\item \emph{Commonsense}.\quad Specialised commonsense theories about multi-sensory integration, multi-agent belief merging, incorporation of contextual knowledge and situational norms within the declarative framework of ASP merits individual strands of further development.

	\item \emph{Tracking by detection}.\quad Given the modularity of the developed framework, incorporating and experimenting with specialised / emerging low-level visual computing methods becomes feasible with relative ease. For instance, in this paper we have not attempted to develop a new tracking algorithm as such; instead, one of our aims has been to showcase the manner in which perceptual sensemaking by visual abduction can be integrated into a standard ``tracking by detection'' paradigm, which is most widely used approach in state of the art tracking (Sec \ref{sec:related-work}). Nevertheless, extensions and variations of this approach deserve further investigation where tracking itself takes a centre-stage. 
	
	\item \emph{Uncertainty}.\quad The present work handles the uncertainty involved in low-level object tracking using a naive approach, which suffices for the present purposes, i.e., a full-scale systematic formalisation of a probabilistic model has not been attempted herein. However, handling uncertainty calls for its systematic treatment, e.g., requiring either integrating a declarative probabilistic model directly within the answer set programming framework, or possibly independently as a separately module. One seemingly natural approach towards this would be to explore possibilities with probabilistic ASP \citep{probab-ASP-2015}.
	

\end{itemize}

\textbf{Towards a Dataset: Reasoning and Scenario Visuospatial Complexity Coverage}.\quad The application demonstrations of this paper have been conducted in the backdrop of select safety-critical situations (Table \ref{tbl:safety-critical-situations};  e.g., Fig. \ref{fig:safety-critical-sits}), without aiming to achieve an exhaustive collection (if at all it is even possible to be comprehensive in this respect). The scenarios and corresponding safety-criticality are exemplary, with the selections emanating from a behavioural study of human-factors in everyday driving situations, and safety criticality determined based on analysis of empirical data about roadside accidents / hazardous situations from publicly available data published in accident research reports, e.g., by the German Insurance Association (``\emph{Unfallforschung der Versicherer}'') \cite{GDV2017}. Work is presently in progress to develop novel benchmark datasets (in synergy with behavioural human studies; refer below) that centralise range and distribution vis-vis commonsense explainability and visuospatial complexity (\citep{Kondyli-DHM-2020,KondyliSTAIRS20}) criteria classes within a dataset, as opposed to merely collecting accumulating ``mileage'' / ``big data''.

\medskip

\textbf{Human-Factors in Autonomous Driving:\\A Cognitive Methodology Combining Behavioural and Computational Approaches}\quad 

In addition to continuing (aforediscussed) technical developments in computational cognitive vision pertaining to the integration of ``vision \& AI'', our ongoing focus is to develop a novel dataset emphasising (visuospatial) semantics and (commonsense) explainability. For instance, we develop a methodology ---focussing on \emph{visuospatial complexity}  \citep{KondyliSTAIRS20} of stimuli and \emph{multimodal interactions} \citep{Kondyli-DHM-2020} in ecologically valid naturalistic \citep{angrosino2007naturalistic,social-interaction-ecologica-2016} driving conditions--- for establishing human-centred benchmarks and corresponding testing \& validation datasets for visual sensemaking primarily from a human cognitive factors viewpoint. Our particular focus  here is on embodied multimodal interactions (e.g., gestures, joint attention, visual search complexity) amongst drivers, pedestrians, cyclists etc under ecologically valid naturalistic conditions. This initiative is driven by bottom-up interdisciplinary research --encompassing AI, Psychology, HCI, and Design-- for the study of driving behaviour particularly in diverse low-speed, complex urban environments possibly with unstructured traffic. Such interdisciplinary studies --\emph{at the confluence of Cognition, AI, Interaction, and Design}-- are needed to better appreciate the complexity and spectrum of varied human-centred challenges in autonomous driving, and demonstrate the significance of integrated vision \& AI  solutions \citep{BhattECAI2020} in those contexts.


\medskip
\medskip
\medskip


\newpage
{\small\sffamily
\uppercase{\textbf{Acknowledgements}}.

\medskip
\medskip

We thank the reviewers at IJCAI 2019 \citep{out-of-sight-2019} for their constructive feedback and support of our work; all reviewer suggestions that could not be included in the conference length paper have been fully incorporated in the present article. We are also grateful to the anonymous reviewers and editors at the AIJ journal whose comments have helped us further improve the (final published) paper; we remain especially appreciative for their timely service during the special times of 2020.

We acknowledge partial funding by the German Research Foundation (DFG) via the Collaborative Research Center 1320 EASE -- ``Everyday Activity Science and Engineering'' (www.ease-crc.org) project:

\medskip

$~$\quad\quad\textbf{Spatial Reasoning in Everyday Activity}., Number 329551904.

$~$\quad\quad{\small\href{http://gepris.dfg.de/gepris/projekt/374123335?language=en}{http://gepris.dfg.de/gepris/projekt/374123335}}
}

\medskip
\medskip
\medskip
\medskip
\medskip

{\small\sffamily
\uppercase{\textbf{APPENDICES}}.\quad 
}

\medskip
\medskip

\makeBlue{
\textbf{A1}.\quad\textbf{Select Answer Set Programming Code}
}

\medskip

\makeBlue{
\textbf{A2}.\quad\textbf{Additional Examples}
}

\medskip

\makeBlue{
\textbf{A2}.\quad\textbf{Example Data}
}

\newpage

\appendix


\makeBlue{
\section{SELECT ANSWER SET PROGRAMMING CODE}\label{appA}
}

\appendixNote{We envisage that all applicable electronic material (data sets, programs, videos....) will be published as a supplement in an archival format in due course. As for this appendix, select code snippets in support of the examples in the paper are included; a full implementation will be linked and released (also including a light-weight execution environment to be determined) upon final publication of the paper independent of the proposed supplementary publication.}

\medskip
\medskip
\medskip
\medskip

\makeBlue{
\subsection{\upshape{\bf Abduction Based Association}}
}

Following the generate and test paradigm of ASP, \aspMeta{choice rules} are used to generate all assignments between detections and tracks to resulting on all possible assignments; assignments are tested using \aspMeta{integrity constraints}.

\appSkip

\noindent $\bullet~~~$Choice rules for generating assignment actions generate the set of assignments actions for all tracks and all detections; for example:

\appSkip

\footnotesize
\begin{minted}[bgcolor=blue!5!white]{prolog}
1{
    assign(Trk, Det): det(Det, _, _);
    end(Trk);
    ignore_trk(Trk);
    halt(Trk);
    resume(Trk, Det): det(Det, _, _)
}1
:- trk(Trk, Trk_Type).

1{
    assign(Trk, Det): trk(Trk, _);
    start(Det);
    ignore_det(Det);
    resume(Trk, Det): trk(Trk, _)
}1
:- det(Det, Det_Type, Conf).
\end{minted}
\normalsize

\noindent $\bullet~~~$Generated assignments are tested based on (spatio-temporal) constraints for each assignment action. Assignments not consistent with these constraints are eliminated from the set of answers using integrity constraints:

\appSkip

\footnotesize
\begin{minted}[bgcolor=blue!5!white]{prolog}
:- assign(Trk, Det), not assignment_constraints(Trk, Det).
:- start(Det), not start_constraints(Det).
:- end(Trk), not trk_state(Trk, halted).
:- ignore_trk(Trk), not trk_state(Trk, halted).
:- halt(Trk), not trk_state(Trk, active).
:- resume(Trk, Det), not resume_constraints(Trk, Det).

assignment_constraints(Trk, Det) :-
    trk(Trk, Trk_Type), det(Det, Det_Type, Conf),
    trk_state(Trk, active), match_type(Trk_Type, Det_Type),
    Conf > conf_thresh_assign,
    iou(Trk, Det, IOU), IOU > iou_thresh.

resume_constraints(Trk, Det) :-
    trk(Trk, Trk_Type),
    det(Det, Det_Type, Conf), Conf > conf_thresh_resume,
    match_type(Trk_Type, Det_Type),
    trk_state(Trk, halted).

start_constraints(Det) :-
    det(Det, _, Conf), Conf > conf_thresh_new_track,
    size(bigger, Det, size_threshold).
\end{minted}
\normalsize

This results in the set of all possible assignments, which further gets optimized based on \aspMeta{optimization statements} in \ref{sec:app:opt}.

\medskip
\medskip

\makeBlue{
\subsection{\upshape{\bf Abducible High-Level Events}}
}

Event hypotheses with respect to background fluents and events are generated to explain assignment actions. 

\appSkip


\noindent $\bullet~~~$Functional fluent $visibility$ of a track can be $fully\_visible,  partially\_visible$, or $not\_visible$.

\appSkip

\footnotesize
\begin{minted}[bgcolor=blue!5!white]{prolog}

fluent(visibility(Trk)) :- trk(Trk, _).

possVal(visibility(Trk), fully_visible) :- trk(Trk, _).
possVal(visibility(Trk), partially_visible) :- trk(Trk,_).
possVal(visibility(Trk), not_visible) :- trk(Trk, _).

\end{minted}
\normalsize

\noindent $\bullet~~~$Boolean fluent $hidden\_by$ for two tracks can be $true$ or $false$.

\appSkip

\footnotesize
\begin{minted}[bgcolor=blue!5!white]{prolog}

fluent(hidden_by(Trk1, Trk2)) :- trk(Trk1,_), trk(Trk2,_).

possVal(hidden_by(Trk1, Trk2), true) :- trk(Trk1, _), trk(Trk2, _).
possVal(hidden_by(Trk1, Trk2), false) :- trk(Trk1, _), trk(Trk2, _).

\end{minted}
\normalsize

\noindent $\bullet~~~$Boolean fluent $clipped$ for a track can be $true$ or $false$.

\appSkip

\footnotesize
\begin{minted}[bgcolor=blue!5!white]{prolog}

fluent(clipped(Trk)) :- trk(Trk,_).

possVal(clipped(Trk), true) :- trk(Trk, _).
possVal(clipped(Trk), false) :- trk(Trk, _).
\end{minted}
\normalsize 

\noindent $\bullet~~~$Fluents corresponding to all tracks and pairs of tracks are initialised as follows: all tracks are initialised as fully visible, not hidden by another track, and not clipped (however, note that it can be the case that events occurring with the start of a track have an effect on initialised fluent values, e.g., an event for a track starting partially occluded).

\appSkip

\footnotesize
\begin{minted}[bgcolor=blue!5!white]{prolog}
holds_at(clipped(Trk), false, mintime) :- trk(Trk, _).
holds_at(hidden_by(Trk1, Trk2), false, mintime) :- trk(Trk1, _), trk(Trk2, _).
holds_at(visibility(Trk), fully_visible, mintime) :- trk(Trk, _).
\end{minted}
\normalsize

\noindent $\bullet~~~$Events and causal effects are defined to describe changes in the fluents as effects of events occurring in the world. Here we show examples for the events \emph{hides\_behind} and \emph{missing\_detections}.

The event \emph{hides\_behind} is defined on two tracks as follows:

\appSkip
\footnotesize
\begin{minted}[bgcolor=blue!5!white]{prolog}
event(hides_behind(Trk1,Trk2)) :- trk(Trk1,_),trk(Trk2,_).

\end{minted}
\normalsize 

One object hiding behind another object causes the visibility fluent for the hidden object to change its value to $not\_visible$. Further the fluent $hidden\_by$ for the two tracks changes its value to $true$.  
\appSkip

\footnotesize
\begin{minted}[bgcolor=blue!5!white]{prolog}
causesValue(hides_behind(Trk1, Trk2), visibility(Trk1), not_visible, T) :-
trk(Trk1,_), trk(Trk2,_), time(T).

causesValue(hides_behind(Trk1, Trk2), hidden_by(Trk1, Trk2), true, T) :-
trk(Trk1,_), trk(Trk2,_), time(T).
\end{minted}
\normalsize 

The event $missing\_detections$ is defined on a single track as follows.

\footnotesize
\begin{minted}[bgcolor=blue!5!white]{prolog}
event(missing_detections(Trk)) :- trk(Trk,_).

causesValue(missing_detections(Trk), clipped(Trk), true, T) :-
trk(Trk,_), time(T).
\end{minted}
\normalsize 

\medskip
\medskip

\makeBlue{
\subsection{\upshape{\bf Abducing High-Level Events Explaining Assignments}}
}

Possible explanations are generated using choice rules for explaining association actions, i.e., for each association a possible explanation in terms of high-level events is generated based on preconditions and causal effects. Here we show examples for the events $hides\_behind$ and $missing\_detections$.

\noindent $\bullet~~~$Choice rule (snippet) for explaining halted tracks:

A track can be halted because it is hiding behind another track, or there are missing detections within the track.

\appSkip

\footnotesize
\begin{minted}[bgcolor=blue!5!white]{prolog}
1{ 
    occurs_at(hides_behind(Trk, Trk2), curr_time): trk(Trk2,_);
    occurs_at(missing_detections(Trk), curr_time)
}1
:- halt(Trk).
\end{minted}
\normalsize

\noindent $\bullet~~~$Constraints for events are defined using integrity constraints for each event:

Integrity constraint for event \emph{hides\_behind} can not occur if \emph{poss(hides\_behind(\_ , \_))} is not true.

\appSkip

\footnotesize
\begin{minted}[bgcolor=blue!5!white]{prolog}
:- occurs_at(hides_behind(Trk1, Trk2), curr_time), not poss(hides_behind(Trk1, Trk2)).
\end{minted}
\normalsize 

\noindent $\bullet~~~$The event $hides\_behind$ is possible if the tracks are overlapping and both tracks visible.
 
\footnotesize
\begin{minted}[bgcolor=blue!5!white]{prolog}
poss(hides_behind(Trk1, Trk2)) :-
    trk(Trk1, _), trk(Trk2, _),
    position(overlapping_top, Trk1, Trk2),
    not holds_at(visibility(Trk1), not_visible, curr_time),
    not holds_at(visibility(Trk2), not_visible, curr_time).
\end{minted}
\normalsize

\noindent $\bullet~~~$Integrity constraint for event $missing\_detections$.

\appSkip

\footnotesize
\begin{minted}[bgcolor=blue!5!white]{prolog}
:- occurs_at(missing_detections(Trk), curr_time), not poss(missing_detections(Trk)).
\end{minted}
\normalsize 

\noindent $\bullet~~~$The event $missing\_detections$ is possible if the track is not clipped and it is visible.

\appSkip

\footnotesize
\begin{minted}[bgcolor=blue!5!white]{prolog}
poss(missing_detections(Trk)) :-
    holds_at(clipped(Trk), false, curr_time),
    not holds_at(visibility(Trk), not_visible, curr_time).
\end{minted}
\normalsize

\medskip
\medskip

\makeBlue{
\subsection{\upshape{\bf Optimisation}}\label{sec:app:opt}
}

\noindent $\bullet~~~$Finding best fitting hypothesis on assignments and high-level events is achieved using ASP optimisation statements as follows: 

\appSkip

{\small\textbf{---}}$~~$Matching likelihood is maximised to ensure matching of best fitting detections to tracks, i.e., here maximising IoU between bounding rectangles of predicted tracks and the detections:

\footnotesize
\begin{minted}[bgcolor=blue!5!white]{prolog}
matching_likelihood(Trk, Det, IOU) :- det(Det, _, _), trk(Trk, _), iou(Trk, Det, IOU).
\end{minted}
\normalsize 

\footnotesize
\begin{minted}[bgcolor=blue!5!white]{prolog}
A#maximize {(ML)@10,Trk,Det : assign(Trk, Det), matching_likelihood(Trk, Det, ML)}.
\end{minted}
\normalsize

{\small\textbf{---}}$~~$Maximising assignment of detections to tracks to avoid segmented tracks, i.e., assign detections to tracks whenever possible:

\appSkip

\footnotesize
\begin{minted}[bgcolor=blue!5!white]{prolog}

A#maximize {1@10, Trk, Det: assign(Trk, Det)}.

\end{minted}
\normalsize

{\small\textbf{---}}$~~$Resume tracks if possible; start / end tracks if resuming is not possible:

\footnotesize
\begin{minted}[bgcolor=blue!5!white]{prolog}

A#minimize {1@2, Trk, Det: resume(Trk, Det)}.

A#minimize {5@2,Trk: end(Trk)}.
A#minimize {5@2,Det: start(Det)}.

\end{minted}
\normalsize

{\small\textbf{---}}$~~$Only if no other explanation can be found, tracks and detections are ignored:

\footnotesize
\begin{minted}[bgcolor=blue!5!white]{prolog}

A#minimize {10@3,Det: ignore_det(Det)}.
A#minimize {10@3,Trk: ignore_trk(Trk)}.

\end{minted}
\normalsize

\newpage

\newpage

\makeBlue{
\section{ADDITIONAL EXAMPLES}\label{appB}
}

\medskip
\medskip

\makeBlue{
\subsection{\upshape{\bf Occlusion Example}}
}

Abduced event sequence for scene 04 from the MOT 2017 benchmark, involving people moving in a crowded environment, with various occlusions.

\begin{figure}[h]
\centering


\includegraphics[width=1.0\textwidth]{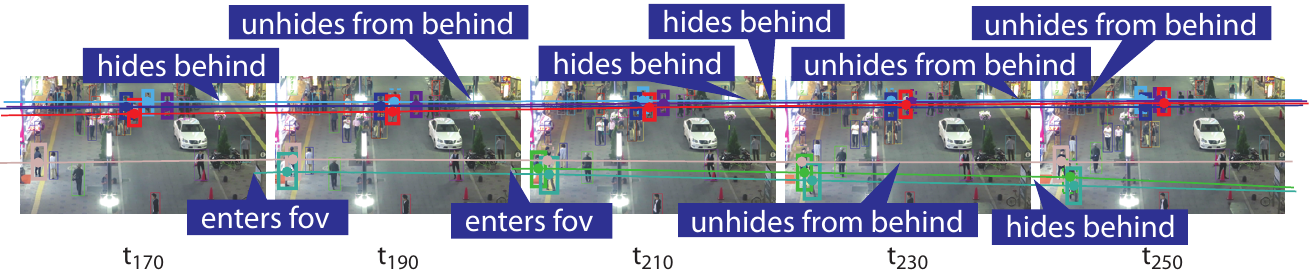}

\footnotesize
\begin{minted}[bgcolor=green!25!white]{prolog}
...
occurs_at(enters_fov(trk_30),172) 
occurs_at(hides_behind(trk_14,trk_22),188) 
occurs_at(hides_behind(trk_16,trk_30),191) 
occurs_at(enters_fov(trk_32),203) 
occurs_at(unhides_from_behind(trk_14,trk_22),205) 
occurs_at(hides_behind(trk_14,trk_20),222) 
occurs_at(hides_behind(trk_8,trk_22),230) 
occurs_at(unhides_from_behind(trk_16,trk_32),238) 
occurs_at(hides_behind(trk_32,trk_30),239) 
occurs_at(unhides_from_behind(trk_14, trk_20),245) 
occurs_at(unhides_from_behind(trk_8,trk_22),250)
...
\end{minted}
\normalsize 

\caption{{\sffamily\footnotesize Abduced events for scene MOT17-04 between time point 270 and time point 310.}}
\label{fig:app:example_mot}
\end{figure}

\newpage

\makeBlue{
\subsection{\upshape{\bf Results for Select (Complete) Scenes}}
}

\begin{figure}[t!]
\centering
\includegraphics[width=0.24\textwidth]{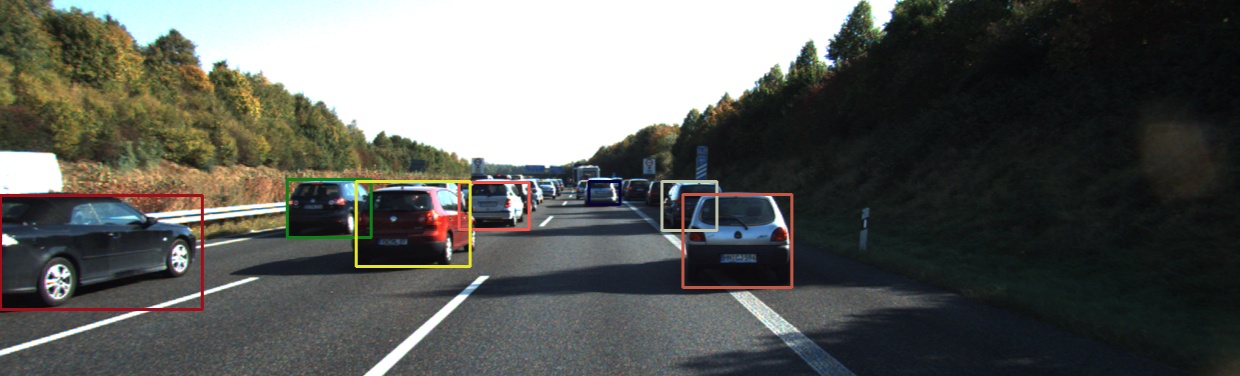}
\includegraphics[width=0.24\textwidth]{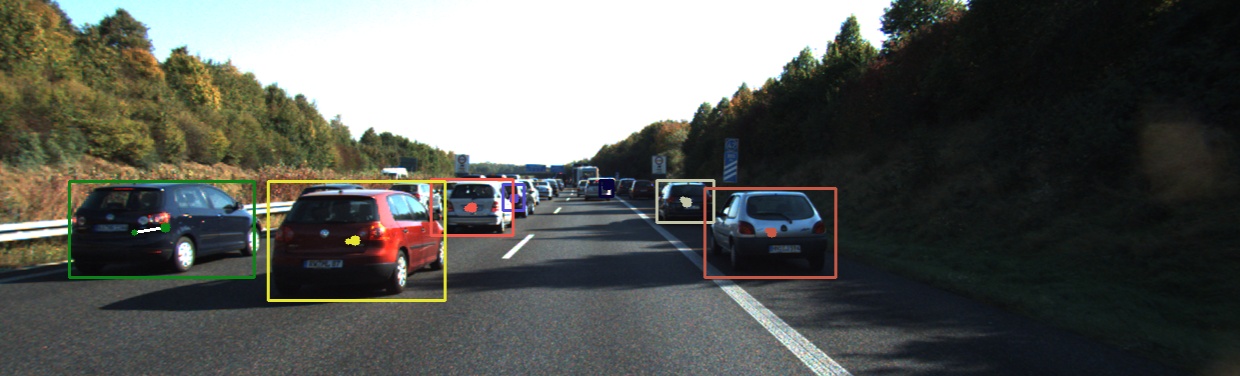}
\includegraphics[width=0.24\textwidth]{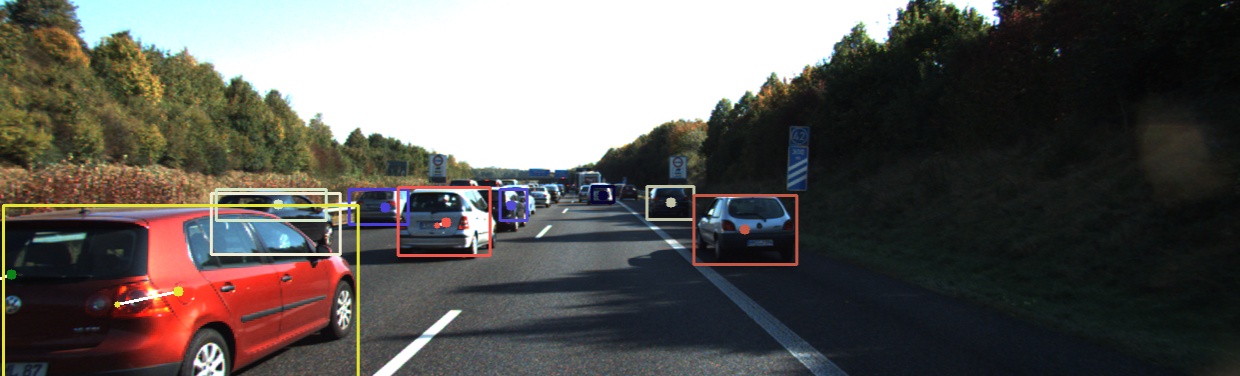}
\includegraphics[width=0.24\textwidth]{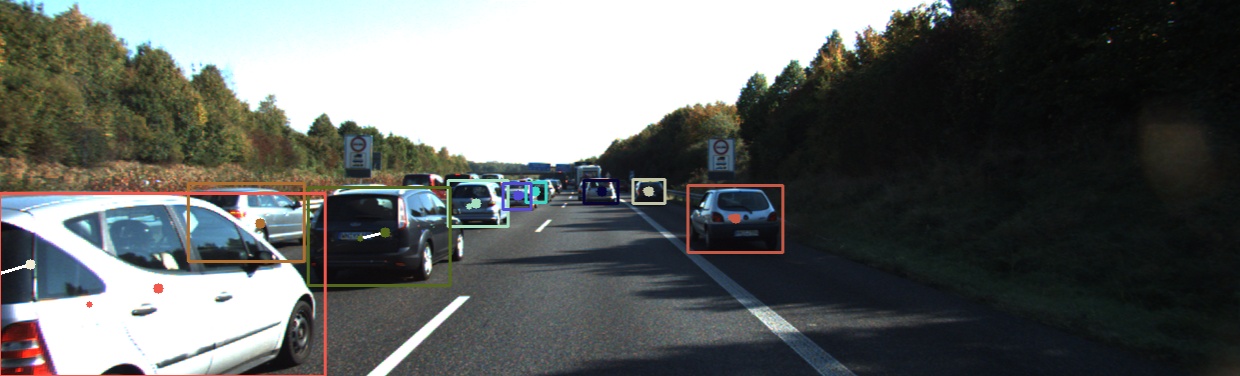}

\includegraphics[width=0.24\textwidth]{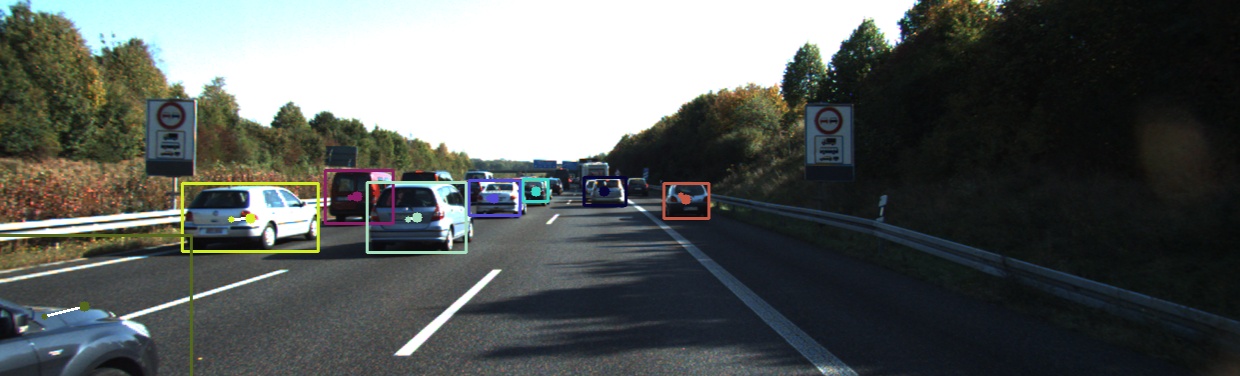}
\includegraphics[width=0.24\textwidth]{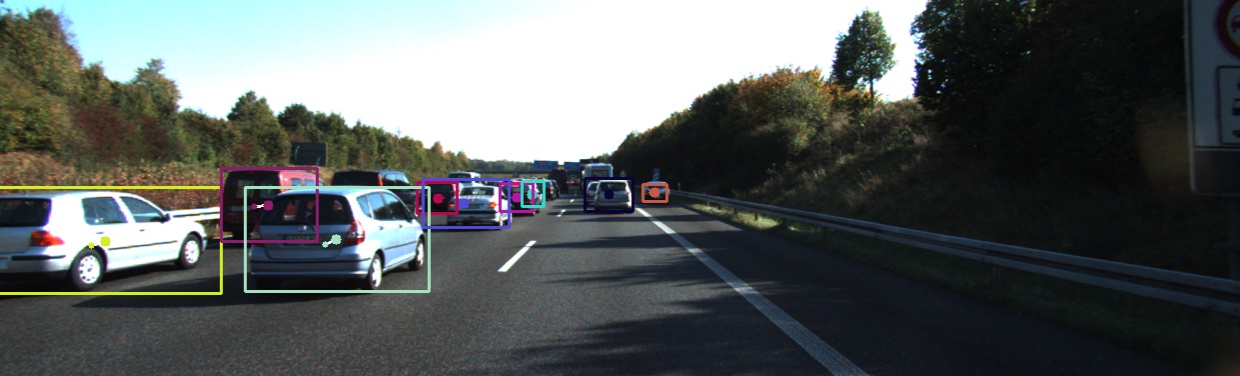}
\includegraphics[width=0.24\textwidth]{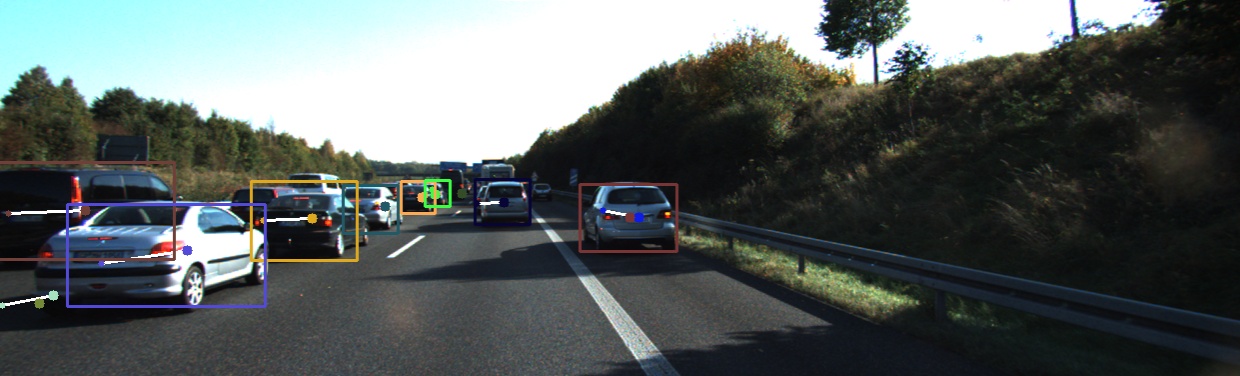}
\includegraphics[width=0.24\textwidth]{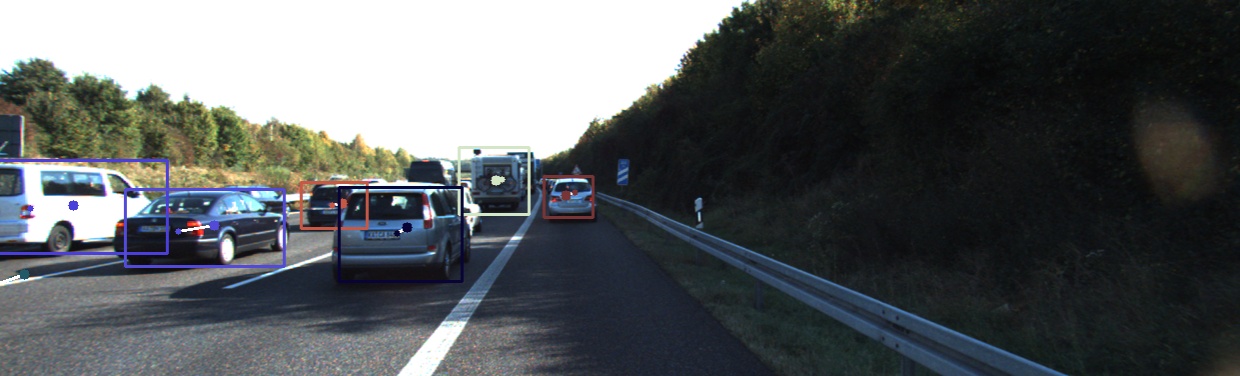}

\includegraphics[width=0.24\textwidth]{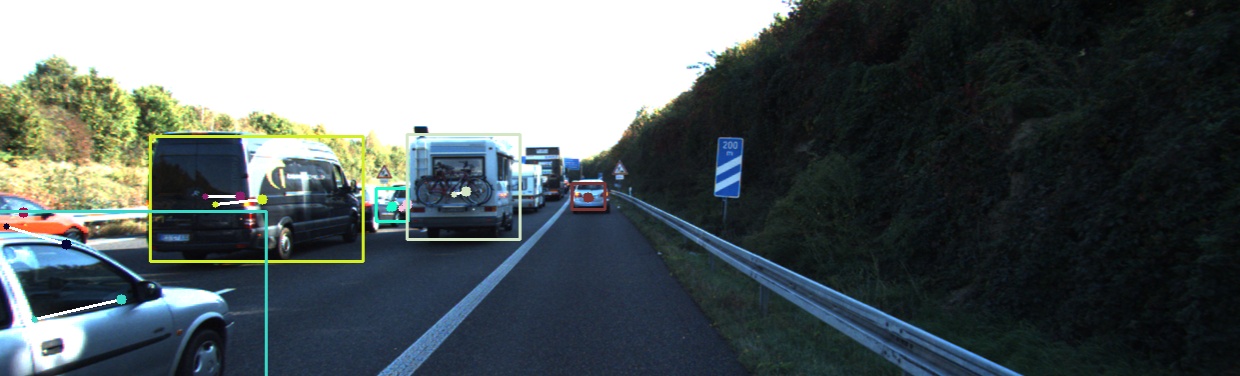}
\includegraphics[width=0.24\textwidth]{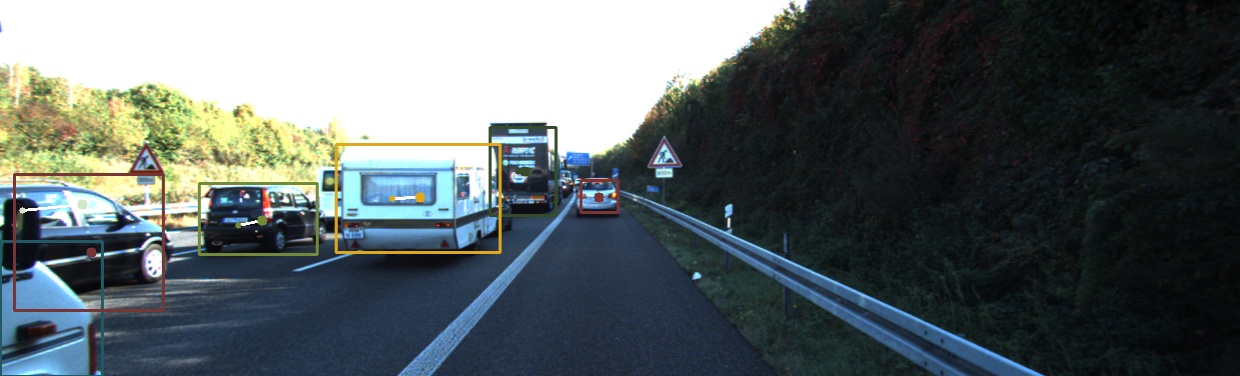}
\includegraphics[width=0.24\textwidth]{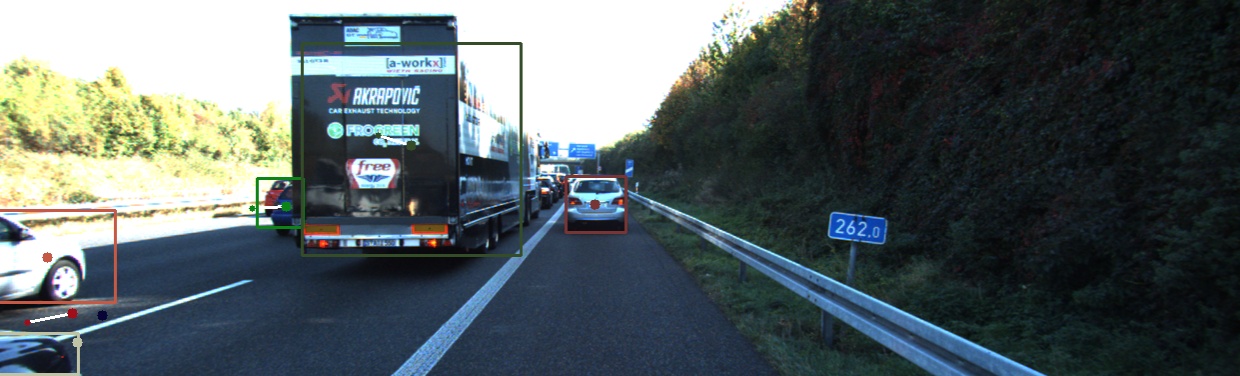}
\includegraphics[width=0.24\textwidth]{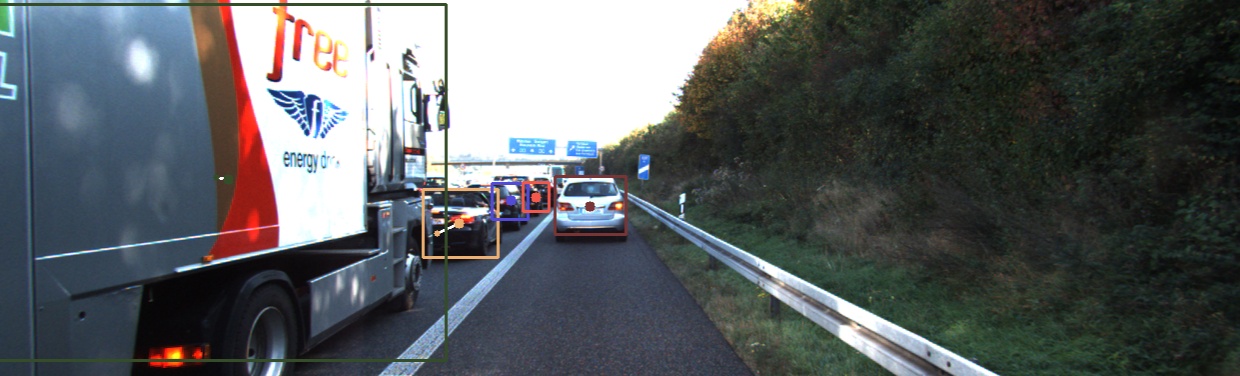}

\includegraphics[width=0.24\textwidth]{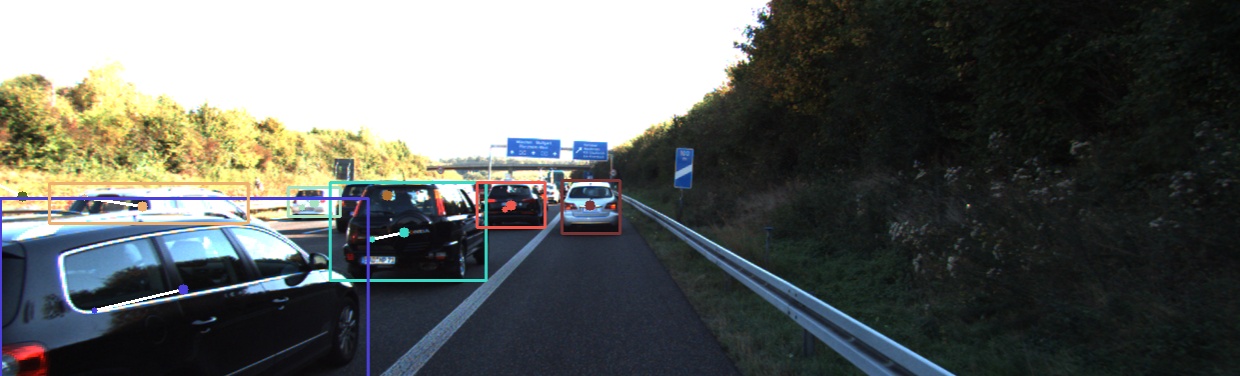}
\includegraphics[width=0.24\textwidth]{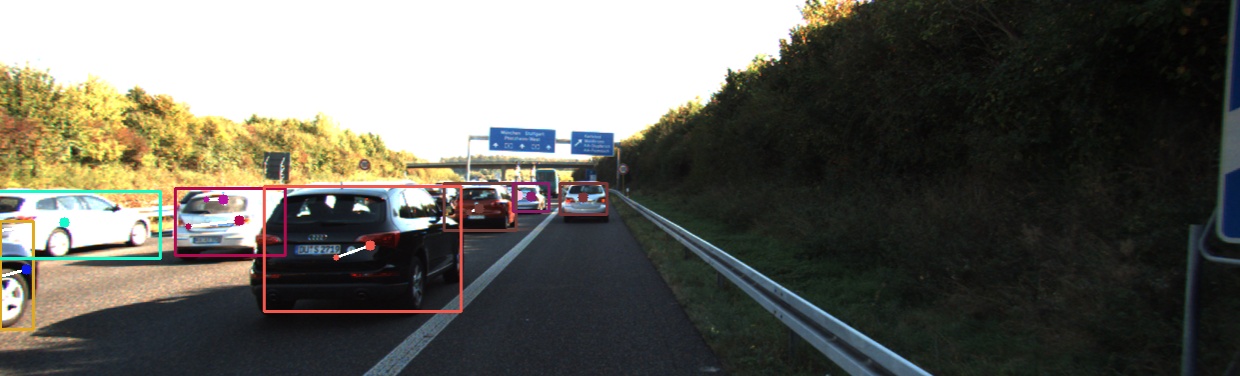}
\includegraphics[width=0.24\textwidth]{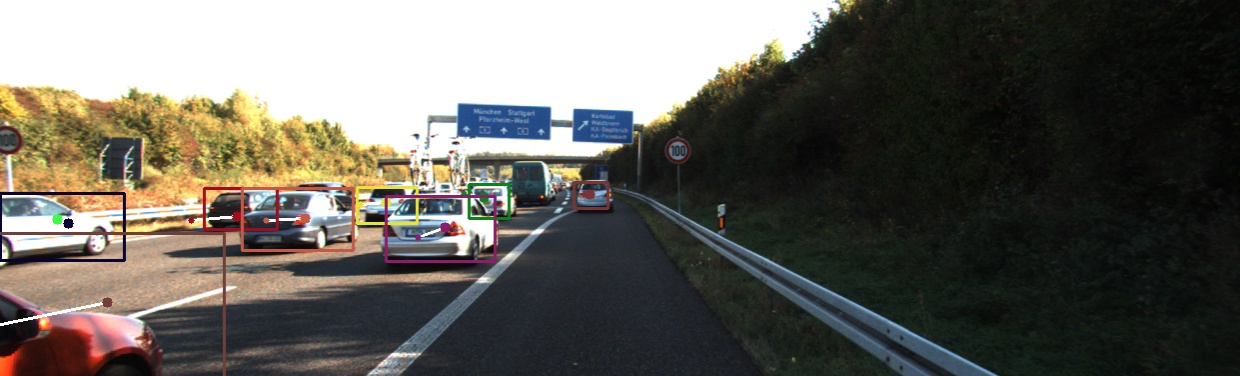}
\includegraphics[width=0.24\textwidth]{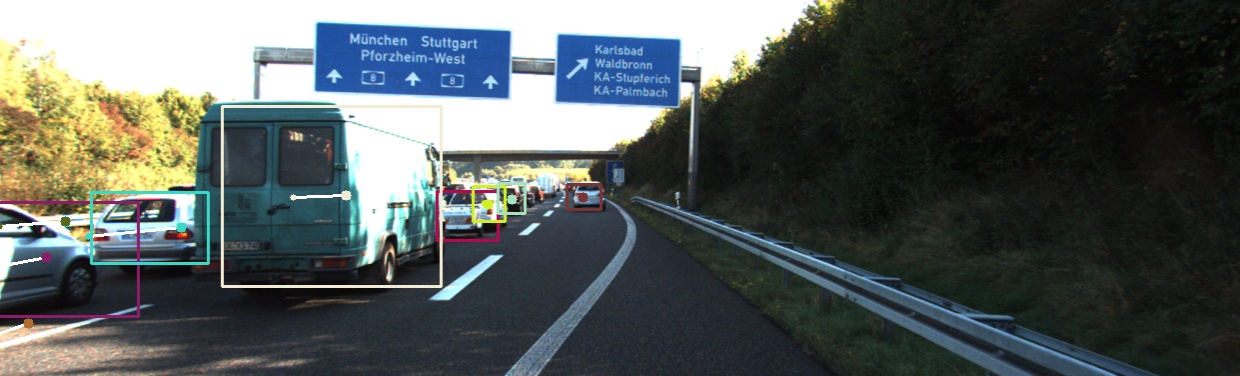}

\includegraphics[width=0.24\textwidth]{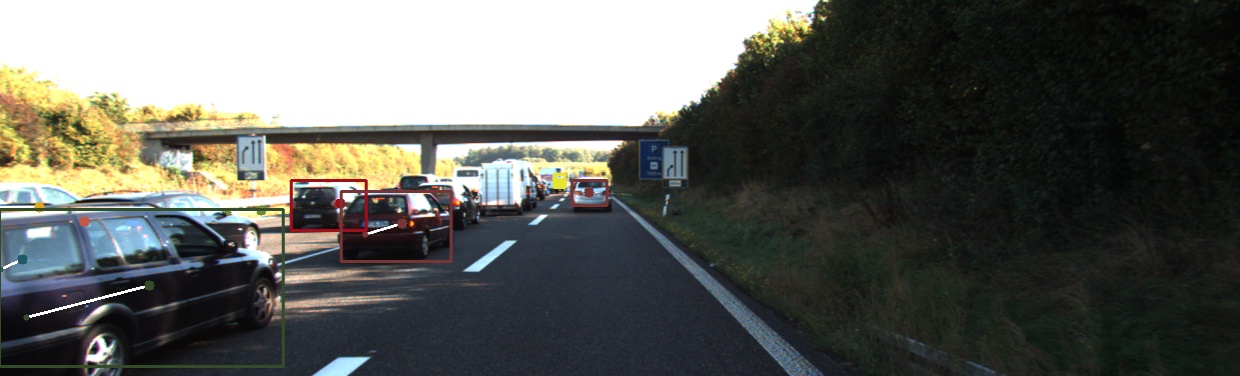}
\includegraphics[width=0.24\textwidth]{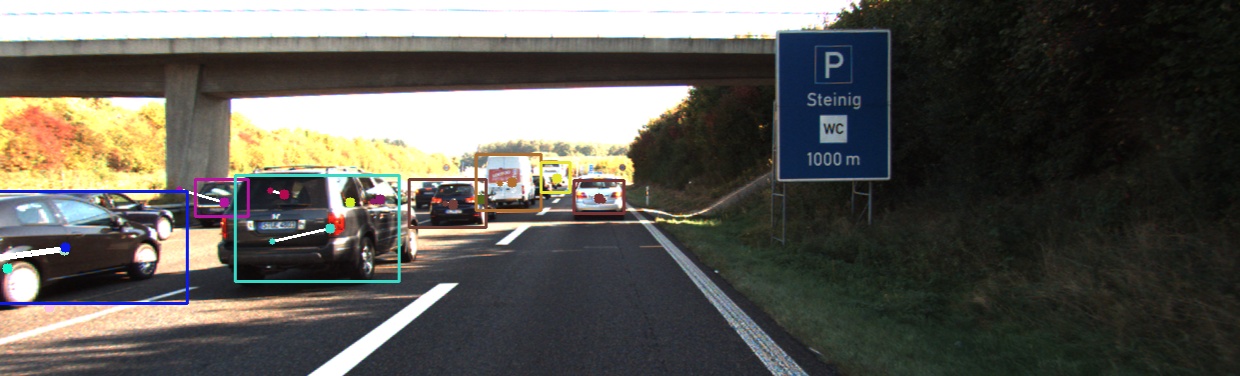}
\includegraphics[width=0.24\textwidth]{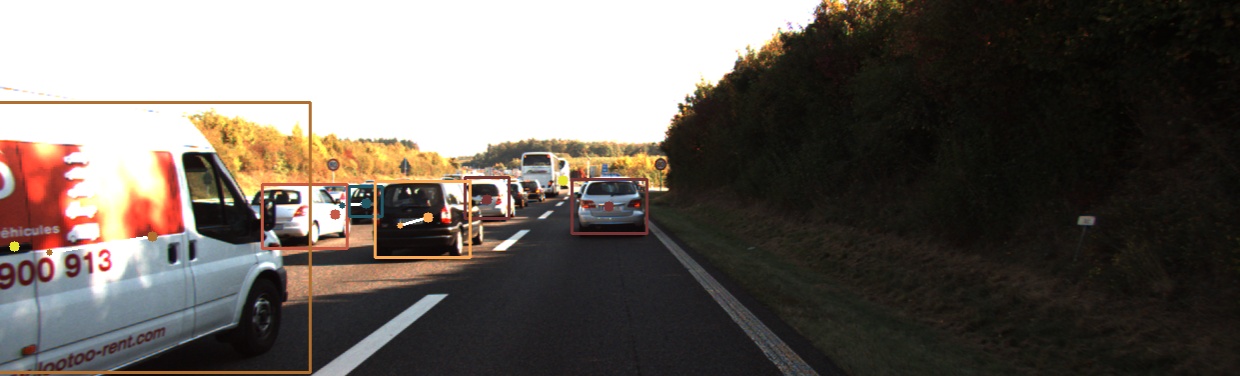}
\includegraphics[width=0.24\textwidth]{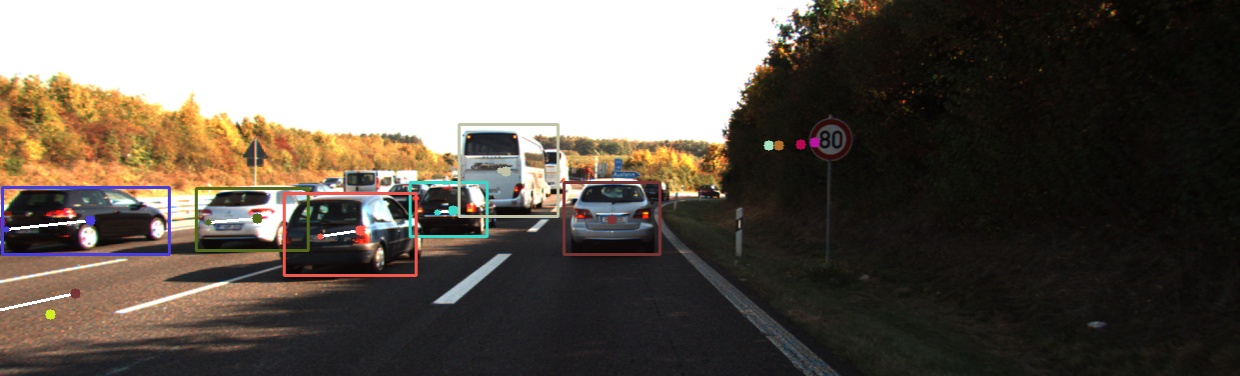}

\includegraphics[width=0.24\textwidth]{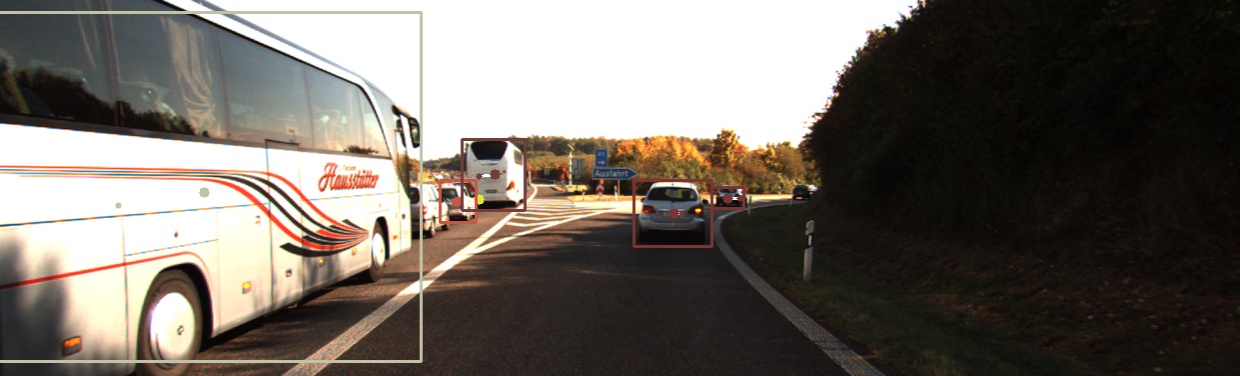}
\includegraphics[width=0.24\textwidth]{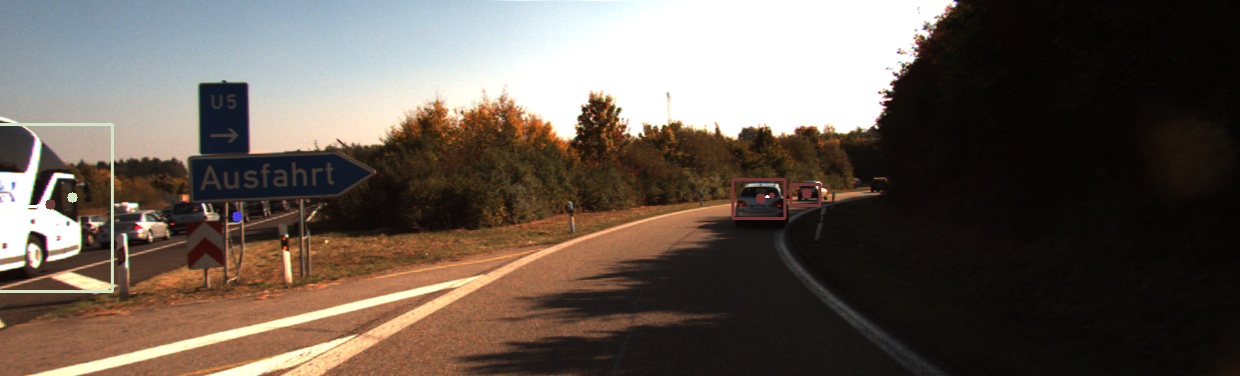}
\includegraphics[width=0.24\textwidth]{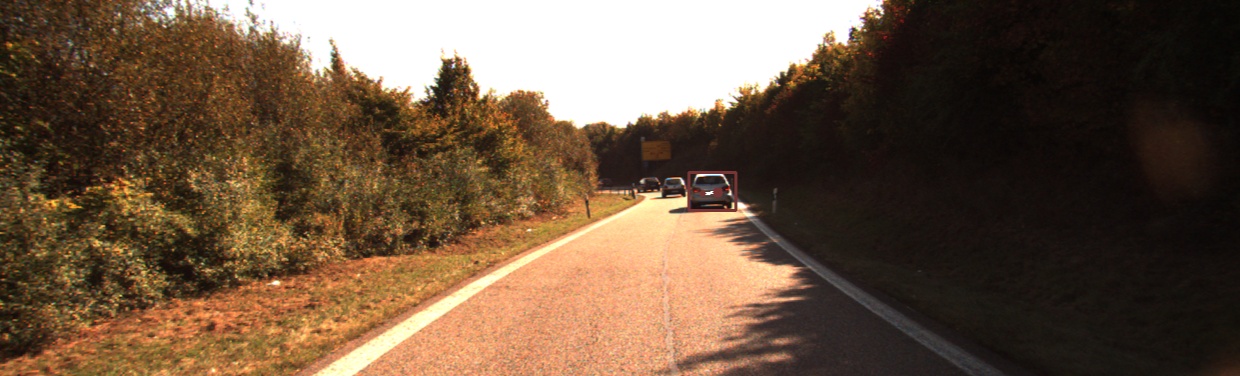}
\includegraphics[width=0.24\textwidth]{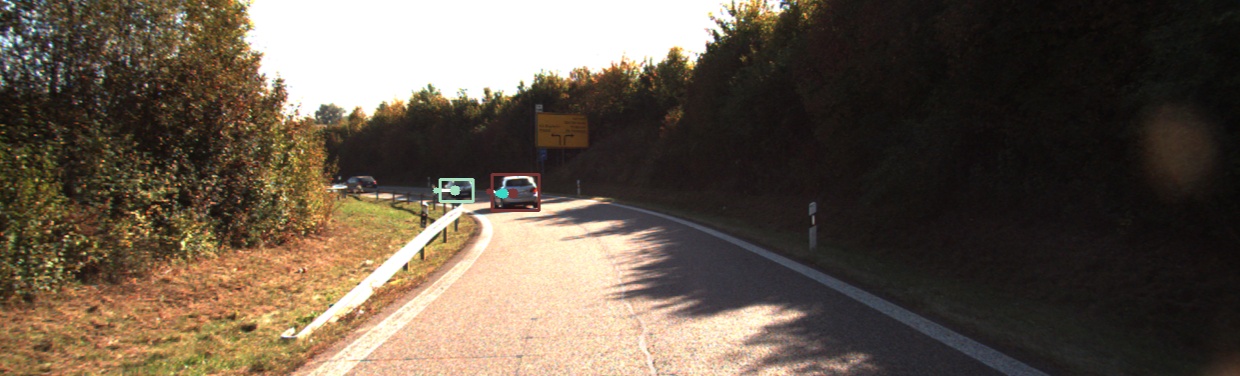}

\includegraphics[width=0.24\textwidth]{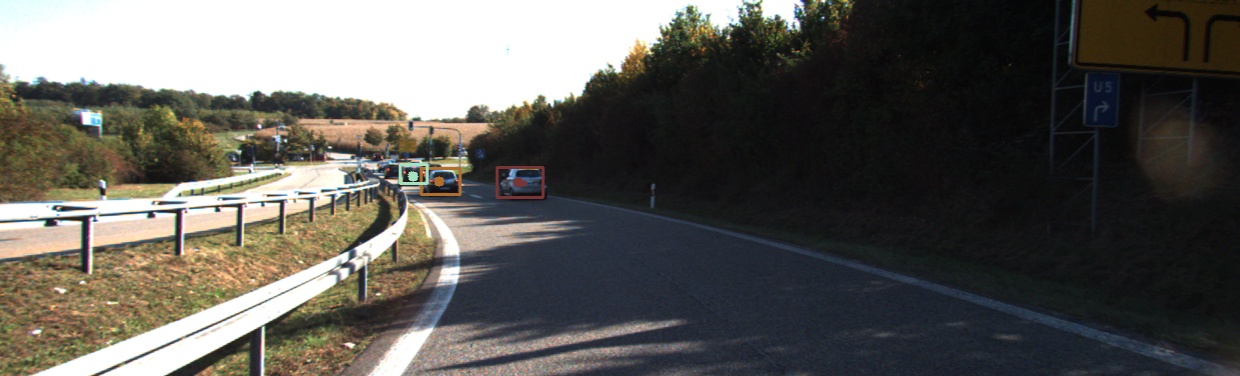}
\includegraphics[width=0.24\textwidth]{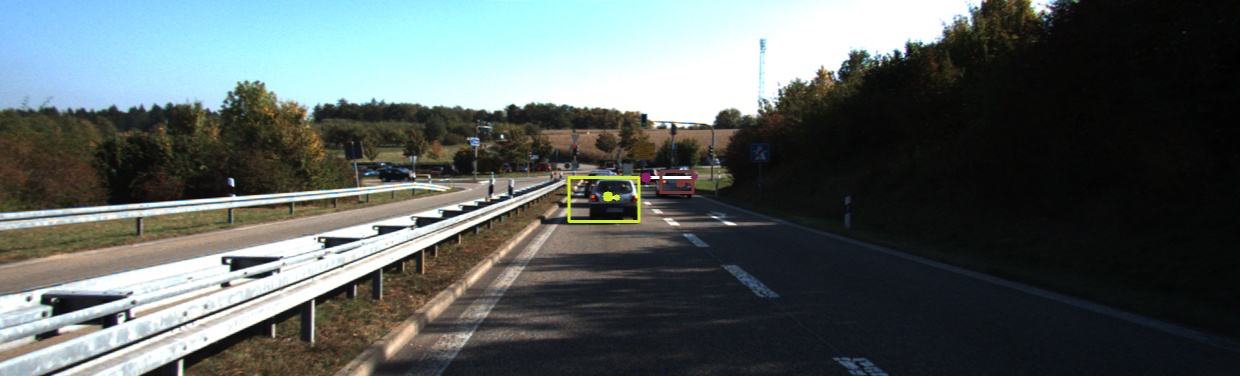}
\includegraphics[width=0.24\textwidth]{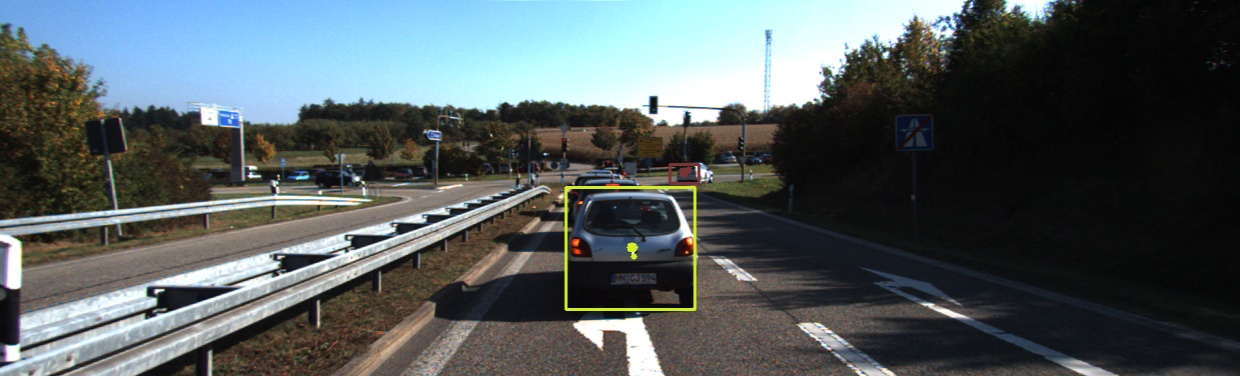}
\includegraphics[width=0.24\textwidth]{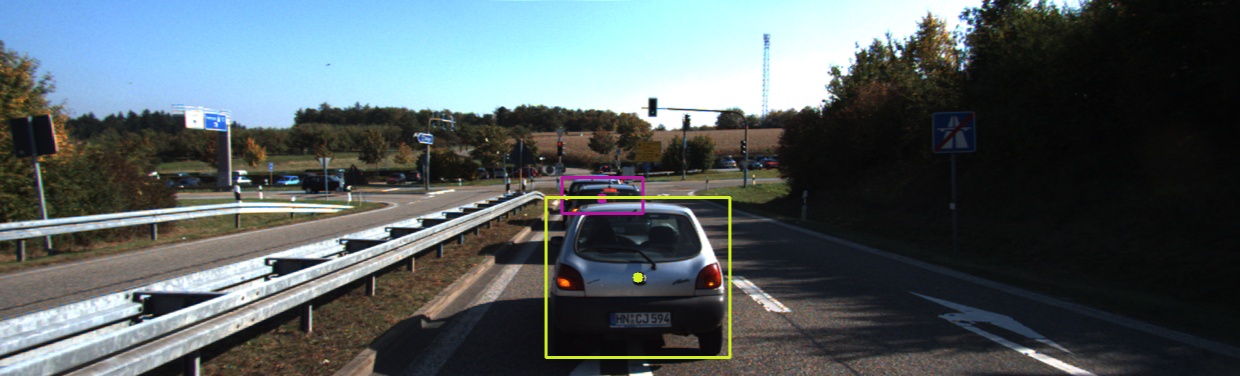}

\caption{{\sffamily\footnotesize Complete example scene form KITTIMOD tracking benchmark. High traffic highway situation including high and low speed driving}.}
\label{fig:app:example_kitti}
\end{figure}

\begin{figure}[b!]
\centering

\centering
\includegraphics[width=0.24\textwidth]{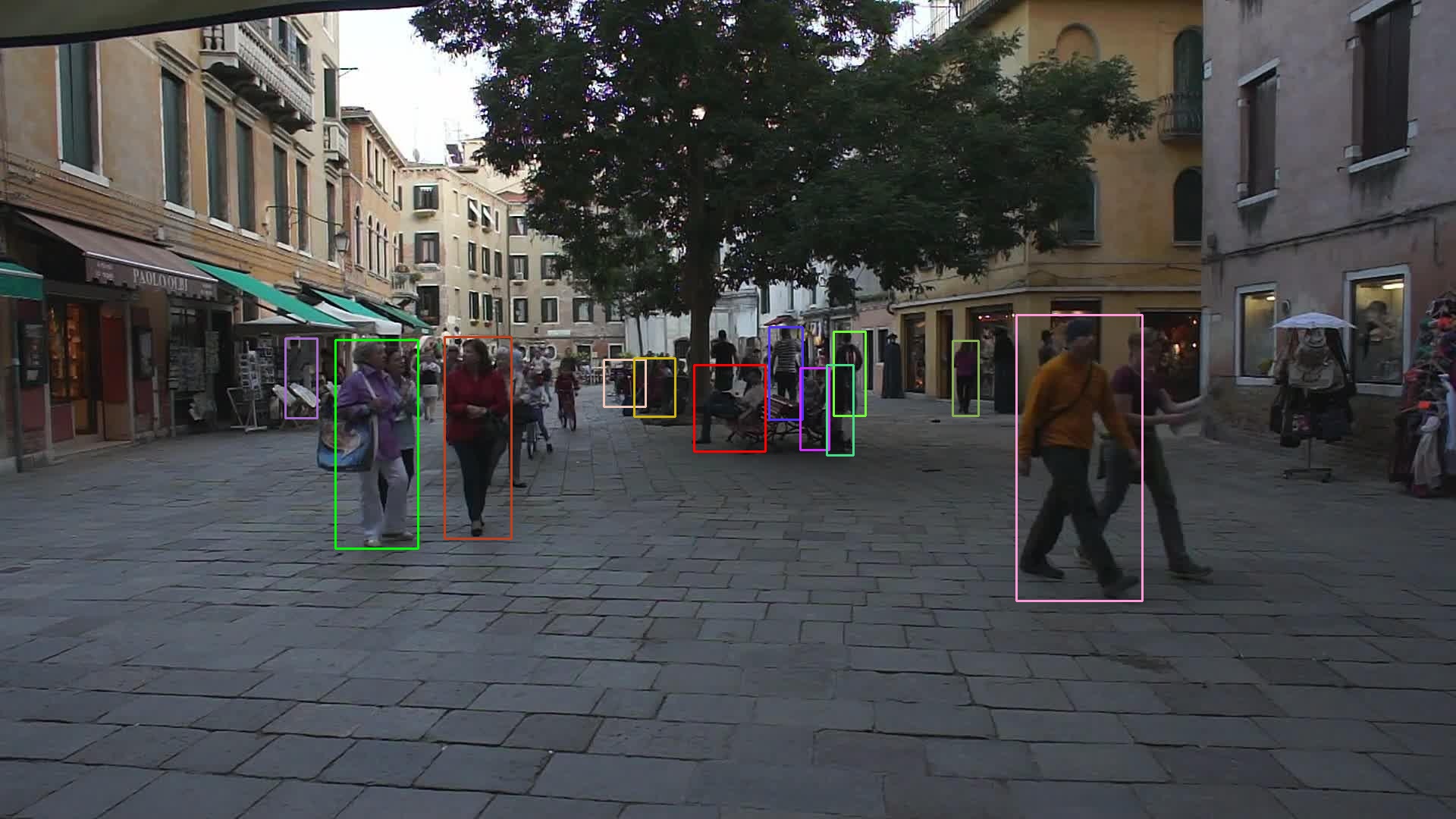}
\includegraphics[width=0.24\textwidth]{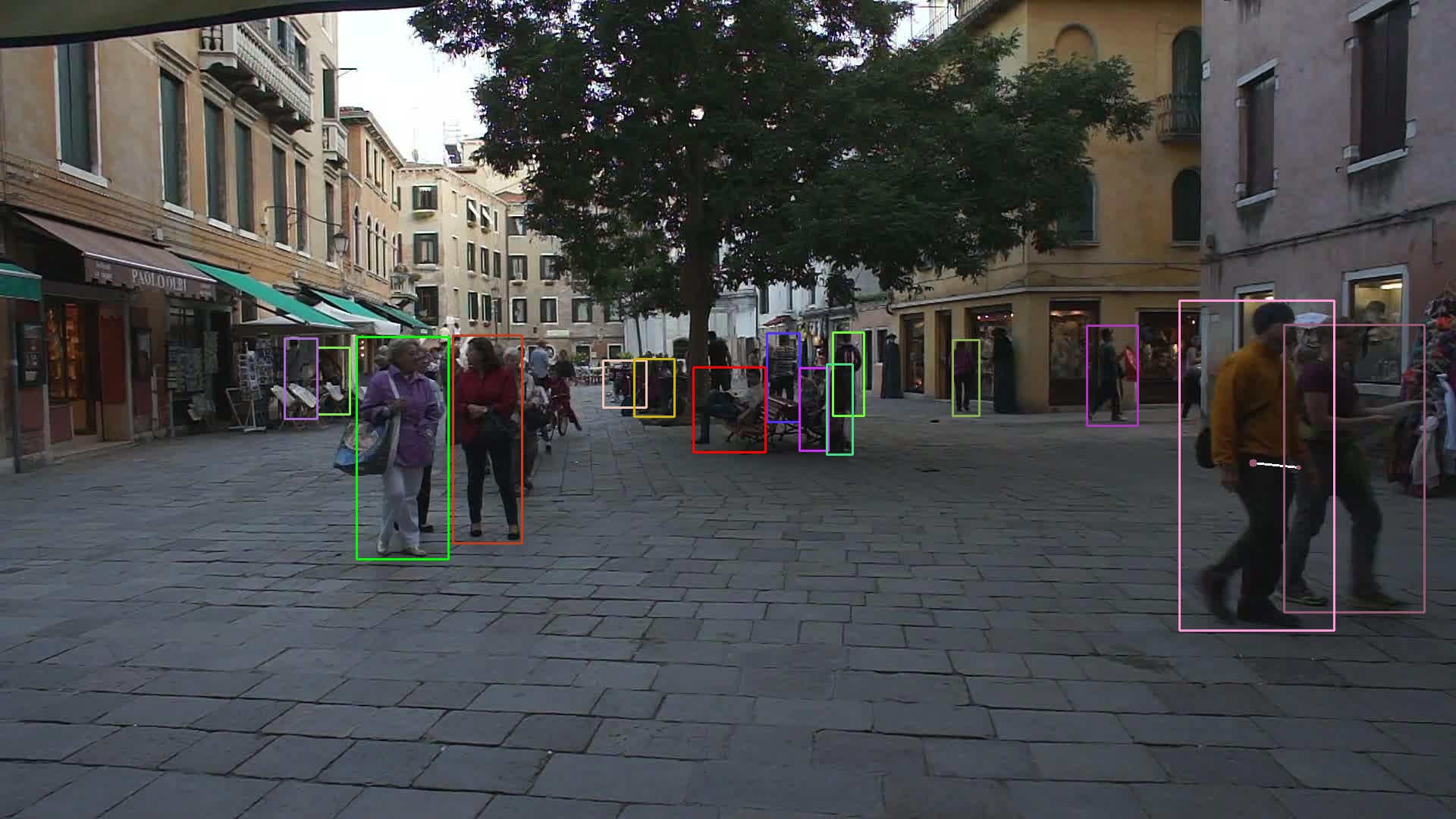}
\includegraphics[width=0.24\textwidth]{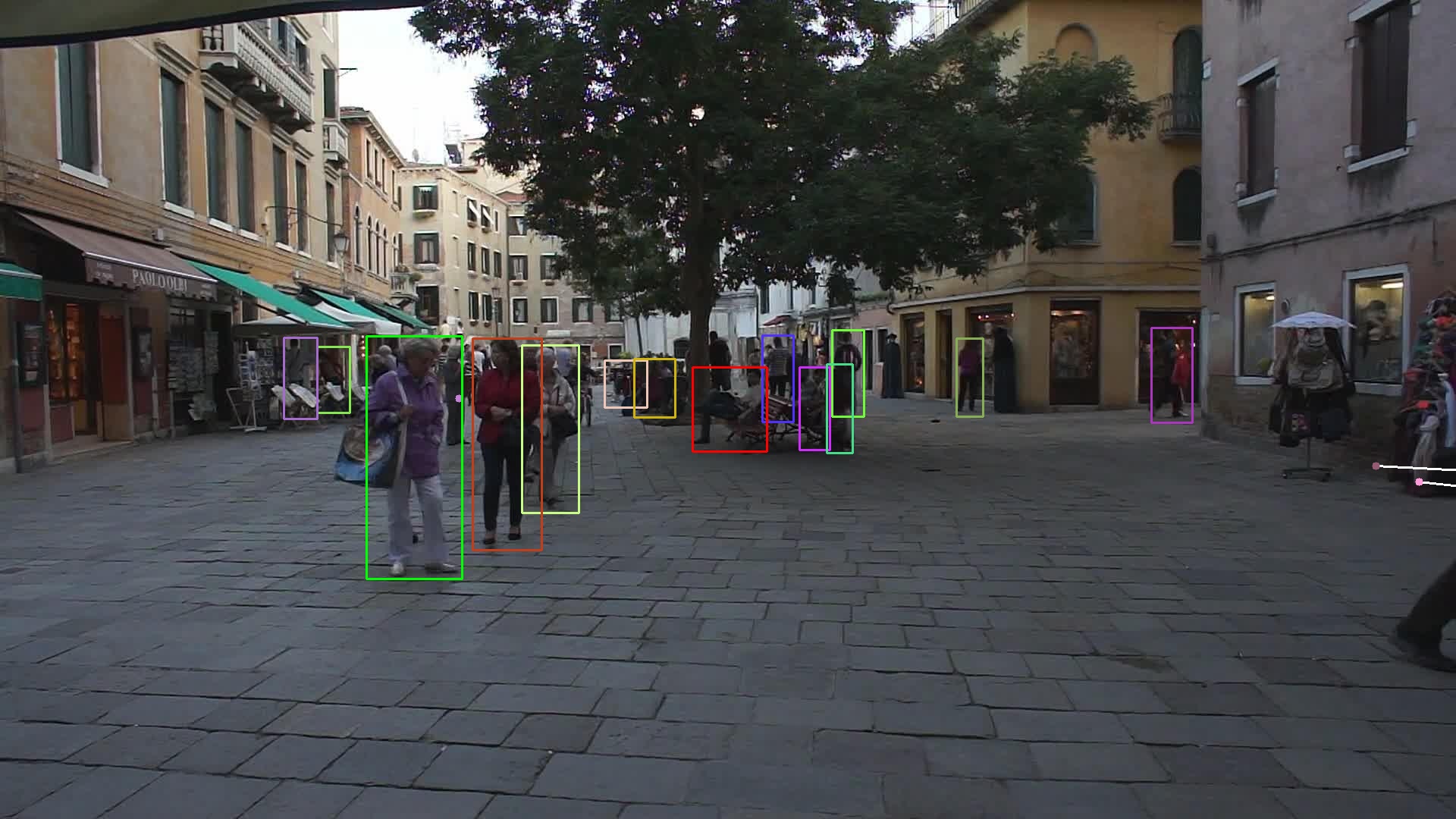}
\includegraphics[width=0.24\textwidth]{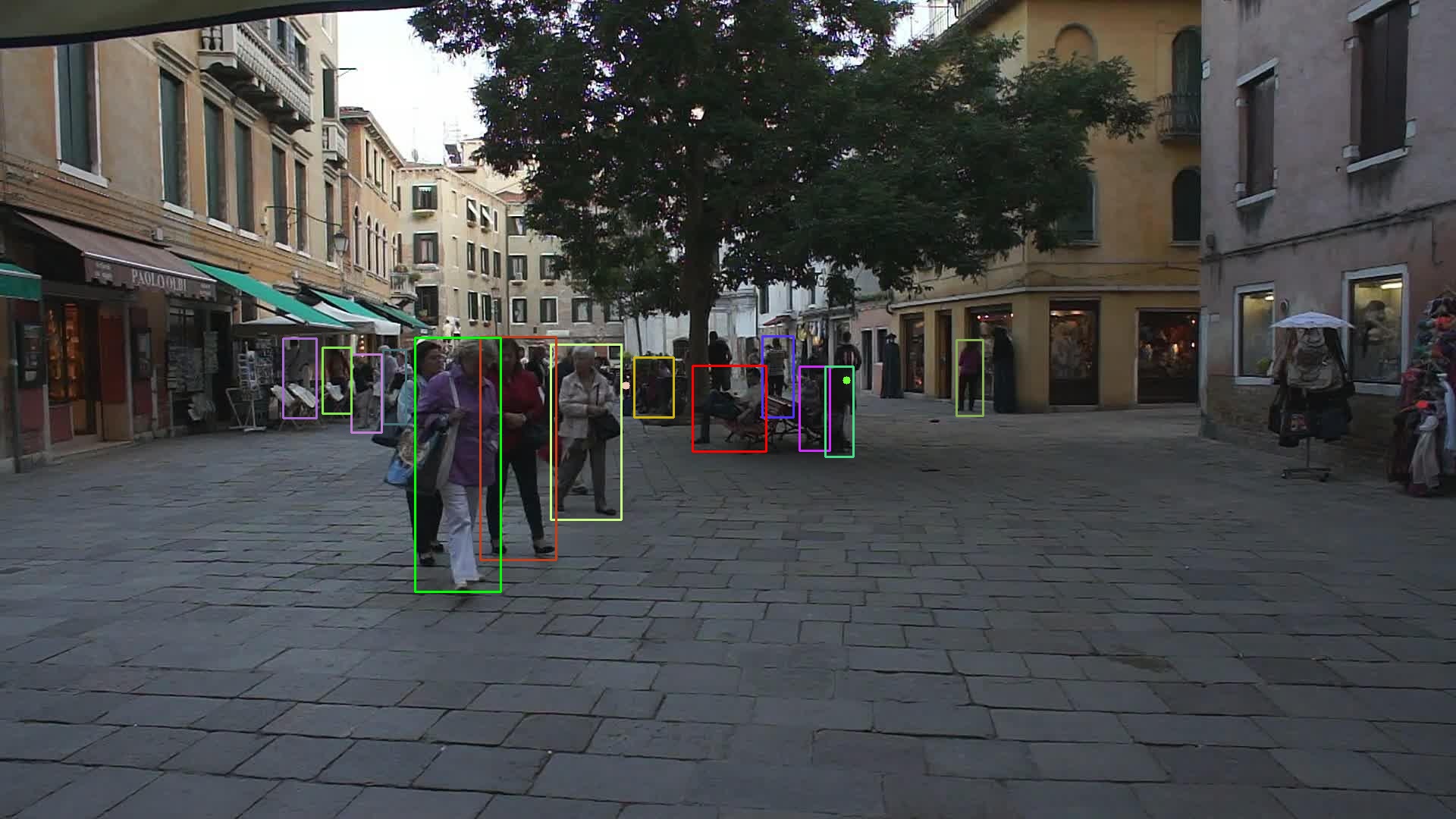}

\includegraphics[width=0.24\textwidth]{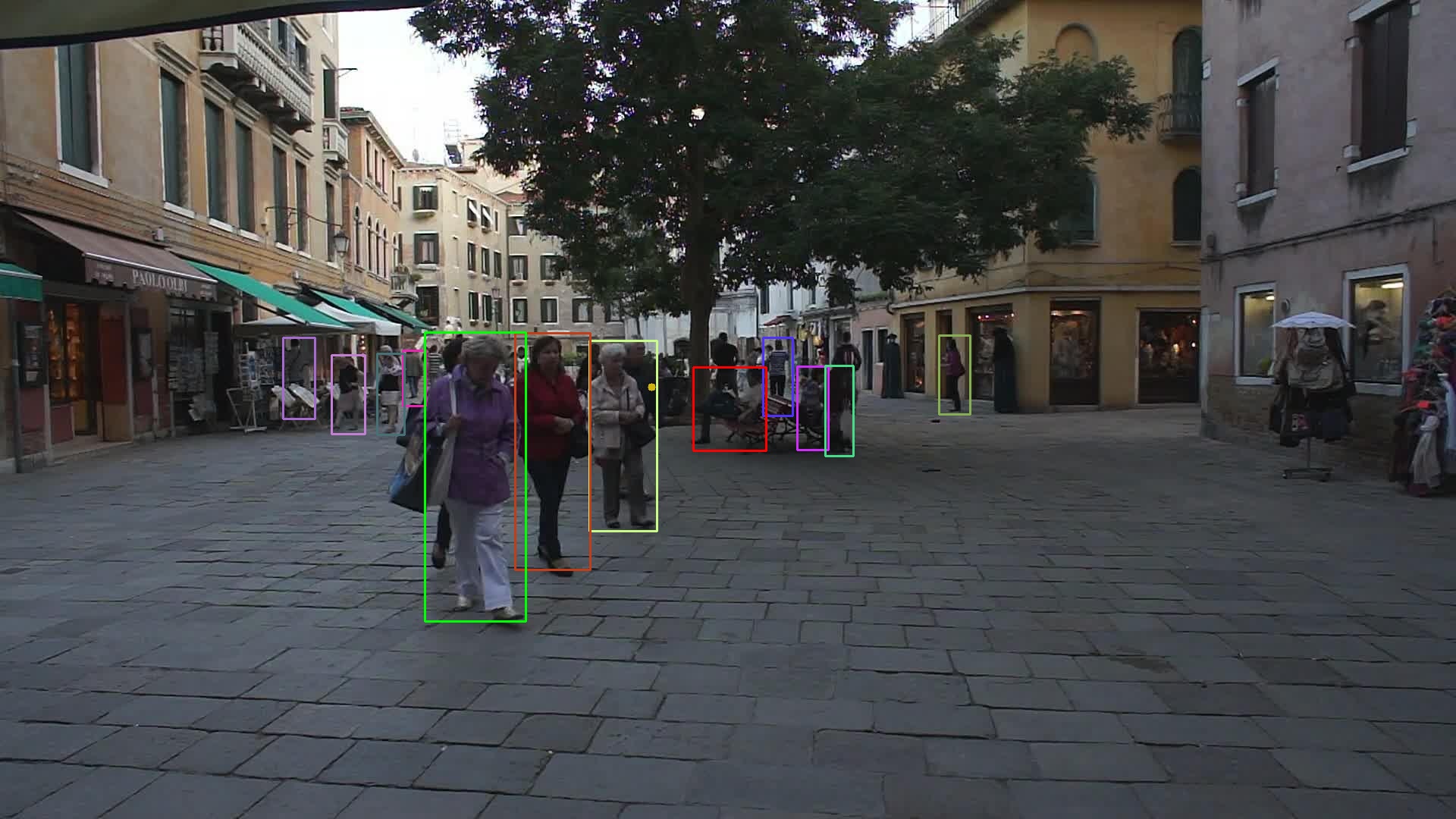}
\includegraphics[width=0.24\textwidth]{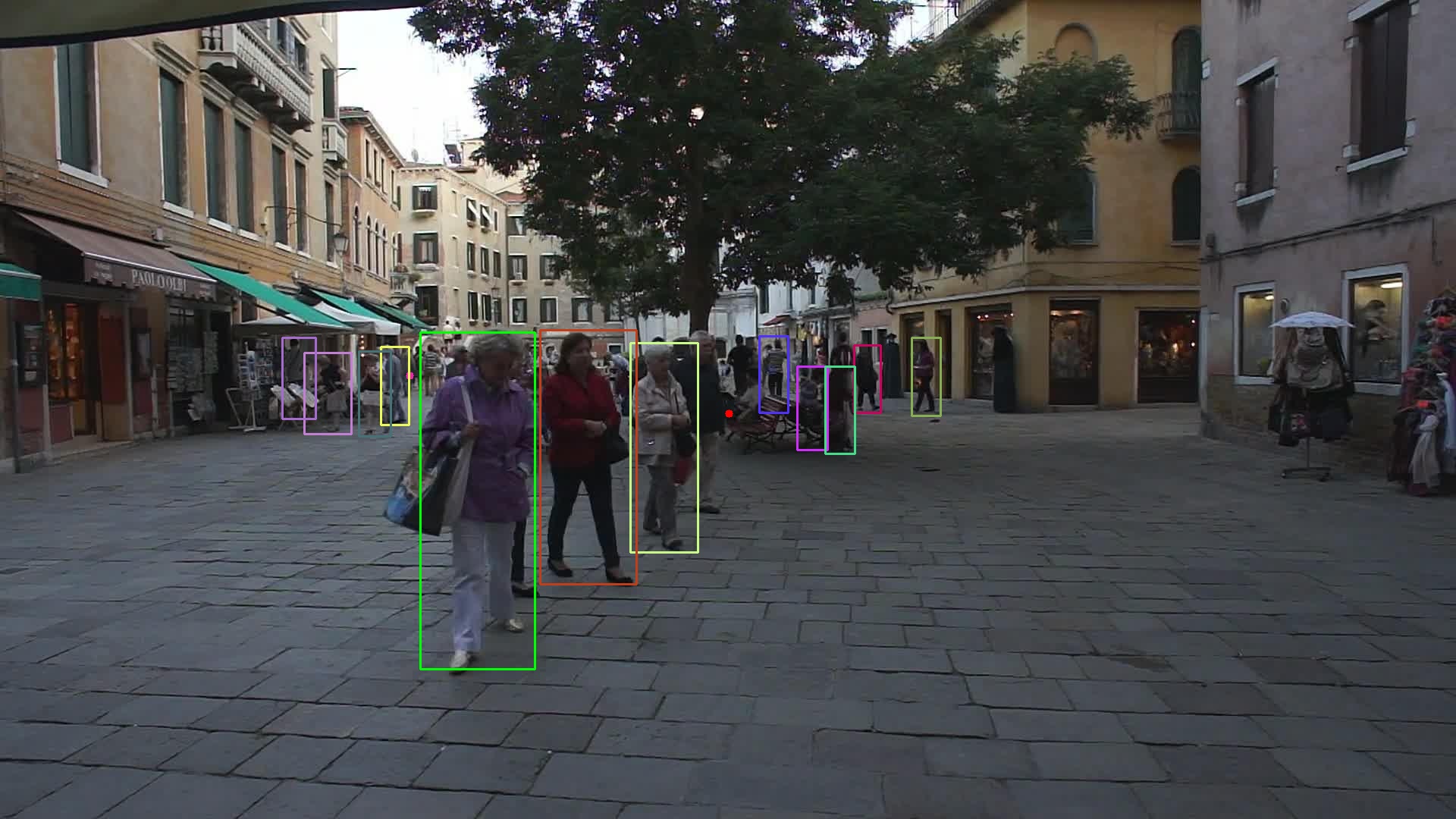}
\includegraphics[width=0.24\textwidth]{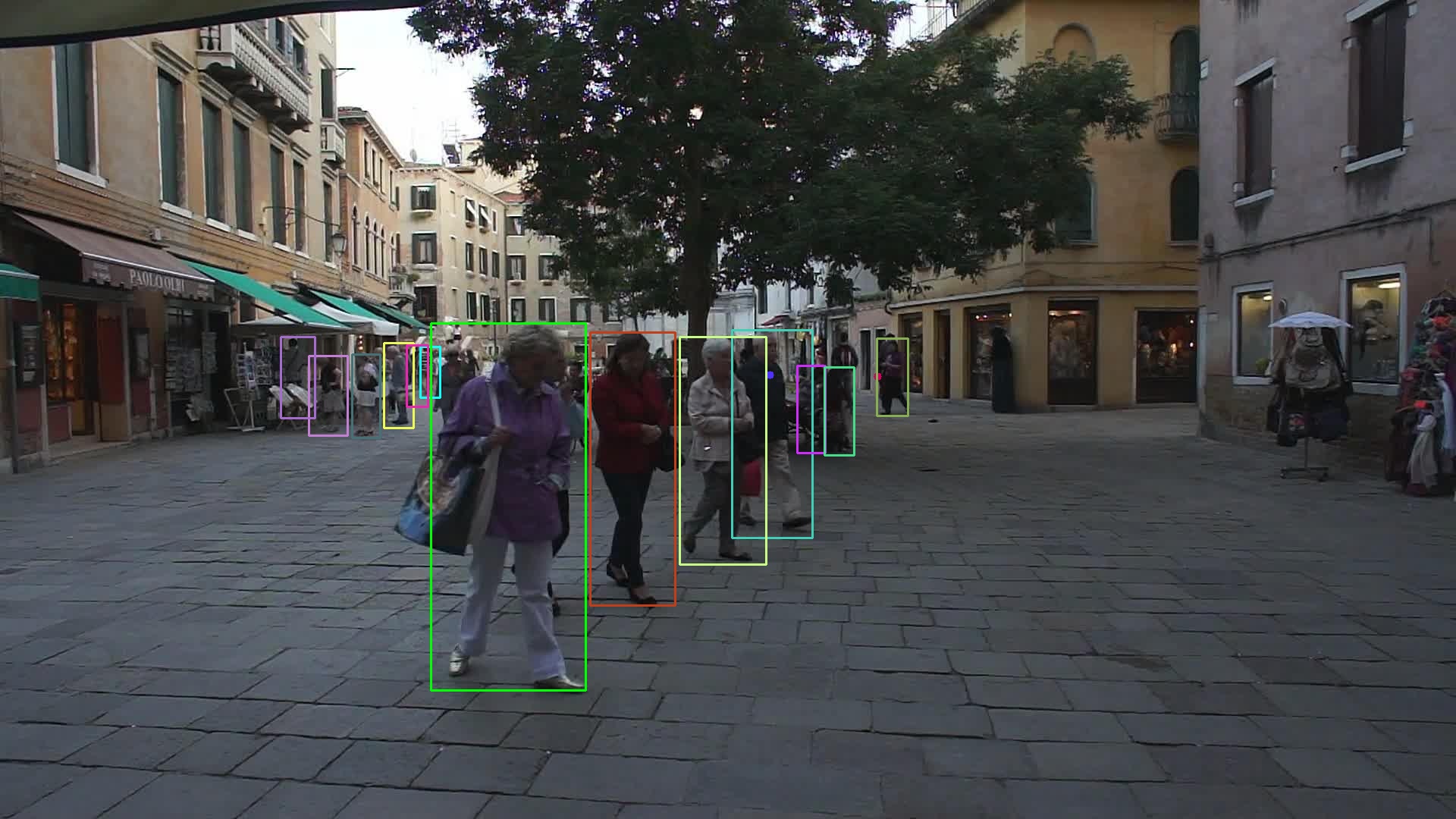}
\includegraphics[width=0.24\textwidth]{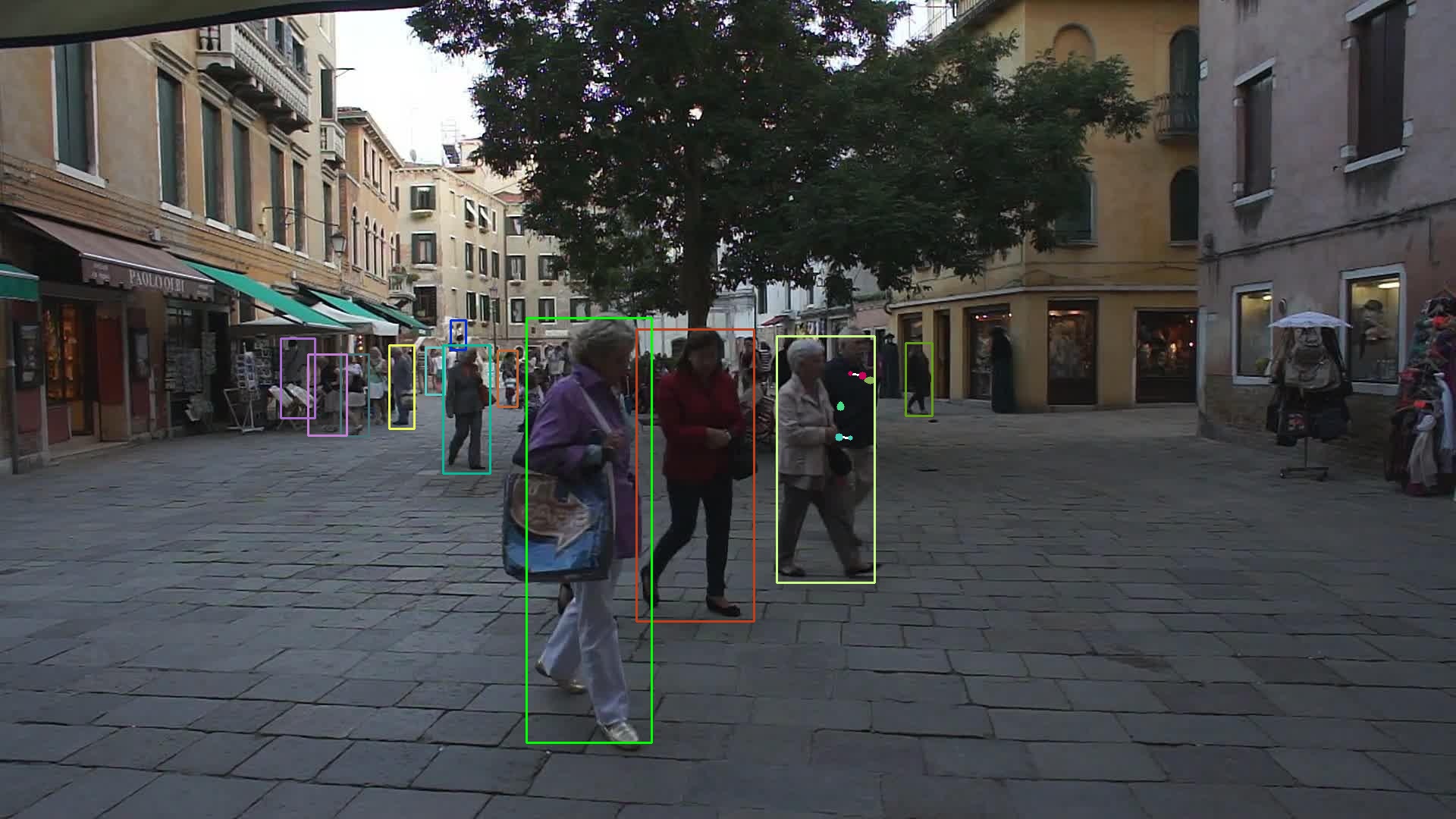}

\includegraphics[width=0.24\textwidth]{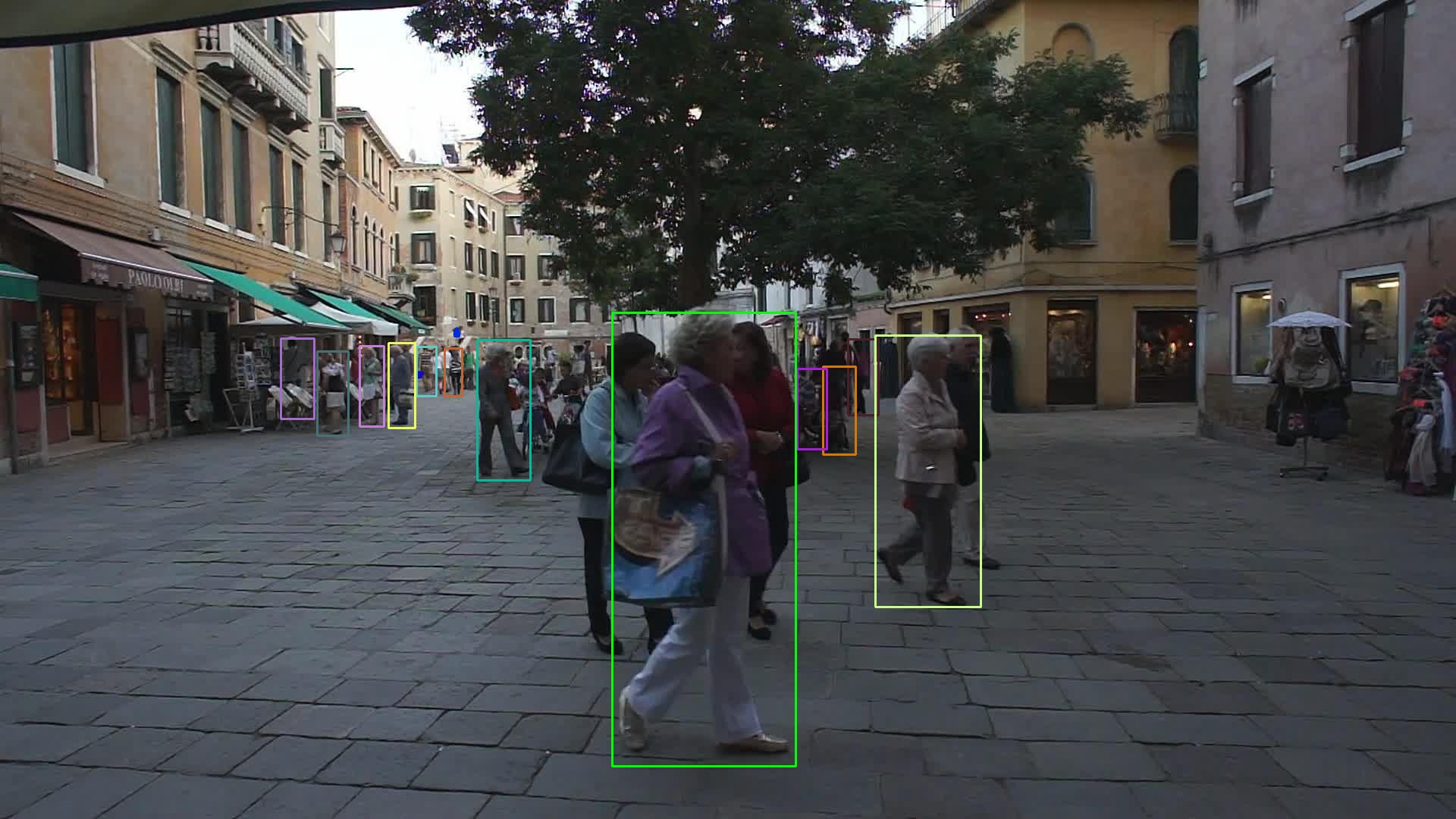}
\includegraphics[width=0.24\textwidth]{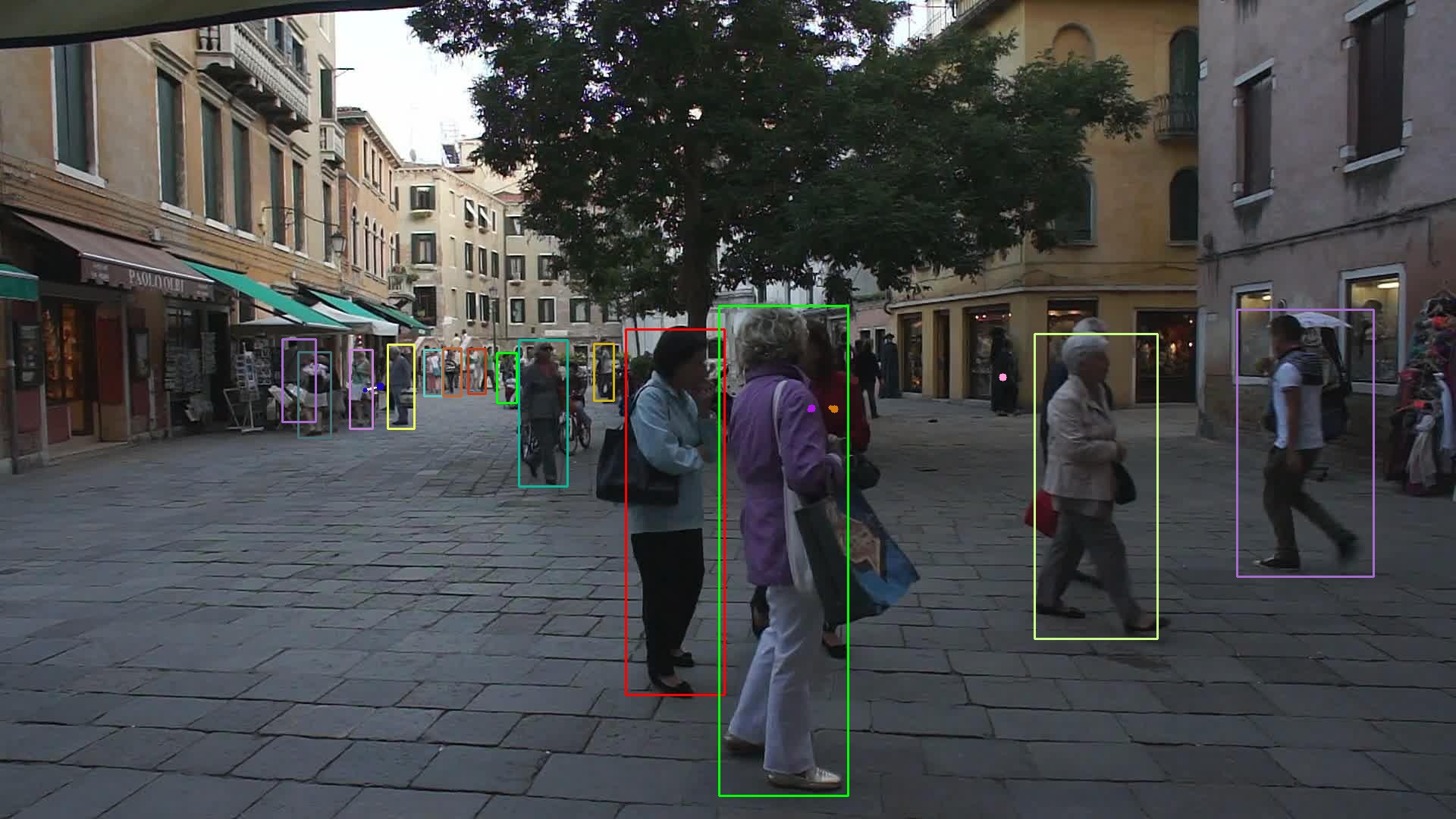}
\includegraphics[width=0.24\textwidth]{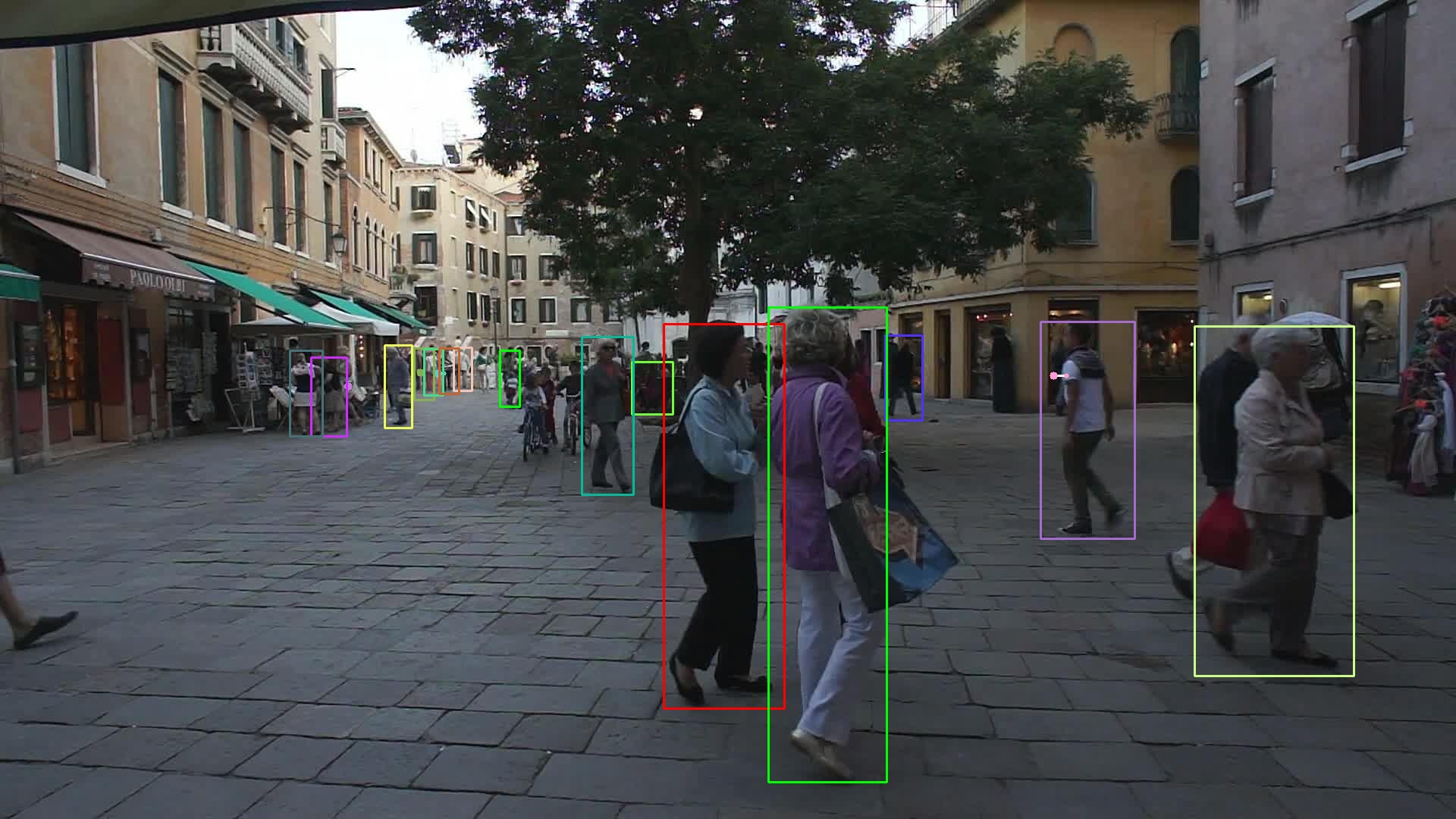}
\includegraphics[width=0.24\textwidth]{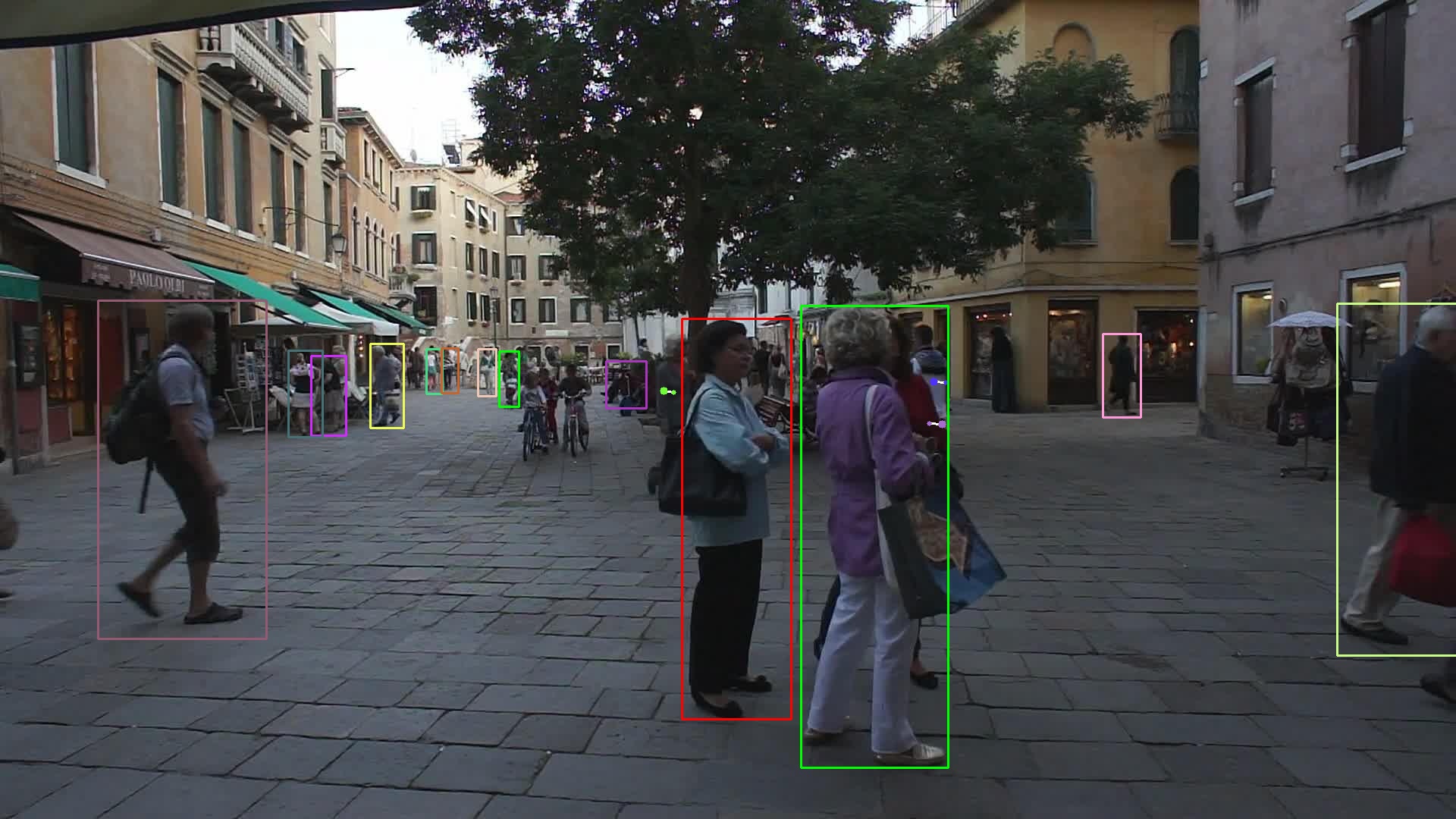}

\includegraphics[width=0.24\textwidth]{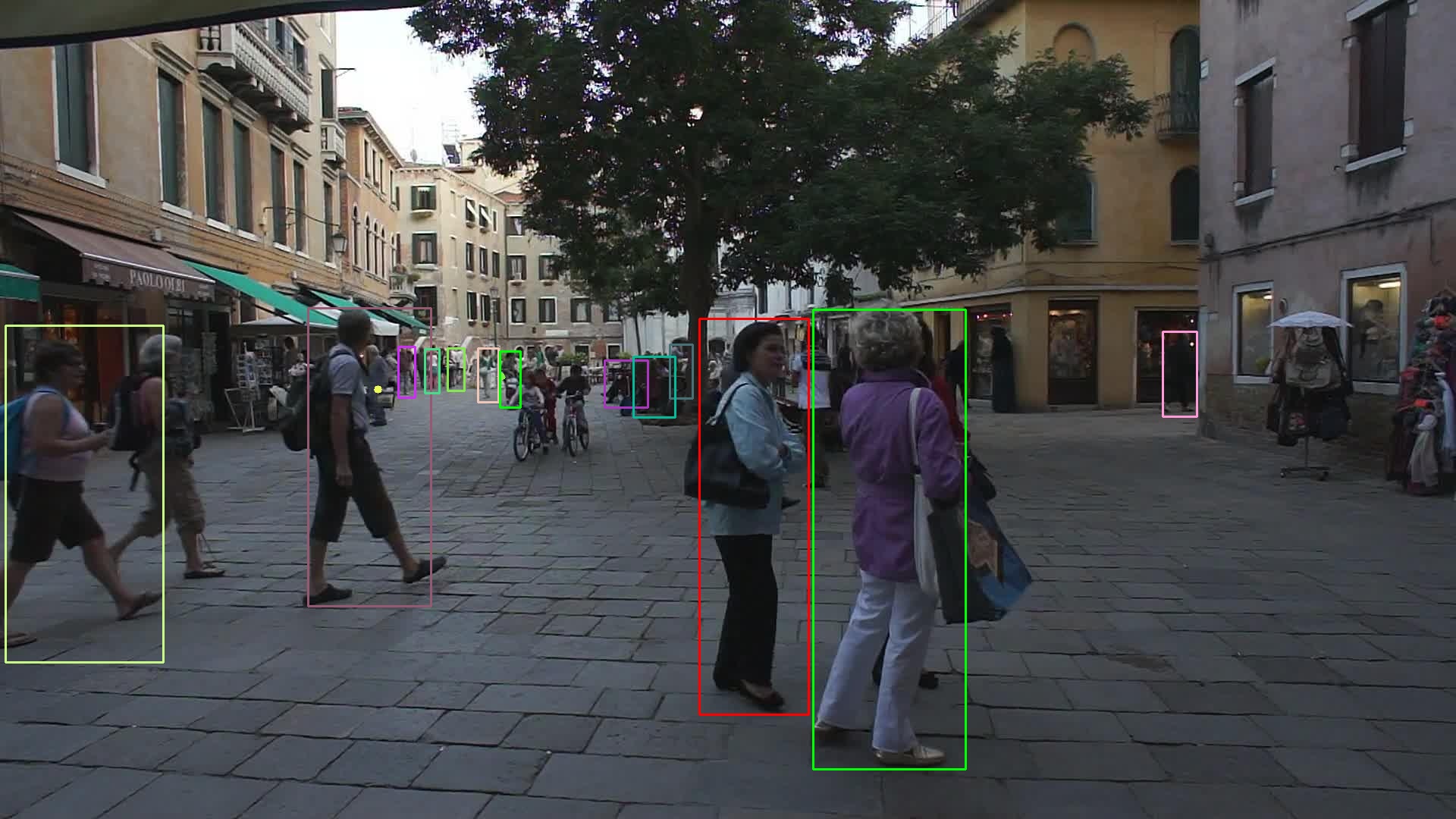}
\includegraphics[width=0.24\textwidth]{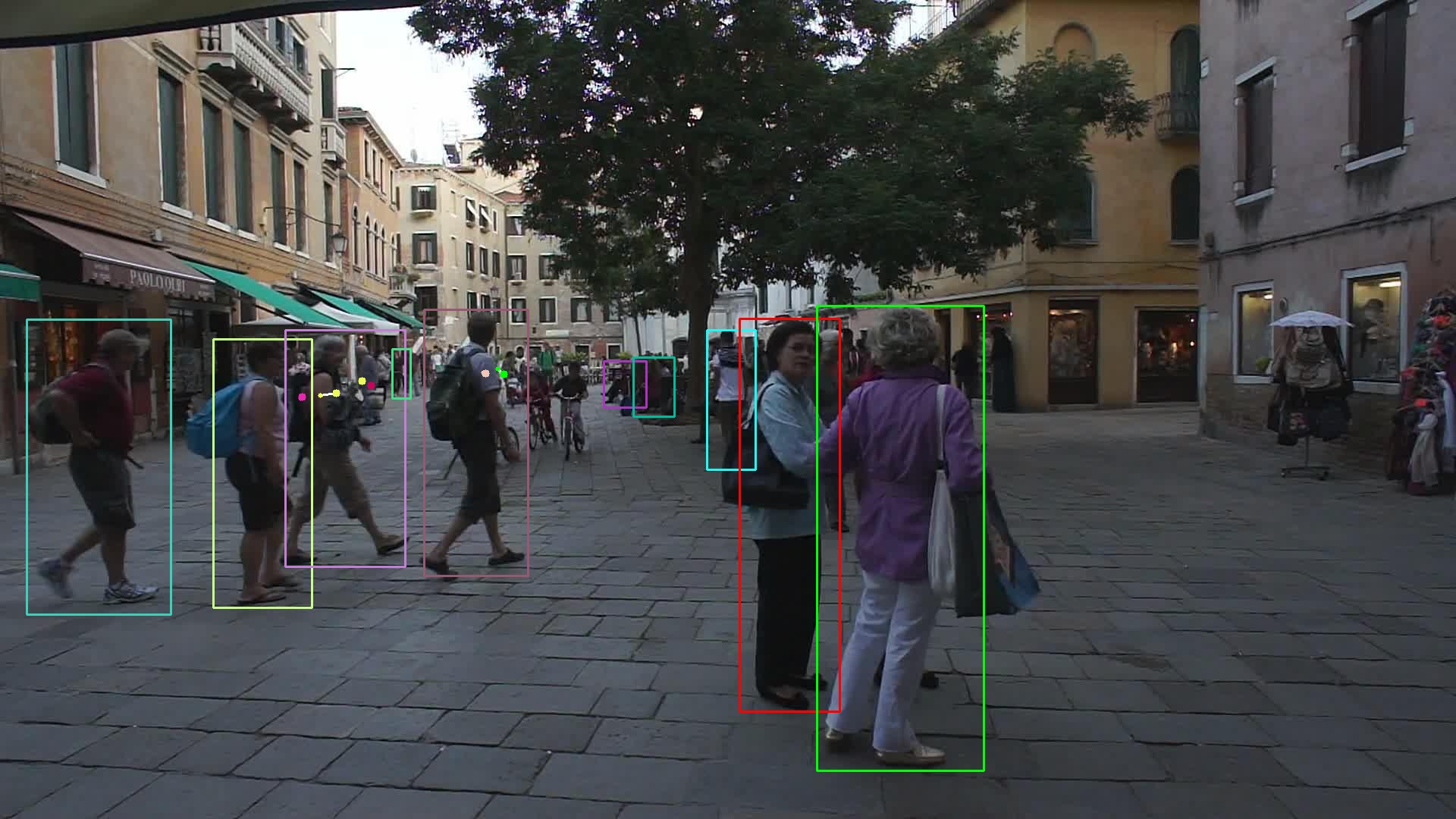}
\includegraphics[width=0.24\textwidth]{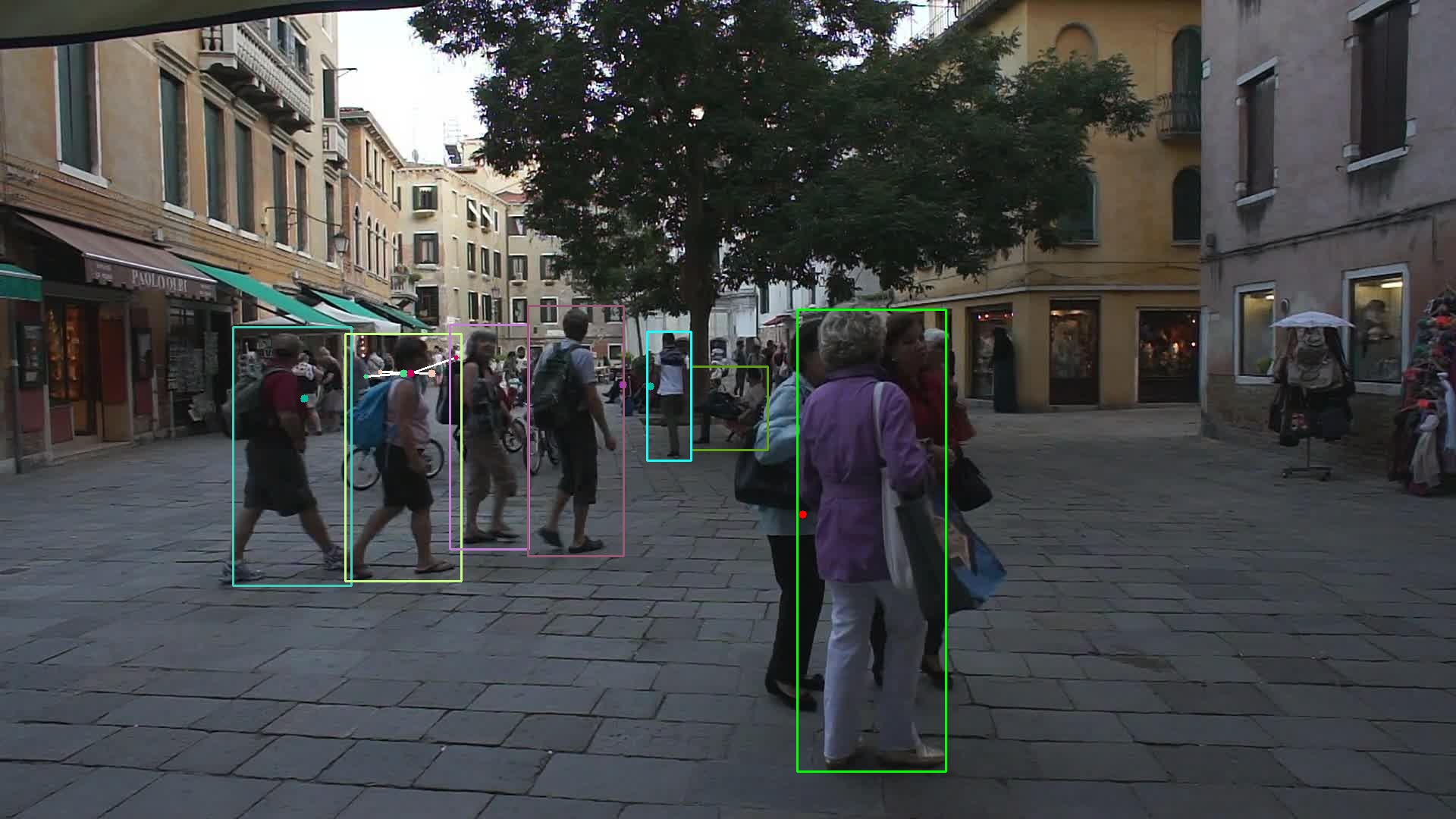}
\includegraphics[width=0.24\textwidth]{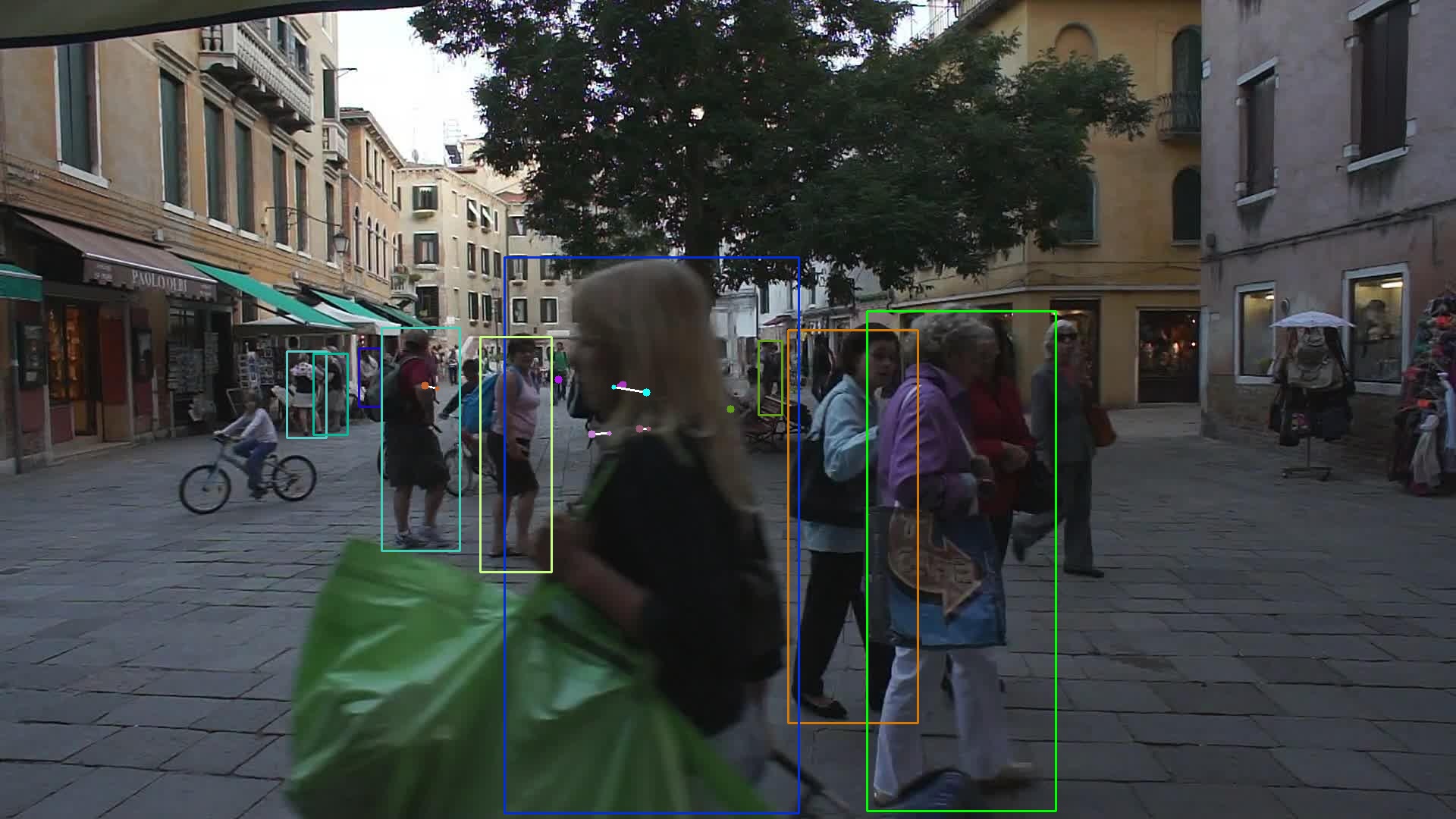}

\includegraphics[width=0.24\textwidth]{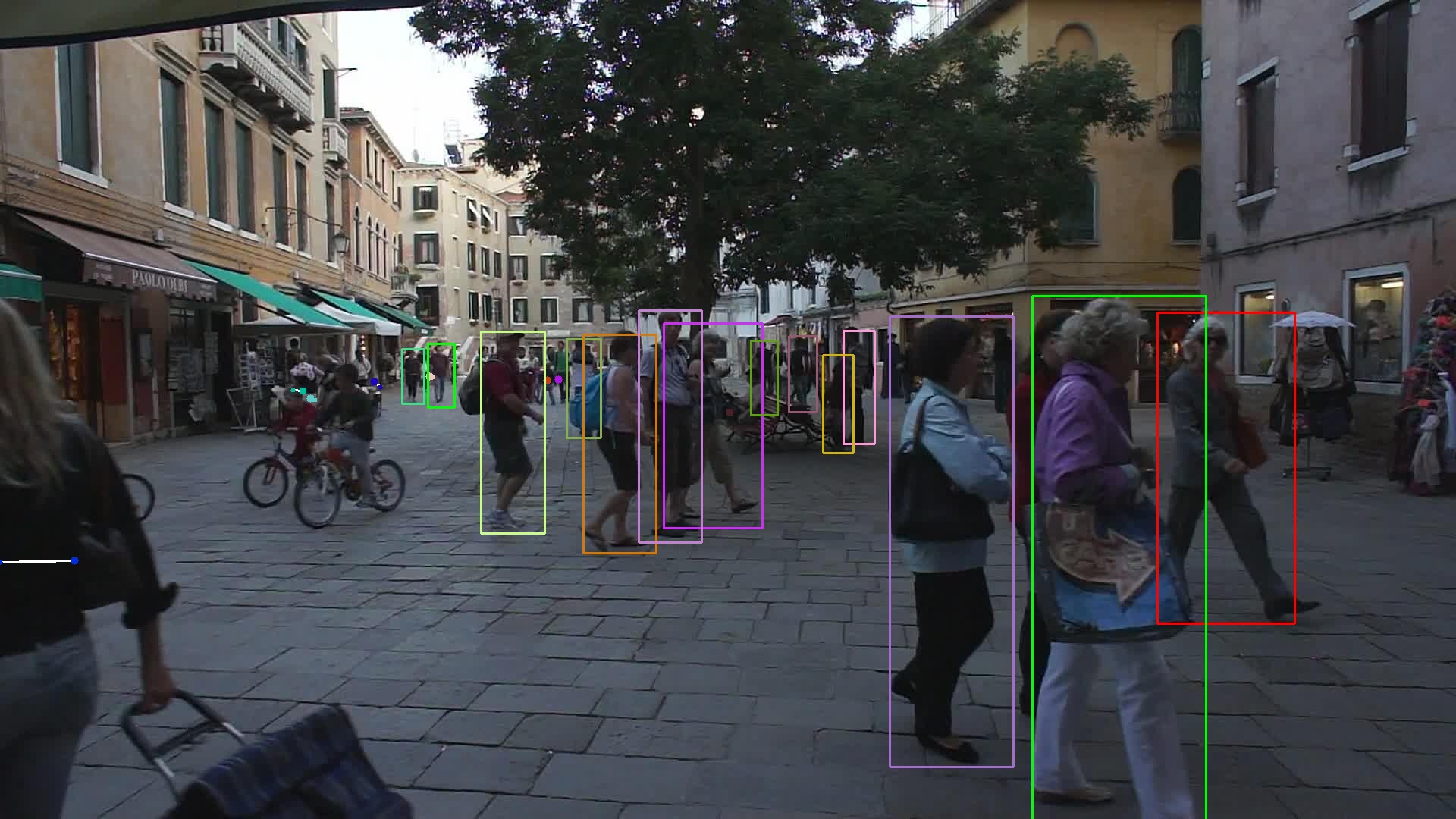}
\includegraphics[width=0.24\textwidth]{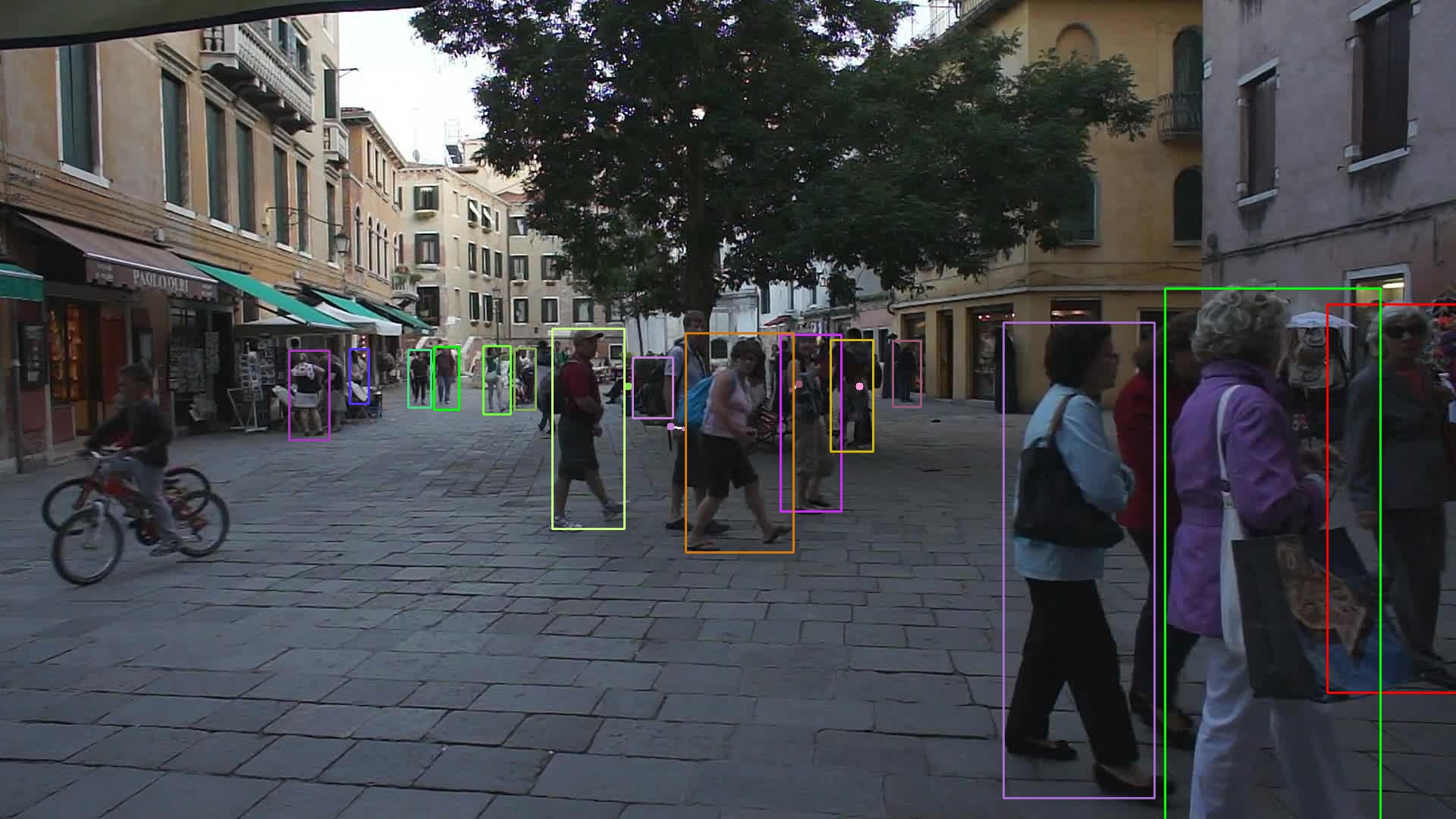}
\includegraphics[width=0.24\textwidth]{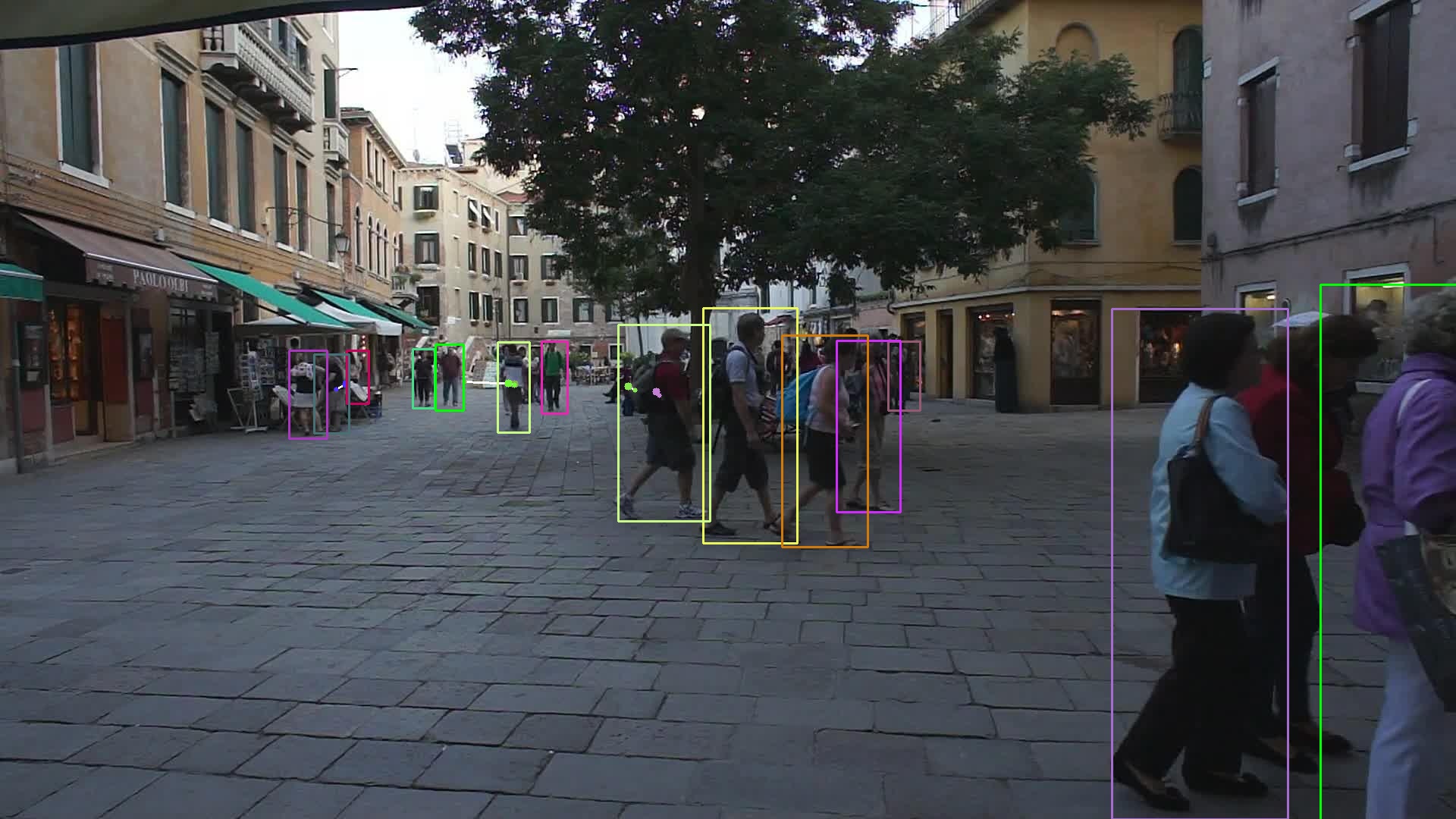}
\includegraphics[width=0.24\textwidth]{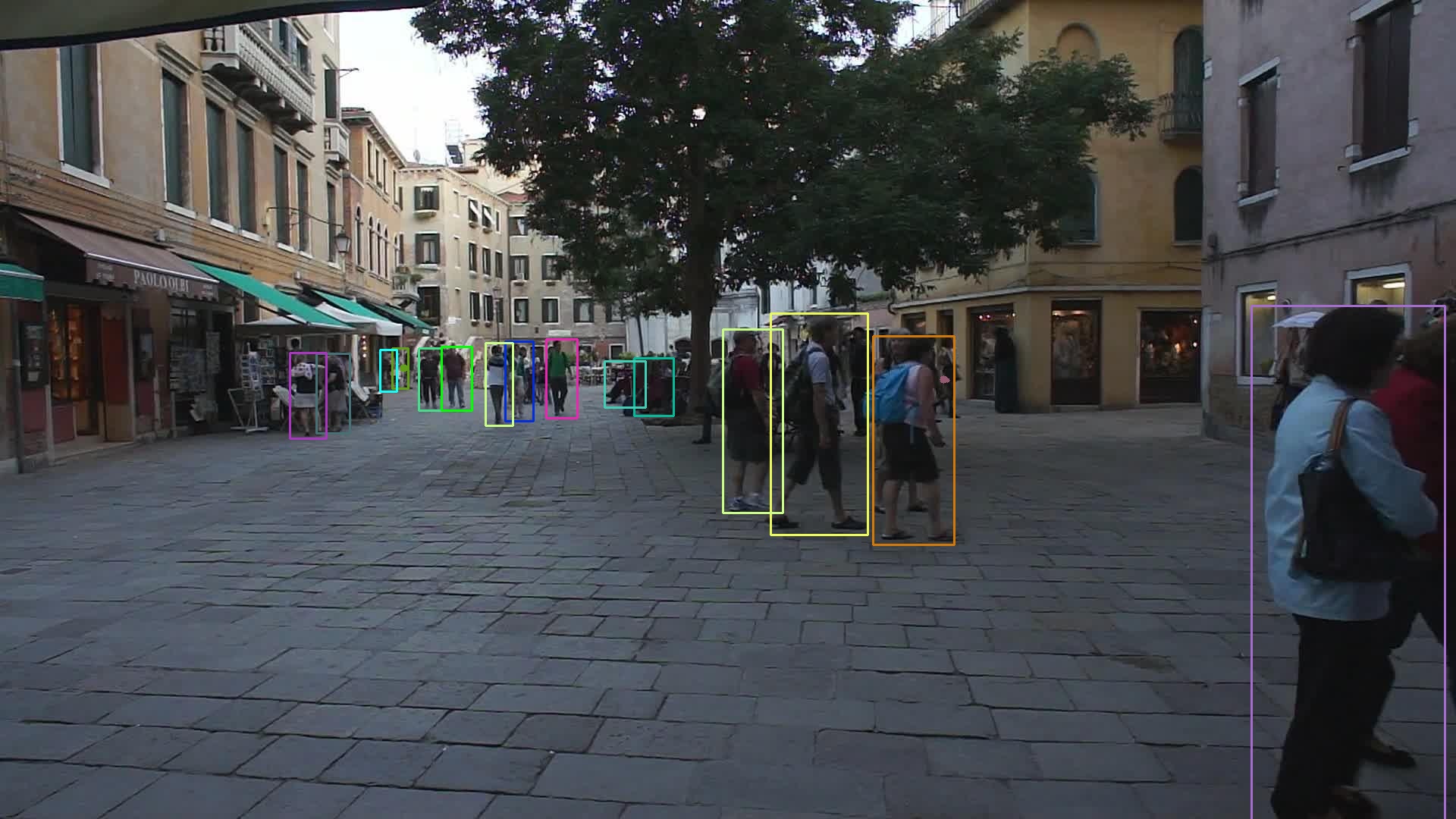}

\caption{{\sffamily\footnotesize Complete example scene from MOT16 benchmark dataset for people tracking}.}
\label{fig:app:example_MOT}
\end{figure}

\begin{figure}[h!]
\centering

\centering
\includegraphics[width=0.24\textwidth]{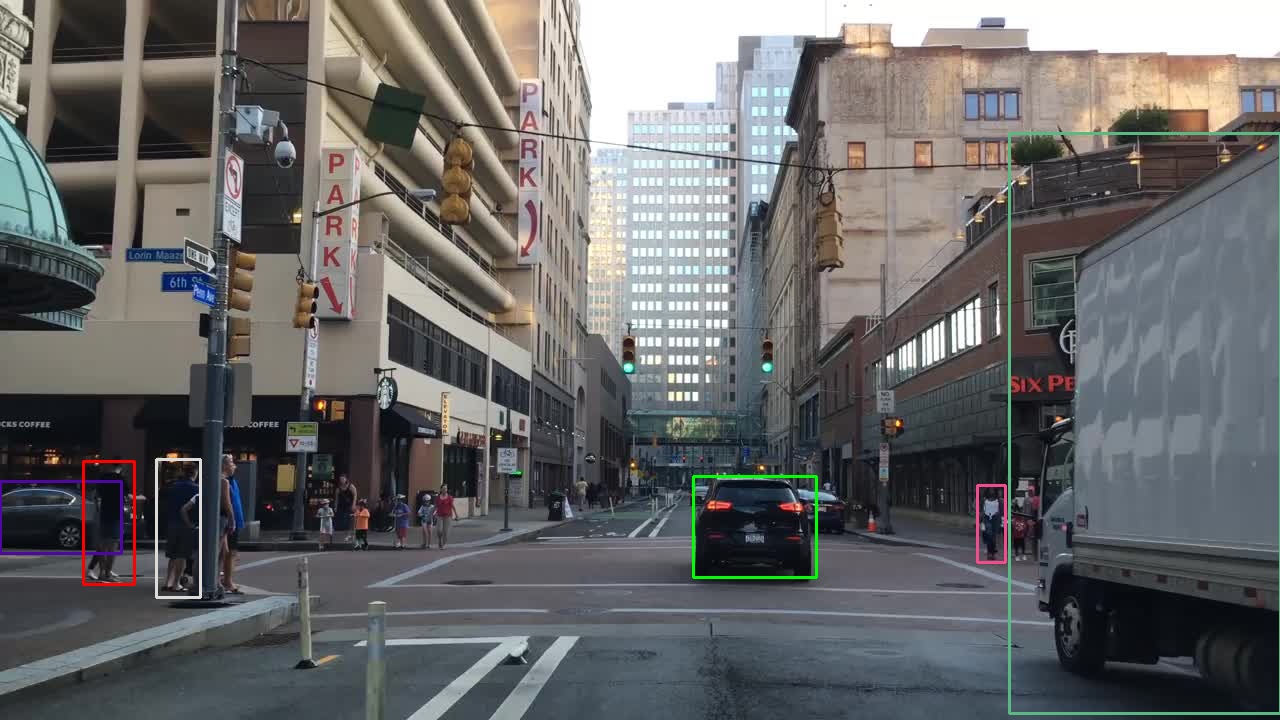}
\includegraphics[width=0.24\textwidth]{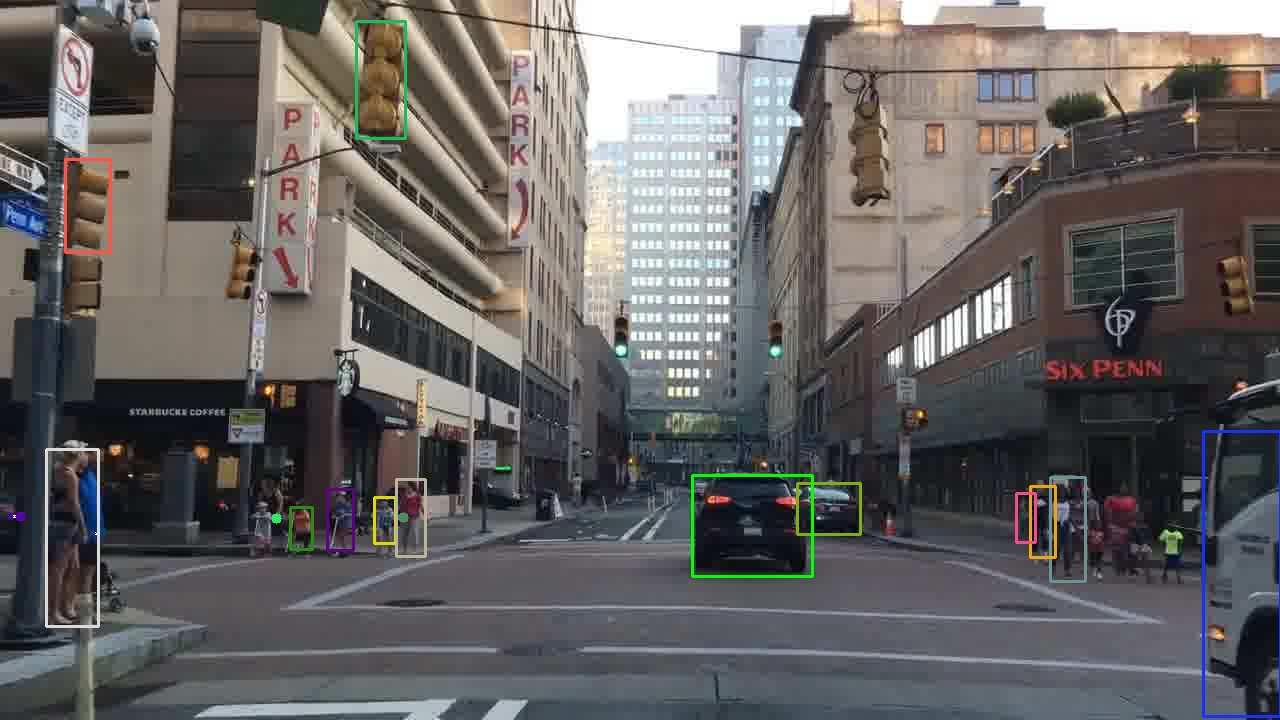}
\includegraphics[width=0.24\textwidth]{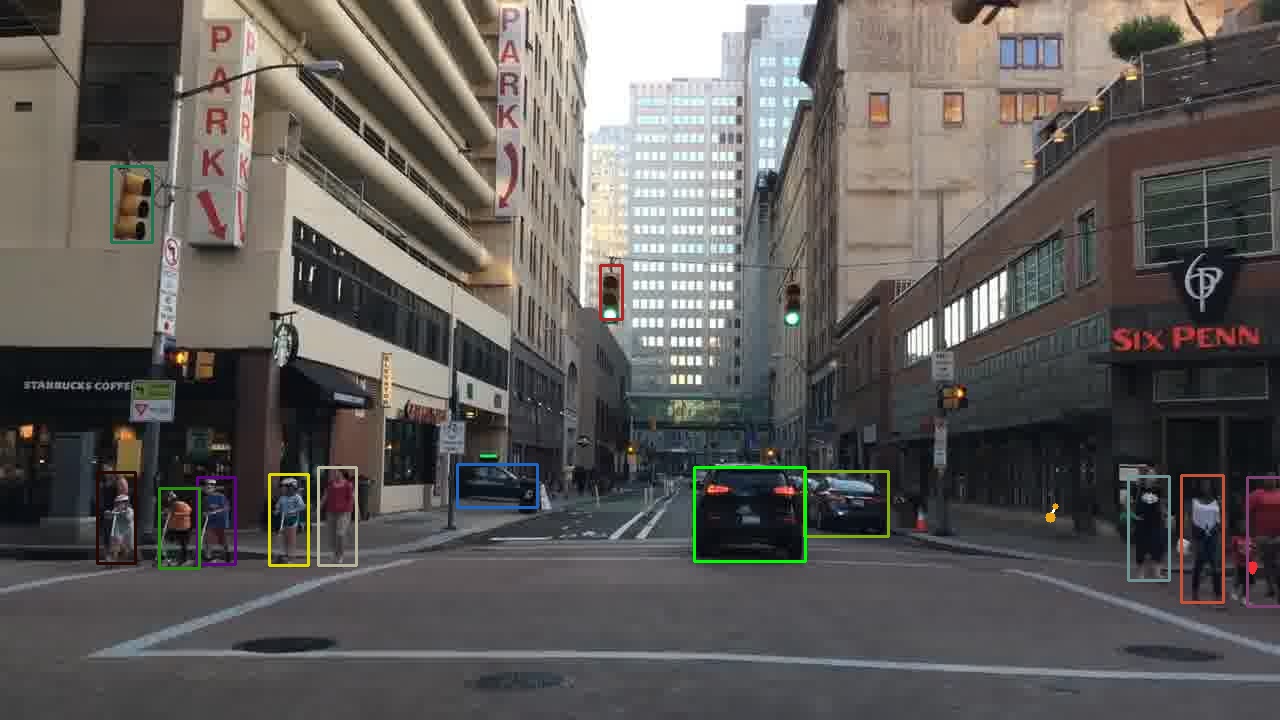}
\includegraphics[width=0.24\textwidth]{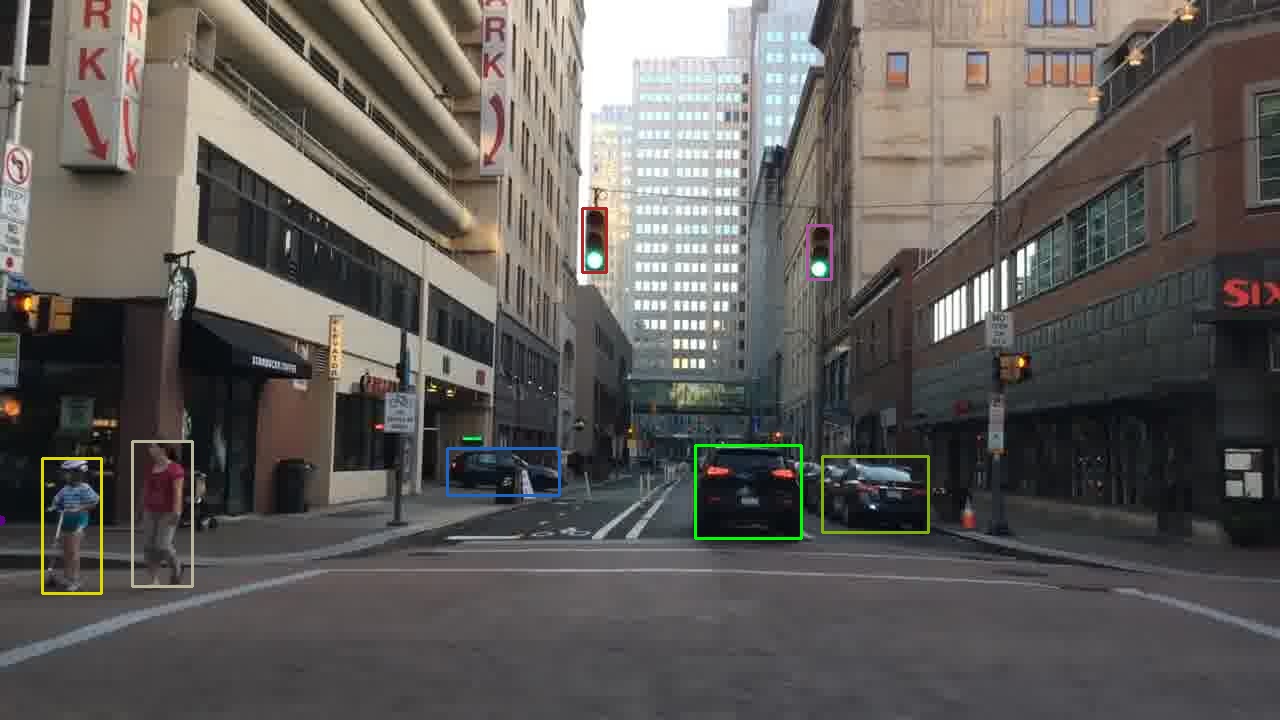}

\includegraphics[width=0.24\textwidth]{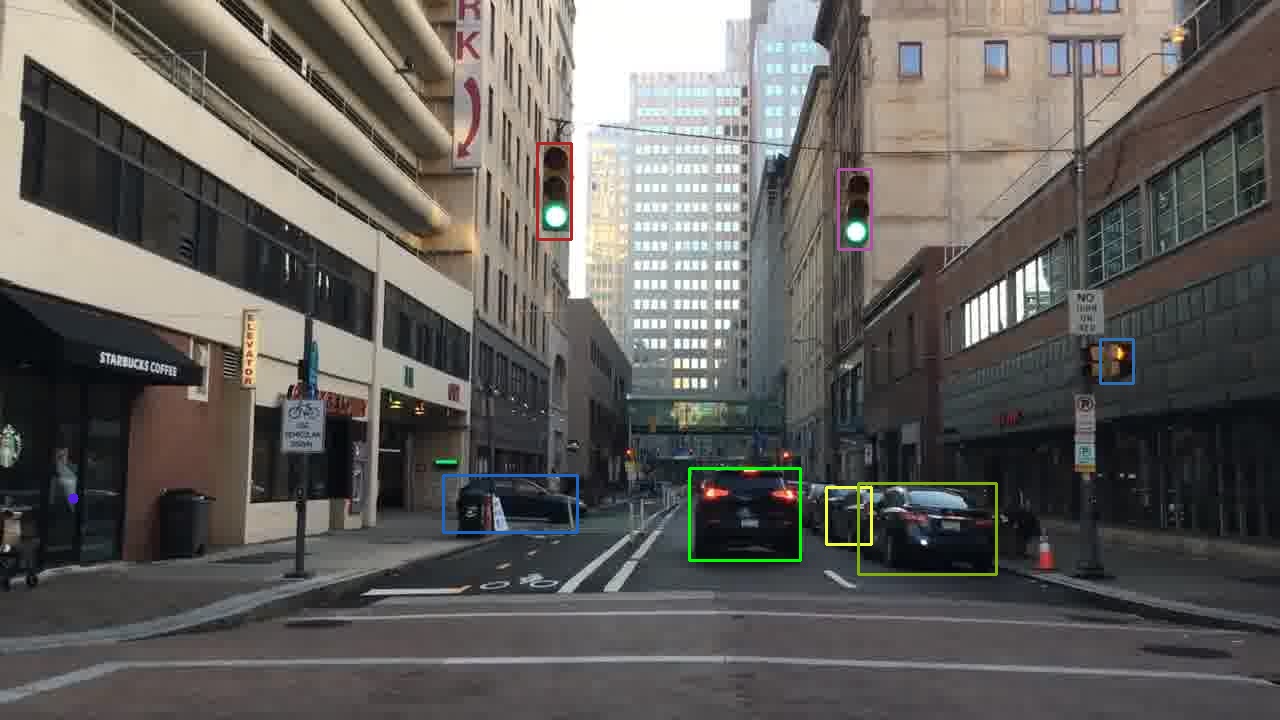}
\includegraphics[width=0.24\textwidth]{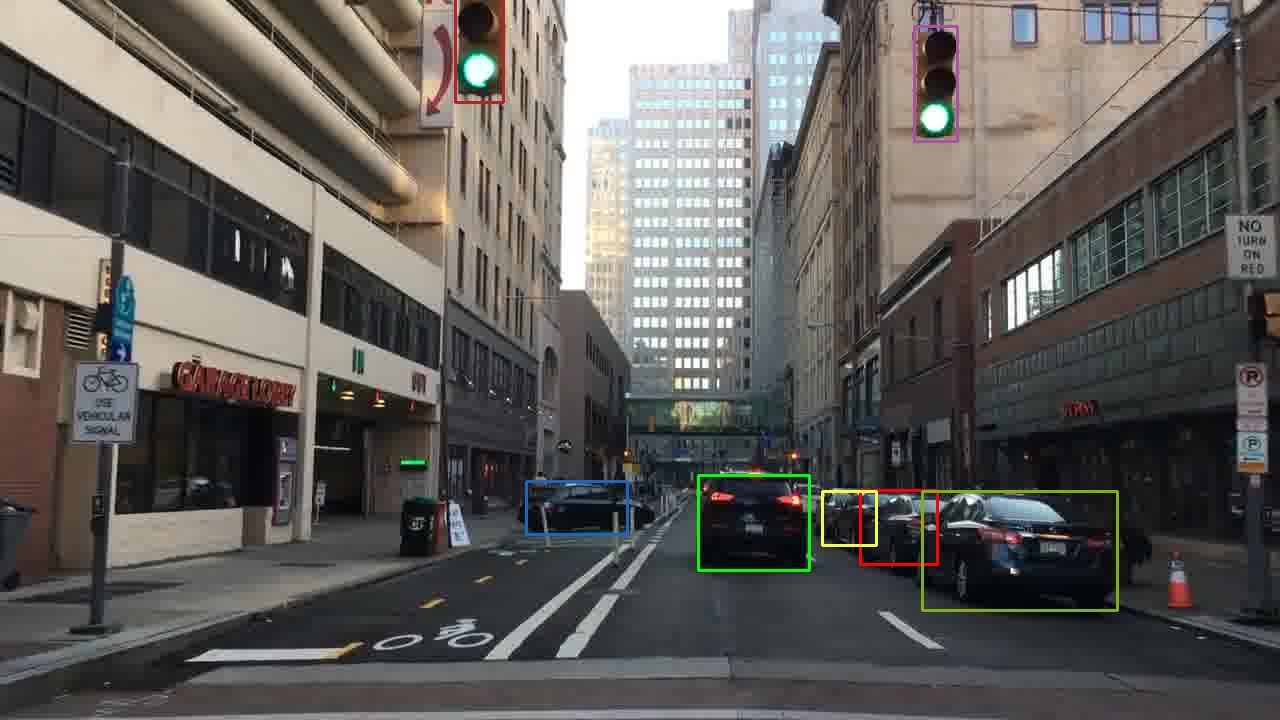}
\includegraphics[width=0.24\textwidth]{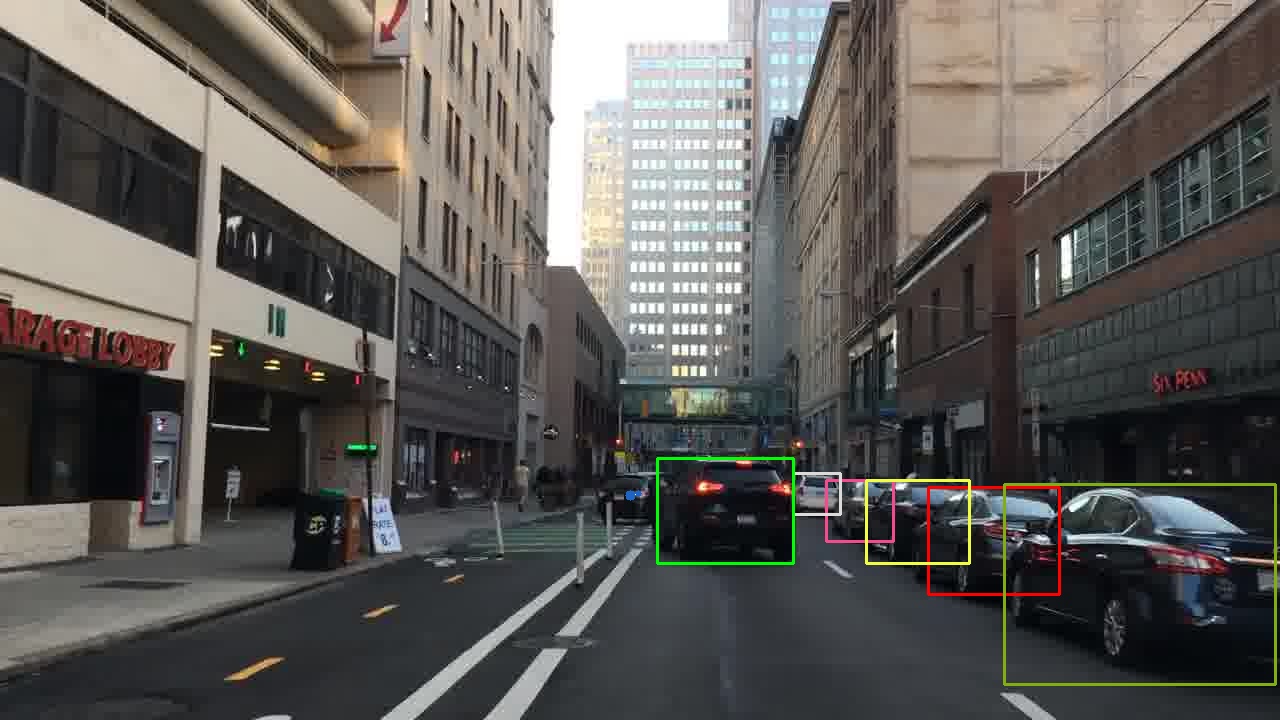}
\includegraphics[width=0.24\textwidth]{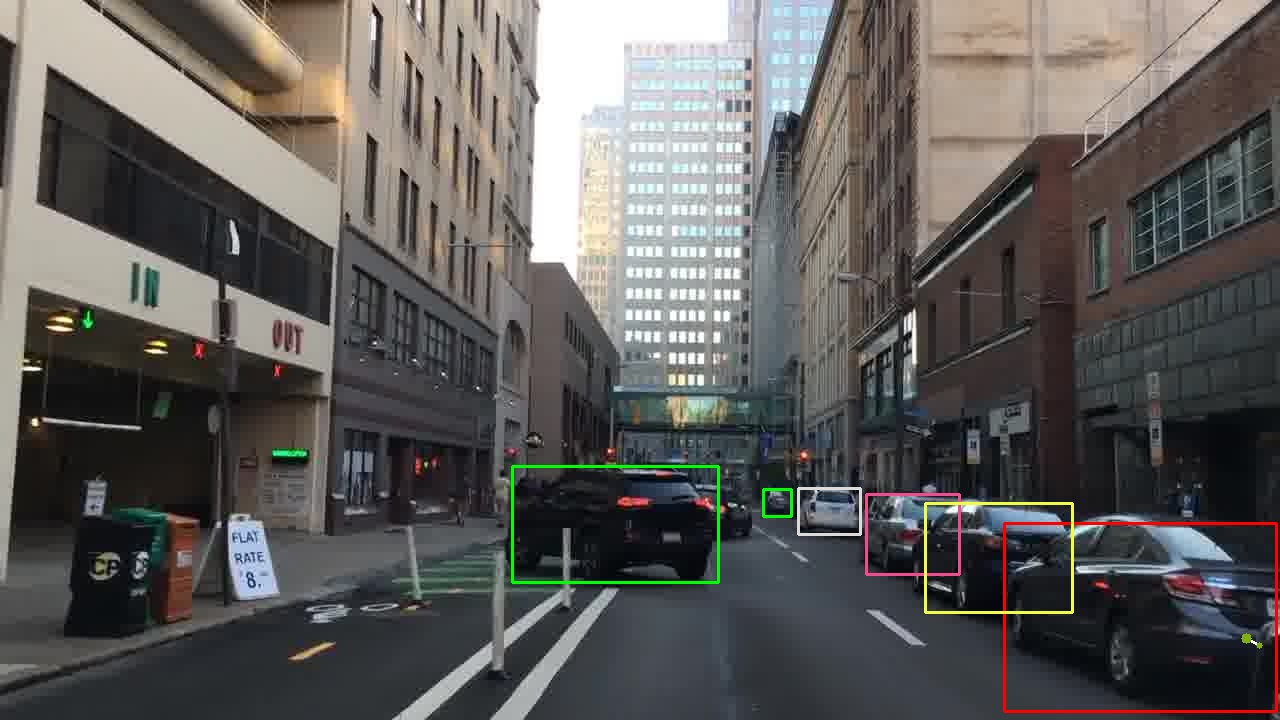}

\includegraphics[width=0.24\textwidth]{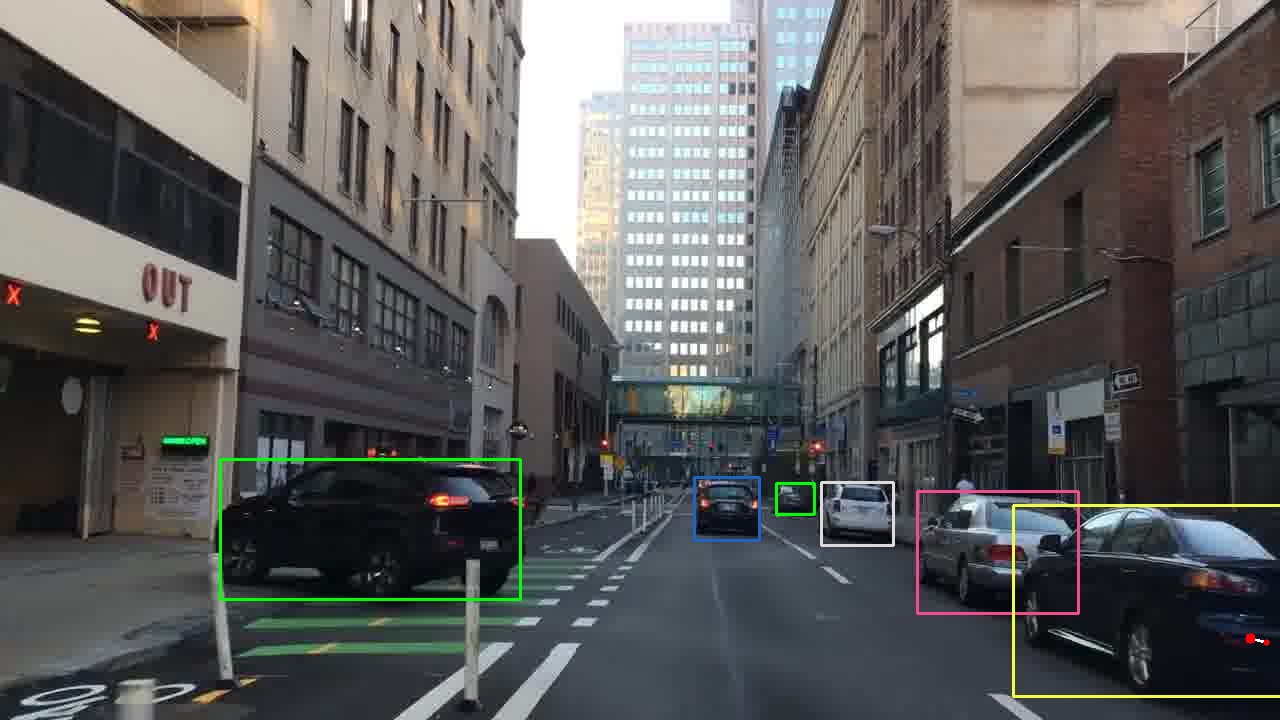}
\includegraphics[width=0.24\textwidth]{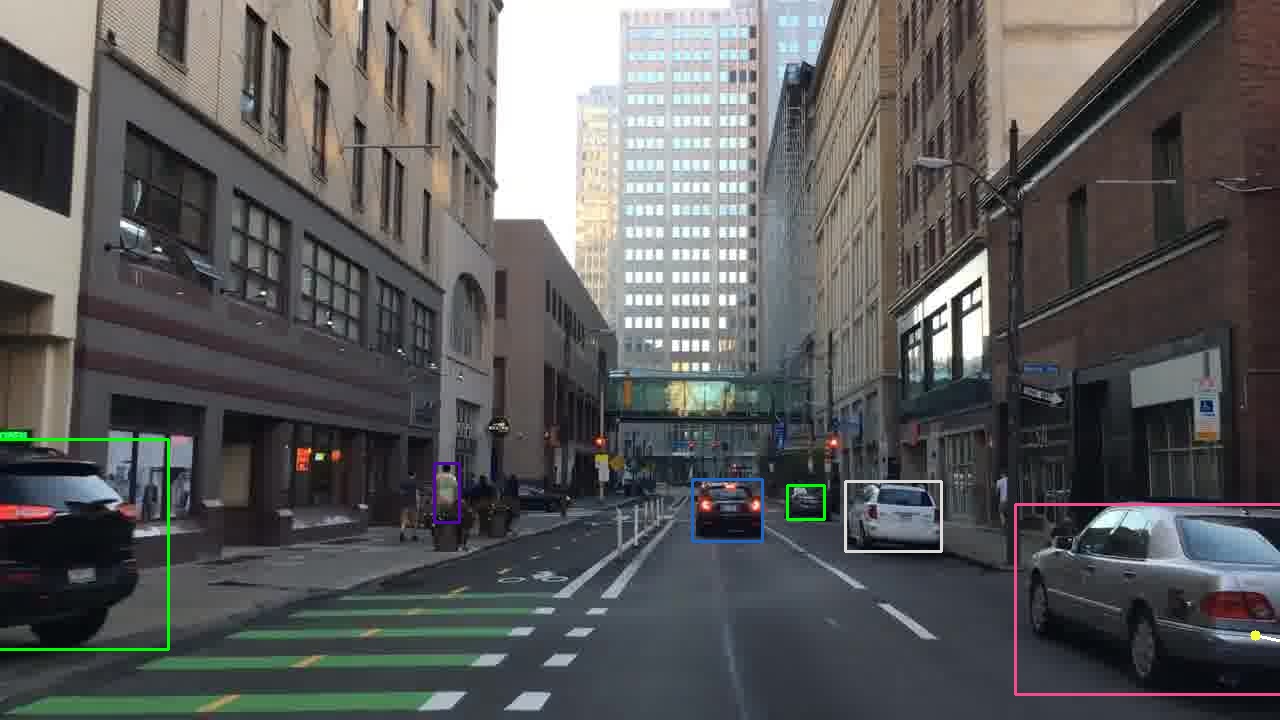}
\includegraphics[width=0.24\textwidth]{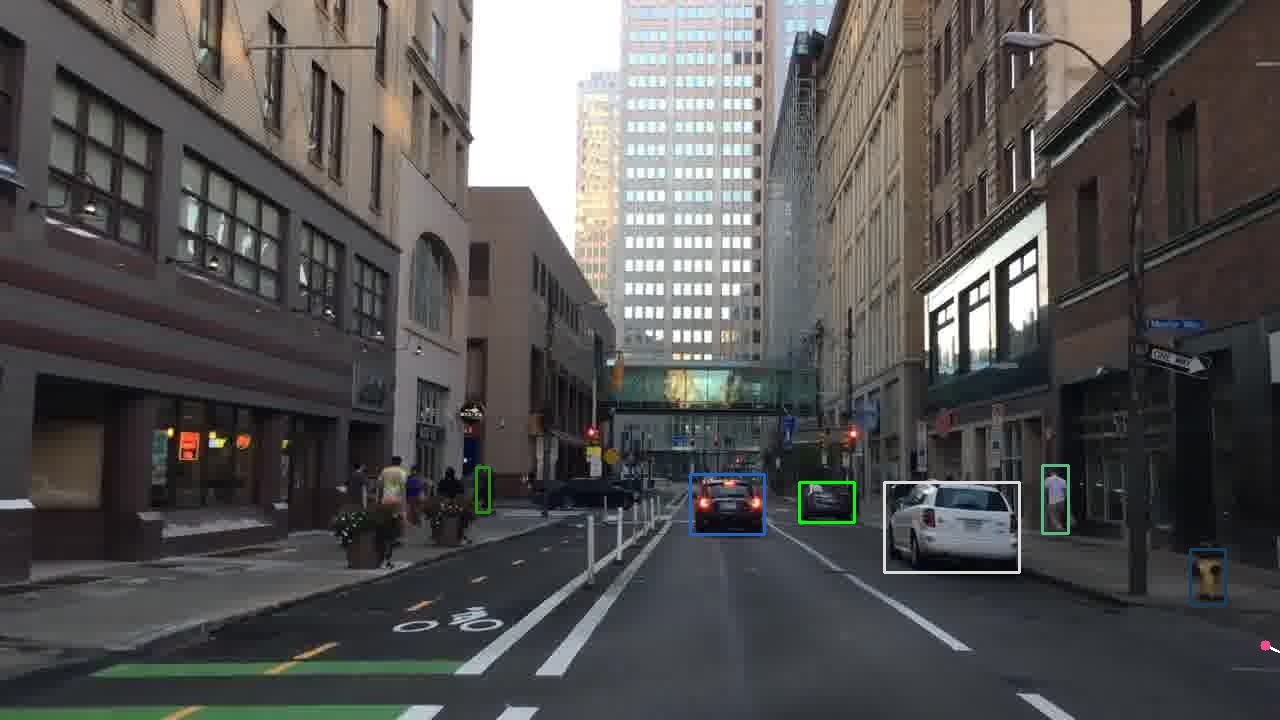}
\includegraphics[width=0.24\textwidth]{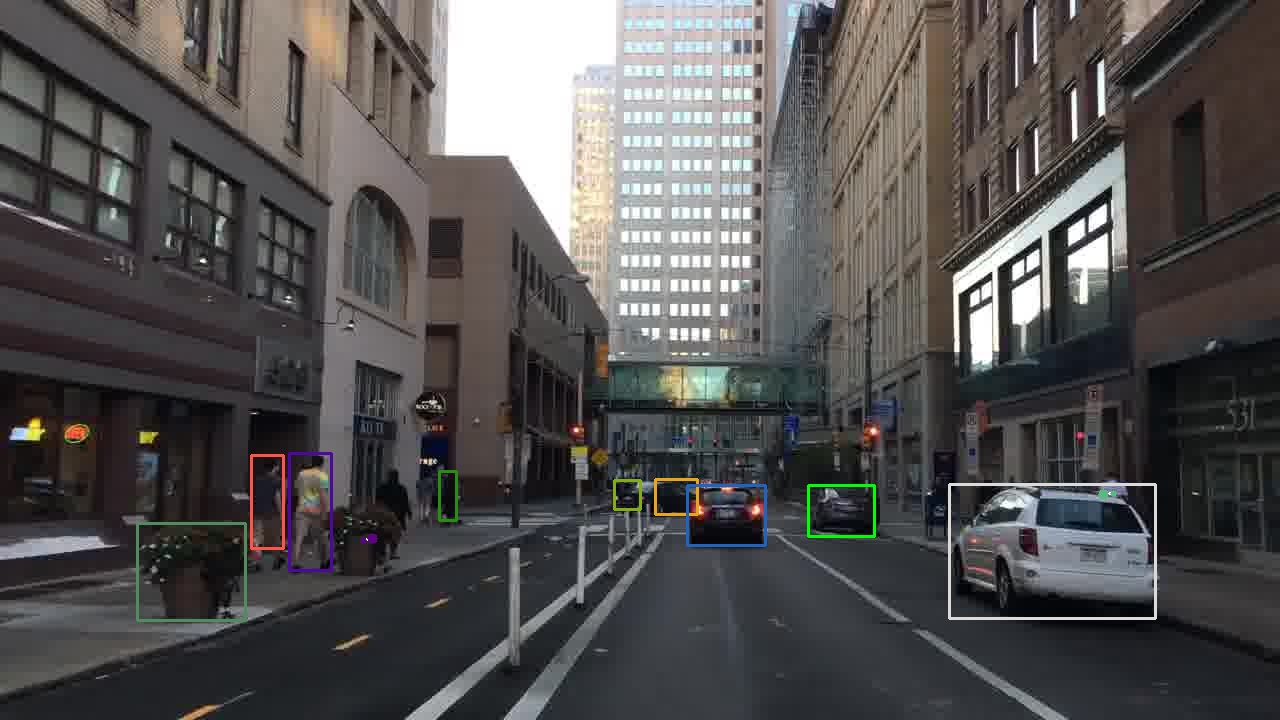}

\includegraphics[width=0.24\textwidth]{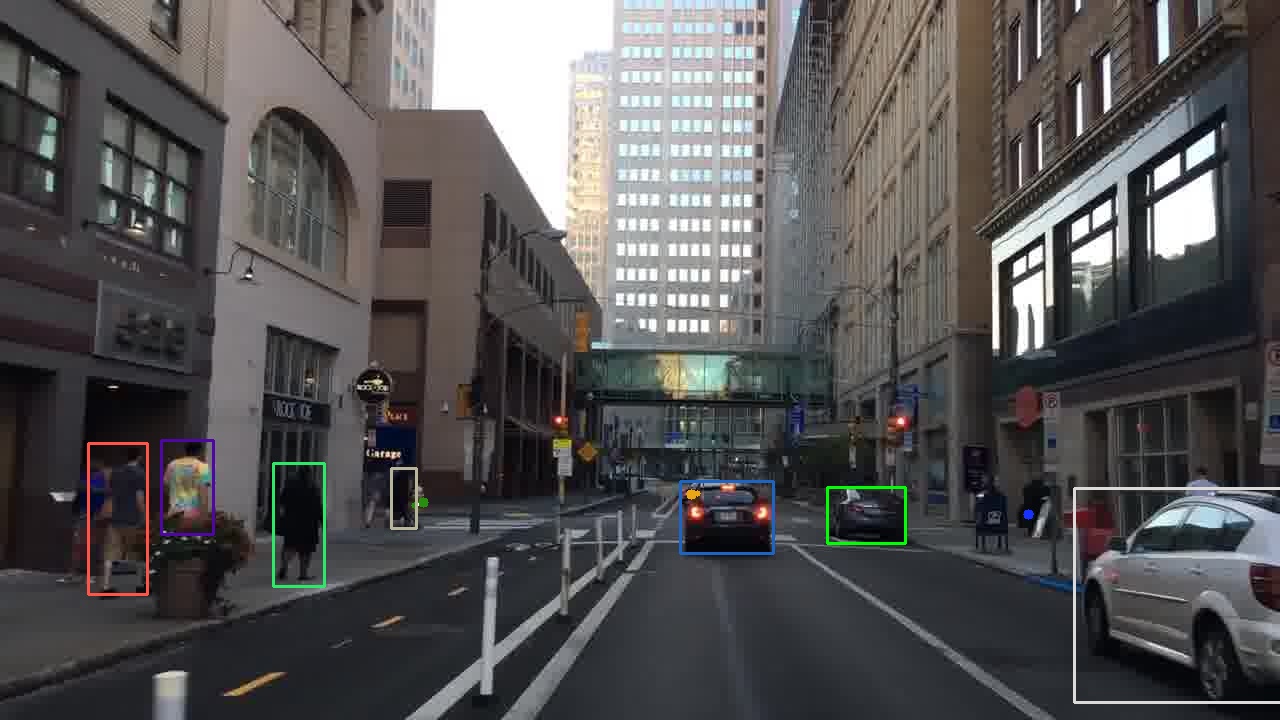}
\includegraphics[width=0.24\textwidth]{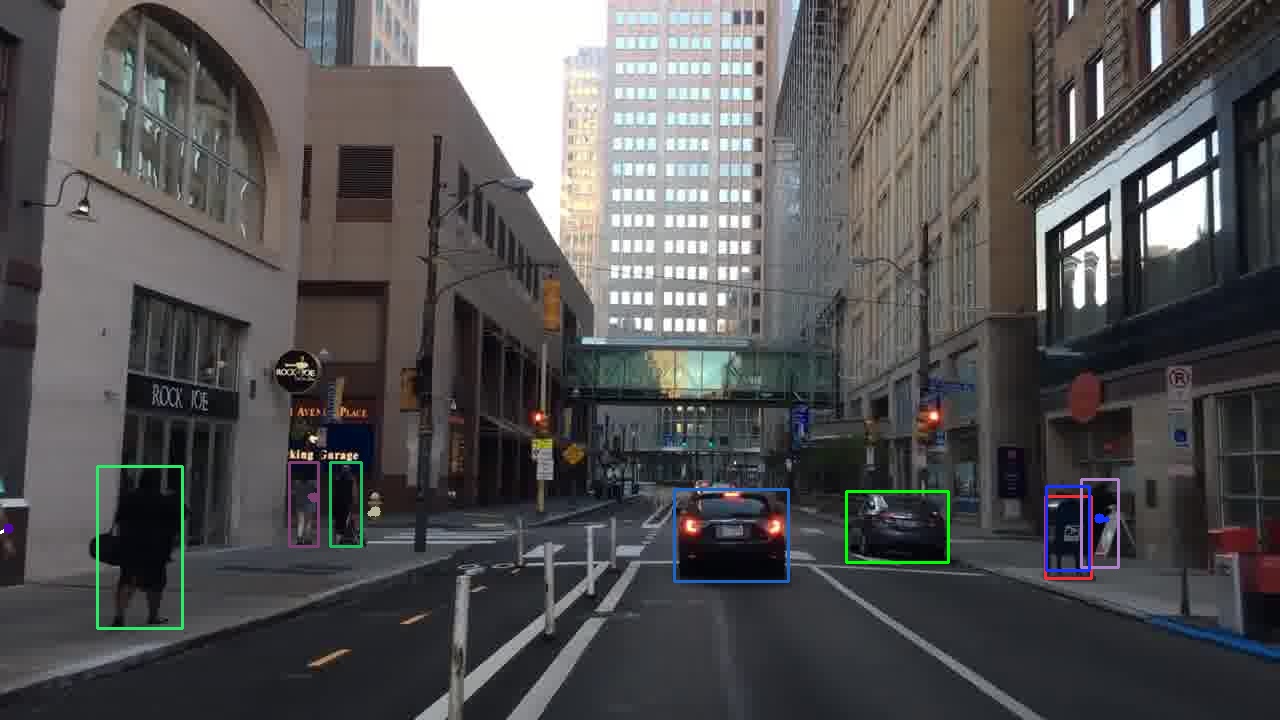}
\includegraphics[width=0.24\textwidth]{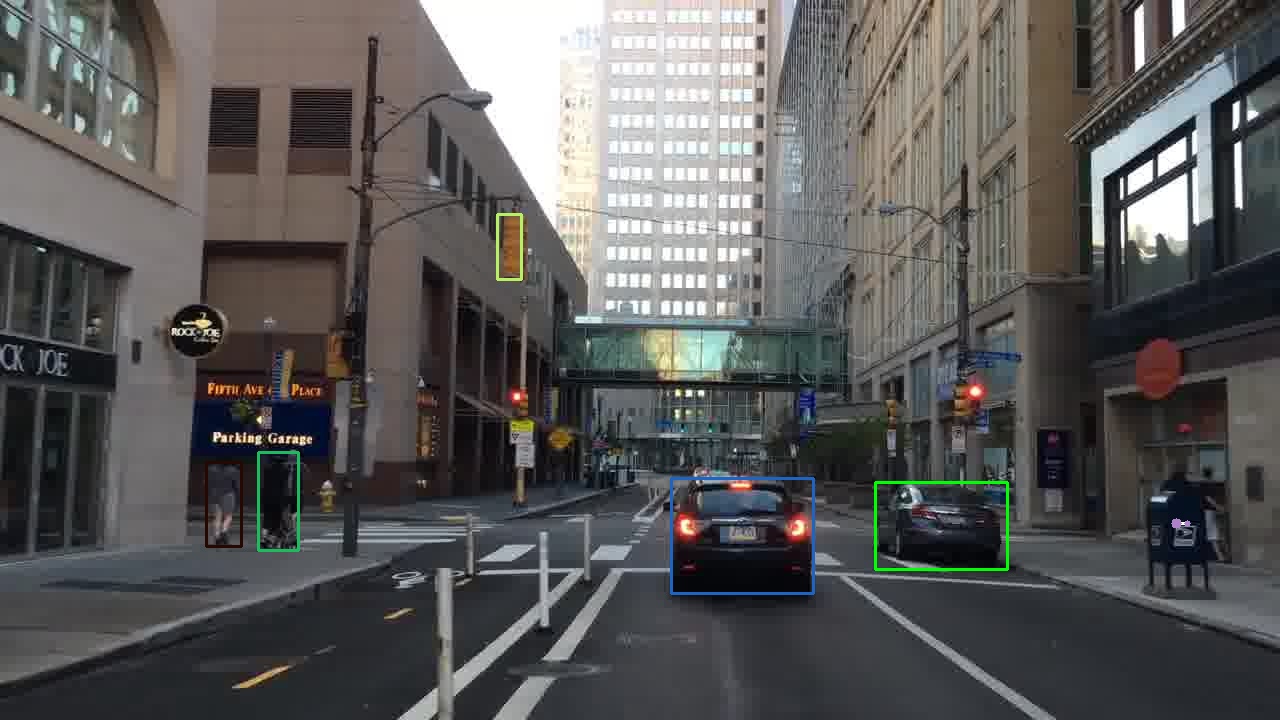}
\includegraphics[width=0.24\textwidth]{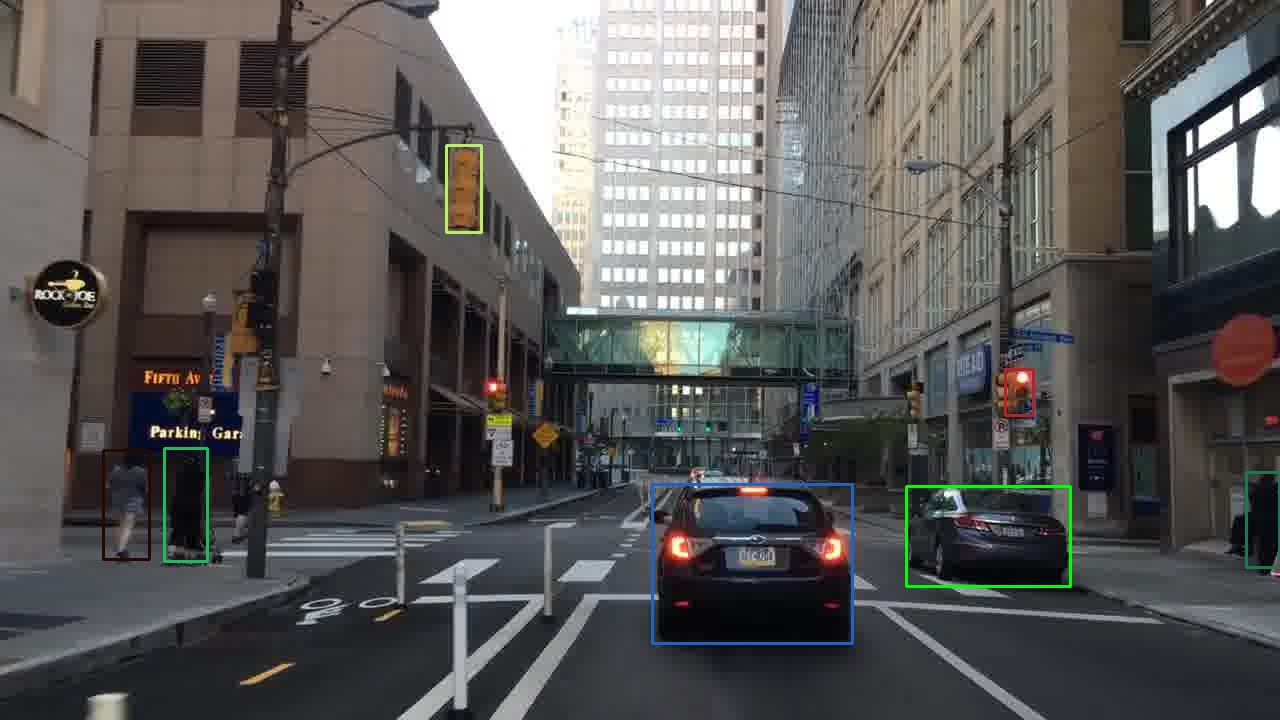}

\includegraphics[width=0.24\textwidth]{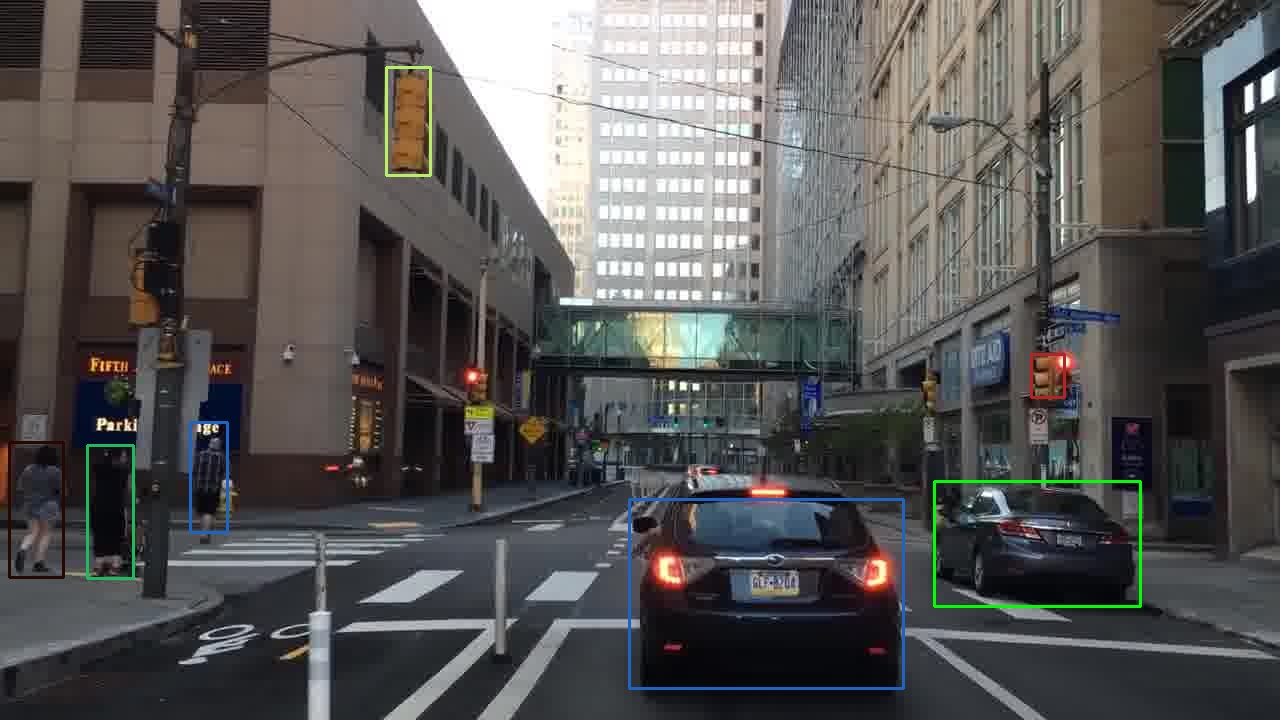}
\includegraphics[width=0.24\textwidth]{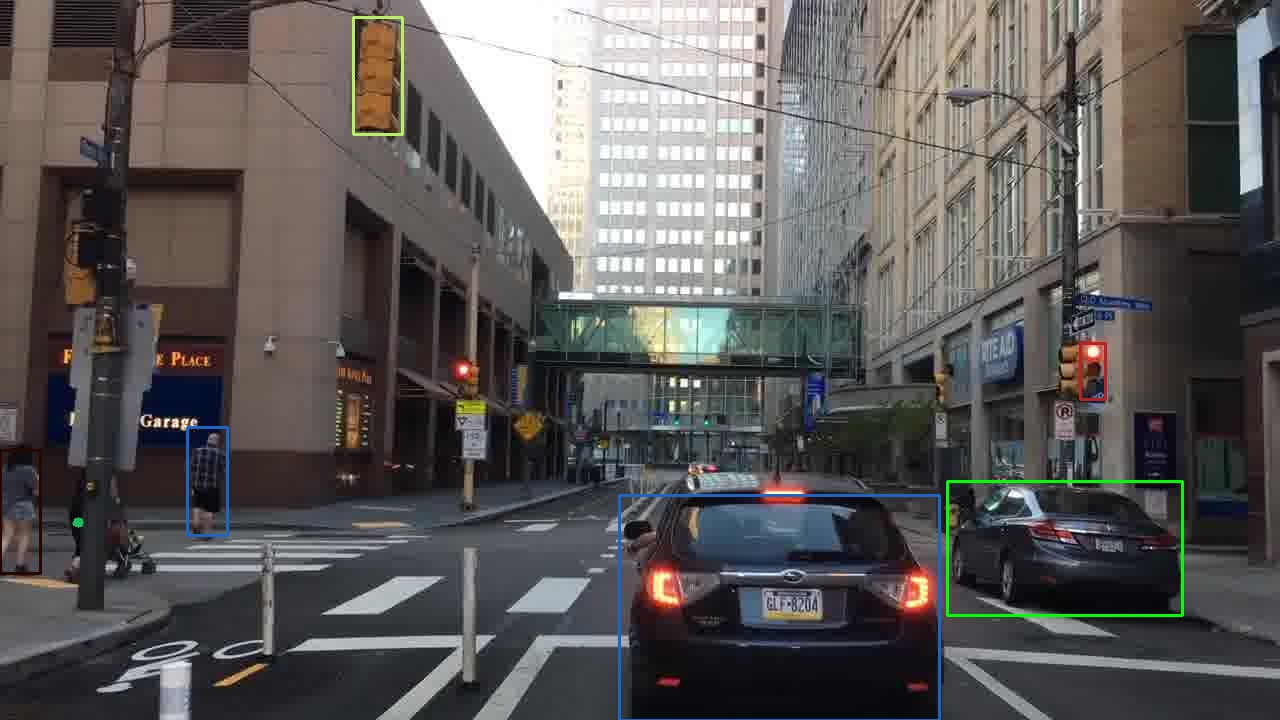}
\includegraphics[width=0.24\textwidth]{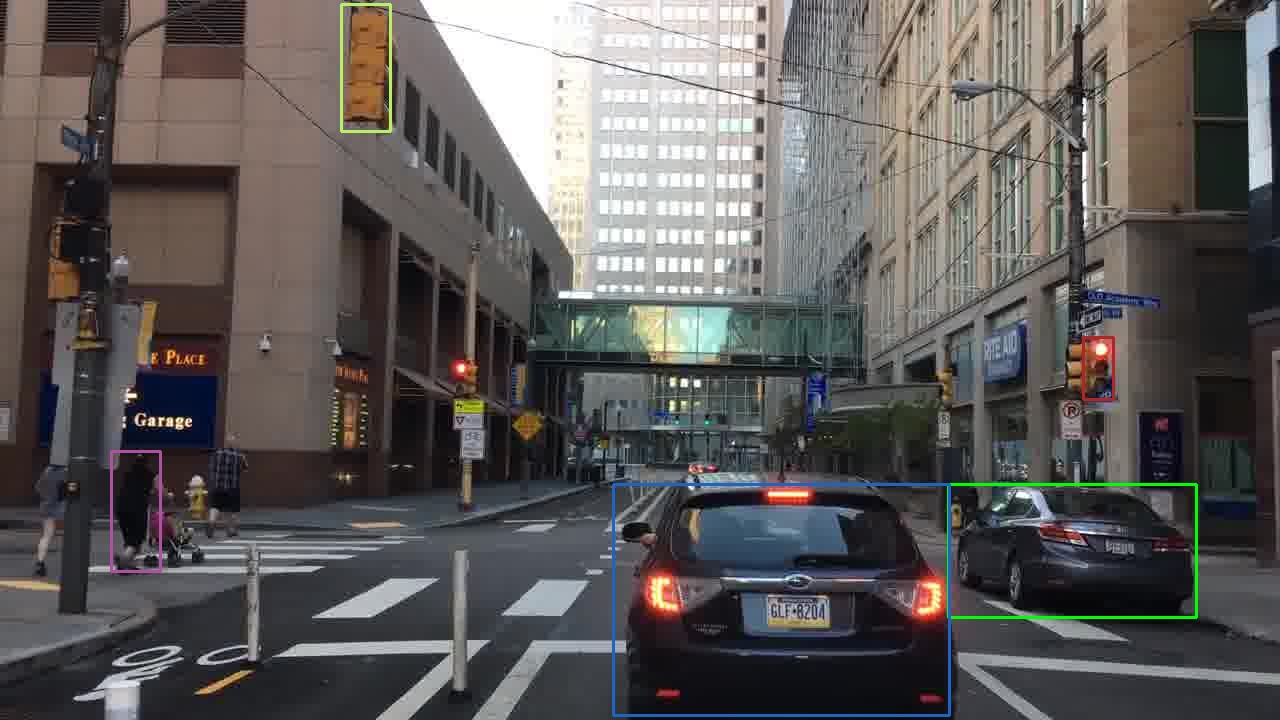}
\includegraphics[width=0.24\textwidth]{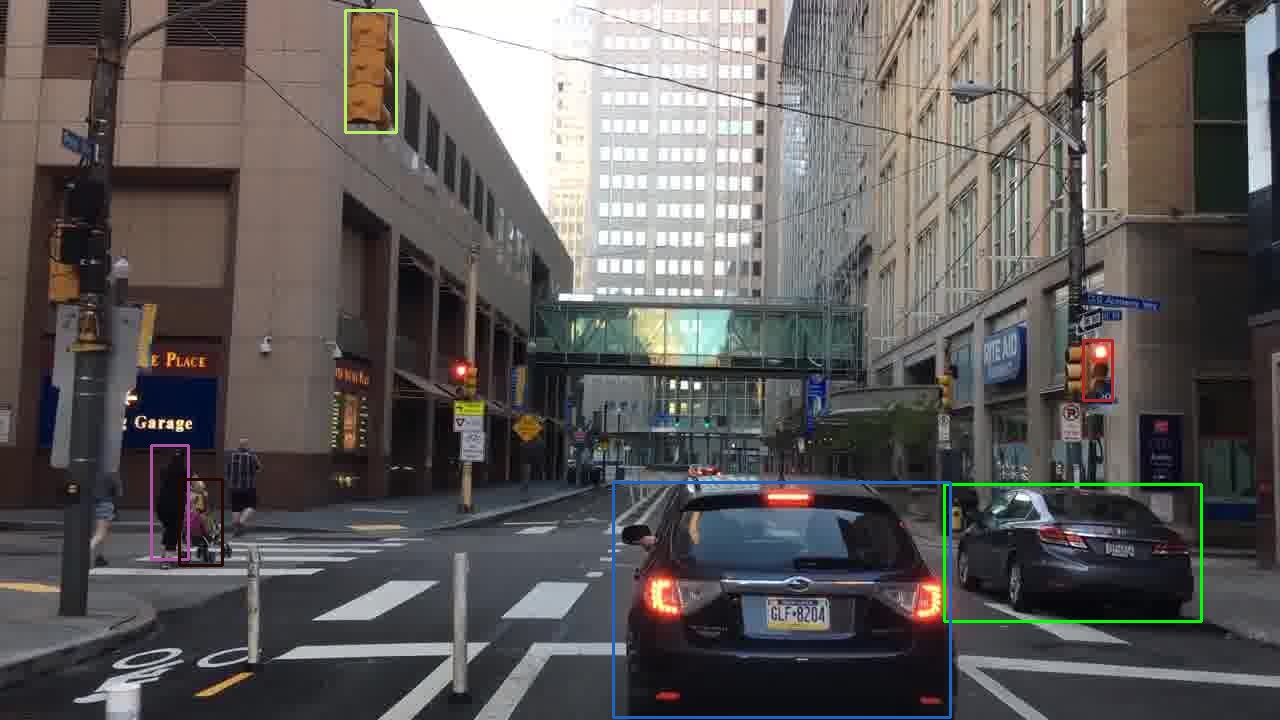}

\caption{{\sffamily\footnotesize Tracking results for the complete scene of the occlusion example (Fig \ref{fig:example_occlusion}; Section \ref{sec:reasoning-hidden-entities}) involving tracking of cars, pedestrians, and traffic lights}}
\label{fig:app:example_pgh}
\end{figure}

The following are results for select scenes from the datasets being used in the evaluation (Sec \ref{sec:application-sec}): KITTIMOD, MOT, and safety-criticality set of scenarios developed as part of this work. For lack of space, we only choose to illustrate one select frames per sec of input stimuli:

\medskip
\medskip

\begin{itemize}

	\item Figure \ref{fig:app:example_kitti}:\quad  Scene 20 from { KITTIMOD} \citep{Geiger2012CVPR}  tracking dataset

	\item Figure \ref{fig:app:example_MOT}:\quad Scene 02 from the MOT Challenge \citep{MOT16-Benchmark}
	
	\item Figure \ref{fig:app:example_pgh}:\quad Scene from safety-critical scenario dataset (Sec \ref{sec:reasoning-hidden-entities})
\end{itemize}

\newpage

\newpage

$~$

\newpage

\makeBlue{
\section{EXAMPLE DATA}\label{app:data}
}

The problem specification for each time point $t$, which is the input data for the answer set programming based abduction, is generated online based on the visual stimuli; because of the size of the data (visual observations, predictions, and matching likelihood for each frame of the video) we only include a snippet for one frame to illustrate the nature of the data.

Example problem specification ($< \mathcal{VO}_t, \mathcal{P}_t, \mathcal{ML}_t >$) generated for KITTI 0020, time point $79$.

\medskip
\medskip

\footnotesize
\begin{minted}[bgcolor=blue!5!white]{prolog}

A#const curr_time=79.

\end{minted}
\normalsize

$\mathcal{VO}_{79}$ $~~$---$~~$ Spatial entities of detected objects as bounding boxes:

\footnotesize
\begin{minted}[bgcolor=blue!5!white]{prolog}

%% detections
det(det_0, car, 99).
det(det_1, car, 99).
det(det_2, car, 99).
det(det_3, car, 99).
det(det_4, car, 95).
det(det_5, car, 91).
det(det_6, car, 84).
det(det_7, car, 75).
det(det_8, car, 72).
det(det_9, car, 46).
det(det_10, car, 40).
det(det_11, car, 25).
det(det_12, car, 22).
det(det_13, truck, 52).
det(det_14, truck, 52).

% boxes for detections
box2d(det_0, 0, 189, 208, 119).
box2d(det_1, 697, 187, 105, 68).
box2d(det_2, 220, 178, 215, 138).
box2d(det_3, 401, 183, 89, 72).
box2d(det_4, 640, 179, 38, 28).
box2d(det_5, 520, 179, 27, 23).
box2d(det_6, 473, 182, 39, 33).
box2d(det_7, 588, 179, 30, 22).
box2d(det_8, 494, 184, 29, 29).
box2d(det_9, 557, 176, 11, 14).
box2d(det_10, 475, 173, 28, 18).
box2d(det_11, 422, 174, 39, 13).
box2d(det_12, 453, 176, 24, 12).
box2d(det_13, 586, 174, 32, 22).
box2d(det_14, 579, 172, 21, 20).
\end{minted}
\normalsize

$\mathcal{P}_{79}$$~~$---$~~$Spatial entities of predicted tracks for time-point $79$ as bounding boxes:

\footnotesize
\begin{minted}[bgcolor=blue!5!white]{prolog}
% tracks and track states
trk(trk_0, car).
trk_state(trk_0, halted).
trk(trk_1, car).
trk_state(trk_1, active).
trk(trk_4, car).
trk_state(trk_4, active).
trk(trk_5, car).
trk_state(trk_5, active).
trk(trk_6, car).
trk_state(trk_6, active).
trk(trk_7, car).
trk_state(trk_7, active).
trk(trk_9, car).
trk_state(trk_9, halted).
trk(trk_11, car).
trk_state(trk_11, active).
trk(trk_12, car).
trk_state(trk_12, active).
trk(trk_13, car).
trk_state(trk_13, active).

% boxes for tracks
box2d(trk_0, -42, 227, 249, 159).
box2d(trk_1, 698, 186, 102, 68).
box2d(trk_4, 590, 179, 26, 21).
box2d(trk_5, 639, 179, 39, 27).
box2d(trk_6, 245, 187, 182, 115).
box2d(trk_7, 495, 181, 27, 31).
box2d(trk_9, 319, 184, 54, 41).
box2d(trk_11, -26, 188, 235, 113).
box2d(trk_12, 404, 181, 85, 70).
box2d(trk_13, 522, 179, 23, 22).
\end{minted}
\normalsize


$\mathcal{ML}_{79}$ $~~$---$~~$ Matching likelihood for pairs of tracks and detections at time point $79$ given by the IoU between them.

\footnotesize
\begin{minted}[bgcolor=blue!5!white]{prolog}
% IoU for overlapping tracks and detections
iou(trk_0,det_0,34625).
iou(trk_11,det_0,83400).
iou(trk_1,det_1,96879).
iou(trk_6,det_2,71071).
iou(trk_9,det_2,7556).
iou(trk_12,det_2,6388).
iou(trk_6,det_3,7330).
iou(trk_12,det_3,90971).
iou(trk_5,det_4,94824).
iou(trk_7,det_5,4551).
iou(trk_13,det_5,84391).
iou(trk_7,det_6,30697).
iou(trk_12,det_6,8403).
iou(trk_4,det_7,84365).
iou(trk_7,det_8,86273).
iou(trk_13,det_8,1145).
iou(trk_7,det_10,5146).
iou(trk_12,det_10,2138).
iou(trk_12,det_11,3087).
iou(trk_12,det_12,2341).
iou(trk_4,det_13,51257).
iou(trk_4,det_14,13495).
\end{minted}
\normalsize

\medskip
\medskip

Abduced Event Sequence  for time point $79$ (snippet for 10 time points)

\footnotesize
\begin{minted}[bgcolor=green!25!white]{prolog}
...
occurs_at(missing_detections(trk_9),69)
occurs_at(missing_detections(trk_0),70)
occurs_at(missing_detections(trk_7),72)
occurs_at(noise(trk_10),73)
occurs_at(recover(trk_7),73)
\end{minted}
\normalsize

\newpage

\normalsize
\section*{Bibliography}

\nocite{Bhatt-NeSy-2019,Suchan-ECAI-2020-Driving,ASPMT-ST-2018,clpqs-2009,geoabduction-2014,aspmtqs-tplp-2017,postdictive-asp1,postdictive-asp2,DBLP:conf/wacv/SuchanB16,DBLP:conf/eccv/SuchanBS14,DBLP:conf/cmn/BhattSS13,moral-machine2018,Bonnefon1573}

\normalsize
\bibliographystyle{abbrvnat}

\normalsize

\end{document}